\documentclass[nohyperref]{article}

\usepackage{microtype}
\usepackage{graphicx}
\usepackage{subfigure}
\usepackage{booktabs} % for professional tables

\usepackage{hyperref}

\usepackage[cmex10]{amsmath}
\usepackage{amsthm}
\usepackage[dvipsnames]{xcolor}
\usepackage{tikz} 
\usetikzlibrary{arrows.meta, calc, decorations.markings, math, arrows}

\usepackage{subfigure}

\newtheorem{theorem}{Theorem}

\newtheorem{lemma}[theorem]{Lemma}
\newtheorem{corollary}[theorem]{Corollary}

\newtheorem{claim}{Claim}

\newtheorem{proposition}{Proposition}

\usepackage{amssymb}
\usepackage{bbm}

\newcommand{\field}[1]{\mathbb{#1}}
\newcommand{\suchthat}{\;\ifnum\currentgrouptype=16 \middle\fi|\;}

\newcommand{\cF}{{\cal F}}

\DeclareMathAlphabet{\pazocal}{OMS}{zplm}{m}{n}
\newcommand{\bn}{{\bf n}}

\newcommand{\bs}{{\bf s}}

\newcommand{\bw}{{\bf w}}

\newcommand{\fR}{\field{R}} % real numbers
\newcommand{\diag}{\mbox{diag}}

\newcommand{\mathify}[1]{\ifmmode{#1}\else\mbox{$#1$}\fi}
\renewcommand{\Pr}[1]{\mathify{\mbox{{Pr}}\left[#1\right]}}

\newcommand{\Epin}[1]{\mathbb{E}_{x \sim p^{in}}\left[{#1}\right]}
\newcommand{\ab}[1]{\left\|{#1}\right\|}
\newcommand{\EP}[1]{\left\|{#1}\right\|_{p^{in}}}
\newcommand{\EE}[1]{\mathbb{E}_{x, x' \sim p^{in}}\left[{#1}\right]}
\newcommand{\epin}[1]{\mathbb{E}_{x' \sim p^{in}}\left[{#1}\right]}
\newcommand{\para}[1]{\left({#1}\right)}
\newcommand{\ve}[1]{\mathrm{vec}\left({#1}\right)}
\newcommand{\loss}{\mathcal{L}}

\newcommand{\CL}{C_L}

\newcommand{\ful}{f^l_{\theta^{(l)}}}
\newcommand{\fult}{f^l_{\theta^{(l)}(t)}}
\newcommand{\fulz}{f^l_{\theta^{(l)}(0)}}

\newcommand{\fultm}{f^{l-1}_{\theta^{(l-1)}(t)}}
\newcommand{\fulzm}{f^{l-1}_{\theta^{(l-1)}(0)}}
\newcommand{\W}{\mathfrak{W}}

\newcommand{\dit}{\delta^i_t}
\newcommand{\dLt}{\delta^L_t}
\newcommand{\convinp}{\stackrel{p}{\rightarrow}}
\newcommand{\bzero}[1]{0}

\newcommand{\idconv}{{\boldsymbol{\iota}}}

\usepackage[accepted]{icml2022}

\usepackage[capitalize,noabbrev]{cleveref}

\usepackage[textsize=tiny]{todonotes}

\icmltitlerunning{
NTK Analysis of Deep Narrow Neural Networks
}

\begin{document}

\twocolumn[
\icmltitle{Neural Tangent Kernel Analysis of Deep Narrow Neural Networks}

% It is OKAY to include author information, even for blind
% submissions: the style file will automatically remove it for you
% uN(L-1)ess you've provided the [accepted] option to the icml2022
% package.

% List of affiliations: The first argument should be a (short)
% identifier you will use later to specify author affiliations
% Academic affiliations should list Department, University, City, Region, Country
% Industry affiliations should list Company, City, Region, Country

% You can specify symbols, otherwise they are numbered in order.
% Ideally, you should not use this facility. Affiliations will be numbered
% in order of appearance and this is the preferred way.
\icmlsetsymbol{equal}{*}

\begin{icmlauthorlist}
\icmlauthor{Jongmin Lee}{ed}
\icmlauthor{Joo Young Choi}{ed}
\icmlauthor{Ernest K. Ryu}{ed}
\icmlauthor{Albert No}{HU}
\end{icmlauthorlist}

\icmlaffiliation{HU}{Department of Electronic and Electrical Engineering, Hongik University, Seoul, Korea}
\icmlaffiliation{ed}{Department of Mathematical Sciences, Seoul National University, Seoul, Korea}

\icmlcorrespondingauthor{Albert No}{albertno@hongik.ac.kr}
\icmlcorrespondingauthor{Ernest K. Ryu}{ernestryu@snu.ac.kr}

% You may provide any keywords that you
% find helpful for describing your paper; these are used to populate
% the "keywords" metadata in the PDF but will not be shown in the document
\icmlkeywords{Deep learning theory, Generative adversarial networks, Infinitely wide networks, Universal approximation theory, Random feature learning}

\vskip 0.3in
]
% this must go after the closing bracket ] following \twocolumn[ ...

% This command actually creates the footnote in the first column
% listing the affiliations and the copyright notice.
% The command takes one argument, which is text to display at the start of the footnote.
% The \icmlEqualContribution command is standard text for equal contribution.
% Remove it (just {}) if you do not need this facility.

\printAffiliationsAndNotice{}  % leave blank if no need to mention equal contribution
% \printAffiliationsAndNotice{\icmlEqualContribution} % otherwise use the standard text.

\begin{abstract}
The tremendous recent progress in analyzing the training dynamics of overparameterized neural networks has primarily focused on wide networks and therefore does not sufficiently address the role of depth in deep learning. In this work, we present the first trainability guarantee of infinitely deep but narrow neural networks. We study the infinite-depth limit of a multilayer perceptron (MLP) with a specific initialization and establish a trainability guarantee using the NTK theory. We then extend the analysis to an infinitely deep convolutional neural network (CNN) and perform brief experiments.
\end{abstract}

%%%%%%%%%%%%%%%%%%%%%%%%%%%%%%%%%%%%%%%%%%%%
%%% NTK review
%%%%%%%%%%%%%%%%%%%%%%%%%%%%%%%%%%%%%%%%%%%%
% 1.Introduction
\section{Introduction}

Despite the remarkable experimental advancements of deep learning in many domains, a theoretical understanding behind this success remains elusive. Recently, significant progress has been made by analyzing limits of infinitely large neural networks to obtain provable guarantees. The neural tangent kernel (NTK) and the mean-field (MF) theory are the most prominent results.

However, these prior analyses primarily focus on the infinite width limit and therefore do not sufficiently address the role of depth in deep learning. After all, substantial experimental evidence indicates that depth is indeed an essential component to the success of modern deep neural networks. Analyses directly addressing the limit of infinite depth may lead to an understanding of the role of depth.

In this work, we present the first trainability guarantee of infinitely deep but narrow neural networks. We study the infinite-depth limit of a multilayer perceptron (MLP) with a very specific initialization and establish a trainability guarantee using the NTK theory. The MLP uses ReLU activation functions and has width on the order of input dimension + output dimension. Furthermore, we extend the analysis to an infinitely deep convolutional neural network (CNN) and perform brief experiments.

%% 1.1 prior works
\subsection{Prior works}
The classical universal approximation theorem establishes that wide 2-layer neural networks can approximate any continuous function \cite{Cybenko1989_approximation,Funahashi1989_approximate}.
Extensions and generalizations \cite{Hornik1991_approximation, LeshnoLinPinkusSchocken1993_multilayer,Jones1992_simple,Barron1993_universal,Pinkus1999_approximation} and random feature learning \cite{RahimiRecht2007_random,RahimiRecht2008_uniform,RahimiRecht2008_weighted}, a constructive version of the universal approximation theorem, use large width in their analyses.
As overparameterization got recognized as a key component in understanding the performance of deep learning \cite{ZhangBengioHardtRechtVinyals2017_understanding}, analyses of large neural networks started to appear in the literature \cite{SoltanolkotabiJavanmardLee2019_theoretical,Allen-ZhuLiSong2019_convergence,pmlr-v97-du19c,DBLP:conf/iclr/DuZPS19,zou2020gradient, DBLP:conf/nips/LiL18a}, and their infinite-width limits such as neural network as Gaussian process (NNGP) \cite{Neal1994_priors,Neal1996_priors,Williams1997_computing,LeeBahriNovakSchoenholzPenningtonSohl-Dickstein2018_deep,MatthewsHronRowlandTurnerGhahramani2018_gaussian,NovakXiaoBahriLeeYangHronAbolafiaPenningtonSohl-dickstein2019_bayesian}, neural tangent kernel (NTK) \cite{JacotGabrielHongler2018_neural}, and mean-field (MF) \cite{MeiMontanariNguyen2018_mean,ChizatBach2018_global,RotskoffVanden-Eijnden2018_parameters,RotskoffJelassiBrunaVanden-Eijnden2019_neuron,SirignanoSpiliopoulos2020_mean,SirignanoSpiliopoulos2020_meana,NguyenPham2021_rigorous,PhamNguyen2021_global} were formulated.
NNGP characterizes the neural network at initialization, while NTK and MF analyses provide guarantees of trainability with SGD.
% The mean-field limit was used to establish 
% MF stuff
%  \citet{ChizatBach2020_implicit} showed that wide two-layer neural networks learn the a maximum margin classifier.
% \citet{NguyenPham2021_rigorous,PhamNguyen2021_global} extended the mean-field results to $L$-layers for $L\ge 3$
This line of research naturally raises the question of whether very deep neural networks also enjoy similar properties as wide neural networks, especially given the importance of depth in modern deep learning.

The analogous line of research for very deep neural networks has a shorter history.
Universality of deep narrow neural networks \cite{LuPuWangHuWang2017_expressive,HaninSellke2017_approximating, LinJegelka2018_resnet,Hanin2019_universal, KidgerLyons2020_universal,ParkYunLeeShin2021_minimum,TabuadaGharesifard2021_universal},
lower bounds on the minimum width necessary for universality \cite{LuPuWangHuWang2017_expressive,HaninSellke2017_approximating,Johnson2019_deep,ParkYunLeeShin2021_minimum}, and
quantitative analyses showing the benefit of depth over width in approximating certain functions \cite{Telgarsky2016_benefits} are all very recent developments.
On the other hand, the neural ODE \cite{ChenRubanovaBettencourtDuvenaud2018_neural} is a continuous-depth model that can be considered an infinite-depth limit of a neural network with residual connections. Also stochastic extensions of
neural ODE, viewing infinitely deep ResNets as diffusion processes, have been considered in \cite{PeluchettiFavaro2020_infinitely,peluchetti2021doubly}. However, these continuous-depth limits do not come with any trainability or generalization guarantees. 
In the infinite width \emph{and} depth limit, a quantitative universal approximation result has been established \cite{LuShenYangZhang2021_deep} and a trainability guarantee was obtained in setups combining the MF limit with the continuous-depth limit inspired by the neural ODE \cite{LuMaLuLuYing2020_mean,DingChenLiWright2021_global,DingChenLiWright2021_overparameterization}.

% \jy{For infinite width and depth, this paper from SIAM could also be cited \cite{LuShenYangZhang2021_deep}:  the optimal approximation error characterization of deep rectified linear unit (ReLU) networks for smooth functions in terms of both width and depth simultaneously.}

% Finally, there has been work analyzing the networks in the limit where both the depth and width are infinite 

% One can make the argument that equilibrium models are models with infinite depth.
% \cite{BaiKolterKoltun2019_deep,WinstonKolter2020_monotone}

Efforts to understand the trainability of deep non-wide neural networks have been made.
\citet{AroraCohenGolowichHu2019_convergence,Shamir2019_exponential} establishes trainability guarantees for linear deep networks (no activation functions).
\citet{PenningtonSchoenholzGanguli2017_resurrecting} studied the so-called dynamic isometry property, \citet{Hanin2018_which} studied the exploding and vanishing gradient problem for ReLU MLPs, and \citet{HuangWangTaoZhao2020_why} studied the NTK of deep MLPs and ResNets at initialization, but these results are limited to the state of the neural network at initialization and therefore do not directly establish guarantees on the training dynamics.
To the best of our knowledge, no prior work has yet established a trainability guarantee on (non-linear) deep narrow neural networks.

Many extensions and variations of the NTK have been studied:
NTK analysis with convolutional layers \cite{aroraExactComputationInfinitely2019, DBLP:journals/corr/abs-1902-04760},
further refined analyses and experiments \cite{LeeXiaoSchoenholzBahriNovakSohl-DicksteinPennington2019_wide},
NTK analysis with regularizers and noisy gradients \cite{ChenCaoGuZhang2020_generalized},
finite-width NTK analysis \cite{HaninNica2020_finite}, 
quantitative universality result of the NTK \cite{JiTelgarskyXian2020_neural},
generalization properties of overparameterized neural networks \cite{BiettiMairal2019_inductive,WoodworthGunasekarLeeMoroshkoSavareseGolanSoudrySrebro2020_kernel},
unified analysis of NTK and MF \cite{GeigerSpiglerJacotWyart2020_disentangling},
notion of lazy training generalizing linearization effect of the NTK regime \cite{DBLP:conf/nips/ChizatOB19},
closed-form evaluations of NTK kernel values \cite{ChoSaul2009_kernel,aroraExactComputationInfinitely2019},
analyses of distribution shift and meta learning in the infinite-width regime \cite{AdlamLeeXiaoPenningtonSnoek2021_exploring,NguyenNovakXiaoLee2021_dataset},
library implementations of infinitely wide neural networks as kernel methods with NTK \cite{NovakXiaoHronLeeAlemiSohl-DicksteinSchoenholz2020_neural},
empirical evaluation of finite vs.\ infinite neural networks \cite{LeeSchoenholzPenningtonAdlamXiaoNovakSohl-Dickstein2020_finite},
and constructing better-performing kernels with improved extrapolation of the standard initialization \cite{Sohl-DicksteinNovakSchoenholzLee2020_infinite}.
Such results have mostly focused on the analysis and application of wide, rather than deep, neural networks.

% \cite{TzenRaginsky2020_meanfield} XX??XX

%% 1.2 contribution
\subsection{Contribution}
In this work, we analyze the training dynamics of a deep narrow MLP and CNN without residual connections and establish trainability guarantees. To the best of our knowledge, this is the first trainability guarantee of deep narrow neural networks, with or without residual connections; prior trainability guarantees consider the infinite width or the infinite depth and width limits. Essential to our analysis is the particular choice of initialization, as very deep networks with the usual initialization schemes are known to not be trainable \citep[Section 4.4]{HeSun2015_convolutional},
\citep[Section 3.1]{SrivastavaGreffSchmidhuber2015_highway}, \citep[Section~4.1]{HuangWangTaoZhao2020_why}.
By considering the limit in which only the depth is infinite, we demonstrate that infinite depth is sufficient to obtain a neural network with trainability guarantees.

% Our proof technique largely follows the NTK theory's framework, but 
The key technical challenge of our work arises from bounding the variation of the scaled NTK in the infinite-depth limit via a delicate Lyapunov analysis. Our infinite-depth analysis requires that we control an infinite composition of layers and an infinite product of matrices, which is significantly more technical than the prior infinite-width analyses that can more directly apply the central limit theorem and law of large numbers.

The MLP we analyze is narrow in the sense that it is within a factor of 2 of the width lower bound for universal approximation \cite{ParkYunLeeShin2021_minimum}. In particular, the width does not depend on the number of data points.

\section{Preliminaries and notation}
In this section, we review the necessary background and set up the notation. We largely follow the notions of \citet{JacotGabrielHongler2018_neural}, although our notation has some minor differences.
% We review the notion of kernel gradient descent as defined \citep[Section~3.1]{JacotGabrielHongler2018_neural}

Given a function $f_\theta(x)\in \mathbb{R}^n$ with $\theta\in \mathbb{R}^p$ and $x\in \fR^d_+$, write $\partial_\theta f_\theta(x)\in \mathbb{R}^{n\times p}$ to denote the Jacobian matrix with respect to $\theta$. 
If $f_\theta(x)\in \mathbb{R}$ is scalar-valued, write $\nabla_\theta f_\theta(x)\in \mathbb{R}^{p\times 1}$ for the gradient.
The gradient and Jacobian matrices are, by convention, transposes of each other, i.e., $\nabla_\theta f_\theta(x)=(\partial_\theta f_\theta(x))^\intercal$.
Write $\|\cdot\|$ to denote the standard Euclidean norm for vectors and the standard operator norm for matrices.
Write $\left \langle  \cdot,\cdot \right \rangle$ for the vector and Frobenius  inner products.
Write $\fR^d_+\subset \fR^d$ for the strict positive orthant, i.e., $\fR^d_+$ is the set of vectors with element-wise positive entries.
Write $\convinp$ to denote convergence in probability. 
Write $g \sim \mathcal{GP}(\mu,\Sigma)$ to denote that $g$ is a Gaussian process with mean $\mu(x)$ and covariance kernel $\Sigma(x,x')$ \cite{Rasmussen2004_gaussian,RasmussenWilliams2005_gaussian}.

% Write % $\stackrel{p}{\rightarrow}$ to denote convergence in probability.

%% 2.2 gradient flow training
Let $p^{in}$ be the empirical distribution on a training dataset $x_1, x_2, \dots, x_N\in \fR^{d_{in}}_+ $, which we assume are element-wise positive.
Let $\cF=\{f\colon\fR^{d_{in}}_+ \rightarrow \fR^{d_{out}}\}$ be the function space with a seminorm $\EP{\cdot}$
induced by the bilinear map
\begin{align*}
\langle f, g \rangle_{p^{in}} &= \mathbb{E}_{x\sim p^{in}} [f(x)^\intercal g(x)]. \\
&
\end{align*}

If $W$ is a matrix-valued function, write
\[
\EP{W}^2=\mathbb{E}_{x\sim p^{in}} \|W(x)\|^2,
\]
where $\|\cdot\|$ here is the operator norm. 

% \begin{figure}[h]
% \vspace{-0.25in}
%     \centering
%     \includegraphics{figures/mlp_arch.pdf}
% \vspace{-0.25in}
% \caption{Initialization of deep MLP with $d_{in} = 3$ and $d_{out} = 2$. Line styles indicate types of weight initializations (solid: 1, double: $C_L$, dash: Gaussian, no line: 0). Box styles indicate types of bias initializations (solid box: 0, dash: Gaussian, double: $C_L$, double dash: $-C_L$).}
%     \label{fig:mlp_arch}
% \end{figure}

\paragraph{Kernel gradient flow.}
Let $\loss\colon\cF\rightarrow \fR$ be a functional loss.
We primarily consider $\loss(f) = \frac{1}{2}\EP{f-f^\star}^2$ with a given target function $f^\star$.
We train a neural network $f_\theta$ by solving
\[
\begin{array}{ll}
\underset{\theta }{\mbox{minimize}}&
\loss(f_\theta).
\end{array}
\]
Let $\cF^*$ denote the dual of $\cF$ with respect to $p^{in}$.
So $\cF^*$ consists of linear maps $\langle \delta, \cdot\rangle_{p^{in}}$ for some $\delta\in\cF$.
% define a map $\Xi_K:\cF^*\rightarrow\cF$ where $\langle \delta, \cdot\rangle_{p_{in}} \mapsto \epin{K(x, x')\delta(x')}$.
Let $\partial_f \loss\vert_{f_0}$ denote the functional derivative of the loss at $f_0$.
Since $\partial_f \loss\vert_{f_0}\in \cF^*$, there exists a corresponding dual element $\delta\vert_{f_0}\in \cF$, where $ \partial_f \loss\vert_{f_0} = \langle \delta\vert_{f_0}, \cdot\rangle_{p^{in}}$.
To clarify, $\delta\vert_{f_0}(x)$ is defined to be a length $d_{out}$ column vector.
% $\delta\vert_{f_0}(x)\in\mathbb{R}^{d_{out}\times 1}$ for all $x\in \mathbb{R}^{d_{in}}$.

A multi-dimensional kernel 
$K\colon\fR^{d_{\mathrm{in}}}_+\times \fR^{d_{\mathrm{in}}}_+\to \fR^{d_{\mathrm{out}}\times d_{\mathrm{out}}}$ is a function such that $K(x,x')=K(x',x)^\intercal$ for all $x,x'\in \fR^{d_{\mathrm{in}}}_+$.
A multi-dimensional kernel is positive semidefinite if 
\[
\mathbb{E}_{x,x'\sim p^{in}}\left[f(x)^\intercal K(x,x') f(x')\right]
\ge 0
\]
for all $f\in \cF$ and (strictly) positive definite if the inequality holds strictly when $\EP{f}\ne 0$.
To clarify, $\mathbb{E}_{x,x'\sim p^{in}}$ denotes the expectation with respect to $x$ and $x'$ sampled independently from $p^{in}$.
The kernel gradient of $\loss$ at $f_0$ with respect to the kernel $K$ is defined as
\begin{align*}
\nabla_K\loss\vert_{f_0}(x) = \epin{K(x, x')\delta\vert_{f_0}(x')}
\end{align*}
for all $x\in \fR_+^{d_{\mathrm{in}}}$.
% The kernel gradient is not restricted to the dataset,
% and it can be viewed as a generalization of functional gradient which is oN(L-1)y defined on the dataset.
We say a time-dependent function $f_t$ follows the kernel gradient flow with respect to $K$ if 
\[
\partial_tf_t(x) = -\nabla_K\loss\vert_{f_t}(x)
\]
for all $t>0$ and $x\in \fR^{d_{\mathrm{in}}}_+$. During the kernel gradient flow, the loss $\loss(f_t)$ evolves as
\begin{align*}
    \partial_t\loss\vert_{f_t} =& -\mathbb{E}_{x,x'\sim p^{in}}\left[\delta\vert_{f_t}(x)^\intercal K(x,x') \delta\vert_{f_t}(x')\right].
\end{align*}
If $K$ is positive definite and if certain regularity conditions hold, then kernel gradient flow converges a critical point and it converges to a global minimum if $\loss$ is convex and bounded from below.

\paragraph{Neural tangent kernel.}
Given a neural network $f_{\theta(t)}$, \citet{JacotGabrielHongler2018_neural} defines the neural tangent kernel (NTK) at time $t$ as
\[
\Theta_t(x, x') = \partial_{\theta} f_{\theta(t)}(x)\para{\partial_{\theta}f_{\theta(t)}(x')}^{\intercal},
\]
which, by definition, is a positive semidefinite kernel.
\citet{JacotGabrielHongler2018_neural} pointed out that a
neural network trained with gradient flow, 
which we define and discuss in Section~\ref{ss:gradient_flow}, 
can be viewed as kernel gradient flow with respect to $\Theta_t$, i.e.,
\[
\partial_tf_{\theta(t)}(x) = -\nabla_{\Theta_t} \loss\vert_{f_t}(x).
\]
However, even though $\Theta_t$ is always positive semidefinite, the time-dependence of $\Theta_t$ makes the training dynamics non-convex and prevents one from establishing trainability guarantees in general.
The contribution of \citet{JacotGabrielHongler2018_neural} is showing that $\Theta_t\convinp\Theta$ in an appropriate infinite-width limit,
where $\Theta$ is a fixed limit that does not depend on time.
Then, since $\Theta$ is fixed, kernel gradient flow generically converges provided that the loss $\loss$ is convex.

% 2. mlp
\section{NTK analysis of infinitely deep MLP}
Consider an $L$-layer multilayer perceptron (MLP) $f^L_{\theta}\colon\fR^{d_{\mathrm{in}}}_+ \rightarrow \fR^{d_{\mathrm{out}}}$ parameterized by $\theta$,
where the input $x\in \fR^{d_{\mathrm{in}}}_+$ is a $d_{\mathrm{in}}$-dimensional vector with positive entries and the output is a $d_{\mathrm{out}}$-dimensional vector.
The network consists of $L-1$ fully connected hidden layers with uniform width $d_{\mathrm{in}}+d_{\mathrm{out}}+1$, 
each followed by the ReLU activation function. The final output layer has width $d_{\mathrm{out}}$ and is not followed by an activation function.
% Each layer contains  neurons followed by a ReLU activations.

Let us set up specific notation.
Define the pre-activation values as 
\begin{align*}
f^1_{\theta^{(1)}}(x) &= W^1x+b^1\\
    \ful(x) &= W^{l}\sigma(f^{l-1}_{\theta^{(l-1)}}(x))+b^l ,
    \qquad
    2\leq l\leq L.
\end{align*}
We use ReLU for the activation $\sigma$.
The weight matrices have dimension $W^1\in \fR^{(d_{\mathrm{in}} + d_{\mathrm{out}} + 1)\times{d_{\mathrm{in}}}}$, $W^2,\dots,W^{L-1}\in \fR^{(d_{\mathrm{in}} + d_{\mathrm{out}} + 1)\times (d_{\mathrm{in}} + d_{\mathrm{out}} + 1)}$, and $W^L \in \fR^{d_{\mathrm{out}}\times(d_{\mathrm{in}} + d_{\mathrm{out}} + 1)}$.
The bias vectors have dimension $b^1,\dots,b^{L-1}\in\fR^{(d_{\mathrm{in}} + d_{\mathrm{out}} + 1)\times 1}$ and $b^L\in\fR^{d_{\mathrm{out}}\times 1}$.
For $1\le l\le L$, write $\theta^{(l)} = \{W^i, b^i : i\leq l\}$
to denote the collection of parameters up to $l$-th layer.
Let $f^L_{\theta}=f^L_{\theta^{(L)}}$.

%% 2.1 initialization
\subsection{Initialization} \label{sec:mlp_initialization}
\begin{figure}
\vspace{0.1cm}
\centering
\tikzset{every picture/.style={line width=1pt}}
\def\bw{19.5} % box width / height
\def\bs{9.5} % space between boxes
\def\lw{17.5} % weight width
\begin{tikzpicture}[x=0.8pt,y=0.8pt,yscale=-1,xscale=1]

% first layer
\foreach \bn in {0, 1, 2}{
{\draw  ({\bn * (\bs + \bw)}, {8 * (\lw + \bw)}) -- ({\bn * (\bs + \bw) + \bw}, {8 * (\lw + \bw)}) -- ({\bn * (\bs + \bw) + \bw}, {8 * (\lw + \bw) + \bw}) -- ({\bn * (\bs + \bw)}, {8 * (\lw + \bw) + \bw}) -- cycle ;
}
\draw    ({\bn * (\bs + \bw) + 0.5 * \bw - 1.5},{7 * (\lw + \bw) + \bw}) -- ({\bn * (\bs + \bw) + 0.5 * \bw - 1.5}, {8 * (\lw + \bw)})({\bn * (\bs + \bw) + 0.5 * \bw + 1.5},{8 * (\lw + \bw)}) -- ({\bn * (\bs + \bw) + 0.5 * \bw + 1.5},{7 * (\lw + \bw) + \bw}) ;
\draw[densely dashed]  ({3 * (\bs + \bw) + 0.5 * \bw},{7 * (\lw + \bw) + \bw}) -- ({\bn * (\bs + \bw) + 0.5 * \bw}, {8 * (\lw + \bw)}) ;
}

\foreach \bn in {4, 5}{
\draw  ({\bn * (\bs + \bw) - 0.1 * \bw}, {7 * (\lw + \bw) - 0.1 * \bw}) -- ({\bn * (\bs + \bw) + 1.1 * \bw}, {7 * (\lw + \bw) - 0.1 * \bw}) -- ({\bn * (\bs + \bw) + 1.1 * \bw}, {7 * (\lw + \bw) + 1.1 * \bw}) -- ({\bn * (\bs + \bw) - 0.1 * \bw}, {7 * (\lw + \bw) + 1.1 * \bw}) -- cycle ;
\draw    ({\bn * (\bs + \bw) + 0.5 * \bw},{7 * (\lw + \bw) - 0.1 * \bw}) -- ({\bn * (\bs + \bw) + 0.5 * \bw},{7 * (\lw + \bw) - \lw}) ;
}

\foreach \hn in {7, 6, 5, 2}{
{\foreach \bn in {0, 1, 2}{
\draw  ({\bn * (\bs + \bw)}, {\hn * (\lw + \bw)}) -- ({\bn * (\bs + \bw) + \bw}, {\hn * (\lw + \bw)}) -- ({\bn * (\bs + \bw) + \bw}, {\hn * (\lw + \bw) + \bw}) -- ({\bn * (\bs + \bw)}, {\hn * (\lw + \bw) + \bw}) -- cycle ;
\draw    ({\bn * (\bs + \bw) + 0.5 * \bw},{\hn * (\lw + \bw)}) -- ({\bn * (\bs + \bw) + 0.5 * \bw},{\hn * (\lw + \bw) - \lw}) ;
\draw[densely dashed]  ({3 * (\bs + \bw) + 0.5 * \bw},{(\hn - 1) * (\lw + \bw) + \bw}) -- ({\bn * (\bs + \bw) + 0.5 * \bw}, {\hn * (\lw + \bw)}) ;
}
}

\draw[densely dashed]  ({3 * (\bs + \bw)}, {\hn * (\lw + \bw)}) -- ({3 * (\bs + \bw) + \bw}, {\hn * (\lw + \bw)}) -- ({3 * (\bs + \bw) + \bw}, {\hn * (\lw + \bw) + \bw}) -- ({3 * (\bs + \bw)}, {\hn * (\lw + \bw) + \bw}) -- cycle ;

{\foreach \bn in {4, 5}
\draw  ({\bn * (\bs + \bw)}, {\hn * (\lw + \bw)}) -- ({\bn * (\bs + \bw) + \bw}, {\hn * (\lw + \bw)}) -- ({\bn * (\bs + \bw) + \bw}, {\hn * (\lw + \bw) + \bw}) -- ({\bn * (\bs + \bw)}, {\hn * (\lw + \bw) + \bw}) -- cycle ;
}
}
\foreach \hn in {1, 2, 3, 4}{
\node[right, font=\small] at ({5* (\bs + \bw) + 1. 5 * \bw}, {(8 - \hn) * (\lw + \bw) + \bw - 0.45 * \bw}) {Layer ${\hn}$} ;
}

\node[right, font=\small] at ({5* (\bs + \bw) + 1.5 * \bw}, {0 * (\lw + \bw) + \bw - 0.45 * \bw}) {Layer $L$} ;

\node[right, font=\small] at ({5* (\bs + \bw) + 1.5 * \bw}, {1 * (\lw + \bw) + \bw - 0.45 * \bw }) {Layer $L-1$} ;

\node[right, font=\small] at ({5* (\bs + \bw) + 1.5 * \bw}, {2 * (\lw + \bw) + \bw - 0.45 * \bw }) {Layer $L-2$} ;

%vertical dots
\draw[very thick, dotted]   ({0.5 * (5 * (\bw + \bs) + \bw)},{3 * (\bw + \lw)}) -- ({0.5 * (5 * (\bw + \bs) + \bw)},{3 * (\bw + \lw) + 1.1 * \bw}) ;

\foreach \hn in {6, 5, 2}
{\foreach \bn in {4, 5}
\draw    ({\bn * (\bs + \bw) + 0.5 * \bw},{\hn * (\lw + \bw)}) -- ({\bn * (\bs + \bw) + 0.5 * \bw},{\hn * (\lw + \bw) - \lw}) ;
}

\foreach \hn in {4, 1}{
{\foreach \bn in {0, 1, 2}
\draw  ({\bn * (\bs + \bw)}, {\hn * (\lw + \bw)}) -- ({\bn * (\bs + \bw) + \bw}, {\hn * (\lw + \bw)}) -- ({\bn * (\bs + \bw) + \bw}, {\hn * (\lw + \bw) + \bw}) -- ({\bn * (\bs + \bw)}, {\hn * (\lw + \bw) + \bw}) -- cycle ;
}
\draw[densely dashed]  ({3 * (\bs + \bw)}, {\hn * (\lw + \bw)}) -- ({3 * (\bs + \bw) + \bw}, {\hn * (\lw + \bw)}) -- ({3 * (\bs + \bw) + \bw}, {\hn * (\lw + \bw) + \bw}) -- ({3 * (\bs + \bw)}, {\hn * (\lw + \bw) + \bw}) -- cycle ;
{\foreach \bn in {4, 5}
\draw  ({\bn * (\bs + \bw)}, {\hn * (\lw + \bw)}) -- ({\bn * (\bs + \bw) + \bw}, {\hn * (\lw + \bw)}) -- ({\bn * (\bs + \bw) + \bw}, {\hn * (\lw + \bw) + \bw}) -- ({\bn * (\bs + \bw)}, {\hn * (\lw + \bw) + \bw}) -- cycle ;
}
}

\foreach \bn in {4, 5}{
\draw[densely dashed]  ({\bn * (\bs + \bw)}, {0 * (\lw + \bw)}) -- ({\bn * (\bs + \bw) + \bw}, {0 * (\lw + \bw)}) -- ({\bn * (\bs + \bw) + \bw}, {0 * (\lw + \bw) + \bw}) -- ({\bn * (\bs + \bw)}, {0 * (\lw + \bw) + \bw}) -- cycle ;
\draw[densely dashed]  ({\bn * (\bs + \bw) - 0.1 * \bw}, {0 * (\lw + \bw) - 0.1 * \bw}) -- ({\bn * (\bs + \bw) + 1.1 * \bw}, {0 * (\lw + \bw) - 0.1 * \bw}) -- ({\bn * (\bs + \bw) + 1.1 * \bw}, {0 * (\lw + \bw) + 1.1 * \bw}) -- ({\bn * (\bs + \bw) - 0.1 * \bw}, {0 * (\lw + \bw) + 1.1 * \bw}) -- cycle ;
\draw    ({\bn * (\bs + \bw) + 0.5 * \bw},{\bw + 0.1 * \bw}) -- ({\bn * (\bs + \bw) + 0.5 * \bw},{1 * (\lw + \bw)}) ;
}

\end{tikzpicture}
\hspace{-1.9cm}
\caption{Initialization of deep MLP with $d_{in} = 3$ and $d_{out} = 2$. 
Intermediate layers have width $d_{in}+1+d_{out}$.
Line styles indicate types of weight initializations (solid:1, double:$C_L$, dash:Gaussian, none:0). Box styles indicate types of bias initializations (solid:0, dash:Gaussian, double:$C_L$, double-dash:$-C_L$).}
    \label{fig:mlp_arch}
\end{figure}
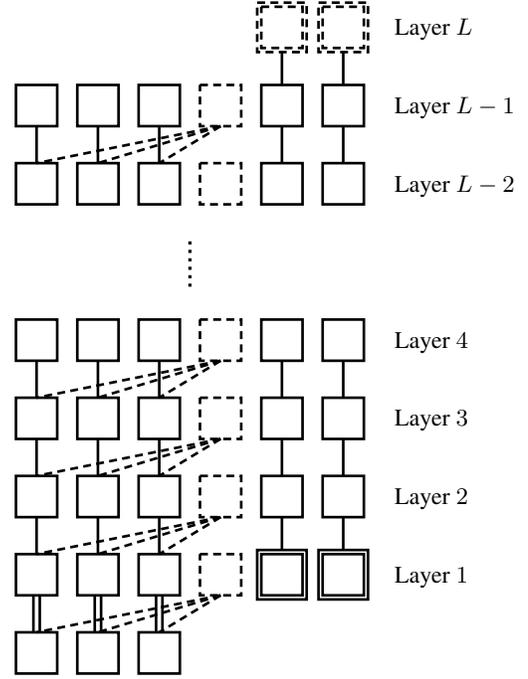

Motivated by \citet{KidgerLyons2020_universal}, initialize the weights of our $L$-layer MLP $f^L_\theta$ as follows:
\begin{align*}
W^1 &=\begin{bmatrix} C_L I_{d_{\mathrm{in}}}\\ u^1\\0_{d_{\mathrm{out}}\times d_{\mathrm{in}}} \end{bmatrix} \\
W^l &=\begin{bmatrix} I_{d_{\mathrm{in}}} & 0_{d_{\mathrm{in}}\times 1} &0_{d_{\mathrm{in}}\times d_{\mathrm{out}}} \\ u^l & 0_{1\times 1} & 0_{1\times d_{\mathrm{out}}} \\ 0_{d_{\mathrm{out}}\times d_{\mathrm{in}}} & 0_{d_{\mathrm{out}}\times 1} & I_{d_{\mathrm{out}}} \end{bmatrix} \\
W^L &=\begin{bmatrix} 0_{d_{\mathrm{out}}\times d_{\mathrm{in}}} & 0_{d_{\mathrm{out}}\times 1} & I_{d_{\mathrm{out}}} \end{bmatrix}
\end{align*}
for $2\le l\le L-1$,
where $u^l \in \fR^{1 \times d_{\mathrm{in}}}$ is randomly sampled $u^l_i \stackrel{iid}{\sim}\mathcal{N}(0, \frac{1}{d_{\mathrm{in}}}\rho^2)$ for $1 \le l \le L-1$.
Here, $I_d$ is the $d \times d$ identity matrix and $0_{m\times n} $ is the $m$ by $n$ matrix with all zero entries, $\CL>0$ is a scalar growing as a function of $L$ at a rate satisfying $L^2/\CL\rightarrow 0$, and $\rho>0$ is a fixed variance parameter.
Initializes the biases as follows: 
\begin{align*}
b^1 &=\begin{bmatrix} 0_{d_{\mathrm{in}}\times 1}\\ v^1\\ \CL \mathbbm{1}_{d_{\mathrm{out}}} \end{bmatrix} \\
b^l &=\begin{bmatrix} 0_{d_{\mathrm{in}}\times 1} \\ v^l\\ 0_{d_{\mathrm{out}}\times 1} \end{bmatrix} \\
b^L &=\begin{bmatrix} -\CL\mathbbm{1}_{d_{\mathrm{out}}} \end{bmatrix}
\end{align*}
for $2\le l\le L-1$, where $v^l \in \mathbb{R}$ is randomly sampled $v^l \stackrel{iid}{\sim}\mathcal{N}(0, \CL^2 \beta^2)$ for $1 \le l \le L-1$.
Here, $\beta>0$ is a fixed variance parameter.
Note that $\mathbbm{1}_k\in \fR^{{ k\times 1}}$ is the vector whose entries are all $1$. Figure~\ref{fig:mlp_arch} illustrates this initialization scheme.

We clarify that while we use many specific non-random initializations, those parameters are not fixed throughout training. In other words, all parameters are trainable, just as one would expect from a standard MLP.

\citet{KidgerLyons2020_universal} used a similar construction to establish a universal approximation result for deep MLPs by showing that their deep MLP mimics a 2-layer wide MLP.
However, their main concern is the existence of a weight configuration that approximates a given function, which does not guarantee that such a configuration can be found through training. On the other hand, we propose an explicit initialization and establish a trainability guarantee;
our network outputs 0 at initialization and converges to the desired configuration through training.

% However, they did not show whether such configurations can be found through training.
% , and our primary concern is obtaining trainability guarantees. 
% Deep neural networks and 2-layer wide neural networks have different training dynamics even if they are equivalent at initialization.

\subsection{Gradient flow and neural tangent kernel}
\label{ss:gradient_flow}
We are now ready to describe the training of our neural network $f^L_\theta$ via gradient flow, a continuous-time model of gradient descent.
Since our initialization scales the input by $\CL$ at the first layer, we scale the learning rate accordingly, both in the continuous-time analysis of Section~\ref{s:cnn} and in the discrete-time experiments of Section~\ref{s:experiments}, so that we get meaningful limits as $L\rightarrow\infty$.

Train $f^L_\theta$ with
\begin{align*}
    \partial_t \theta(t) &
    \stackrel{\text{(a)}}{=}
    -\frac{1}{L\CL^2}\nabla_\theta \text{Loss}(\theta)\Big|_{\theta=\theta(t)}\\
    &\stackrel{\text{(b)}}{=} -\frac{1}{L\CL^2}\left(\partial_{\theta} \loss( f^{L}_\theta)\right)^\intercal\Big|_{\theta=\theta(t)}\\
    &\stackrel{\text{(c)}}{=} -\frac{1}{L\CL^2}\Epin{(\partial_{\theta} f_\theta^{L}(x))^\intercal \delta\vert_{f_{\theta}^L}(x)}\Big|_{\theta=\theta(t)},
\end{align*}
where (a) defines the $\theta$-update to be gradient flow with learning rate $1/(LC_L^2)$, (b) plugs in our notation, and (c) follows from the chain rule.

This gradient flow defines $\theta(t)$ to be a function of time, but we will often write $\theta$ rather than $\theta(t)$ for notational conciseness.
% where $C$ is cost function, $\theta$ is vector of trainable parameters, and $f^L$ is $L$-layer neural networks.
This gradient flow and the chain rule induces the functional dynamics of the neural network:\vspace{-0.08in}
\begin{align*}
 \partial_t f_\theta^{L}(x) &=\partial_{\theta} f_\theta^{L}(x)\partial_t \theta   \\  
&
\!\!\!\!\!
=-\frac{1}{L\CL^2}\epin{ \partial_{\theta}f_\theta^{L}(x) \para{\partial_{\theta} f_\theta^{L}(x')}^{\intercal} \delta\vert_{f_\theta^L}(x')}.\vspace{-0.1in}
\end{align*}
Since we use a scaling factor in our gradient flow, we define the \emph{scaled NTK} at time $t$ as\vspace{-0.05in}
\begin{align*}
\tilde{\Theta}^L_t(x, x') = \frac{1}{L\CL^2}\partial_{\theta} f_\theta^{L}(x)\para{\partial_{\theta}f_\theta^{L}(x')}^{\intercal}.\\[-0.35in]
\end{align*}
Then, \vspace{-0.1in}
\begin{align*}
 \partial_t f^{L}_\theta(x)
&=-\epin{\tilde{\Theta}^L_{t}(x, x') \delta\vert_{f_\theta^L}(x')}\\
&= -\nabla_{\tilde{\Theta}^L_{t}}\loss\vert_{f_\theta^L}(x).
\end{align*}

%% 3.2 main results
\subsection{Convergence in infinite-depth limit}
We now analyze the convergence of the infinitely deep MLP.

Theorem~\ref{thm:1} establishes that the scaled NTK at initialization (before training) of the randomly initialized MLP converges to a deterministic limit with a closed-form expression as the depth $L$ becomes infinite.

% Our first main result characterizes the deterministic kernel which $\tilde{\Theta}^L_0$ converges in infinite-depth limit.
\begin{theorem}[Scaled NTK at initialization] \label{thm:1}
Suppose $f^{L}_\theta$ is initialized as in Section \ref{sec:mlp_initialization}.
For any $x, x'\in\fR_+^{d_{in}}$, 
\[
\tilde{\Theta}^L_0(x,x')
\convinp
\tilde{\Theta}^\infty(x, x')
\]
as $L \rightarrow \infty$, where 
\begin{align*}
\tilde{\Theta}^\infty(x, x')=
\para{x^{\intercal} x'+1+ \mathbb{E}_{ g }[\sigma(g(x))\sigma(g(x'))]}I_{d_{\mathrm{out}}},
\end{align*}
and $g\sim \mathcal{GP} ( 0 ,\frac{\rho^2}{d_{\mathrm{in}}} x^{\intercal}x'+\beta^2)$.
\end{theorem}
% Obviously, $\Theta(x,x')$ is positive constant, positive semi-definite.  
% \begin{remark}
% Our NTK is not infinite width and depth limit. Our NTK assumed bounded width. 
% \end{remark}

When $L<\infty$, the scaled NTK $\tilde{\Theta}^L_t(x,x')$ depends on time through its dependence on $\theta(t)$.
Theorem~\ref{thm:2} establishes that $\tilde{\Theta}^L_t(x,x')$ becomes independent of $t$ as $L\rightarrow\infty$.
% For the sake of simplicity, we let $\delta_t^L = \partial_f C = \delta\vert_{f_\theta^L}$.
\begin{theorem}[Invariance of scaled NTK] 
\label{thm:2}
Let $T>0$.
Suppose $\int_0^T\EP{\delta\vert_{f_\theta^L}}\!\!\!\!dt$ is stochastically bounded as $L\rightarrow\infty$.
Then, for any $x, x'\in\fR_+^{d_{in}}$, 
\[\tilde{\Theta}^L_t(x,x')
\convinp
\tilde{\Theta}^{\infty}(x,x')
\]
uniformly for $t\in [0,T]$ as $L \rightarrow \infty$.
\end{theorem}

Let $\loss(f) = \frac{1}{2}\EP{f-f^\star}^2$ be the quadratic loss. In this case, the stochastic boundedness assumption of Theorem~\ref{thm:2} holds, as we show in Lemma~\ref{lemm:stochastically bound} of the appendix, and we characterize the training dynamics explicitly. For the sake of notational simplicity, assume the MLP's prediction is a scalar, i.e., assume $d_{out}=1$. The generalization to multi-dimensional outputs is straightforward, following the arguments of \citep[Section~5]{JacotGabrielHongler2018_neural}.

Theorem~\ref{thm:3} concludes the analysis by characterizing the trained MLP as $L\rightarrow\infty$.
Define the kernel regression predictor as   
\begin{align*}
    f_{\mathrm{ntk}}(x)=\para{\tilde{\Theta}^{\infty}(x,x_1), \dots, \tilde{\Theta}^{\infty}(x,x_N)}K^{-1}f^\star(X),
\end{align*}
where $K_{i,j}=\tilde{\Theta}^{\infty}(x_i,x_j)$ and $[f^\star(X)]_i=f^\star(x_i)$ for $i,j\in\{1,\dots,N\}$.
Let $f_t$ be trained with the limiting kernel gradient flow of $f_{\theta(t)}^L$ as $L\rightarrow\infty$, i.e., $f_0=0$ and $f_t$ follows
\[
\partial_t f_t= -\nabla_{\tilde{\Theta}^\infty}\loss\vert_{f_t}(x).
\]

Then the infinitely deep training dynamics converge to $f_{\mathrm{ntk}}$ in the following sense.
\begin{theorem}[Equivalence between deep MLP and kernel regression]
\label{thm:3}
Let $\loss(f) = \frac{1}{2}\EP{f-f^\star}^2$. Let $\tilde{\Theta}^{\infty}$ be positive definite.
If $f_t$ follows the kernel gradient flow with respect to $\tilde{\Theta}^{\infty}$, then for any $x \in \fR^{d_{in}}_+$,
\[
\lim_{t \rightarrow \infty}f_t (x) = f_{\mathrm{ntk}}(x).
\]
\end{theorem}

% \begin{figure}[h]
% \vspace{-0.25in}
%     \centering
%     \includegraphics{figures/cnn_arch.pdf}
% \vspace{-0.25in}
% \caption{Initialization of deep CNN. The grids represent channels, and the box at the top represents the scalar output of the final average pool. Line styles indicate types of weight initializations (solid: 1, double: $C_L$, dash: Gaussian, no line: 0). Box styles indicate types of bias initializations (solid box: 0, dash: Gaussian, double: $C_L$, double dash: $-C_L$).}
%     \label{fig:cnn_arch}
% \end{figure}

%% 3.3 outline of proofs
\subsection{Proof outline}
At a high level, our analysis follows the same line of argument as that of the original NTK paper \cite{JacotGabrielHongler2018_neural}:
Theorem~\ref{thm:1} characterizes the limiting NTK at initialization, 
Theorem~\ref{thm:2} establishes that the NTK remains invariant throughout training, and
Theorem~\ref{thm:3} establishes convergence in the case of quadratic loss functions.
The proof of Theorem~\ref{thm:3} follows from arguments similar to those of \citep[Theorem~3]{JacotGabrielHongler2018_neural}.
The proof of Theorem~\ref{thm:1} follows from identifying the recursive structure and noticing that the initialization is designed to simplify this recursion.

The key technical challenge of this work is in Theorem~\ref{thm:2}.
The analysis is based on defining the Lyapunov function 
\begin{align*}
     \Gamma_L(t)&= \Psi_{L,2}(t)+ \Psi_{L,8}(t)+\sum_{j=1}^8\Phi_{L,j}(t)
\end{align*}
(the individual terms will be defined soon),
establishing
\[
     \Gamma_L(t) \le \Gamma_L(0)+ \int_0^{t} \mathcal{O}(L/\CL)\Gamma_L(s)^4\;ds,
\]
and appealing to Gr\"onwall's lemma to show that $\Gamma_L(t)$ is invariant, i.e., $\Gamma_L(t)\rightarrow\Gamma_{\infty}(0)$ as $L\rightarrow\infty$ uniformly in $t\in[0,T]$.
For $j=1,\dots,8$, the first term is defined as
\[
\!\!
\Phi_{L, j}\!
=
\!
\frac{1}{(L-1) \CL^j}\!\sum^{L-1}_{l=1} \!\!\left(\EP{\fulz}\!\!\!\!+\EP{\fult-\fulz}\!\!\right)^j
\]
and its invariance implies that the average variation of the layers vanish for inputs $x\in\{x_1,\dots,x_N\}$ to the MLP.

Because we study the infinite-depth regime, our Lyapunov analysis is significantly more technical compared to the prior NTK analyses studying the infinite-width regime. Prior work dealt with sums of infinitely many terms, which were analyzed with the central limit theorem and law of large numbers. In contrast, the infinite depth of our setup leads to an infinite composition of layers and an infinite product of matrices, which must be controlled through a more delicate Lyapunov analysis.

For $L \ge j\ge k\ge 2$, define
\[
\!\!
\W^k_j(x,t)
=
W^j \diag(\dot{\sigma}(f^{j-1}_{\theta^{(j - 1)}}))\cdots W^k \diag(\dot{\sigma}(f^{k-1}_{\theta^{(k-1)}}))\]
(where $W^j$ and $\theta^{(j)}$ depend on $t$ and $f^{l}_{\theta^{(l)}}$ depends on $x$ and $t$)
and we bound its change by incorporating the following terms into the Lyapunov function: 
\begin{align*}
\scriptstyle
\!\!\!
    \Psi_{L, 2}& \scriptstyle
= \frac{1}{(L-1)^2}\sum^{ L}_{l=2}\!\! \frac{L-1}{l-1} \sum_{ i=2}^{l} \left(\EP{\W_l^i(\cdot, 0)}\!+\EP{\W_l^i(\cdot, t)-\W_l^i(\cdot, 0)}\!\right)^2\\
\scriptstyle
\!\!\!
\Psi_{L,8}& \scriptstyle
=\frac{1}{L-1} \sum_{l= 2}^L \left(\EP{\W_L^l(\cdot, 0)}+\EP{\W_L^l(\cdot, t)-\W_L^l(\cdot, 0)}\right)^8.
\end{align*}

Establishing the invariance of $\Gamma_L(t)$ as $L\rightarrow\infty$ is the first step of the analysis, but, by itself, it does not characterize the limiting MLP for inputs $x\notin \{x_1,\dots,x_N\}$ and it only bounds the average variation of the layers, rather than the layer-wise variation. Hence, we generalize the invariance results with two additional Lyapunov analyses and combine these results to establish the scaled NTK's invariance.

% In our analysis, we encounter $\ddot{\sigma}(s)$, where $\sigma$ is ReLU. 
Our Lyapunov analysis significantly is simplified by setting $\ddot{\sigma}(s)=0$ and thereby removing terms involving $\ddot{\sigma}$.
However, while $\ddot{\sigma}(s)=0$ for $s\ne 0$, we cannot ignore the fact that $\ddot{\sigma}(0)\ne 0$ (in fact undefined).
We resolve this issue by showing that all instances of $\sigma(s)$ never encounter the input $s=0$ throughout training with probability approaching $1$ as $L\rightarrow\infty$.
We outline this argument below.
% Since $\sigma$ is an absolutely continuous function, we can simply define $\dot{\sigma}(0)=0$.

Due to the randomness of the initialization, the pre-activation values $\fulz(x)$ are element-wise nonzero for all $x\in \{x_1,\dots,x_N\}$ with probability $1$. If $(\fult(x))_r= 0$ for some $l$, $r$, and $t$, i.e., if a \emph{zero-crossing} happens for a neuron at time $t$, then the analysis must somehow deal with the behavior of $\sigma(s)$ at $s=0$. However, if zero-crossing happens for no neurons for all $t\in[0,T]$, then we can safely set $\ddot{\sigma}=0$ in our analysis. We prove that the probability of a zero crossing (over all neurons of all layers and all $t\in [0,T]$) vanishes as $L\rightarrow\infty$. We specifically establish this claim by showing that as $L\rightarrow\infty$,
\begin{align*}
    \sup_{1\leq l\leq L-1, t\in[0, T]} \ab{\fult(x)-\fulz(x)} \le
    KL
\end{align*}
with high probability and
\begin{align*}
    \Pr{\displaystyle \inf_{l,r}\left|\para{\fulz(x)}_r\right| > (K+1)L}
    \rightarrow 1.
\end{align*}
for some constant $K>0$. These two results establish that the zero-crossing probability vanishes as $L\rightarrow\infty$.
We provide the details in Section~\ref{app:not crossing}.
% , where $\min_{l,r}$ is taken over $1\le l\le L$ and $r=1,\dots,n_l$.

% First, we define similar Lyapunov functions and again appeal to Gr\"onwall's lemma. 
% Second, we define, for any given $x$,
% \begin{align*}
%     \tilde{\Psi}_L^{k}& = \ab{\W_L^k(x,0)}+\ab{\W_L^k(x,t)-\W_L^k(x,0)}\\
%     \tilde{\Phi}_L^{l} & = \frac{1}{ \CL}\para{\ab{\fulz(x)}+\ab{\fult(x)-\fulz(x)}}
% \end{align*}
% satisfies
% $\tilde{\Psi}^{k}_L(x,t) \rightarrow \tilde{\Psi}^{k}_{\infty}(x,0)$ and $\tilde{\Phi}^{l}_{L}(x,t) \rightarrow \tilde{\Phi}^{l}_{\infty}(x,0)$.

% \begin{align*}
%     \tilde{\Phi}_L^{l}(x,t) & = \frac{1}{ \CL}\para{\ab{\fulz(x)}+\ab{\fult(x)-\fulz(x)}}
% \end{align*}
% that 
% $ \tilde{\Phi}^{l}_{L}(x,t) \rightarrow \tilde{\Phi}^{l}_{\infty}(x,0)$.

\begin{figure}
\centering
\tikzset{every picture/.style={line width=1pt}}
\def\bw{20} % box width / height
\def\bs{20} % space between boxes
\def\lw{15} % weight width
\begin{tikzpicture}[x=0.8pt,y=0.8pt,yscale=-1,xscale=1]
% first layer
\foreach \bn in {0}{
{\draw  ({\bn * (\bs + \bw)}, {7 * (\lw + \bw) + \bw}) -- ({\bn * (\bs + \bw) + \bw}, {7 * (\lw + \bw) + \bw}) -- ({\bn * (\bs + \bw) + \bw}, {7 * (\lw + \bw) + 2 * \bw}) -- ({\bn * (\bs + \bw)}, {7 * (\lw + \bw) + 2 * \bw}) -- cycle ;
}
\draw    ({\bn * (\bs + \bw) + 0.5 * \bw - 1.5},{7 * (\lw + \bw) + \bw}) -- ({\bn * (\bs + \bw) + 0.5 * \bw - 1.5}, {6 * (\lw + \bw) + 2 * \bw})({\bn * (\bs + \bw) + 0.5 * \bw + 1.5},{6 * (\lw + \bw) + 2 * \bw}) -- ({\bn * (\bs + \bw) + 0.5 * \bw + 1.5},{7 * (\lw + \bw) + \bw}) ;
\draw[densely dashed]  ({1 * (\bs + \bw) + 0.5 * \bw},{6 * (\lw + \bw) + 2 * \bw}) -- ({\bn * (\bs + \bw) + 0.5 * \bw}, {7 * (\lw + \bw) + \bw}) ;
{\foreach \g in {1, 2, 3}
{
\draw    ({\bn * (\bs + \bw) + 0.25 * \g * \bw}, {7 * (\lw + \bw) + \bw}) -- ({\bn * (\bs + \bw) + 0.25 * \g * \bw},{7 * (\lw + \bw) + 2 * \bw}) ;
\draw    ({\bn * (\bs + \bw)}, {7 * (\lw + \bw) + \bw + 0.25 * \g * \bw}) -- ({\bn * (\bs + \bw) + \bw},{7 * (\lw + \bw) + \bw + 0.25 * \g * \bw}) ;
}
}
}

\foreach \bn in {2}{
{\draw  ({\bn * (\bs + \bw) - 0.1 * \bw}, {6 * (\lw + \bw) + 0.9 * \bw}) -- ({\bn * (\bs + \bw) + 1.1 * \bw}, {6 * (\lw + \bw) + 0.9 * \bw}) -- ({\bn * (\bs + \bw) + 1.1 * \bw}, {6 * (\lw + \bw) + 2.1 * \bw}) -- ({\bn * (\bs + \bw) - 0.1 * \bw}, {6 * (\lw + \bw) + 2.1 * \bw}) -- cycle ;
}
\draw    ({\bn * (\bs + \bw) + 0.5 * \bw},{6 * (\lw + \bw) + 0.9 * \bw}) -- ({\bn * (\bs + \bw) + 0.5 * \bw},{5 * (\lw + \bw) + 2 * \bw}) ;
{\foreach \g in {1, 2, 3}
{
\draw    ({\bn * (\bs + \bw) + 0.25 * \g * \bw}, {6 * (\lw + \bw) + \bw}) -- ({\bn * (\bs + \bw) + 0.25 * \g * \bw},{6 * (\lw + \bw) + 2 * \bw}) ;
\draw    ({\bn * (\bs + \bw)}, {6 * (\lw + \bw) + \bw + 0.25 * \g * \bw}) -- ({\bn * (\bs + \bw) + \bw},{6 * (\lw + \bw) + \bw + 0.25 * \g * \bw}) ;
}
}
}

\foreach \hn in {6, 5, 2}{
{\foreach \bn in {0}{
\draw  ({\bn * (\bs + \bw)}, {\hn * (\lw + \bw) + \bw}) -- ({\bn * (\bs + \bw) + \bw}, {\hn * (\lw + \bw) + \bw}) -- ({\bn * (\bs + \bw) + \bw}, {\hn * (\lw + \bw) + 2 * \bw}) -- ({\bn * (\bs + \bw)}, {\hn * (\lw + \bw) + 2 * \bw}) -- cycle ;
\draw    ({\bn * (\bs + \bw) + 0.5 * \bw},{\hn * (\lw + \bw) + \bw}) -- ({\bn * (\bs + \bw) + 0.5 * \bw}, {(\hn - 1) * (\lw + \bw) + 2 * \bw});
\draw[densely dashed]  ({1 * (\bs + \bw) + 0.5 * \bw},{(\hn - 1) * (\lw + \bw) + 2 * \bw}) -- ({\bn * (\bs + \bw) + 0.5 * \bw}, {\hn * (\lw + \bw) + \bw});
{\foreach \g in {1, 2, 3}
{
\draw    ({\bn * (\bs + \bw) + 0.25 * \g * \bw}, {\hn* (\lw + \bw) + \bw}) -- ({\bn * (\bs + \bw) + 0.25 * \g * \bw},{\hn * (\lw + \bw) + 2 * \bw}) ;
\draw    ({\bn * (\bs + \bw)}, {\hn * (\lw + \bw) + \bw + 0.25 * \g * \bw}) -- ({\bn * (\bs + \bw) + \bw},{\hn * (\lw + \bw) + \bw + 0.25 * \g * \bw}) ;
}
}
}
}

\draw[densely dashed]  ({1 * (\bs + \bw)}, {\hn * (\lw + \bw) + \bw}) -- ({1 * (\bs + \bw) + \bw}, {\hn * (\lw + \bw) + \bw}) -- ({1 * (\bs + \bw) + \bw}, {\hn * (\lw + \bw) + 2 * \bw}) -- ({1 * (\bs + \bw)}, {\hn * (\lw + \bw) + 2 * \bw}) -- cycle ;
{\foreach \g in {1, 2, 3}
{
\draw    ({1 * (\bs + \bw) + 0.25 * \g * \bw}, {\hn* (\lw + \bw) + \bw}) -- ({1 * (\bs + \bw) + 0.25 * \g * \bw},{\hn * (\lw + \bw) + 2 * \bw}) ;
\draw    ({1 * (\bs + \bw)}, {\hn * (\lw + \bw) + \bw + 0.25 * \g * \bw}) -- ({1 * (\bs + \bw) + \bw},{\hn * (\lw + \bw) + \bw + 0.25 * \g * \bw}) ;
}
}

{\foreach \bn in {2}
{\draw  ({\bn * (\bs + \bw)}, {\hn * (\lw + \bw) + \bw}) -- ({\bn * (\bs + \bw) + \bw}, {\hn * (\lw + \bw) + \bw}) -- ({\bn * (\bs + \bw) + \bw}, {\hn * (\lw + \bw) + 2 * \bw}) -- ({\bn * (\bs + \bw)}, {\hn * (\lw + \bw) + 2 * \bw}) -- cycle ;

{\foreach \g in {1, 2, 3}
{
\draw    ({\bn * (\bs + \bw) + 0.25 * \g * \bw}, {\hn* (\lw + \bw) + \bw}) -- ({\bn * (\bs + \bw) + 0.25 * \g * \bw}, {\hn * (\lw + \bw) + 2 * \bw}) ;
\draw    ({\bn * (\bs + \bw)}, {\hn * (\lw + \bw) + \bw + 0.25 * \g * \bw}) -- ({\bn * (\bs + \bw) + \bw}, {\hn * (\lw + \bw) + \bw + 0.25 * \g * \bw}) ;
}
}
}
}
}

%vertical dots
\draw[very thick, dotted]   ({0.5 * (2 * (\bw + \bs) + \bw)},{3 * (\bw + \lw) + 1.1 * \bw}) -- ({0.5 * (2 * (\bw + \bs) + \bw)},{3 * (\bw + \lw) + 2 * \bw}) ;

\foreach \hn in {5, 2}
{\foreach \bn in {2}{
\draw    ({\bn * (\bs + \bw) + 0.5 * \bw},{\hn * (\lw + \bw) + \bw}) -- ({\bn * (\bs + \bw) + 0.5 * \bw},{(\hn - 1) * (\lw + \bw) + 2 * \bw}) ;
{\foreach \g in {1, 2, 3}
{
\draw    ({\bn * (\bs + \bw) + 0.25 * \g * \bw}, {\hn* (\lw + \bw) + \bw}) -- ({\bn * (\bs + \bw) + 0.25 * \g * \bw}, {\hn * (\lw + \bw) + 2 * \bw}) ;
\draw    ({\bn * (\bs + \bw)}, {\hn * (\lw + \bw) + \bw + 0.25 * \g * \bw}) -- ({\bn * (\bs + \bw) + \bw}, {\hn * (\lw + \bw) + \bw + 0.25 * \g * \bw}) ;
}
}
}
}

\foreach \hn in {4, 1}{
\foreach \bn in {0}{
\draw  ({\bn * (\bs + \bw)}, {\hn * (\lw + \bw) + \bw}) -- ({\bn * (\bs + \bw) + \bw}, {\hn * (\lw + \bw) + \bw}) -- ({\bn * (\bs + \bw) + \bw}, {\hn * (\lw + \bw) + 2 * \bw}) -- ({\bn * (\bs + \bw)}, {\hn * (\lw + \bw) + 2 * \bw}) -- cycle ;
\draw[densely dashed]  ({1 * (\bs + \bw)}, {\hn * (\lw + \bw) + \bw}) -- ({1 * (\bs + \bw) + \bw}, {\hn * (\lw + \bw) + \bw}) -- ({1 * (\bs + \bw) + \bw}, {\hn * (\lw + \bw) + 2 * \bw}) -- ({1 * (\bs + \bw)}, {\hn * (\lw + \bw) + 2 * \bw}) -- cycle ;
{\foreach \g in {1, 2, 3}
{
\draw    ({1 * (\bs + \bw) + 0.25 * \g * \bw}, {\hn* (\lw + \bw) + \bw}) -- ({1 * (\bs + \bw) + 0.25 * \g * \bw}, {\hn * (\lw + \bw) + 2 * \bw}) ;
\draw    ({1 * (\bs + \bw)}, {\hn * (\lw + \bw) + \bw + 0.25 * \g * \bw}) -- ({1 * (\bs + \bw) + \bw}, {\hn * (\lw + \bw) + \bw + 0.25 * \g * \bw}) ;
}
}
{\foreach \g in {1, 2, 3}
{
\draw    ({\bn * (\bs + \bw) + 0.25 * \g * \bw}, {\hn* (\lw + \bw) + \bw}) -- ({\bn * (\bs + \bw) + 0.25 * \g * \bw}, {\hn * (\lw + \bw) + 2 * \bw}) ;
\draw    ({\bn * (\bs + \bw)}, {\hn * (\lw + \bw) + \bw + 0.25 * \g * \bw}) -- ({\bn * (\bs + \bw) + \bw}, {\hn * (\lw + \bw) + \bw + 0.25 * \g * \bw}) ;
}
}
}
{\foreach \bn in {2}{
\draw  ({\bn * (\bs + \bw)}, {\hn * (\lw + \bw) + \bw}) -- ({\bn * (\bs + \bw) + \bw}, {\hn * (\lw + \bw) + \bw}) -- ({\bn * (\bs + \bw) + \bw}, {\hn * (\lw + \bw) + 2 * \bw}) -- ({\bn * (\bs + \bw)}, {\hn * (\lw + \bw) + 2 * \bw}) -- cycle ;
{\foreach \g in {1, 2, 3}
{
\draw    ({\bn * (\bs + \bw) + 0.25 * \g * \bw}, {\hn* (\lw + \bw) + \bw}) -- ({\bn * (\bs + \bw) + 0.25 * \g * \bw}, {\hn * (\lw + \bw) + 2 * \bw}) ;
\draw    ({\bn * (\bs + \bw)}, {\hn * (\lw + \bw) + \bw + 0.25 * \g * \bw}) -- ({\bn * (\bs + \bw) + \bw}, {\hn * (\lw + \bw) + \bw + 0.25 * \g * \bw}) ;
}
}
}
}
}

\foreach \bn in {2}{
{\draw[densely dashed]  ({\bn * (\bs + \bw) - 0.1 * \bw}, {0.9 * \bw}) -- ({\bn * (\bs + \bw) + 1.1 * \bw}, {0.9 * \bw}) -- ({\bn * (\bs + \bw) + 1.1 * \bw}, {2.1 * \bw}) -- ({\bn * (\bs + \bw) - 0.1 * \bw}, {2.1 * \bw}) -- cycle ;
}
\draw[densely dashed]  ({\bn * (\bs + \bw)}, {\bw}) -- ({\bn * (\bs + \bw) + \bw}, {\bw}) -- ({\bn * (\bs + \bw) + \bw}, {2 * \bw}) -- ({\bn * (\bs + \bw)}, {2 * \bw}) -- cycle ;
\draw  ({\bn * (\bs + \bw) + 0.5 * \bw}, {2.1 * \bw}) -- ({\bn * (\bs + \bw) + 0.5 * \bw}, {2 * \bw + \lw});
{\foreach \g in {1, 2, 3}
{
\draw    ({\bn * (\bs + \bw) + 0.25 * \g * \bw}, {\bw}) -- ({\bn * (\bs + \bw) + 0.25 * \g * \bw}, {2 * \bw}) ;
\draw    ({\bn * (\bs + \bw)}, {\bw + 0.25 * \g * \bw}) -- ({\bn * (\bs + \bw) + \bw}, {\bw + 0.25 * \g * \bw}) ;
}
}
}

\draw ({2 * (\bs + \bw) + 0.375 * \bw}, 0) -- ({2 * (\bs + \bw) + 0.375 * \bw}, {0.25 * \bw}) -- ({2 * (\bs + \bw) + 0.625 * \bw}, {0.25 * \bw}) -- ({2 * (\bs + \bw) + 0.625 * \bw}, 0) -- cycle ;

\draw ({2 * (\bs + \bw) + 0.375 * \bw}, {0.25 * \bw}) -- ({2 * (\bs + \bw) - 0.1 * \bw}, {0.9 * \bw}) ;
\draw ({2 * (\bs + \bw) + 0.625 * \bw}, {0.25 * \bw}) -- ({2 * (\bs + \bw) + 1.1 * \bw}, {0.9 * \bw}) ;

\foreach \hn in {1, 2, 3}{
\node[right, font=\small] at ({2 * (\bs + \bw) + \bw + 0.5 * \bw}, {(7 - \hn) * (\lw + \bw) + 2 * \bw - 0.45 * \bw}) {Layer ${\hn}$} ;
}
\node[right, font=\small] at ({2 * (\bs + \bw) + \bw + 0.5 * \bw}, {2 * (\lw + \bw) + 2 * \bw - 0.45 * \bw}) {Layer $L - 2$} ;

\node[right, font=\small] at ({2 * (\bs + \bw) + \bw + 0.5 * \bw}, {1 * (\lw + \bw) + 2 * \bw - 0.45 * \bw}) {Layer $L - 1$} ;

\node[right, font=\small] at ({2 * (\bs + \bw) + \bw + 0.5 * \bw}, {0 * (\lw + \bw) + 2 * \bw - 0.45 * \bw}) {Layer $L $} ;

\node[right, font=\small] at ({2 * (\bs + \bw) + \bw + 0.5 * \bw}, {0.2 * \bw}) {Average Pool} ;

\end{tikzpicture}
\hspace{-2cm}
\caption{Initialization of deep CNN with $4\times 4$ input. The 3 grids per row represent the 3 channels per layer, and the box at the top represents the scalar output of the final average pool. Line styles indicate types of weight initializations (solid:$\diag(0, 1, 0)$, double:$\diag(0, C_L, 0)$, dash:Gaussian, none:$0_{3\times 3}$). Box styles indicate types of bias initializations (solid:0, dash:Gaussian, double:$C_L$, double-dash:$-C_L$).}
    \label{fig:cnn_arch}

\end{figure}
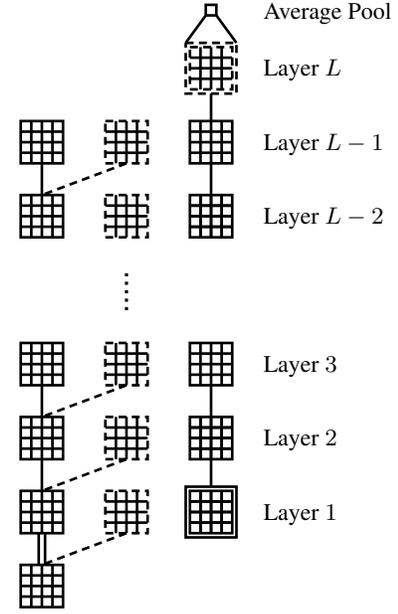

\section{NTK analysis of infinitely deep CNN}
\label{s:cnn}
Consider an $L$-layer convolutional neural network (CNN) $f^{L+1}_{\theta}\colon \fR^{d\times d}_+ \rightarrow \fR$ parameterized by $\theta$,
where the input $x\in \fR^{{d\times d}}_+$ is a $d\times d$ image with positive entries and the output is a scalar.
The network consists of $L-1$ convolutional layers using $3\times 3$ filters and zero-padding of $1$ with 3 output channels, each followed by the ReLU activation function.
The $L$-th convolutional layer uses a $3\times 3$ filter with zero-padding of $1$ and has a single output channel.
This is followed by a global average pool with no activation function applied before or after the average pool.
All convolutional layers use stride of $1$.

Let us set up specific notation.
For a convolutional filter $w \in \mathbb{R}^{3 \times 3}$ and a (single-channel) image $x \in \mathbb{R}^{d \times d}_+$, denote the convolution operation with zero padding as 
\vspace{-0.01in}
\[
[w*x]_{i,j} = \left \langle w, \phi_{i,j}(x) \right \rangle\vspace{-0.01in}
\]
% \[ [w*x]_{i,j} = \sum_{q=-1}^1 \sum_{p=-1}^1 w_{p+2, q+2}x_{p+i, q+j}.\]
for $1 \le i, j \le d$, where $\phi_{i,j}(x) = [x]_{i-1:i+1, j-1:j+ 1}$ and $x_{pq}=0$ if $p$ or $q$ is less than 1 or greater than $d$, i.e., $x_{pq}=0$ if the index is out of bounds.
Define $\idconv_{3\times 3}=\diag(0, 1, 0)$ to be the $3\times3$ filter serving as the identity map. So
$\idconv_{3\times 3} * x=x$.
Define the pre-activation values as 
\vspace{-0.02in}
\begin{align*}
    \!\!\!\!\para{f^1_{\theta^{(1)}}}_{r,:,:} &=  w^1 _{r,1,:,:}*x+\mathbbm{1}_{d \times d}b^1_r,\\ 
    \!\!\!\!\para{f^l_{\theta^{(l)}}}_{r,:,:} &= \sum_{s=1}^{n_{l-1}} w^l _{r,s,:,:}*\sigma\para{\para{f^{l-1}_{\theta^{(l-1)}}}_{s,:,:}}+\mathbbm{1}_{d \times d}b^l_r,
    \vspace{-0.09in}
\end{align*} 
for $2\leq l\leq L$,
where $\mathbbm{1}_{d \times d}$ is the $d$ by $d$ matrix with all unit entries, $n_l$ is the number of channels of the $l$-th layer, and $r=1,\dots,n_l$.
%  (and $n_0=1$)
Our notation indexing the 3D and 4D tensors resembles the PyTorch convention and is defined precisely in Section~\ref{pre::index}.
We use ReLU for the activation $\sigma$.
The number of channels are $3=n_1=\dots=n_{L-1}$ and $n_0=n_L=1$.
So,  $f^l_{\theta^{(l)}}(x) \in \mathbb{R}^{n_{l} \times d \times d} $,
$w^l \in \mathbb{R}^{ n_{l}\times n_{l-1} \times 3\times 3}$,
and $b^l\in\fR^{n_l}$ for $1\le l\le L$.
Write average pool as $S(A)= \frac{1}{d^2}\sum^d_{i=1}\sum^d_{j=1}A_{i,j}$ for $A \in \fR^{d \times d}$.
The CNN outputs 
\vspace{-0.05in}
% \[
%  f^{L+1}_\theta=\frac{1}{d^2}\sum_{i,j}f^L_{i,j,\theta^{(L)}}.
% \]
\[
 f^{L+1}_{\theta}=S(f^L_{\theta^{(L)}}).
\]
For $1\le l\le L$, write $    \theta^{(l)} = \{w^i, b^i : i\leq l\}$
to denote the collection of parameters up to $l$-th layer.

% 4. cnn

%% 4.1 initialization
\subsection{Initialization} \label{sec:cnn_initialization}
Initialize the filters of our $L$-layer CNN $f^{L+1}_\theta$ as follows:
\vspace{-0.05in}
% \begin{align*}
% w^1_{1,1,:,:} &=\begin{bmatrix} 0 & 0 &0\\ 0 & C_L & 0 \\ 0 & 0 & 0  \end{bmatrix}, \quad  w^1_{2,1,:,:}= \begin{bmatrix} u^1_{1,1} & u^1_{1,2} &u^1_{1,3}\\ u^1_{2,1} & u^1_{2,2} & u^1_{2,3} \\ u^1_{3,1} & u^1_{3,2} & u^1_{3,3}  \end{bmatrix},\\\quad  w^1_{3,1,:,:}&= 0_{3 \times 3}\\
% w^l_{1,:,:,:} &=\begin{bmatrix} 0 & 0 &0\\ 0 & 1 & 0 \\ 0 & 0 & 0  \end{bmatrix}, 0_{3 \times 3}, 0_{3 \times 3} \quad \\w^l_{2,:,:,:}&=\begin{bmatrix}  u^l_{1,1} & u^l_{1,2} &u^l_{1,3}\\ u^l_{2,1} & u^l_{2,2} & u^l_{2,3} \\ u^l_{3,1} & u^l_{3,2} & u^l_{3,3}    \end{bmatrix}, 0_{3 \times 3}, 0_{3 \times 3} \quad  \\w^l_{3,:,:,:}&=0_{3 \times 3}, 0_{3 \times 3}, \begin{bmatrix} 0 & 0 &0\\ 0 & 1 & 0 \\ 0 & 0 & 0  \end{bmatrix} \\
% w^L_{1,:,:,:} &= 0_{3 \times 3}, 0_{3 \times 3}, \begin{bmatrix} 0 & 0 &0\\ 0 & 1 & 0 \\ 0 & 0 & 0  \end{bmatrix}
% \end{align*}
\begin{align*}
w^1_{1,1,:,:} &=\begin{bmatrix} 0 & 0 &0\\ 0 & C_L & 0 \\ 0 & 0 & 0  \end{bmatrix}, \,\,\,  w^1_{2,1,:,:}= \begin{bmatrix} u^1_{1,1} & u^1_{1,2} &u^1_{1,3}\\ u^1_{2,1} & u^1_{2,2} & u^1_{2,3} \\ u^1_{3,1} & u^1_{3,2} & u^1_{3,3}  \end{bmatrix},\\\quad  w^1_{3,1,:,:}&= 0_{3 \times 3}\\
w^l_{1,:,:,:} &=\idconv_{3\times 3}, 0_{3 \times 3}, 0_{3 \times 3} \quad \\w^l_{2,:,:,:}&=\begin{bmatrix}  u^l_{1,1} & u^l_{1,2} &u^l_{1,3}\\ u^l_{2,1} & u^l_{2,2} & u^l_{2,3} \\ u^l_{3,1} & u^l_{3,2} & u^l_{3,3}    \end{bmatrix}, 0_{3 \times 3}, 0_{3 \times 3} \quad  \\w^l_{3,:,:,:}&=0_{3 \times 3}, 0_{3 \times 3}, \idconv_{3\times 3} \\
w^L_{1,:,:,:} &= 0_{3 \times 3}, 0_{3 \times 3},\idconv_{3\times 3}
\end{align*}
for $2\le l\le L-1$, where $u^l_{i,j} \in \fR^{3 \times 3}$ is randomly sampled $u^l_{i,j} \stackrel{iid}{\sim} \mathcal{N}(0,\rho^2)$ for $1 \le l \le L-1$.
Here, $0_{3\times 3}$ is the $3$ by $3$ matrix with all zero entries, $\CL>0$ is a scalar growing as a function of $L$ at a rate satisfying $L^2/\CL\rightarrow 0$, and $\rho>0$ is a fixed variance parameter.
Initialize the biases as follows:\vspace{-0.18in}
\begin{align*}
(b_1^1,b_2^1,b_3^1) &=(0, v^1, \CL ) \\
(b_1^l,b_2^l,b_3^l) &=(0,v^l,0)
\\
b^L &=-\CL
\end{align*}
for $2\le l\le L-1$,
where $v^l \in \mathbb{R}$ is randomly sampled $v^l \stackrel{iid}{\sim} \mathcal{N}(0, C_L^2 \beta^2)$ for $1 \le l \le L-1$.
Here, $\beta>0$ is a fixed variance parameter.

\begin{figure*}[ht]
\centering
\begin{tabular}{ccc}
% \hspace{-7mm}
% \hspace{-9mm}
% \subfigure[Convergence of the loss function $L$]{
%       \includegraphics[width=0.45\textwidth]{fig2}}
% \end{tabular}\\
% \vspace{-7mm}
% \hspace{-4mm}
\subfigure [True function and trained MLP after 2000 iterations.]{
      \raisebox{0. \height} {
      \includegraphics[height=0.28\textwidth]{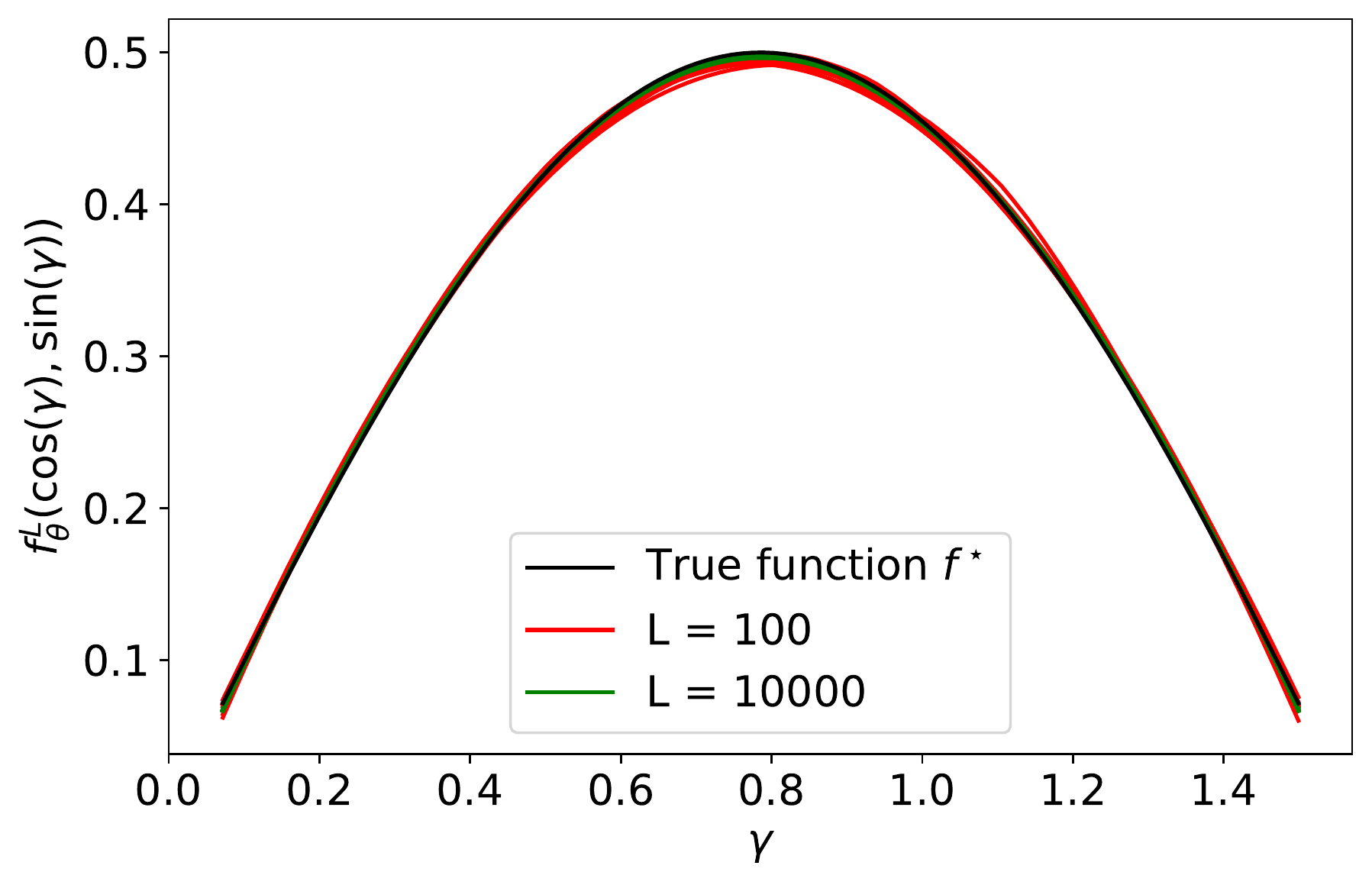}
      }
      \label{fig:ntk_convergence_a}
      }
% \hspace{-6mm}
\subfigure[Scaled NTK values]{
      \raisebox{0. \height} {
      \includegraphics[height=0.28\textwidth]{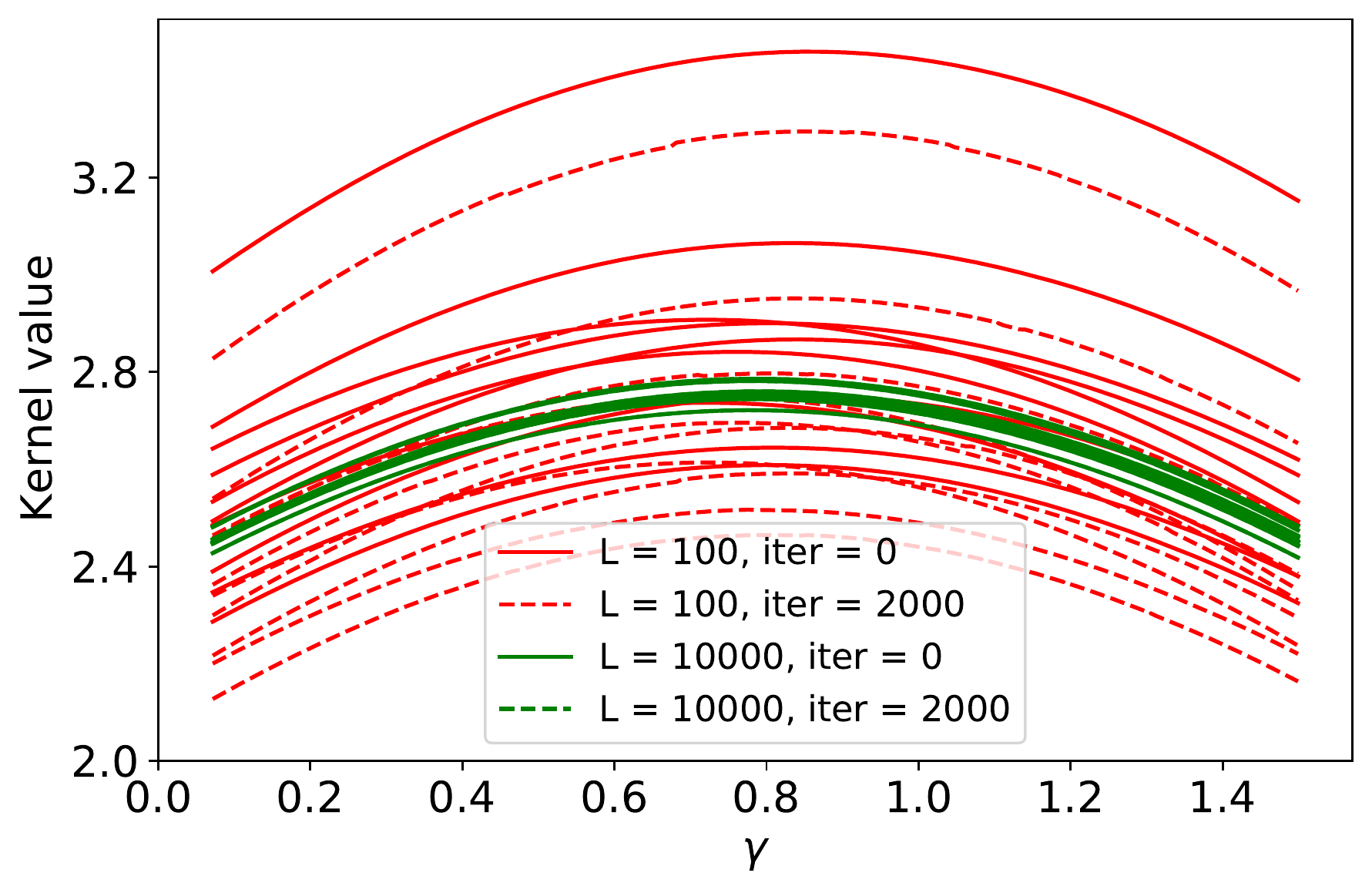}
      }
      \label{fig:ntk_convergence_b}
      }
\end{tabular}
 \vspace{-0.05in}
\caption{
Depth $L$ MLPs learning a toy function with 2D inputs as described in Section~\ref{ss:toy_example},
(Left) Trained deep MLP approximates the true function well, i.e., training succeeds.
(Right) Kernel values evaluated at initialization and after training with 10 independent initialization-training trials each for $L=100$ and $L=10000$.
As $L$ grows, initialization becomes less random, as Theorem~\ref{thm:1} predicts, and the kernel changes less throughout training, as Theorem~\ref{thm:2} predicts.
}
\label{fig:ntk_convergence}
\end{figure*}

\begin{figure*}[ht]
\centering
\begin{tabular}{ccc}
% \hspace{-4mm}
\subfigure [Depth 1000 MLP]{
      \raisebox{0.0 \height} {
      \includegraphics[width=0.45\linewidth]{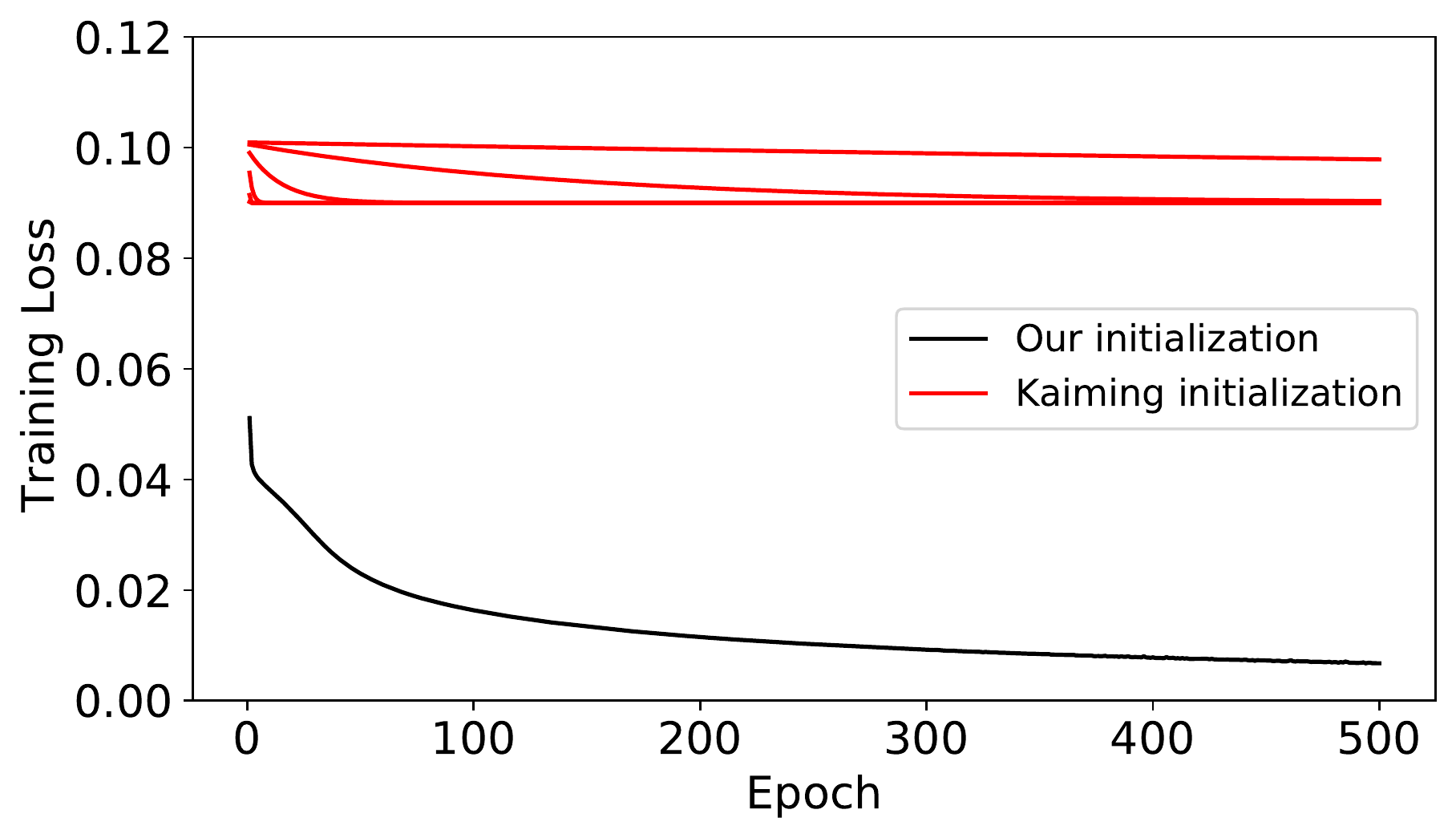}
      }
      \label{fig:train_comp_mlp}
      }
&
% \hspace{-6mm}
\subfigure[Depth 1000 CNN]{
      \raisebox{0.0 \height} {
      \includegraphics[width=0.45\linewidth]{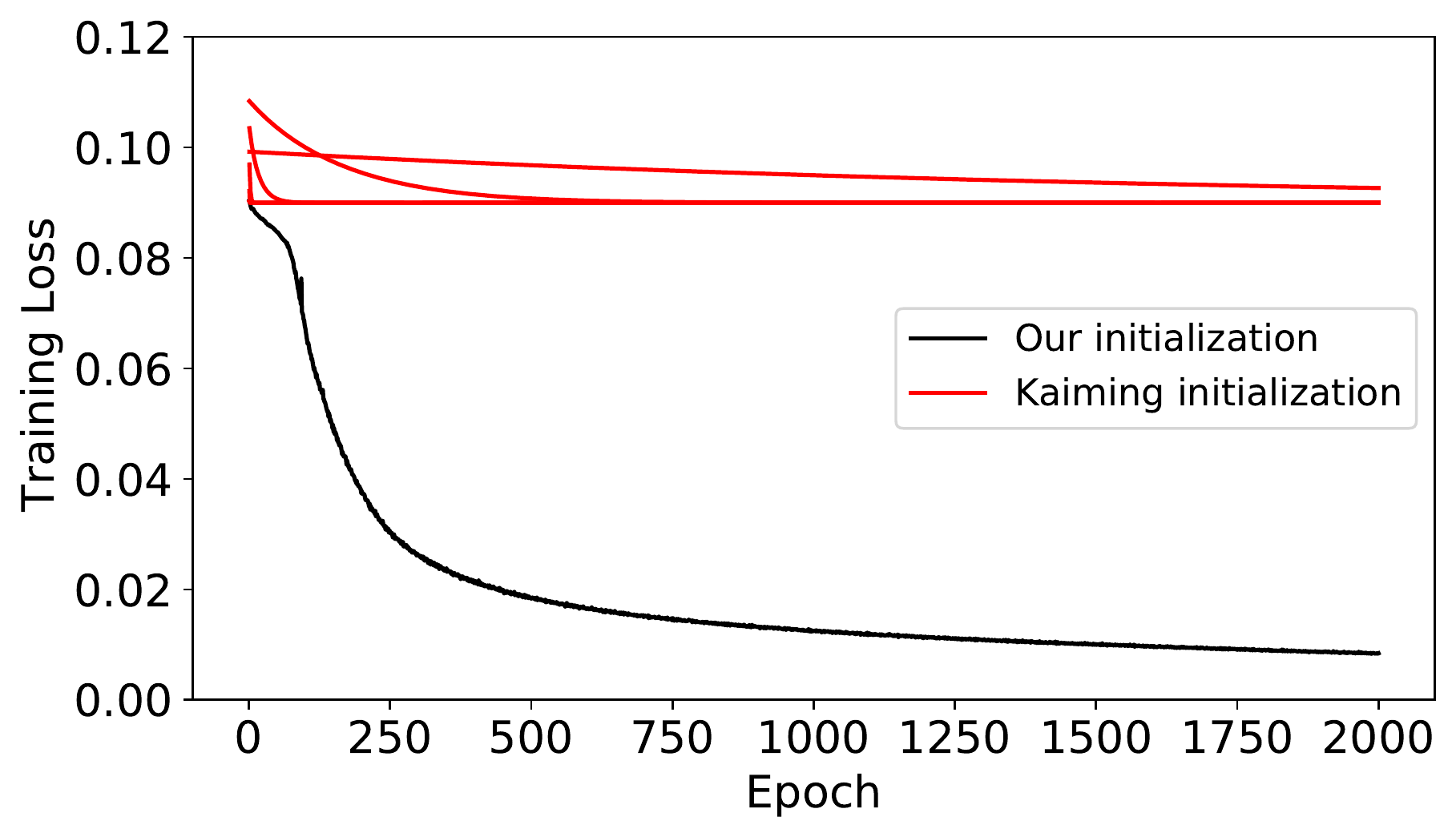}
      }
      \label{fig:train_comp_cnn}
      }
\end{tabular}
 \vspace{-0.15in}
\caption{
Depth 1000 MLPs and CNNs with MNIST are trainable with our proposed initialization but not with the standard Kaiming He initialization. For Kaiming initialization, we show trials with learning rates $1\times 10^{-5}$, $1\times 10^{-4}$, $1\times 10^{-3}$, $0.01$, $0.1$, and $1$.}
\label{fig:train_loss_comp}
\end{figure*}
\begin{table*}[!htbp]
\centering
\begin{tabular}{|c c c c c c c c c c|} 
 \hline
 Task & Architecture & Depth & $C_L$ & $\rho$ & $\beta$ & Learning rate & Epochs  & Training loss   & Test accuracy \\ 
 \hline
 10-class & MLP & 4000 & $4$ & 1 & 1 & $1\times 10^{-5}$ & 1500 & 0.0055 & 97.48\% \\
 10-class & CNN & 4000 & $4$ & $1 / \sqrt{3}$ & 1 & $1\times 10^{-5}$ & 2000 & 0.014578 & 94.02\% \\
%  10-class & CNN & 12000 & $12$ & $1 /\sqrt{3}$ & $5\times 10^{-7}$ &X&X&X\\ 
 Binary & CNN & 20000 & $20$ & $1 /\sqrt{3}$ & 1 & $1\times 10^{-8}$ & 1000 & 0.031257 & 98.87\% \\
 \hline
\end{tabular}
 \vspace{-0.05in}
\caption{Very deep MLP and CNNs trained with MNIST. As the theory predicts, the very deep networks are trainable.}
\label{table:very_deep_exp_mnist}
\end{table*}
% \begin{table*}[!htbp]
% \centering
% \begin{tabular}{|c c c c c c c c c c|} 
%  \hline
%  Task & Architecture & Depth & $C_L$ & $\rho$ & $\beta$ & Learning rate & Epochs  & Training loss   & Test accuracy \\ 
%  \hline
%  10-class & MLP & 1000 & $1$ & 1 & 1 & $1\times 10^{-4}$ & XXX & XXX & XXX\% \\
%  10-class & CNN & 1000 & $1$ & $1 / \sqrt{3}$ & 1 & $1\times 10^{-4}$ & XXX & XXX & XXX\% \\
% %  10-class & CNN & 12000 & $12$ & $1 /\sqrt{3}$ & $5\times 10^{-7}$ &X&X&X\\ 
%  \hline
% \end{tabular}
%  \vspace{-0.05in}
% \caption{Very deep MLP and CNNs trained with CIFAR10. As the theory predicts, the very deep networks are trainable.}
% \label{table:very_deep_exp_cifar}
% \end{table*}

%% 4.2 main results
\subsection{Convergence in infinite-depth limit}
\label{ss:cnn-convergence}
We now analyze the convergence of the infinitely deep CNN.

% Theorem~\ref{thm:4} establishes that the scaled NTK of the randomly initialized CNN has a closed-form expression at initialization, before training, as the depth $L$ becomes infinite.
\begin{theorem}[Scaled NTK at initialization]
\label{thm:4}
Suppose $f^{L+1}_\theta$ is initialized as in Section \ref{sec:cnn_initialization}.
For any $x, x'\in\fR_+^{d\times d}$, 
\[
S\para{\tilde{\Theta}^L_0(x,x')}
\convinp
\tilde{\Theta}^\infty(x, x')
\]
as $L \rightarrow \infty$, where
\begin{align*}
&\tilde{\Theta}^\infty(x, x')=
\frac{1}{d^2}\sum_{s=1}^3\sum_{u=1}^3\bigg(p_d(s,u)+d^2S(x_{\psi_{s,u}})S(x'_{\psi_{s,u}})\\&
\qquad
+\sum_{i,j\in \psi_{s,u}}\sum_{i',j'\in \psi_{s,u}}\mathbb{E}[\sigma(g(\phi_{i,j}(x)))\sigma(g(\phi_{i',j'}(x')))]\bigg),
\end{align*}
$g \sim \mathcal{GP} ( 0 ,\rho^2 \left\langle x, x'\right\rangle +\beta^2)$, $\psi_{s,u}$ is the set of coordinate which satisfy $x_{\psi_{s,u}} = [x]_{s-1:d+s-2, u-1:d+u-2}$ and $x_{pq}=0, \phi_{mn}(x)=0$ if the index is out of bounds, and 
\begin{align*}
p_d(s,u)=\left\{
\begin{array}{ll}
d^4 & (s,u)=(2,2),\\
d^2(d-1)^2 & |s-u|=1,\\
(d-1)^4 & \text{else}.
\end{array}\right.
\end{align*}
% \begin{align*}
    
% \end{align*}
\end{theorem}
% Obviously, $\Theta(x,x')$ is positive constant, positive semi-definite.  
% \begin{remark}
% Our NTK is not infinite width and depth limit. Our NTK assumed bounded width. 
% \end{remark}

% The scaled CNTK depends on $\theta$ and evolves during training.
% Similar to the CNTK with infinite width,  we show that CNTK stays at the initial deterministic kernel in the infinite-depth limit.

% Theorem~\ref{thm:5} establishes that the scaled NTK is invariant throughout the training as $L\rightarrow\infty$.

% For the sake of simplicity, we let $\delta_t^L = \partial_f C = \delta\vert_{f_\theta^L}$.
\begin{theorem}[Invariance of scaled NTK] 
\label{thm:5}
Let $T>0$.
Suppose $\int_0^T\EP{\delta\vert_{f_\theta^L}}\!\!\!\!dt$ is stochastically bounded as $L\rightarrow\infty$.
Then, for any $x, x'\in\fR^{d\times d}_+$, 
\[
S\para{\tilde{\Theta}^L_t(x,x')}
\convinp
\tilde{\Theta}^{\infty}(x,x')
\]
uniformly for $t\in [0,T]$ as $L \rightarrow \infty$.
\end{theorem}

% Also Let the kernel predictor defined as follows
% \begin{align*}
%     f_{\mathrm{ntk}}(x)=\frac{1}{d^2}(\tilde{\Theta}_0^{\infty}(x,x_1), \cdots, \tilde{\Theta}_0^{\infty}(x,x_N))K^{-1}f^\star(X).
% \end{align*}
% where $K_{i,j}=\tilde{\Theta}_0^{\infty}(x_i,x_j)$ and $[f^\star(X)]_i=f^\star(x_i)$.

% Theorem~\ref{thm:6} concludes the analysis by characterizing the trained CNN as $L\rightarrow\infty$ in the case of the quadratic loss.
Again, define the kernel regression predictor as
\begin{align*}
    f_{\mathrm{ntk}}(x)=\para{\tilde{\Theta}^{\infty}(x,x_1), \dots, \tilde{\Theta}^{\infty}(x,x_N)}K^{-1}f^\star(X),
\end{align*}
where $K_{i,j}=\tilde{\Theta}^{\infty}(x_i,x_j)$ and $[f^\star(X)]_i=f^\star(x_i)$ for $i,j\in\{1,\dots,N\}$.
\begin{theorem}[Equivalence between deep CNN and kernel regression]
\label{thm:6}
Let $\loss(f) = \frac{1}{2}\EP{f-f^\star}^2$. Let $\tilde{\Theta}^{\infty}$ be positive definite.
If $f_t$ follows the kernel gradient flow with respect to $\tilde{\Theta}^{\infty}$, then for any $x \in \fR^{d\times d}_+$,
% \vspace{-0.02in}
\[
\lim_{t \rightarrow \infty}f_t (x) = f_{\mathrm{ntk}}(x).
\]
\end{theorem}
\vspace{-0.1in}

% The definitions of this section and the trainability results of Section~\ref{ss:cnn-convergence} can be generalized as follows.
% In this section, we do not pursue the following two forms of generality for the sake of notational simplicity. 
\paragraph{Generalizations.}
% \emph{Remark.}
At the expense of slight notational complications, we can generalize our results as follows. We assumed the convolutional filter size is $3\times 3$, but we can use larger filters by assigning a symbol for the filter size and managing the summation indices with care. We assumed the number of input channels and output scalar dimension are $1$, but we can have $c$ input channels and $k$ outputs, i.e., $f^{L+1}_{\theta}\colon \fR^{c\times d\times d}_+ \rightarrow \fR^k$, by letting the intermediate layers have $c+1+k$ channels. In fact, Section~\ref{ss:trainability_experiment} presents a 10-class classification of MNIST with a deep CNN using $1+1+10=12$ channels in intermediate layers.

% 5. experiments
\section{Experiments}
\label{s:experiments}

In this section, we experimentally demonstrate the invariance of the scaled NTK and the trainability of deep neural networks. The code is available at \url{https://github.com/lthilnklover/deep-narrow-NTK}
% The code is provided as supplementary material.

\subsection{Convergence of the scaled NTK}
\label{ss:toy_example}
Our first experiment, inspired by \citet{JacotGabrielHongler2018_neural},
trains $L$-layer MLPs on 2-dimensional inputs and shows that the network and its scaled NTK converges as the depth $L$ increases.
For $L = 100$ and $L= 10000$, we initialize 10 independent MLP instances and train them to approximate $f^\star(x_1, x_2) = x_1x_2$ using the quadratic loss.
To verify that the networks are indeed successfully trained, we compare the trained MLP against the true function $f^\star$ in Figure~\ref{fig:ntk_convergence_a}.
We then plot the scaled NTK $\tilde{\Theta}^{L}(x_0, x)$ for fixed $x_0 = \big{(}\frac{1}{\sqrt{2}}, \frac{1}{\sqrt{2}}\big{)}$ and $x = (\cos(\gamma), \sin(\gamma))$ for $0 < \gamma < \pi / 2$ in Figure~\ref{fig:ntk_convergence_b}.
The kernels are plotted at initialization ($t = 0$), and after 2000 iterations of gradient descent.

\subsection{Trainability of the deep narrow neural network}
\label{ss:trainability_experiment}
Next, we demonstrate the empirical trainability of the deep narrow networks on the MNIST dataset.

% Note that as the depth and $C_L$ increases, the learning rate should be decreased to properly train the network. Hence, in very deep networks, we use double precision floating-point format (float64) for the computational stability.

\paragraph{Very deep MLP.}
We train $L$-layer MLPs with $d_{\mathrm{in}} = 784$ and $d_{\mathrm{out}} = 10$ using the quadratic loss with one-hot vectors as targets.
To establish a point of comparison, we attempt to train a 1000-layer MLP with the typical Kaiming He uniform initialization \cite{HeZhangRenSun2015_delving}.
We tuned the learning rate via a grid search from $0.00001$ to $1.0$, but the network was untrainable, as one would expect based on the prior findings of \cite{HeSun2015_convolutional, SrivastavaGreffSchmidhuber2015_highway,HuangWangTaoZhao2020_why}. 
In contrast, when we use the initialization defined in Section~\ref{sec:mlp_initialization}, the deep MLP was trainable. Figure~\ref{fig:train_comp_mlp} reports these results. To push the depth, we also trained a 4000-layer MLP and report the results in Table~\ref{table:very_deep_exp_mnist}. To the best of our knowledge, this 4000-layer MLP holds the record for the deepest trained MLP.

% \paragraph{Very deep MLP (CIFAR10).}
%  We also train a 1000-layer MLP on the CIFAR10 dataset. To reduce computational cost, we insert a $4 \times  4$ average pool before the first layer to reduce the CIFAR10 input size to $3\times 8 \times  8 $: $d_{in} = 192, d_{out}=10$. The results and parameter settings are reported in Table~\ref{table:very_deep_exp_cifar}. 

\paragraph{Very deep CNN.} We train $L$-layer CNNs using the quadratic loss with one-hot probability vectors as targets.
To reduce the computational cost, we insert a $4\times 4$ average pool before the first layer to reduce the MNIST input size to $7\times 7$.
As in the MLP experiment, we attempt to train a 1000-layer CNN with the typical Kaiming He uniform initialization, but the network was untrainable.
In contrast, when we use the initialization defined in Section~\ref{sec:cnn_initialization}, the deep CNN was trainable. Figure~\ref{fig:train_comp_cnn} reports these results.
To push the depth, we also trained a 4000-layer CNN and report the results in Table~\ref{table:very_deep_exp_mnist}. To further push the depth, we simplify the problem to binary classification between digits 0 and 1 and use values $0$ and $1$ as targets. We then trained a 20000-layer CNN and report the results in Table~\ref{table:very_deep_exp_mnist}. This CNN surpasses the 10000-depth CNN of \cite{XiaoBahriSohl-DicksteinSchoenholzPennington2018_dynamical} and, to the best of our knowledge, holds the record of the deepest trained CNN.

\subsection{Accumulation of the layer-wise effect}
\begin{figure}
    \centering
    \includegraphics[width=0.45\textwidth ]{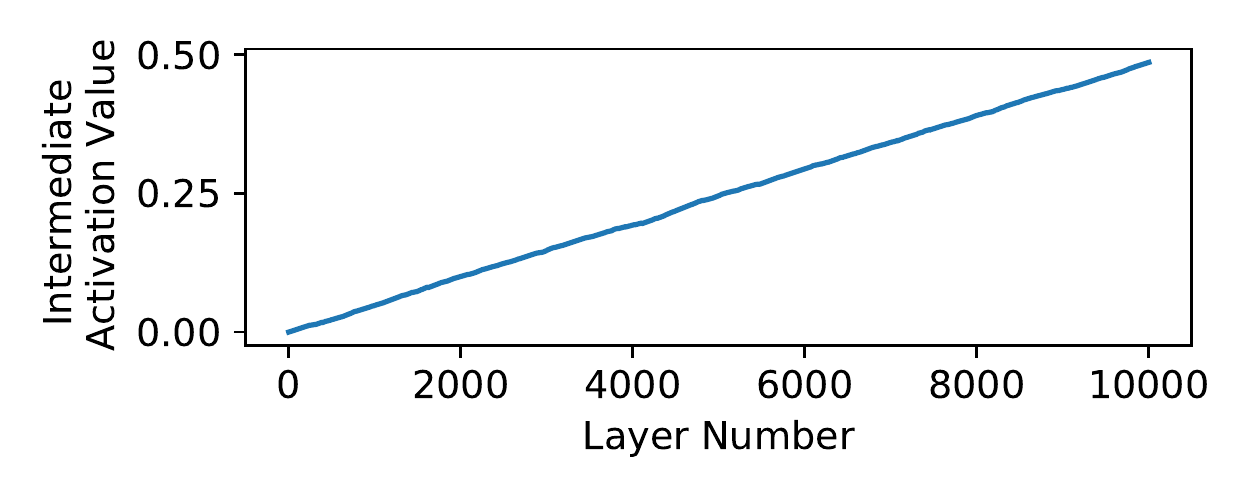}
    \vspace{-6 mm}
    \caption{Intermediate activation values of the rightmost neurons of each layer after training. (The effect of the bias $C_L$ is compensated for to help visualize the change.) 
    }
     \vspace{-0.10in}
    \label{fig:output_evo}
\end{figure}

To observe the accumulation of the output value throughout the depth, we plot the intermediate activation values of the rightmost neurons of each layer from the $10000$-layer MLP trained in Section \ref{ss:toy_example}. Precisely, we plot $(\sigma(f^{(l)}(x)))_{d_{\mathrm{in}} + d_{\mathrm{out}} + 1}-C_L$ for $1\le l \le L-1$ and $(f^{(L)}(x))_{d_{\mathrm{in}} + d_{\mathrm{out}} + 1}$. Figure~\ref{fig:output_evo} shows that the target output value is achieved through the accumulation of the effect of $10000$-layers.
% The gradual change of intermediate activation value to the target value $0.5$ shows that each layer is contributing infinitesimally, as is the case with any NTK setup, although the aggregate variation is not infinitesimal.

\section{Conclusion}
This work presents an NTK analysis of a deep narrow MLP and CNN in the infinite-depth limit and establishes the first trainability guarantee on deep narrow neural networks. Our results serve as a demonstration that infinitely deep neural networks can be made provably trainable using the right initialization, just as the infinitely wide counterparts are. However, our results do have the following limitations. First, while our proposed initialization is straightforwardly implementable, it is far from the initializations used in practice. Second, our results do not indicate any benefit of using deep neural networks compared to wide neural networks. Further investigating the trainability of overparameterized deep neural networks to address these questions would be an interesting direction of future work.

% \section*{Acknowledgements}
% TBD

\section*{Acknowledgements}
JL, JYC, and EKR were supported by the National Research Foundation of Korea (NRF) Grant funded by the Korean Government (MSIP) [No. 2022R1C1C1010010], the National Research Foundation of Korea (NRF) Grant funded by the Korean Government (MSIP) [No. 2022R1A5A6000840], and the Samsung Science and Technology Foundation (Project Number SSTF-BA2101-02). AN was supported by Basic Science Research Program through the National Research Foundation of Korea (NRF) funded by the Ministry of Education (2021R1F1A1059567). We thank Jisun Park and TaeHo Yoon for providing valuable feedback. We also thank anonymous reviewers for giving thoughtful comments.

% \newpage

%%%%%%%%%%%%%%%%%%%%%%%%%%%%%%%%%%%%%%%%%%%%%%%%%%
%%% Bibliography
%%%%%%%%%%%%%%%%%%%%%%%%%%%%%%%%%%%%%%%%%%%%%%%%%%
\bibliography{infinitedepth}
\bibliographystyle{icml2022}

%%%%%%%%%%%%%%%%%%%%%%%%%%%%%%%%%%%%%%%%%%%%%%%%%%%%%%%%%%%%%%%%%%%%%%%%%%%%%%%
%%%%%%%%%%%%%%%%%%%%%%%%%%%%%%%%%%%%%%%%%%%%%%%%%%%%%%%%%%%%%%%%%%%%%%%%%%%%%%%
% DELETE THIS PART. DO NOT PLACE CONTENT AFTER THE REFERENCES!
%%%%%%%%%%%%%%%%%%%%%%%%%%%%%%%%%%%%%%%%%%%%%%%%%%%%%%%%%%%%%%%%%%%%%%%%%%%%%%%
%%%%%%%%%%%%%%%%%%%%%%%%%%%%%%%%%%%%%%%%%%%%%%%%%%%%%%%%%%%%%%%%%%%%%%%%%%%%%%%

\onecolumn
\newpage
\appendix

%%%%%%%%%%%%%%%%%%%%%%%%%%%%%%%%%%%%%%%%%%%%%%%%%%%%%
%%% Proof of Theorems
%%%%%%%%%%%%%%%%%%%%%%%%%%%%%%%%%%%%%%%%%%%%%%%%%%%%%

\newpage

\section{Preliminaries}
\subsection{Gr\"onwall's lemma}
The following is a generalized version of Gr\"onwall's lemma, which is essential to prove the invariance of kernel.
\begin{lemma}[\citealt{Dragomir2003_gronwall}] 
% \jy{XXX citation issue. Maybe also have to cite \cite{Dragomir2003_gronwall}}
Let the $x, a, b$ and $k$ be continuous and nonnegative functions on $J=[\alpha, \beta]$.
Let $n\ge 2$ be a positive integer $\frac{a}{b}$ be a nondecreasing function. 
If
\[z(t) \le a(t) +b(t) \int_\alpha^{t} k(s) z^n(s)\, ds, \,\,t\in J \]
then
\[z(t) \le a(t) \left\{1- (n-1) \int_\alpha^{t} k(s) b(s) a^{n-1}(s) \,ds \right\}^{\frac{1}{1-n}},\]
where $\alpha \le t \le \beta_n$, for $\beta_n$ defined by
\[\beta_n = \mathrm{sup} \left\{ t \in J: (n-1) \int_\alpha^{t} kba^{n-1}\,ds<1 \right\}.\]
\end{lemma}
% The similar result holds in $n=1$ case.
% \begin{lemma}
% For a constant $K$, if
% \[z(t) \le K +\int^t_a k(s)z(s)\,ds \]
% for all $t\in[\alpha, \beta]$, then
% \[z(t) \le K\mathrm{exp}\para{ \int^t_\alpha k(s)\,ds}. \]
% \end{lemma}

\section{Proof of Theorems of NTK for MLP}
%%% Gronwall's lemma

%%% Proof of Theorem 1
\subsection{Proof of Theorem~\ref{thm:1}}
The following lemma shows the recursive relation of the neural tangent kernel (without the scaling factor).
\begin{lemma}\label{lem:recursive}
For $t\geq 0$, $l\ge 2$ and $x, x'\in\fR_+^{d_{in}}$, 
\begin{align*}
    \Theta^{l}_t(x,x')
    &=W^l(t) \diag(\dot{\sigma}(f^{l-1}_{\theta^{(l-1)}}(x)))\Theta^{l-1}_t(x,x') \diag(\dot{\sigma}(\fultm(x')))(W^l(t))^{\intercal}\\
    &\quad+(\sigma(\fultm(x'))^{\intercal}\sigma(\fultm(x))+1) I_{d_{\mathrm{in}}+d_{\mathrm{out}}+1},
\end{align*}
where $\Theta^{1}_t(x,x')=(x^{\intercal}x'+1)I_{d_{\mathrm{in}}+d_{\mathrm{out}}+1}$ if $l=1$. 
\end{lemma}
\begin{proof}[Proof of Lemma~\ref{lem:recursive}]
\begin{align*}
    \Theta^l_t(x, x') 
    &= \partial_{\theta^{(l)}} \fult(x)\para{\partial_{\theta^{(l)}}\fult(x')}^{\intercal} \\
    &=\sum_{\theta_i \in \theta^{(l)} } \partial_{\theta_i} \fult(x)\para{\partial_{\theta_i}\fult(x')}^{\intercal}\\    
    &=\sum_{\theta_i \in \theta^{(l-1)} } \partial_{\theta_i} \fult(x)\para{\partial_{\theta_i}\fult(x')}^{\intercal}\\
    &\quad +\sum_{W^l_{ij}} \partial_{W^l_{ij}} \fult(x)\para{\partial_{W^l_{ij}}\fult(x')}^{\intercal}
    +\sum_{b^l_i} \partial_{b^l_i} \fult(x)\para{\partial_{b^l_i}\fult(x')}^{\intercal}.
\end{align*}
The last two terms of RHS is
\begin{align*}
    &\sum_{W^l_{ij}} \para{\partial_{W^l_{ij}} \fult(x)}_k\para{\partial_{W^l_{ij}}\fult(x')}_{k'}
    +\sum_{b^l_i} \para{\partial_{b^l_i} \fult(x)}_k\para{\partial_{b^l_i}\fult(x')}_{k'}\\
    &=(\sigma(\fultm(x) )^{\intercal}\sigma(\fultm(x'))+1) \delta_{k,k'},
\end{align*}
where $(\cdot)_k$ denote the $k$-th component of the vector, and $\delta_{k,k'}=1$ if $k=k'$ and $0$ otherwise. 
Also, since 
\begin{align*}
    \partial_{\theta^{(l-1)}} \fult(x)= W^l(t) \diag(\dot{\sigma}(\fultm(x)))\partial_{\theta^{(l-1)}} \fultm(x),
\end{align*}
the first term of RHS is
\begin{align*}
\sum_{\theta_i \in \theta^{(l-1)} } \partial_{\theta_i} \fult(x)\para{\partial_{\theta_i}\fult(x')}^{\intercal}
=& \partial_{\theta^{(l-1)}} \fult(x)\para{\partial_{\theta^{(l-1)}}\fult(x')}^{\intercal}\\
=&W^l(t) \diag(\dot{\sigma}(\fultm(x)))\Theta^{l-1}_t(x,x') \diag(\dot{\sigma}(\fultm(x')))(W^l(t))^{\intercal}.
\end{align*}
Thus, we have a recursive relation for kernel
\begin{align*}
    \Theta^{l}_t(x,x')
    &=W^l(t) \diag(\dot{\sigma}(f^{l-1}_{\theta^{(l-1)}}(x)))\Theta^{l-1}_t(x,x') \diag(\dot{\sigma}(\fultm(x')))(W^l(t))^{\intercal}\\
    &\quad+(\sigma(\fultm(x))^{\intercal}\sigma(\fultm(x'))+1) I_{d_{\mathrm{in}}+d_{\mathrm{out}}+1}.
\end{align*}
for $l \ge 2$, where $\Theta^{1}_t(x,x')=(x^{\intercal}x'+1)I_{d_{\mathrm{in}}+d_{\mathrm{out}}+1}$ if $l=1$. 
\end{proof}

For notational simplicity, we let
\[
\W^k_j(x,t)=W^j(t) \diag\left(\dot{\sigma}\left(f^{j - 1}_{\theta^{(j - 1)}(t)}(x)\right)\right)W^{j-1}(t) \diag\left(\dot{\sigma}\left(f^{j-2}_{\theta^{(j-2)}(t)}(x)\right)\right)
\cdots W^k(t) \diag\left(\dot{\sigma}\left(f^{k-1}_{\theta^{(k-1)}(t)}(x)\right)\right)
\]
% \jy{may for simplicity as in the Appendix C, \[
% \W^k_j(x,t)= \prod_{l=j}^j W^l(t)\diag\left(\dot{\sigma}\left(f^{l-1}_{\theta^{(l-1)}(t)}(x)\right)\right),
% \]
% or in inverted order as in Appendix C\\}
for $2 \le k, j \le L$, where $\W^k_j(x,t)=1$ if $k > j$. 
By applying the above recursive relation inductively, we get the following corollary.
\begin{corollary}\label{cor:recursive}
For any $x, x'\in\fR_+^{d_{in}}$, and $t\ge 0$,
\begin{align*}
    \tilde{\Theta}^L_t(x, x')
    &=\frac{1}{L}\sum_{l=0}^{L-1} \left[\left( \frac{1}{\CL}\sigma \left(\fult(x) \right)^{\intercal} \frac{1}{\CL}\sigma \left(\fult(x') \right)\right) \W_L^{l+2}(x,t)\W_L^{l+2}(x',t)^{\intercal}\right]\\
    &\quad+\frac{1}{L\CL^2}\sum_{l=0}^{L-1} \left[\W_L^{l+2}(x,t)\W_L^{l+2}(x',t)^{\intercal}\right],
\end{align*}
where we define $\sigma(f^{0}(x))=x$ for convenience.
\end{corollary}

Now, we are ready to prove the Theorem~\ref{thm:1}.
\begin{proof}[Proof of Theorem~\ref{thm:1}]
At initialization, we can simplify the kernel further. 
Since we assumed positive input, we have
\begin{align*}
W^{l}(0) \diag(\dot{\sigma}(f^{l-1}_{\theta^{(l-1)}(0)}(x))) =&\begin{bmatrix} I_{d_{\mathrm{in}}} & 0_{d_{\mathrm{in}}\times 1} &0_{d_{\mathrm{in}}\times d_{\mathrm{out}}} \\ u^l & 0_{1\times 1} & 0_{1\times d_{\mathrm{out}}} \\ 0_{d_{\mathrm{out}}\times d_{\mathrm{in}}} & 0_{d_{\mathrm{out}}\times 1} & I_{d_{\mathrm{out}}} \end{bmatrix}\quad \mbox{for $1\leq l\leq L-1$,}\\
W^{L}(0) \diag(\dot{\sigma}(f^{L-1}_{\theta^{(L-1)}(0)}(x))) =&\begin{bmatrix} 0_{d_{\mathrm{out}}\times d_{\mathrm{in}}} & 0_{d_{\mathrm{out}}\times 1} & I_{d_{\mathrm{out}}} \end{bmatrix}.
\end{align*}

This implies 
\begin{align}
\W^k_l(x,0) =W^{l}(0) \diag(\dot{\sigma}(f^{l-1}_{\theta^{(l-1)}(0)}))= W^l(0) \label{eq:initial W}
\end{align}
for $l \ge k$.
Also, our initialization implies
\begin{align}
f^1_{\theta^{(1)}(0)}(x) = \begin{bmatrix} \CL x\\ (u^{1})^{\intercal}x+v^1\\ \CL\mathbbm{1}_{d_{\mathrm{out}}} \end{bmatrix},
\,\, f^l_{\theta^{(l)}(0)}(x) =  \begin{bmatrix} \CL x\\ \CL (u^{l})^{\intercal} x+v^{l}\\ \CL\mathbbm{1}_{d_{\mathrm{out}}} \end{bmatrix}, \,\, 
f^L_{\theta^{(L)}(0)}=0_{d_{\mathrm{in}} \times 1} \label{eq:initialization hidden layer}
\end{align}
for $2 \le l \le L - 1$. 
Thus, the scaled NTK is given by
\begin{align*}
     \tilde{\Theta}^{L}_0(x,x')=\frac{1}{L\CL^2}\para{x^{\intercal} x'+1+\sum_{j=0}^{L-1} (\CL^2x^{\intercal} x'+\sigma(\CL (u^{j})^{\intercal} x+v^{j})\sigma(\CL (u^{j})^{\intercal} x'+v^{j})+\CL^2+1)}I_{d_{\mathrm{out}}}.
\end{align*}
Let  $g^j(x)=u^{j} x+v^{j}/\CL$, then $\{g^j(x)\}_{j=0}^{L-1}$ are i.i.d.\ Gaussian process,
where $g^j  \sim \mathcal{GP} ( 0 ,\frac{\rho^2}{d_{in}} x^{\intercal}x'+\beta^2)$.
This is because
\[\mathbb{E}_{u^j,v^j}[g^j]=0\]
and
\begin{align*}
\mathbb{E}_{u^j,v^j}[g^j(x)g^j(x')]&=\mathbb{E}_{u^j,v^j}\left[\left((u^{j})^{\intercal} x+v^{j}/\CL\right) \left((u^{j})^{\intercal} x'+v^{j}/\CL \right)\right]\\
&=\mathbb{E}_{u^j}\left[\left((u^j)^{\intercal} x\right)\left((u^j)^{\intercal} x' \right) \right]+\beta^2\\
&= \frac{\rho^2}{d} x^{\intercal}x'+\beta^2.
\end{align*}
Hence, we have 
 \[\tilde{\Theta}^{L}_0(x,x') 
 \convinp \para{x^{\intercal} x'+1+ \mathbb{E}_{ g \sim \mathcal{GP} ( 0 ,\frac{\rho^2}{d_{\mathrm{in}}} x^{\intercal}x'+\beta^2)}[\sigma(g(x))\sigma(g(x'))]}I_{d_{\mathrm{out}}}\]
as $L \rightarrow \infty$ by the law of large number, which concludes the proof.
 \end{proof}

%%% Proof of Theorem 2
\subsection{Proof of Theorem~\ref{thm:2}}
For proving our scaled NTK stays constant during training, we need to show that intermediate weight and pre-activation values are effectively unchanged.
In this paper, we also consider Lyapunov functions similar to \cite{JacotGabrielHongler2018_neural};
however, we need more delicate Lyapunov function to handle infinite depth network.

For the sake of simplicity, we set the scaling factor of gradient flow by $\frac{1}{(L-1)\CL^2}$ throughout the proof.
Also, without loss of generality, we assume the norms of data points $\ab{x_i}$ are bounded by 1 for all $1\leq i\leq N$,
and we further assume that all other inputs $x\notin\{x_1, \dots, x_N\}$ also have bounded norm, i.e., $\ab{x}\leq 1$.

We define Lyapunov functions to track intermediate weights and pre-activation values of the network.
For $ 1 \le j \le 8$ and  $\CL>0$ satisfying $L^2/\CL\rightarrow 0$,
\begin{align*}
    \Phi_{L, j}(t) & = \frac{1}{(L-1) \CL^j}\sum^{L-1}_{l=1} \left(\EP{\fulz(x)}+\EP{\fult(x)-\fulz(x)}\right)^j\\
    \Psi_{L, 2}(t)& = \frac{1}{(L-1)^2}\sum^{ L}_{l=2} \frac{L-1}{l-1} \sum_{ i=2}^{l} \left(\EP{\W_l^i(x, 0)}+\EP{\W_l^i(x, t)-\W_l^i(x, 0)}\right)^2\\
    \Psi_{L,8}(t)& = \frac{1}{L-1} \sum_{l= 2}^L \left(\EP{\W_L^l(x, 0)}+\EP{\W_L^l(x, t)-\W_L^l(x, 0)}\right)^8,
\end{align*}
where $\Phi_{L, 0} =1$.

At initialization ($t=0$), we have
\begin{align*}
    \Phi_{L, j}(0) & = \frac{1}{(L-1) \CL^j}\sum^{L-1}_{l=1} \left(\EP{\fulz(x)}\right)^j\\
    \Psi_{L, 2}(0)& = \frac{1}{(L-1)^2}\sum^{ L}_{l=2} \frac{L-1}{l-1} \sum_{ i=2}^{l} \left(\EP{\W_l^i(x, 0)}\right)^2\\
    \Psi_{L,8}(0)& = \frac{1}{L-1} \sum_{l= 2}^L \left(\EP{\W_L^l(x, 0)}\right)^8.
\end{align*}
From \eqref{eq:initial W} and \eqref{eq:initialization hidden layer},  imply that
$\Phi_{L,j}(0)$ and $\Psi_{L,2}(0)$ converges (in probability) to constant values $\Phi_{\infty, j}(0)$ and $\Psi_{\infty, 2}(0)$ 
by the law of large number, respectively.
Furthermore, $\Psi_{L, 8}(0)$ also converges (in probability) to constant value $\Psi_{\infty, 8}(0)$ since $\W_L^l(x, 0) = W^L(0)$.
Then, the following proposition implies Lyapunov function remains constant during training.

% prp 1
\begin{proposition}\label{prp:pin-norm}
For $1 \le j \le 8$,
\[ \Phi_{L,j}(t)    \convinp\Phi_{\infty, j}(0), 
\quad \Psi_{L,2}(t) \convinp \Psi_{\infty, 2}(0), 
\quad \Psi_{L, 8}(t)\convinp \Psi_{\infty, 8}(0)\]
uniformly for $t \in [0,T]$ as $ L \rightarrow \infty$.
\end{proposition}

The following lemma is a key step to prove the proposition which allows us to apply Gr\"onwall's Lemma. For the sake of simplicity, we let $\delta_t^L = \partial_f C = \delta\vert_{f_\theta^L}$. 
% Lemma
\begin{lemma}\label{lem:pin-bound}
For $t \geq 0$, if $f^l_{\theta^{(l)}(t)}(x)$ is element-wise nonzero for $1 \le l \le L-1$ and $x\in\{x_1,\dots, x_N\}$, then
\begin{align*}
    \partial_t \Phi_{L,j}(t) &\le \frac{2jN^{3/2}(L-1)}{\CL}\Phi_{L,j-1}(t)\Phi_{L,8}(t)\Psi_{L, 2}(t)\Psi_{L,8}(t)\EP{\dLt}\\
    \partial_t \Psi_{L,2}(t)& \le \frac{2N(L-1)}{\CL}\Phi_{L,4}(t) \para{\Psi_{L,2}(t)}^{2}(t)\Psi_{L,8}(t)\EP{\dLt}\\
    \partial_t \Psi_{L,8}(t) & \le \frac{8N(L-1)}{\CL}\Phi_{L,4}(t) \Psi_{L,2}(t)\left(\Psi_{L,8}(t)\right)^{2}\EP{\dLt},
\end{align*}
for $1\le j\leq 8$. 
\end{lemma}
The proof of Lemma~\ref{lem:pin-bound} is given in Section~\ref{app:prp1}.

\begin{proof}[Proof of Proposition~\ref{prp:pin-norm}]
In order to apply Lemma~\ref{lem:pin-bound} for all $t\in[0, T]$,
we need to show that $\fult(x)$ is element-wise nonzero for all $t\in[0, T]$, $1 \le l \le L-1$, and $x\in\{x_1,\dots, x_N\}$.
This is indeed true when $L$ is large enough, which we proved in Section~\ref{app:not crossing}.
From Lemma~\ref{lem:pin-bound}, we get 
\begin{align*}
    \partial_t \left(\sum_{j=1}^8\Phi_{L,j}(t)+ \Psi_{L,2}(t)+  \Psi_{L,8}(t)\right) 
    &\le \frac{16N^{3/2}(L-1)}{\CL} \left(\sum_{j=1}^8\Phi_{L,j}(t)+ \Psi_{L,2}(t)+ \Psi_{L,8}(t)\right)^4\EP{\dLt},
\end{align*}
which implies 
\begin{align*}
     \Gamma_L(t) &\le \Gamma_L(0)+ \frac{16N^{3/2}(L-1)}{\CL}\int_0^{t} \Gamma_L(s)^4 \EP{\delta_s^L}\,ds,
\end{align*}
where $\Gamma_L(t)= \sum_{j=1}^8\Phi_{L,j}(t) + \Psi_{L,2}(t)+ \Psi_{L,8}(t)$.
By Gr\"onwall's lemma with $z(t)=\Gamma_L(t), \, a(t)=\Gamma_L(0), \, b(t)=\frac{16N^{3/2}(L-1)}{\CL}, \, k(s)=\EP{\delta^L_s} $,
\[\Gamma_L(t) \le \Gamma_L(0) \left\{ 1- \frac{48N^{3/2}(L-1)}{\CL}\Gamma_L(0)^{3} \int_0^{t} \EP{\delta^L_s}\,ds  \right\}^{-\frac{1}{3}} \]
for $0 \le t\leq \beta_L$.
Recall that $\int_0^{T} \EP{\delta_s^L}\,ds$ is stochastically bounded and $\Gamma_L(0)$ converges to a constant.
Thus, for $L^2/\CL\rightarrow 0$, we get $\frac{48N^{3/2}(L-1)}{\CL}\Gamma_L(0)^{3} \int_0^{t} \EP{\delta^L_s}\, ds\convinp 0$.
This implies the upper bound of $\Gamma_\infty(t)$ converges to  $\Gamma_\infty(0)$ in probability.
On the other hand, $\Gamma_L(t) \ge \Gamma_L(0)$ by construction and  $\beta_L\rightarrow \infty$ as $L \rightarrow \infty$.
Thus, $\Gamma_L(t)\convinp \Gamma_\infty(0)$ uniformly for $t\in[0, T]$.
This concludes the proof.
\end{proof}

Next, we consider similar Lyapunov functions but under $\ell_2$ norm.
This is because scaled NTK is not restricted to the dataset.
For $1 \le j \le 8$,
\begin{align*}
    \tilde{\Phi}_{L, j}(x, t) & = \frac{1}{(L-1) \CL^j}\sum^{L-1}_{l=1} \left(\ab{\fulz(x)}+\ab{\fult(x)-\fulz(x)}\right)^j\\
    \tilde{\Psi}_{L, 2}(x,t)& = \frac{1}{(L-1)^2}\sum^{ L}_{l=2} \frac{L-1}{l-1} \sum_{ i=2}^{l} \left(\ab{\W_l^i(x, 0)}+\ab{\W_l^i(x, t)-\W_l^i(x, 0)}\right)^2\\
    \tilde{\Psi}_{L,8}(x, t)& = \frac{1}{L-1} \sum_{l= 2}^L \left(\ab{\W_L^l(x,0)}+\ab{\W_L^l(x,t)-\W_L^l(x,0)}\right)^8,
\end{align*}
where $\tilde{\Phi}_{L, 0} =1$. 

Note that Lyapunov functions under $\ell_2$-norm are functions of $(x, t)$.
Similar to Lyapunov functions under $p^{in}$-norm, 
at initialization ($t=0$), we have $\tilde{\Phi}_{L,j}(x, 0)\convinp \tilde{\Phi}_{\infty, j}(x, 0), \, \tilde{\Psi}_{L,2}(x, 0) \convinp\tilde{\Psi}_{\infty, 2}(x, 0), \, \tilde{\Psi}_{L, 8}(x, 0) \convinp \tilde{\Psi}_{\infty, 8}(x, 0)$, where $\tilde{\Phi}_{L, \infty}(x, 0), \tilde{\Psi}_{\infty, 2}(x, 0), \tilde{\Psi}_{\infty, 8}(x, 0)$ are constant by the law of large number.
The similar proposition holds, which implies the intermediate weights and pre-activation values are invariant during training in $\ell_2$-norm sense.
% prp 2
\begin{proposition}\label{prp:l2-norm}
For $1 \le j \le 8$ and any $T>0$,
\[ \tilde{\Phi}_{L,j}(x, t) \convinp \tilde{\Phi}_{\infty, j}(x, 0),
\quad \tilde{\Psi}_{L,2}(x, t) \convinp \tilde{\Psi}_{\infty, 2}(x, 0), \quad \tilde{\Psi}_{L, 8}(x, t) \convinp \tilde{\Psi}_{\infty, 8}(x, 0)\]
uniformly  for $t \in [0,T]$ as $ L \rightarrow \infty$.
\end{proposition}
The following lemma, which corresponds to Lemma~\ref{lem:pin-bound}, allows us to apply Gr\"onwall's lemma.
% Lemma
\begin{lemma}\label{lem:l2-bound}
For $t \geq 0$ and $x\in\fR_+^{d_{in}}$, if $f^l_{\theta^{(l)}(t)}(x)$ is element-wise nonzero for $1 \le l \le L-1$, then
\begin{align*}
    \partial_t \tilde{\Psi}_{L,2}(x,t)&\le \frac{2(L-1)}{\CL} \para{\tilde{\Psi}_{L,2}(x,t)}^{2}\Phi_{L, 4}(t)\Psi_{L,8}(t)\EP{\dLt}\\
    \partial_t \tilde{\Phi}_{L, j}(x,t) &\le \frac{2jN^{1/2}(L-1)}{\CL}\tilde{\Phi}_{L, j-1}(t)\tilde{\Phi}_{L,8}(x,t)\tilde{\Psi}_{L, 2}(x,t)\Phi_{L,8}(t)\Psi_{L,8}(t)\EP{\dLt}\\
    \partial_t \tilde{\Psi}_{L, 8}(x,t)& \le \frac{8(L-1)}{\CL}\tilde{\Phi}_{L,4}(x,t) \tilde{\Psi}_{L,2}(x,t)\para{\tilde{\Psi}_{L,8}(x,t)}^{2}\Psi_{L,8}(t)\EP{\dLt},
\end{align*}
for $1\leq j\leq 8$.
\end{lemma}
The proof of Lemma~\ref{lem:l2-bound} is given in Section~\ref{app:prp2}.

\begin{proof}[Proof of Proposition~\ref{prp:l2-norm}]
Similar to the proof of Proposition~\ref{prp:pin-norm}, we need to show that 
$\fult(x)$ is element-wise nonzero for all $t\in[0, T]$, $1 \le l \le L-1$, and $x\in\fR_+^{d_{in}}$.
Again, the element-wise nonzero assumption also holds if $L$ is large enough, and the proof is given in Section~\ref{app:not crossing}.
From Lemma~\ref{lem:l2-bound},
\begin{align*}
\tilde{\Psi}_{L,2}(x, t) \le \tilde{\Psi}_{L,2}(x, 0) +\frac{2(L-1)}{\CL}\int^{t}_0{\para{\tilde{\Psi}_{L,2}(x, s)}^{2}\Phi_{L, 4}(s)\Psi_{L,8}(s)\EP{\delta^L_s}\,ds.}
\end{align*}
Then using Gr\"onwall's lemma where $z(t)=\tilde{\Psi}_{L,2}(x, t), \, a(t)=\tilde{\Psi}_{L,2}(x, 0), \, b(t)=\frac{2(L-1)}{\CL}, \, k(s)=\Phi_{L,4}(s)\Psi_{L,8}(s)\EP{\delta^L_s} $, we get
\[\tilde{\Psi}_{L, 2}(x, t) \le \tilde{\Psi}_{L, 2}(x, 0) \left\{ 1-\frac{2(L-1)}{\CL}\tilde{\Psi}_{L, 2}(x, 0) \int^t_0 \Phi_{L, 4}(s)\Psi_{L, 8}(s)\EP{\delta^L_s}\,ds  \right\}^{-1} \]
for all $0 \le t \le \beta_L$.
Similar to the proof of Proposition~\ref{prp:pin-norm}, if $L^2/\CL\rightarrow 0$, then
we get $\tilde{\Psi}_{L,2}(t) \convinp \tilde{\Psi}_{\infty,2}(0)$ uniformly for $t\in[0, T]$.

On the other hand, Lemma~\ref{lem:l2-bound} also implies
\begin{align*}
\sum^8_{j=1}\tilde{\Phi}_{L,j}(x, t) &\le \sum^8_{j=1}\tilde{\Phi}_{L,j}(x, 0) +\int^{t}_0{ \frac{16N^{1/2}(L-1)}{\CL} \para{ \sum^8_{j=1}\tilde{\Phi}_{L, j-1}(x, s)}\tilde{\Phi}_{L,8}(x, s)\tilde{\Psi}_{L, 2}(x, s)\Phi_{L,8}(s)\Psi_{L,8}(s)\EP{\delta^L_s}\,ds}\\
&\le \sum^8_{j=1}\tilde{\Phi}_{L,j}(x, 0) +\int^{t}_0{ \frac{16N^{1/2}(L-1)}{\CL} \para{ \sum^8_{j=1}\tilde{\Phi}_{L, j}(x, s)}^2\tilde{\Psi}_{L, 2}(x,s)\Phi_{L,8}(s)\Psi_{L,8}(s)\EP{\delta^L_s}\,ds}
\\
\tilde{\Psi}_{L,8}(x,t) &\le \tilde{\Psi}_{L,8}(x, 0) +\int^{t}_0{ \frac{8(L-1)}{\CL}\para{\tilde{\Psi}_{L,8}(x, s)}^{2}\tilde{\Phi}_{L,4}(x, s) \tilde{\Psi}_{L,2}(x, s)\Psi_{L,8}(s)\EP{\delta^L_s}\,ds}.
\end{align*}
We can similarly apply Gr\"onwall's lemma to show that $\sum_{j=1}^8 \tilde{\Phi}_{L, j}(x, t)$ and $\tilde{\Psi}_{L, 8}(x, t)$
converge to constant values.
\end{proof}

Now, we consider our last and core Lyapunov function which contains intermediate weight and pre-activation values from a single layer.
Define
\begin{align*}
    \tilde{\Psi}_L^{k}(x,t)& = \ab{\W_L^k(x,0)}+\ab{\W_L^k(x,t)-\W_L^k(x,0)}\\
    \tilde{\Phi}_L^{l}(x,t) & = \frac{1}{ \CL}\para{\ab{\fulz(x)}+\ab{\fult(x)-\fulz(x)}}
\end{align*}
for $1 \le l \le L$ and $2 \le k \le L$.

At initialization, $\tilde{\Psi}^k_L(x, 0)$ and $\tilde{\Phi}_L^l(x, 0)$ are stochastically bounded.
Then the following proposition implies the invariance of Lyapunov functions during training.
% prp 3
\begin{proposition}\label{prp:layerwise}
For $1 \le l \le L$, $2 \le k \le L$ and any $T>0$,
\[\tilde{\Psi}^{k}_L(x,t) \convinp \tilde{\Psi}^{k}_{\infty}(x,0) \quad \tilde{\Phi}^{l}_{L}(x,t) \convinp \tilde{\Phi}^{l}_{\infty}(x,0).\]
uniformly for $t\in[0, T]$ as $L \rightarrow \infty$.
\end{proposition}

This implies individual intermediate weights and pre-activation values are effectively unchanged at each layer. 
Recall that this layer-wise invariance is straightforward in an infinite width network with finite depth.
However, in our setting, an infinitely deep network has composition of infinitely many weights,
which requires much careful analysis including the following lemma as well as previous propositions.
% Lemma
\begin{lemma}\label{lem:layer-bound}
For $t \geq 0$ and $x\in\fR_+^{d_{in}}$, if $f^l_{\theta^{(l)}(t)}(x)$ is element-wise nonzero for $1 \le l \le L-1$, then
\begin{align*}
    \partial_t\tilde{\Psi}^{k}_L(x,t) &\le \frac{L-1}{\CL} \Psi_{L,8}(t)\tilde{\Phi}_{L,4}(x,t)\tilde{\Psi}_{L,2}(x,t)\tilde{\Psi}_{L,8}(x,t)\EP{\dLt}\\
    \partial_t\tilde{\Phi}^{l}_L(x,t) &\le \frac{2N^{1/2}(L-1)}{\CL}\Phi_{L,8}(t)\Psi_{L,8}(t)\tilde{\Phi}_{L,8}(x,t)\tilde{\Psi}_{L, 2}(x,t)\EP{\dLt},
\end{align*}
for $1 \le l \le L$ and $2 \le k \le L$.
\end{lemma}
The proof of Lemma~\ref{lem:layer-bound} is given in Section~\ref{app:prp3}.

\begin{proof}[Proof of Proposition~\ref{prp:layerwise}]
As we discussed in the proof of Proposition~\ref{prp:l2-norm}, if $L$ is large enough,
$\fult(x)$ is element-wise nonzero for all $t\in[0, T]$, $1 \le l \le L-1$, and $x\in\fR_+^{d_{in}}$.
Lemma~\ref{lem:layer-bound} implies
\begin{align*}
\tilde{\Psi}^{k}_L(x, t) &\le \tilde{\Psi}^{k}_L(x, 0) +\int^{t}_0{ \frac{L-1}{\CL} \Psi_{L,8}(t)\tilde{\Phi}_{L,4}(x,t)\tilde{\Psi}_{L,2}(x,t)\tilde{\Psi}_{L,8}(x,t)\EP{\dLt}\,ds}\\
\tilde{\Phi}^{l}_L(x, t)&\le \tilde{\Phi}^{l}_L(x, 0)+ \int_0^{t} \frac{2N^{1/2}(L-1)}{\CL}\Phi_{L,8}(t)\Psi_{L,8}(t)\tilde{\Phi}_{L,8}(x,t)\tilde{\Psi}_{L, 2}(x,t)\EP{\dLt}\,ds.
\end{align*}

Unlike previous proofs, we do not need Gr\"onwall's Lemma.
From Proposition~\ref{prp:pin-norm} and Proposition~\ref{prp:l2-norm}, all terms in RHS 
(such as $\Psi_{L, 8}(t), \tilde{\Phi}_{L, 4}(t), \tilde{\Psi}_{L, 2}(t)$, etc.) converge to constants in probability.
Since $L^2/\CL\rightarrow 0$, integral terms converge to zero as $L \rightarrow \infty$. 
On the other hand,
it is clear that $\tilde{\Psi}_L^k(x, t) \ge \tilde{\Psi}_L^k(x, 0)$ and $\tilde{\Phi}_L^l(x, t) \ge \tilde{\Phi}_L^l(x, 0)$ by construction. 
Thus, $\tilde{\Psi}_L^k(x, t)$ and $\tilde{\Phi}_L^l(x, t)$ converges to $\tilde{\Psi}_L^k(x, 0)$ and $\tilde{\Phi}_L^l(x, 0)$ in probability, respectively.
Since the integral terms are independent from the choice of $k$ and $l$, we get uniform convergence.
\end{proof}

Proposition~\ref{prp:layerwise} implies the following desired result 
which implies that the variation of intermediate pre-activation values and weights must be negligible.

\begin{corollary}\label{cor:finite}
For any $x$, as $L\rightarrow \infty$ , we have
\begin{align*}
    \sup_{1\leq l\leq L, t\in[0, T]} \frac{1}{ \CL}\ab{\fult(x)-\fulz(x)} \convinp& 0\\
    \sup_{2\leq k\leq L, t\in[0, T]} \ab{\W_L^{k}(x,t)-\W_L^{k}(x,0)} \convinp& 0.
\end{align*}
\end{corollary}
With this corollary, we are now ready to prove our main theorem, the invariance of scaled NTK.

\begingroup
\def\thetheorem{\ref{thm:2}}
\begin{theorem}[Invariance of scaled NTK] 
Let $T>0$.
Suppose $\int_0^T\EP{\delta\vert_{f_\theta^L}}\,dt$ is stochastically bounded as $L\rightarrow\infty$.
Then, for any $x, x'\in\fR_+^d$, 
\[\tilde{\Theta}^L_t(x,x')
\convinp
\tilde{\Theta}^{\infty}(x,x')
\]
uniformly for $t\in [0,T]$ as $L \rightarrow \infty$.
\end{theorem}
\addtocounter{theorem}{-1}
\endgroup

\begin{proof}[Proof of Theorem~\ref{thm:2}]
By definition,
\begin{align*}
    \tilde{\Theta}^L_t(x, x')
    &=\frac{1}{L}\sum_{l=0}^{L-1} \left[\left( \frac{1}{\CL}\sigma \left(\fult(x) \right)^{\intercal} \frac{1}{\CL}\sigma \left(\fult(x') \right)\right) \W_L^{l+2}(x,t)\W_L^{l+2}(x',t)^{\intercal}\right]\\
    &\quad+\frac{1}{L\CL^2}\sum_{l=0}^{L-1} \left[\W_L^{l+2}(x,t)\W_L^{l+2}(x',t)^{\intercal}\right].
\end{align*}
Informally, Proposition~\ref{prp:layerwise} implies that $f^l_{\theta^{(l)}(t)}(x)$ and $\W^{l+2}_L(x, t)$ are effectively invariant,
and therefore the kernel $\tilde{\Theta}_t^L(x, x')$ is invariant.
An additional effort is required to handle the summation of $L$ terms since $L$ also increases.
More formal proof is given in the following.

Since ReLU is 1-Lipshitz, Corollary~\ref{cor:finite} implies
\begin{align*}
    \sup_{1\leq l\leq L, t\in[0, T]}
    \ab{ \frac{1}{\CL}\sigma \para{\fult(x)}^{\intercal}\frac{1}{\CL}\sigma \para{\fult(x')}-\frac{1}{\CL}\sigma \para{\fulz(x)}^{\intercal}\frac{1}{\CL}\sigma \para{\fulz(x')}} \convinp& 0\\
    \sup_{2\leq k\leq L, t\in[0, T]} \ab{\W_L^k(t,x)\W_L^k(t,x')-\W_L^k(0,x)\W_L^k(0,x')} \convinp & 0
\end{align*}
as $L\rightarrow \infty$ for all $x$ and $x'$.
On the other hand, the norm of the difference between kernels is bounded by
\begin{align*}
    \ab{\tilde{\Theta}^L_t(x,x')- \tilde{\Theta}^L_0(x,x')} 
    & \le \frac{1}{L}\sum_{l=0}^L\Bigg\|\left( \frac{1}{\CL}\sigma \left(\fult(x) \right)^{\intercal} \frac{1}{\CL}\sigma \left(\fult(x') \right)\right) \W_L^{l+2}(x,t)\W_L^{l+2}(x',t)^{\intercal}\\
    &\quad\quad-\left( \frac{1}{\CL}\sigma \left(\fulz(x) \right)^{\intercal} \frac{1}{\CL}\sigma \left(\fulz(x') \right)\right) \W_L^{l+2}(x,0)\W_L^{l+2}(x',0)^{\intercal}\Bigg\|\\
    &\quad+\frac{1}{L\CL^2}\sum_{l=0}^{L-1}\ab{ \left[\W_L^{l+2}(x,t)\W_L^{l+2}(x',t)^{\intercal}\right]-\left[\W_L^{l+2}(x,0)\W_L^{l+2}(x',0)^{\intercal}\right]}\\
    & \le \max_l\Bigg\|\left( \frac{1}{\CL}\sigma \left(\fult(x) \right)^{\intercal} \frac{1}{\CL}\sigma \left(\fult(x') \right)\right) \W_L^{l+2}(x,t)\W_L^{l+2}(x',t)^{\intercal}\\
    &\quad\quad-\left( \frac{1}{\CL}\sigma \left(\fulz(x) \right)^{\intercal} \frac{1}{\CL}\sigma \left(\fulz(x') \right)\right) \W_L^{l+2}(x,0)\W_L^{l+2}(x',0)^{\intercal}\Bigg\|\\
    &\quad+\frac{1}{\CL^2}\max_l\ab{ \left[\W_L^{l+2}(x,t)\W_L^{l+2}(x',t)^{\intercal}\right]-\left[\W_L^{l+2}(x,0)\W_L^{l+2}(x',0)^{\intercal}\right]}.
\end{align*}
Thus, as $L\rightarrow \infty$, we have
\begin{align*}
   \sup_{t\in[0, T]} \ab{\tilde{\Theta}^L_t(x,x')- \tilde{\Theta}^L_0(x,x')} \convinp 0.
\end{align*}
Finally, Theorem~\ref{thm:1} implies $\tilde{\Theta}^L_0$ converges to $\tilde{\Theta}^{\infty}$ in probability,
and therefore $\tilde{\Theta}_t^L$ converges to $\tilde{\Theta}^{\infty}$ in probability.
\end{proof}

%%% Proof of Theorem 3
\subsection{Proof of Theorem~\ref{thm:3}}
Theorem~\ref{thm:3} implies that the solution, a fully-trained inifinite MLP, matches the kernel regression predictor $f_{\mathrm{ntk}}$. 

\begingroup
\def\thetheorem{\ref{thm:3}}
\begin{theorem}[Equivalence between deep MLP and kernel regression]
Let $\loss(f) = \frac{1}{2}\EP{f-f^\star}^2$. Let $\tilde{\Theta}^{\infty}$ be positive definite.
If $f_t$ follows the kernel gradient flow with respect to $\tilde{\Theta}^{\infty}$, then for any $x \in \fR^{d_{in}}_+$,
\[
\lim_{t \rightarrow \infty}f_t (x) = f_{\mathrm{ntk}}(x).
\]
\end{theorem}
\addtocounter{theorem}{-1}
\endgroup

\begin{proof}[Proof of Theorem~\ref{thm:3}]
Throughout the proof, we implicitly assume $d_{out}=1$ for the sake of simplicity.
Our proof conveys all important idea of the general proof, and the extension to general $d_{out}$ is straightforward.

The gradient flow is
\begin{align*}
 \partial_t f_t(x') &=-\Epin{\tilde{\Theta}^\infty(x', x)\para{f_t(x)-f^\star(x)} }.
\end{align*}
Define an linear operator $D:\cF\rightarrow \cF$ such that $D(g)(x') = \Epin{\tilde{\Theta}^{\infty}(x',x)g(x)}$.
Then, we have 
\begin{align*}
    &\partial_t f_t(x') =-D \para{f_t(x)-f^\star(x)}
\end{align*}
which has an explicit solution:
\begin{align*}
    &f_t(x')=f^\star(x')+e^{-tD}(f_0(x)-f^\star(x)),
\end{align*}
where $e^{-tD} = \sum_{k=0}^{\infty} \frac{(-t)^k}{k!}D^k$.
Then following lemma holds.

\begin{lemma}\label{lem:eigenfunctions}
Any function $f : \mathbb{R}_+^d \rightarrow \mathbb{R}$  is a linear combination of eigenfunctions of $D$.
More precisely, if we let $f(X) = (f(x_1), \dots, f(x_N)) \in \fR^N$, then $f$ can be expressed as
\begin{align*}
f(x) = (\tilde{\Theta}^{\infty}(x,x_1), \cdots, \tilde{\Theta}^{\infty}(x,x_N))K^{-1}f(X) + h(x),
\end{align*}
where $ (\tilde{\Theta}_0^{\infty}(x,x_1), \cdots, \tilde{\Theta}^{\infty}(x,x_N))K^{-1}f(X)$ 
is a linear combination of eigenfunctions of $D$ for nonzero eigenvalues,
and $h(x)$ is an eigenfunction for zero eigenvalue.
\end{lemma}
\begin{proof}
Let $K$ be an $N$ by $N$ matrix, where $K_{i, j} = \tilde{\Theta}^{\infty}(x_i, x_j)$.
Then, $K$ is positive definite, and $K$ has $N$ eigenvectors with nonzero eigenvalues.
Suppose $a=(a_1,\dots, a_N)\in\fR^N$ be an eigenvector of $K$ for eigenvalue $\lambda>0$, then
\begin{align*}
    D\left(\sum_{i=1}^N a_i \tilde{\Theta}^\infty(\cdot, x_i)\right)(x')
    =& \frac{1}{N}\sum_{j=1}^N \left(\tilde{\Theta}^\infty(x', x_j)\sum_{i=1}^N \tilde{\Theta}^{\infty}(x_j, x_i) a_i\right)\\
    =& \frac{1}{N}\sum_{j=1}^N \left(\tilde{\Theta}^\infty(x', x_j)\lambda a_j\right)\\
    =& \frac{\lambda}{N}\sum_{i=1}^N a_i \tilde{\Theta}^\infty(x', x_i).
\end{align*}
In other words, $\sum_{i=1}^N a_i\tilde{\Theta}^{\infty}(\cdot, x_i)$ is an eigenvector of $D$.
Since $K$ is positive definite, any linear combination of $\{\tilde{\Theta}^{\infty}(\cdot, x_i)\}_{i=1}^N$
is a linear combination of eigenfunctions of $D$ for nonzero eigenvalues,
including $(\tilde{\Theta}^{\infty}(x,x_1), \cdots, \tilde{\Theta}^{\infty}(x,x_N))K^{-1}f(X)$ for any $f:\fR^d_+\rightarrow \fR$.

On the other hand, for a given function $f$, 
let $h(x) =f(x)-(\tilde{\Theta}^{\infty}(x,x_1), \cdots, \tilde{\Theta}^{\infty}(x,x_N))K^{-1}f(X)$.
Since $h(x_i)=0$ for all $i$,
\begin{align*}
    D(h)(x') =  \frac{1}{N} \sum_{i=1}^N\tilde{\Theta}^{\infty}(x', x_i) h(x_i) = 0,
\end{align*}
and therefore $h$ is an eigenfunction of $D$ for zero eigenvalue.
Thus, we have
\begin{align*}
f(x) = (\tilde{\Theta}^{\infty}(x',x_1), \cdots, \tilde{\Theta}^{\infty}(x',x_N))K^{-1}f(X) + h(x),
\end{align*}
which is a linear combination of eigenfunctions of $D$.
\end{proof}

From Lemma~\ref{lem:eigenfunctions}, we have
\begin{align*}
f_0(x')-f^\star(x')  
% =&
% (\tilde{\Theta}^{\infty}(x',x_1), \cdots, \tilde{\Theta}^{\infty}(x',x_N))K^{-1}(f_0(X) - f^\star(X))\\
=&\sum a_ig_i(x') + h(x'),
\end{align*}
where $g_i$ are eigenfunctions of $D$ for nonzero eigenvalues and $h$ is an eigenfunction for zero eigenvalue.
Then,
\[f_t(x')=f^\star(x')+h(x')+\sum_i e^{-t\lambda_i}a_ig_i(x').\]
At initialization $t=0$,
\[f_0(x')=f^\star(x')+h(x')+\sum_i a_ig_i(x').\]
On the other hand, as $t \rightarrow \infty$, we have
\[\lim_{t\rightarrow \infty} f_t(x')=f^\star(x')+h(x').\]
By combining two equations, we have
\[\lim_{t\rightarrow \infty} f_t(x')=f_0(x')-\sum_i a_ig_i(x').\]

Finally, we have
\begin{align*}
\lim_{t\rightarrow \infty} f_t(x')
=&f_0(x') -(\tilde{\Theta}^{\infty}(x',x_1), \cdots, \tilde{\Theta}^{\infty}(x',x_N))K^{-1}(f_0(X)-f^\star(X))\\
=& (\tilde{\Theta}^{\infty}(x',x_1), \cdots, \tilde{\Theta}^{\infty}(x',x_N))K^{-1}f^\star(X),
\end{align*}
where the last equality is from $f_0(x')=0$ by initialization.
This concludes the proof.
\end{proof}

%%%%%%%%%%%%%%%%%%%%%%%%%%%%%%%%%%%%%%%%%%%%%%%%%%%%%%%%%%%%%%%%%
%%% Intermediate value does not cross zero during training
%%%%%%%%%%%%%%%%%%%%%%%%%%%%%%%%%%%%%%%%%%%%%%%%%%%%%%%%%%%%%%%%%
\section{Zero-crossing does not happen during training MLP}\label{app:not crossing}
While proving key statements in the proof of Theorem~\ref{thm:2} 
(Proposition~\ref{prp:pin-norm}, Proposition~\ref{prp:l2-norm}, and Proposition~\ref{prp:layerwise}),
we assumed that all elements of $\fult(x)$ are nonzero for $1 \le l \le L-1$, $x\in\{x_1, \dots, x_N\}$, and any input $x\in\fR_+^{d_{in}}$ while training.
In this section, for any $x', x''\in \fR_+^{d_{in}}$, 
we show that all entries of $\{\fult(x_1), \dots \fult(x_N), \fult(x'), \fult(x'')\}$ are nonzero with high probability while training.
For the sake of notational simplicity, we set $x_{N+1}=x'$ and $x_{N+2} = x''$ in the following lemma.

\begin{lemma}\label{lem:not crossing} 
Suppose $\int_0^T\EP{\delta\vert_{f_\theta^L}}\,dt$ is stochastically bounded as $L\rightarrow \infty$.
For any $T>0$, let 
\begin{align*}
    T_L = \bigwedge^{N+2}_{i=1} \inf_t \left\{\para{\fult(x_i)}_r=0 \mbox{ for some $1 \le l \le L-1$ and $r$}\right\} \bigwedge T,
\end{align*}
where $x_{N+1}, x_{N+2}\in\fR^{d_{in}}_+$ are arbitrary inputs.
Then, for any  $\epsilon>0$, there exists $L_{\max}$ such that $\Pr{T_L= T} \ge 1-\epsilon$ for all $L > L_{\max}$.
\end{lemma}
\begin{proof}[Proof of Lemma~\ref{lem:not crossing}]
For $\epsilon'>0$, let $M = 1+\epsilon'$.
Since $\int_0^T\EP{\delta\vert_{f_\theta^L}}\,dt$ is stochastically bounded as $L\rightarrow \infty$,
there exists $K_0>0$ and large enough $L_0>0$ such that
\begin{align*}
\Pr{\int_0^T\EP{\delta\vert_{f_\theta^L}}\,dt > K_0} < \frac{\epsilon}{3}
\end{align*}
for all $L\ge L_0$.
We define a complement of such event by
\begin{align*}
    E_0 =  \left\{\int_0^T\EP{\delta\vert_{f_\theta^L}}\,dt \leq K_0\right\},
\end{align*}
where $\Pr{E_0} \geq 1-\frac{\epsilon}{3}$.

On the other hand, recall that 
$\tilde{\Psi}_{\infty, 2}(x_i, 0), \tilde{\Psi}_{\infty, 8}(x, 0), \tilde{\Phi}_{\infty, j}(x_i, 0), 
\Gamma_{\infty}(0)$ are all (non-random) constants for all $1\leq j\leq 8$ and $1\leq i\leq N+2$.
Since $L^2/\CL\rightarrow 0$ as $L\rightarrow \infty$, we have
\begin{align*}
\frac{48N^{3/2}M^3(L-1)}{\CL}\Gamma_{\infty}(0)^{3}K_0\rightarrow& 0, \\
\frac{2M^5(L-1)}{C_{L}}\tilde{\Psi}_{\infty, 2}(x_i, 0) \Gamma^2_{\infty}(0)K_0 \rightarrow& 0, \\
\frac{M^716N^{1/2}(L-1)}{\CL} \para{ \sum^8_{j=1}\tilde{\Phi}_{\infty, j}(x_i, 0)}\tilde{\Psi}_{\infty, 2}(0)\Gamma^2_{\infty}(0)K_0 \rightarrow& 0,\\
\frac{M^78(L-1)}{\CL}\tilde{\Psi}_{\infty,8}(x_i, 0)\para{\sum^8_{j=1}\tilde{\Phi}_{\infty, j}(x_i, 0)} \tilde{\Psi}_{\infty,2}(x_i, 0)\Gamma_{\infty}(0)K_0\rightarrow& 0,
\end{align*}
as $L\rightarrow \infty$.
Thus, there exists $L_1$ such that 
\begin{align*}
\frac{48N^{3/2}M^3(L-1)}{\CL}\Gamma_{\infty}(0)^{3}K_0 <& 1-M^{-3}, \\
\frac{2M^5(L-1)}{C_{L}}\tilde{\Psi}_{\infty, 2}(x_i, 0) \Gamma^2_{\infty}(0)K_0 <& 1-M^{-1}, \\
\frac{M^716N^{1/2}(L-1)}{\CL} \para{ \sum^8_{j=1}\tilde{\Phi}_{\infty, j}(x_i, 0)}\tilde{\Psi}_{\infty, 2}(0)\Gamma^2_{\infty}(0)K_0 <& 1-M^{-1},\\
\frac{M^78(L-1)}{\CL}\tilde{\Psi}_{\infty,8}(x_i, 0)\para{\sum^8_{j=1}\tilde{\Phi}_{\infty, j}(x_i, 0)} \tilde{\Psi}_{\infty,2}(x_i, 0)\Gamma_{\infty}(0)K_0 <& 1-M^{-1}, 
\end{align*}
for all $L\ge L_1$.

Now, we define following events which indicate Lyapunov functions converge to constant values by the law of large numbers.
\begin{align*}
    E^{(L)}_{1} =& \left\{\Gamma_{L}(0) \leq M\Gamma_{\infty}(0)\right\} \\
    E^{(L)}_{2} =& \bigcup^{N+2}_{i=1}  \left\{\tilde{\Psi}_{L,8}(x_i,0)\leq M\tilde{\Psi}_{\infty,8}(x_i,0)\right\}\\
    E^{(L)}_{3} =& \bigcup^{N+2}_{i=1}  \left\{\tilde{\Psi}_{L,2}(x_i, 0) \leq M\tilde{\Psi}_{\infty,2}(x_i,0)\right\}\\
    E^{(L)}_{4} =& \bigcup^{N+2}_{i=1}\left\{\sum^8_{j=1}\tilde{\Phi}_{L,j}(x_i, 0) \leq M\sum^8_{j=1}\tilde{\Phi}_{\infty,j}(x_i, 0)\right\}.
\end{align*} 
By the law of large numbers, all events $E^{(L)}_{1}, E^{(L)}_{2}, E^{(L)}_{3}, E^{(L)}_{4}$ have probabilities converge to 1 as $L\rightarrow\infty$.
Thus, there exists $L_2$ such that 
\begin{align*}
    \Pr{E^{(L)}_{1}\cap E^{(L)}_{2}\cap E^{(L)}_{3}\cap E^{(L)}_{4}} \ge 1-\frac{\epsilon}{3}
\end{align*}
for all $L\ge L_2$.

Finally, we are interested in the event
\begin{align*}
E_5 = \bigcup^{N+2}_{i=1} \left\{\left|\para{f_{\theta(0)}^l(x_i)}_r\right| \geq \frac{L}{\CL}(K_1+1)\mbox{ for all $1 \le l \le L-1$ and $r$}\right\},
\end{align*} 
where we specify $K_1>0$ later.
Consider a random element of intermediate pre-activation values
\begin{align*}
g^1(x) =& \left(f^1_{\theta(0)}(x)\right)_{d_{in}+1} = \frac{1}{\CL}(u^1)^\intercal x+\frac{1}{\CL}v^1\\
g^l(x) =& \left(\fulz(x)\right)_{d_{in}+1} = (u^l)^\intercal x+\frac{1}{\CL}v^l \mbox{ for $2 \le l \le L-1$}.
\end{align*}
Then, $g^1 \sim \mathcal{N}(0, \sigma_1^2)$ and $g^l \sim \mathcal{N}(0, \sigma_g^2)$ for $2 \le l \le L-1$ 
are independent Gaussian random variables with finite variances.
Since
\begin{align*}
    \Pr{|g^l| < \frac{L}{\CL}(K+1)\mbox{ for some $1 \le l \le L-1$}}
    &\leq 2(K+1)\frac{L(L-2)}{\CL}\frac{1}{\sqrt{2\pi\sigma_g^2}}+2(K+1)\frac{L}{\CL}\frac{1}{\sqrt{2\pi\sigma_{1}^2}}
\end{align*}
by union bound, the probability converges to 0 as $L\rightarrow \infty$.
Thus, there exists $L_3$ such that $\Pr{E_5}\geq 1-\frac{\epsilon}{3}$ holds for all $L\ge L_3$.

Then, we would like to show the following claim.
% claim
\begin{claim}\label{claim1}
If $L\geq L_1$, and all events $E_0, E^{(L)}_1, E^{(L)}_2, E^{(L)}_3, E^{(L)}_4, E_5$ are given,
then all elements of intermediate pre-activation values never cross zeros while training for $t\in[0, T]$.
\end{claim}
If the claim holds, then for $L\geq L_{\max}=\max\{L_0, L_1, L_2, L_3\}$,
\begin{align*}
    \Pr{E_0\cap E^{(L)}_1\cap E^{(L)}_2\cap E^{(L)}_3\cap E^{(L)}_4\cap E_5}\geq 1-\epsilon,
\end{align*}
which concludes the proof of Lemma~\ref{lem:not crossing}.
In the rest of the proof, we will prove the above claim.
Suppose $L\geq L_1$ and all events $E_0, E^{(L)}_1, E^{(L)}_2, E^{(L)}_3, E^{(L)}_4, E_5$ are given.
Since all intermediate pre-activation values at initialization 
\begin{align*}
    f^1_{\theta^{(1)}(0)}(x_i) = \begin{bmatrix} \CL x_i\\ (u^{1})^{\intercal}x_i+v^1\\ \CL\mathbbm{1}_{d_{\mathrm{out}}} \end{bmatrix}, \quad \fulz(x_i) = \begin{bmatrix}
    \CL x_i\\
    \CL(u^l)^\intercal x_i+ v^l\\
    \CL\mathbbm{1}_{d_{\mathrm{out}}}
    \end{bmatrix}
\end{align*}
are nonzero for $2 \le l \le L-1$ and $1 \le i \le N+2$ with probability 1. 
Then, for all $t\in[0,T_L]$, Lemma~\ref{lem:pin-bound} holds which is essential in the proof of Proposition~\ref{prp:pin-norm}.
In the proof, using Gr\"onwall's lemma, we showed  
\[
\Gamma_L(t) \le \Gamma_L(0) \left\{ 1- \frac{48N^{3/2}(L-1)}{\CL}\Gamma_L(0)^{3} \int_0^{\beta_L} \EP{\delta^L_s}\,ds  \right\}^{-\frac{1}{3}}
\]
for $t \in \left[0, \beta^{(1)}_L \right]$, where 
\[
\beta^{(1)}_L = \mathrm{sup}\left\{t: \frac{48N^{3/2}(L-1)}{\CL}\Gamma_L(0)^{3} \int_0^{t} \EP{\delta^L_s}\, ds <1\right\}.
\]
Since $E^{(L)}_1$ is given, $ \Gamma_L(0) \le M \Gamma_{\infty}(0)$.
Then, since $E_0$ is given,
\begin{align*}
\frac{48N^{3/2}(L-1)}{\CL}\Gamma_{L}(0)^{3} \int_0^{t} \EP{\delta^L_s}\, ds  
\leq &\frac{48M^3N^{3/2}(L-1)}{\CL}\Gamma_{\infty}(0)^{3} K_0 \\
<& 1-M^{-3}
\end{align*} 
by definition of $L_1$.
This implies that $\beta_L^{(1)}=T_L$ and $\Gamma_L(t) \le M^2 \Gamma_{\infty}(0)$ if $E_0$ and $E^{(L)}_1$ are given and $L\geq L_1$.
Recall that $\Gamma_L(t)= \sum_{j=1}^8\Phi_{L,j}(t) + \Psi_{L,2}(t)+ \Psi_{L,8}(t)$,
and therefore $\Gamma_L(t)\leq M^2\Gamma_{\infty}(0)$ also implies 
$\Psi_{L,8}(t) \le M^2 \Gamma_{\infty}(0)$, $\Psi_{L,2}(t) \le M^2 \Gamma_{\infty}(0)$,
and $\Phi_{L,j}(t) \le M^2 \Gamma_{\infty}(0)$ for all $1 \le j \le 8$.

Similarly, in the proof of Proposition~\ref{prp:l2-norm}, we proved 
\[
\tilde{\Psi}_{L, 2}(x_i, t) \le \tilde{\Psi}_{L, 2}(x_i, 0) 
\left\{ 1-\frac{2(L-1)}{\CL}\tilde{\Psi}_{L, 2}(x_i, 0) \int^t_0 \Phi_{L, 4}(s)\Psi_{L, 4}(s)\EP{\delta^L_s}\,ds  \right\}^{-1} 
\]
for all $ t \in \left[0, \beta^{(2)}_L \right]$, where
\[
\beta^{(2)}_L 
= \mathrm{sup}\left\{t: \frac{2(L-1)}{\CL}\tilde{\Psi}_{L, 2}(x_i, 0) \int^t_0 \Phi_{L, 4}(s)\Psi_{L, 8}(s)\EP{\delta^L_s}\,ds <1\right\}.
\]
We have $\tilde{\Psi}_{L,8}(x_i,0) < M\tilde{\Psi}_{\infty,2}(x_i,0)$ if $E^{(L)}_3$ is given.
In this case, with a definition of $L_1$,
 \begin{align*}
     \frac{2(L-1)}{\CL}\tilde{\Psi}_{L, 2}(x_i, 0) \int^t_0 \Phi_{L, 4}(s)\Psi_{L, 8}(s)\EP{\delta^L_s}\,ds 
     & \le  \frac{2M^5(L-1)}{C_{L}}\tilde{\Psi}_{\infty, 2}(x_i, 0) \Gamma^2_{\infty}(0)K_0\\ 
     & <1-M^{-1}
 \end{align*}
 if $E_0$ is also given.
This implies that $\beta^{(2)}_L=T_L$ and  $\tilde{\Psi}_{L,2}(x_i, t) \le M^2 \tilde{\Psi}_{\infty,2}(x_i, 0)$,
if $E^{(L)}_3$ and $E_0$ are given and $L\geq L_1$.

In similar way,  we obtain $\sum^8_{j=1}\tilde{\Phi}_{\infty, j}(x_i, t) \le M^2 \sum^8_{j=1}\tilde{\Phi}_{\infty, j}(x_i, 0)$ for $t \in [0, T_L]$ 
if $E^{(L)}_4$ and $E_0$ are given and $L\geq L_1$.
Also, we have $\tilde{\Psi}_{L,8}(x_i, t) \le M^2 \tilde{\Psi}_{\infty,8}(x_i, 0)$ for $t \in [0, T_L]$ 
if $E^{(L)}_2$ and $E_0$ are given and $L\geq L_1$.

Finally, in the proof of Proposition~\ref{prp:layerwise}, we proved 
\begin{align*}
\tilde{\Phi}^{l}_L(x, t)&\le \tilde{\Phi}^{l}_L(x, 0)+ \int_0^{t} \frac{2N^{1/2}(L-1)}{\CL}\Phi_{L,8}(t)\Psi_{L,8}(t)\tilde{\Phi}_{L,8}(x,t)\tilde{\Psi}_{L, 2}(x,t)\EP{\dLt}\,ds.
\end{align*}
Then, if $E_0, E^{(L)}_1, E^{(L)}_2, E^{(L)}_3, E^{(L)}_4$ are given and $L\geq L_1$,
\begin{align*}
    &\int^{T_L}_0{ \frac{2N^{1/2}(L-1)}{\CL}\Phi_{L,8}(t)\Psi_{L,8}(t)\tilde{\Phi}_{L,8}(x,t)\tilde{\Psi}_{L, 2}(x,t)\EP{\dLt}\,ds} \\
    & \quad \le \frac{2N^{1/2}M^8(L-1)}{\CL} \Gamma^2_{\infty}(0)
 \left(\sum_{j=1}^8\tilde{\Phi}_{\infty, j}(x_i, 0)\right) \tilde{\Psi}_{\infty, 2}(x_i, 0)K_0. 
\end{align*} 
If we let $K_1=\sup_{1\le i\le N+2} 2N^{1/2}M^8 \Gamma^2_{\infty}(0)
 \left(\sum_{j=1}^8\tilde{\Phi}_{\infty, j}(x_i, 0)\right) \tilde{\Psi}_{\infty, 2}(x_i, 0)K_0$,
then this inequality implies
\[
\sup_{1\leq l\leq {L-1}, t\in[0, T_L]} \frac{1}{ \CL}\ab{\fult(x_i)-\fulz(x_i)} < \frac{L}{ \CL}K_1
\]
for all $i$.
On the other hand, if $E_5$ is given,
\[
\frac{1}{\CL}\left|\para{f_{\theta(0)}^l(x_i)}_r\right| > \frac{L}{\CL}(K_1+1)
\]
for all $i, l$ and $r$, and therefore
\[
\inf_{1\leq l\leq L-1, t\in[0, T_L]} \left|\para{\fult(x_i)}_r\right| > L
\]
for all $i$ and $r$.
This implies $T_L=T$, which concludes the proof of Claim~\ref{claim1}.
\end{proof}

% prove theorem 2 again
Equipped with Lemma~\ref{lem:not crossing}, we provide a brief outline of the rigorous proof of Theorem~\ref{thm:2}.
For any $T>0$, let 
\begin{align*}
    T_L = \bigwedge^{N+2}_{i=1} \inf_t \left\{\para{\fult(x_i)}_r=0 \mbox{ for some $1 \le l \le L-1$ and $r$}\right\} \bigwedge T,
\end{align*}
where $x_{N+1}, x_{N+2}\in\fR^{d_{in}}_+$ are arbitrary inputs.
Also, for $\epsilon, \epsilon'>0$, where $M=1+\epsilon'$, we can define events $E_0, E^{(L)}_1, E^{(L)}_2, E^{(L)}_3, E^{(L)}_4, E_5$
and constants $K_0, L_0, L_1, L_2, L_3$ as defined in the proof of Lemma~\ref{lem:not crossing}.
We further let 
\begin{align*}
    K_2 =  \max\left\{K_1, \sup_{1\leq i\leq N+2}M^8\Gamma_{\infty}(0)\para{\sum^8_{j=1}\tilde{\Phi}_{\infty, j}(x_i, 0)}\tilde{\Psi}_{\infty,2}(x_i,0)\tilde{\Psi}_{\infty,8}(x_i,0) K_0\right\}.
\end{align*}
and
\begin{align*}
    E_6 = 
    & \Bigg\{\frac{1}{L}\sum_l\left( 1 + \frac{L}{C_L}K_2+\Bigg\| \frac{1}{\CL}\sigma \left(\fulz(x) \right)\Bigg\|\right)
    \left( 1 + \frac{L}{C_L}K_2+\Bigg\| \frac{1}{\CL}\sigma \left(\fulz(x') \right)\Bigg\|\right)\\
    &\quad\quad\times  \left( 1 + \frac{L}{C_L}K_2+\Bigg\| \W_L^{l+2}(x, 0)\Bigg\|\right) \left( 1 + \frac{L}{C_L}K_2+\Bigg\| \W_L^{l+2}(x', 0)\Bigg\|\right)
    \times \frac{L}{C_L}K_2\\
    &\quad+\frac{1}{L\CL^2}\sum_l\left(1 + \frac{L}{\CL}K_2+ \ab{\W_L^{l+2}(x, 0)}\right) 
        \left(1 + \frac{L}{\CL}K_2+ \ab{\W_L^{l+2}(x', 0)}\right)\times\frac{L}{C_L}K_2  < \epsilon''\Bigg\}
\end{align*}
for $\epsilon''>0$.
Since $L^2/\CL\rightarrow 0$ as $L\rightarrow\infty$, the probability of $E^{(L)}_6$ converges to  $1$ by the law of large numbers,
and there exists $L_4$ such that
\begin{align*}
    \Pr{E^{(L)}_6} > 1-\epsilon
\end{align*}
for all $L\ge L_4$.
Suppose all events $E_0, E^{(L)}_1, E^{(L)}_2, E^{(L)}_3, E^{(L)}_4, E_5, E^{(L)}_6$ are given, and $L\geq L_{\max}=\max\{L_0, L_1, L_2, L_3, L_4\}$.
Then, instead of Proposition~\ref{prp:pin-norm}, the following inequalities hold without having $L\rightarrow\infty$:
\begin{align*}
    \Phi_{L, j}(t) & \leq M^2 \Gamma_{\infty}(0)\\
    \Psi_{L, 2}(t) & \leq M^2 \Gamma_{\infty}(0)\\
    \Psi_{L, 8}(t) & \leq M^2 \Gamma_{\infty}(0)
\end{align*}
for all $1\leq j\leq 8$ and $t\in[0, T_L]$ since $\beta_L^{(1)}=T_L$.
Similarly, we can rewrite Proposition~\ref{prp:l2-norm} without having $L\rightarrow\infty$:
\begin{align*}
    \sum_{j=1}^8\tilde{\Phi}_{\infty, j}(x_i, 0)\leq& \sum_{j=1}^8\tilde{\Phi}_{L, j}(x_i, t) \leq M^2 \sum_{j=1}^8\tilde{\Phi}_{\infty, j}(x_i, 0)\\
    \tilde{\Psi}_{\infty, 2}(x_i, 0)\leq& \tilde{\Psi}_{L, 2}(x_i, t) \leq M^2 \tilde{\Psi}_{\infty, 2}(x_i, 0)\\
    \tilde{\Psi}_{\infty, 8}(x_i, 0)\leq& \tilde{\Psi}_{L, 8}(x_i, t) \leq M^2 \tilde{\Psi}_{\infty, 8}(x_i, 0)
\end{align*}
for all $1\leq i\leq N+2$ and $t\in[0, T_L]$.
Finally, Corollary~\ref{cor:finite} can be replaced by
\begin{align*}
    \sup_{1\leq l\leq L-1, t\in[0, T_L]} \frac{1}{\CL}\ab{\fult(x_i)-\fulz(x_i)} < \frac{L}{\CL}K_1
\end{align*}
for all $i$, where $K_1$ is as defined in the proof of Lemma~\ref{lem:not crossing}.
This already implies $T_L=T$.

In addition, by Lemma~\ref{lem:layer-bound},
\begin{align*}
    \sup_{1\leq l\leq L-1, t\in[0, T]} \ab{\W^k_L(x_i, t)-\W^k_L(x_i, 0)} < \frac{L}{C_L}K_2.
\end{align*}
Thus, for $x, x'\in\{x_1, \dots, x_{N+2}\}$, we can bound the deviation of $\tilde{\Theta}^L_t(x, x')$ by
\begin{align*}
    \ab{\tilde{\Theta}^L_t(x,x')- \tilde{\Theta}^L_0(x,x')} 
    & \le \frac{1}{L} \sum_l\Bigg\|\left( \frac{1}{\CL}\sigma \left(\fult(x) \right)^{\intercal} 
        \frac{1}{\CL}\sigma \left(\fult(x') \right)\right) \W_L^{l+2}(x,t)\W_L^{l+2}(x',t)^{\intercal}\\
        &\quad\quad-\left( \frac{1}{\CL}\sigma \left(\fulz(x) \right)^{\intercal} \frac{1}{\CL}\sigma \left(\fulz(x') \right)\right) \W_L^{l+2}(x,0)\W_L^{l+2}(x',0)^{\intercal}\Bigg\|\\
        &\quad+\frac{1}{L\CL^2}\sum_l \ab{ \left[\W_L^{l+2}(x,t)\W_L^{l+2}(x',t)^{\intercal}\right]-\left[\W_L^{l+2}(x,0)\W_L^{l+2}(x',0)^{\intercal}\right]}\\
    & \le \frac{1}{L}\sum_l\Bigg[\left( \frac{L}{C_L}K_2+\Bigg\| \frac{1}{\CL}\sigma \left(\fulz(x) \right)\Bigg\|\right)
        \left(  \frac{L}{C_L}K_2+\Bigg\| \frac{1}{\CL}\sigma \left(\fulz(x') \right)\Bigg\|\right)\\
        &\quad\quad\quad\times  \left(\frac{L}{C_L}K_2+\Bigg\| \W_L^{l+2}(x, 0)\Bigg\|\right) \left(\frac{L}{C_L}K_2+\Bigg\| \W_L^{l+2}(x', 0)\Bigg\|\right)\\
        &\quad\quad\quad-\ab{\frac{1}{\CL}\sigma\left(\fulz(x)\right)} \ab{\frac{1}{\CL}\sigma\left(\fulz(x')\right)} 
            \ab{\W_L^{l+2}(x, 0)} \ab{\W_L^{l+2}(x', 0)} \Bigg]\\
        &\quad+\frac{1}{L\CL^2}\sum_l\Bigg[\left(\frac{L}{\CL}K_2+ \ab{\W_L^{l+2}(x, 0)}\right) \left(\frac{L}{\CL}K_2+ \ab{\W_L^{l+2}(x', 0)}\right) \\
        &\quad\quad\quad\quad -  \ab{\W_L^{l+2}(x, 0)} \ab{\W_L^{l+2}(x', 0)} \Bigg]\\
    & \le \frac{1}{L}\sum_l\Bigg[\left( 1 + \frac{L}{C_L}K_2+\Bigg\| \frac{1}{\CL}\sigma \left(\fulz(x) \right)\Bigg\|\right)
        \left( 1 + \frac{L}{C_L}K_2+\Bigg\| \frac{1}{\CL}\sigma \left(\fulz(x') \right)\Bigg\|\right)\\
        &\quad\quad\quad\times  \left( 1 + \frac{L}{C_L}K_2+\Bigg\| \W_L^{l+2}(x, 0)\Bigg\|\right) \left( 1 + \frac{L}{C_L}K_2+\Bigg\| \W_L^{l+2}(x', 0)\Bigg\|\right)\Bigg] \times \frac{L}{C_L}K_2\\
        &\quad+\frac{1}{L\CL^2}\sum_l\left(1 + \frac{L}{\CL}K_2+ \ab{\W_L^{l+2}(x, 0)}\right) 
            \left(1 + \frac{L}{\CL}K_2+ \ab{\W_L^{l+2}(x', 0)}\right)\times\frac{L}{C_L}K_2\\
    &\leq \epsilon''
\end{align*}
Finally,
\begin{align*}
    \Pr{ \ab{\tilde{\Theta}^L_t(x,x')- \tilde{\Theta}^L_0(x,x')} < \epsilon''} \ge 1-2\epsilon
\end{align*}
for all $L\ge L_{\max}'$.
Since $\epsilon, \epsilon'$, and $\epsilon''$ are arbitrary,
we can conclude $\tilde{\Theta}_t^L(x, x')\convinp \tilde{\Theta}^{\infty}(x, x')$.

%%%%%%%%%%%%%%%%%%%%%%%%%%%%%%%%%%%%%%%%%%%%%%%%%%%
%%% Stochastically bounded assumption of Theorem 2
%%%%%%%%%%%%%%%%%%%%%%%%%%%%%%%%%%%%%%%%%%%%%%%%%%%
\section{Stochastically bounded assumption of Theorem~\ref{thm:2}}
Since the scaled NTK stays constant during training by Theorem~\ref{thm:2}, 
we can find an explicit solution to the differential equation when the loss function is quadratic loss.
However, Theorem~\ref{thm:2} requires stochastically bounded assumption of $\int_0^T \EP{\delta_t^L}\,dt$,
which is provided by the following lemma.
\begin{lemma}\label{lemm:stochastically bound}
If $\loss(f)=\EP{f-f^\star}^2$, then $\int_0^T \EP{\dLt}\, dt$ is stochastically bounded for all $T>0$.
\end{lemma}
\begin{proof}[Proof of Lemma~\ref{lemm:stochastically bound}]
Since $\Theta^{L}(x,x')$ is semidefinite by definition,
\begin{align*}
    \partial_t\loss\vert_{f_t} &= -\Epin{\delta\vert_{f_t}(x)\partial_t f_t(x)} \\&=-\mathbb{E}_{x,x'\sim p^{in}}\left[\delta\vert_{f_t}(x)^\intercal \Theta^{L}(x,x') \delta\vert_{f_t}(x')\right] \le 0.
\end{align*}
This implies $\EP{f-f^\star}^2$ is non increasing during training. Then $\EP{\dLt}=\EP{f-f^\star}$ is also non increasing, and therefore $\int_0^T \EP{\dLt}\, dt$ is stochastically bounded for all $T>0$.
\end{proof}

\newpage
%%%%%%%%%%%%%%%%%%%%%%%%%%%%%%%%%%%%%%%%%%%%%%%%%%%%%%%%%
%%% Equalites and inequalities of sub layers
%%%%%%%%%%%%%%%%%%%%%%%%%%%%%%%%%%%%%%%%%%%%%%%%%%%%%%%%%
\section{Equalites and inequalities of sub layers for MLP}\label{sec:ineq of MLP}
In this section, we introduce some useful equalities and inequalities in our problem setting.

First, we prove some property of $\EP{\cdot}$.

\begin{lemma}
$\EP{\cdot}$ is seminorm for both matrix and vector. 
\end{lemma}
\begin{proof}
Since it is clear that $\EP{\cdot}$ satisfies absolute homogeneity and nonnegativity,
it is enough to show the triangle inequality (which suffices to prove for matrix only).
\begin{align*}
    \EP{A(x)+B(x)} &= \sqrt{\frac{1}{N}\sum_i\|A(x_i)+B(x_i)\|^2} \\
    &\le \sqrt{\frac{1}{N}\sum_i(\|A(x_i)\|+\|B(x_i)\|)^2}\\
    &\le \sqrt{\frac{1}{N}\sum_i\|A(x_i)\|^2}+\sqrt{\frac{1}{N}\sum_i\|B(x_i)\|^2}\\
    & = \EP{A(x)}+\EP{B(x)}
\end{align*}
by Cauchy inequality.
\end{proof}

\begin{lemma}
\[\EP{A(x)B(x)} \le \sqrt{N}\EP{A(x)}\EP{B(x)}\]
\end{lemma}
\begin{proof}
\begin{align*}
    \EP{A(x)B(x)} &= \sqrt{\frac{1}{N}\sum_i\|A(x_i)B(x_i)\|^2} \\
    &\le \sqrt{\frac{1}{N}\sum_i\|A(x_i)\|^2 \max_i \|B(x_i)\|^2}\\
    &\le \sqrt{N}\sqrt{\frac{1}{N}\sum_i\|A(x_i)\|^2}\sqrt{\frac{1}{N}\sum_i\|B(x_i)\|^2}\\
    & = \sqrt{N}\EP{A(x)}\EP{B(x)}
\end{align*}
by the property of $\ell_2$ norm. 
\end{proof}

Note that if $W$ does not depend on $x$, then $\EP{W} = \|W\|$.

For simplicity, we use $f^l\equiv f^l_{\theta^{(l)}}$ if it is clear from the context. As previously mentioned, we set the scaling factor of gradient flow by $\frac{1}{(L-1)\CL^2}$ throughout the proof. Hence the gradient flow is given by
\begin{align*}
    \partial_t \theta^l &=  -\frac{1}{(L-1)\CL^2}\mathbb{E}_{x \sim p^{in}}[(\partial_{\theta^l} \fult(x))^{\intercal}\delta^l_t],
\end{align*}
where $\delta^l_t =  \para{\W^{l+1}_L(x,t)}^{\intercal}\delta^L_t$.
This is because
\begin{align*}
\partial_{\theta^l} f^{L}= \para{\prod_{j=L}^{l+1} W^j \diag(\dot{\sigma}(f^{j-1}))}\partial_{\theta^l} f^{l} \quad \text{for} \, L \ge l.
\end{align*}
We use inverted indexing for $\prod$ to emphasize the order of product, i.e.,
\begin{align*}
\prod_{j=L}^{l+1} W^j \diag(\dot{\sigma}(f^{j-1})) = W^L \diag(\dot{\sigma}(f^{L-1})) \cdots W^{l+1} \diag(\dot{\sigma}(f^{l}))
\end{align*}
Then, by the chain rule, we obtain the following lemma.
\begin{lemma} \label{lem:partial derivative}
For $t\ge 0$ and $1\leq l\leq L$, 
\[\partial_t W^l = -\frac{1}{(L-1)\CL^2}\Epin{ \delta_t^l\para{\sigma(f_{\theta^{(l-1)}}^{l-1}(x))^{\intercal}}},
\quad \partial_t b^l = -\frac{1}{(L-1)\CL^2}\Epin{\delta_t^l}.\]
\end{lemma}

Then, the following lemma provides a bound of derivatives.
\begin{lemma}\label{lem:bound weight derivatives}
For any $t\ge 0$ and $1\leq l\leq L$, 
\[\EP{\partial_t W^l} \le \frac{1}{(L-1)\CL^2}\EP{ \delta^l_t}\EP{f^{l-1}},
\quad 
\EP{\partial_t b^l} \le \frac{1}{(L-1)\CL^2}\EP{ \delta^l_t}.
\]
\end{lemma}
\begin{proof}[Proof of Lemma~\ref{lem:bound weight derivatives}]
From Lemma~\ref{lem:partial derivative},
\begin{align*}
\EP{\partial_t W^l}&=\|\partial_t W^l\|\\
&=  \ab{\frac{1}{(L-1)\CL^2}\Epin{\delta^l_t \sigma(f^{l-1})^{\intercal}}}\\
& \le \frac{1}{(L-1)\CL^2}\Epin{ \|(\delta^l_t \sigma(f^{l-1})^{\intercal}\|}\\
& \le \frac{1}{(L-1)\CL^2}\Epin{\|\delta^l_t\| \|\sigma(f^{l-1})\| }\\
&\le \frac{1}{(L-1)\CL^2}\sqrt{\Epin{\|\delta^l_t\|^2}}\sqrt{\Epin{\|\sigma(f^{l-1})\| ^2}}\\
&= \frac{1}{(L-1)\CL^2}\EP{\delta^l_t}\EP{\sigma(f^{l-1})}\\
&\le \frac{1}{(L-1)\CL^2}\EP{\delta^l_t}\EP{f^{l-1}},
\end{align*}
where the first inequality is from Jensen's inequality and third inequality comes from Cauchy's inequality.
Finally, we can bound $\EP{\partial_t b^l}$ similarly.
\end{proof}

The gradient flow also implies
\begin{align*}
    \partial_t \fult(x') &=-\frac{1}{(L-1)\CL^2}\mathbb{E}_{x \sim p^{ in}}[ \partial_{\theta^l}\fult(x') (\partial_{\theta^l} \fult(x))^{\intercal} \delta^l_t(x)],
\end{align*}
where the following lemma provides a bound of the norm.
\begin{lemma}\label{lem:bound derivative of intermediate layer}
For $1\leq l\leq L$ and $t\ge 0$,
\[\EP{\partial_t f^l(x')} \le \frac{N}{(L-1)\CL^2}\EE{\left\|\partial_{\theta} f_{\theta}^l(x')  \para{\partial_{\theta} f^l_{\theta}(x)}^{\intercal} \delta^l_t(x)\right\|}.\]
\end{lemma}
\begin{proof}[Proof of Lemma~\ref{lem:bound derivative of intermediate layer}]
From previous gradient flow,
\begin{align*}
\EP{\partial_t f^l(x')} 
&= \sqrt{\epin{\|\partial_{{\theta^l}} f_{\theta^l}^l(x')\frac{1}{(L-1)\CL^2}\Epin { (\partial_{\theta^l} f_{\theta^l}^{l}(x))^{\intercal} \delta^l_t(x)}\|^2}}\\
&=  \frac{1}{(L-1)\CL^2}\sqrt{\epin{\left\|\Epin{\partial_{\theta^l} f_{\theta^l}^l(x') (\partial_{\theta^l} f_{\theta^l}^{l}(x))^{\intercal} \delta^l_t(x)}\right\|^2}}\\
&\le \frac{1}{(L-1)\CL^2}\sqrt{\frac{1}{N}\epin{\sum_i\left\|\partial f^l_{\theta^l}(x')(\partial_{\theta^l} f^{l}(x_i))^{\intercal} \delta^l_t(x)\right\|^2}}\\
&\le \frac{1}{(L-1)\CL^2}\sqrt{\frac{1}{N^2}\sum_j\sum_i\left\|\partial f^l_{\theta^l}(x_j')(\partial_{\theta^l} f^{l}(x_i))^{\intercal} \delta^l_t(x_i)\right\|^2}\\
&\le \frac{1}{(L-1)\CL^2}\frac{1}{N}\sum_j\sum_i\left\|\partial f^l_{\theta^l}(x_j')(\partial_{\theta^l} f^{l}(x_i))^{\intercal} \delta^l_t(x_i)\right\|\\
&\le \frac{N}{(L-1)\CL^2}\EE{\left\|\partial_{\theta^l} f^l(x') (\partial_{\theta^l} f^{l}(x))^{\intercal} \delta^l_t(x)\right\|}.
\end{align*}
\end{proof}

The following is the key lemma to prove invariances of Lyapunov functions.
\begin{lemma}\label{lem:bound derivative of pin norm}
Suppose $f^i_{\theta^{(i)}(t)}(x')$ is element-wise nonzero for $1 \le i \le L-1$ and $x'$ in dataset. 
Then, for $t \in [0,T]$, $1\leq k \le j\le L$, and $1\leq i \leq L$, 
\begin{align*}
    \partial_t \EP{\W^k_j(x,t)} 
    &\le \frac{N}{(L-1)\CL^2}\sum^j_{i = k}\EP{\W^{i+1}_{j}(x,t)}\EP{\W^k_{i-1}(x,t)}\EP{\dit} \EP{\sigma\para{f^{i-1}_{\theta^{(i-1)}(t)}(x)}}\\
    \partial_t \EP{f^i_{\theta^{(i)}(t)}(x')} 
    &\le \frac{N^{3/2}}{(L-1)\CL^{2}} \sum_{l=0}^{i-1} \EP{ f^{l}_{\theta^{(l)}(t)}(x)}\EP{ f^{l}_{\theta^{(l)}(t)}(x')}\EP{\W_i^{l+2}}\EP{(\W_L^{l+2})^{\intercal}} \EP{\dLt}\\
    &\quad +\frac{N}{(L-1)\CL^{2}}  \sum_{l=0}^{i-1}\EP{\W_i^{l+2}}\EP{(\W_L^{l+2})^{\intercal}} \EP{\dLt}.
  \end{align*}
\end{lemma}
\begin{proof}[Proof of Lemma~\ref{lem:bound derivative of pin norm}]
By the chain rule,
\begin{align*}
    \partial_t \EP{\W^k_j(x,t)}
    &\le \EP{\partial_t \prod_{l=j}^{k} W^l(t) \diag\left(\dot{\sigma}\left(f^{l - 1}_{\theta^{(l - 1)}(t)}(x)\right)\right)}\\
    &\le  \sum^j_{i = k} \EP{\W^{i + 1}_j \partial_t W^i(t)\diag(\dot{\sigma}(f^{i - 1}_{\theta^{(i - 1)}(t)}(x))) \W_{i - 1}^k },
 %     +\sum^j_{i = k} \EP{\W^{i+1}_j W^i(t)\textcolor{red}{\partial_t\diag(\dot{\sigma}(f^{i}_{\theta^{(i)}(t)}(x))}\W_{i-1}^k}
\end{align*}
where $\W^{i+1}_j$ denotes $\W^{i+1}_j(x, t)$ if it is clear from the context.
Then, 
\begin{align*}
    \partial_t \EP{\W^k_j(x,t)}
    &\le  \sum^j_{i = k} \EP{\W_j^{i+1}\frac{1}{(L-1)\CL^2}\Epin{ \dit \sigma(f^{i-1}_{\theta^{(i-1)}(t)}(x))^{\intercal}}\diag(\dot{\sigma}(f^{i - 1}_{\theta^{(i - 1)}(t)}(x))) \W_{i-1}^k}\\
    &\le \frac{N}{(L-1)\CL^2}\sum^j_{i = k} \EP{\W_j^{i+1}} \ab{\Epin{ \dit \sigma(f^{i-1}_{\theta^{(i-1)}(t)}(x))^{\intercal}}}
    \EP{\diag(\dot{\sigma}(f^{i - 1}_{\theta^{(i - 1)}(t)}(x))) }\EP{\W_{i-1}^k} \\
    & \le \frac{N}{(L-1)\CL^2}\sum^j_{i = k}\EP{\W^{i+1}_{j}(x,t)}\EP{\W^k_{i-1}(x,t)}\EP{\dit} \EP{\sigma(f^{i-1}_{\theta^{(i-1)}(t)}(x)},
  \end{align*}
where first inequality comes from Lemma~\ref{lem:bound weight derivatives}.

On the other hand, from Lemma~\ref{lem:bound derivative of intermediate layer} and Corollary~\ref{cor:recursive},
\begin{align*}
      &\EP{\partial_t  \fult(x)}    \\& \le \frac{N}{(L-1)\CL^2}  \mathbb{E}_{x, x' \sim p^{in}}\left[\left\|\partial_{\theta^l} \fult(x') \para{\partial_{\theta^l} \fult(x)}^{\intercal} \left(\W_L^{l+1}(x)\right)^{\intercal} \dLt(x)\right\|\right]\\
    &\le\frac{N}{(L-1)\CL^2}  \mathbb{E}_{x, x' \sim p^{in}}\bigg[\sum_{i=0}^{l-1} \bigg\|( \sigma(f^i_{\theta^{(i)}(t)}(x'))^{\intercal}\sigma(f^i_{\theta^{(i)}(t)}(x))+1)\W_l^{i+2}(x')(\W_l^{i+2}(x))^{\intercal}(\W_L^{l+1}(x))^{\intercal} \dLt(x)\bigg\|\bigg]\\
    &\le\frac{N}{(L-1)\CL^2}  \sum_{i=0}^{l-1}\mathbb{E}_{x, x' \sim p^{in}} \bigg\| \sigma(f^i(x'))^{\intercal}\sigma(f^i(x))\W_l^{i+2}(x')(\W_L^{i+2}(x))^{\intercal}\dLt(x)\bigg\|\\
    &\quad+\frac{N}{(L-1)\CL^2} \sum_{i=0}^{l-1}\EE{\left\|\W_l^{i+2}(x')(\W_L^{i+2}(x))^{\intercal} \dLt(x)\right\|}\\
    &\le\frac{N}{(L-1)\CL^2}  \sum_{i=0}^{l-1}\mathbb{E}_{x, x' \sim p^{in}} \big[\|( \sigma(f^i(x))\|\|\sigma(f^i(x'))\|\|\W_l^{i+2}(x')\|\|(\W_L^{i+2}(x))^{\intercal}\| \|\dLt(x)\|\big]\\
    &\quad+\frac{N}{(L-1)\CL^2}  \sum_{i=0}^{l-1}\EE{\left\|\W_l^{i+2}(x')\|\|(\W_L^{i+2}(x))^{\intercal}\| \|\dLt(x)\right\|}\\
    &\le\frac{N^{3/2}}{(L-1)\CL^2} \sum_{i=0}^{l-1} \EP{ \sigma(f^i(x))}\EP{ \sigma(f^i(x'))}\EP{\W_l^{i+2}}\EP{(\W_L^{i+2})^{\intercal}} \EP{\dLt(x)}\\
    &\quad+\frac{N}{(L-1)\CL^2}  \sum_{i=0}^{l-1}\EP{\W_l^{i+2}}\EP{(\W_L^{i+2})^{\intercal}} \EP{\dLt(x)},
\end{align*}
where the last inequality holds due to the property of $p^{in}$-norm
\begin{align*}
\Epin{\|( \sigma(f^i(x))\| \|(\W_L^{i+2})^{\intercal}\| \|\dLt(x)\|}
    \le \sqrt{N}\EP{( \sigma(f^i(x))}\EP{(\W_L^{i+2}(x))^{\intercal}} \EP{\dLt(x)}.
\end{align*}
\end{proof}

We have a similar bounds for derivatives of $\ell_2$-norms.
\begin{lemma}\label{lem:bound derivative of l2 norm}
For $x'\in\fR_+^{d_{in}}$, if all elements of $f^i_{\theta^{(i)}(t)}(x')$ are nonzero for $1 \le i \le L-1$, 
then, for $t \in [0,T]$, $1\leq l \le j\le L$, and $1\leq i \leq L$
\begin{align*}
    \partial_t \ab{\W^k_l(x',t)} 
    & \le \frac{1}{(L-1)\CL^2}\sum_{l \ge i \ge k} \ab{\W^{i+1}_{l}(x')}\ab{\W^k_{i-1}(x')}\EP{\delta^i_{t}} \EP{\sigma(f^{i-1})^{\intercal}}\\
    \partial_t \ab{f^i_{\theta^{(l)}(t)}(x')} 
    &\le\frac{N^{1/2}}{(L-1)\CL^2} \sum_{i=0}^{l-1} \EP{ \sigma(f^i(x))}\ab{ \sigma(f^i(x'))}\ab{\W_l^{i+2}(x')}\EP{(\W_L^{i+2})^{\intercal}} \EP{\delta_t^L(x)}\\
    &\quad +\frac{1}{(L-1)\CL^2}  \sum_{i=0}^{l-1}\ab{\W_l^{i+2}(x')}\EP{(\W_L^{i+2})^{\intercal}} \EP{\delta_t^L(x)}.
  \end{align*}
\end{lemma}

\begin{proof}[Proof of Lemma~\ref{lem:bound derivative of l2 norm}]
The proof is similar to that of Lemma~\ref{lem:bound derivative of pin norm}.
\begin{align*}
    \partial_t \ab{\W^k_l} &\le \ab{\partial_t \W_l^k}\\
    &\le  \sum_{l \ge i \ge k} \ab{\W_l^{i+1}\frac{1}{(L-1)\CL^2}\Epin{ \dit \sigma(f^{i-1})^{\intercal}}\diag(\dot{\sigma}(f^{i - 1}(t)))\W_{i-1}^{k}}\\
    &\le \frac{1}{(L-1)\CL^2}\sum_{l \ge i \ge k} \ab{\W_l^{i+1}}\ab{\Epin{ \dit \sigma(f^{i-1})^{\intercal}}}\ab{\diag(\dot{\sigma}(f^{i - 1}(t)))}\ab{\W^k_{i-1}}\\
    & \le \frac{1}{(L-1)\CL^2}\sum_{l \ge i \ge k} \ab{\W^{i+1}_{l}}\ab{\W^k_{i-1}}\EP{\para{\dit}} \EP{\sigma(f^{i-1})^{\intercal}}.
\end{align*}

On the other hand,
\begin{align*}
 \ab{\partial_t  f^l(t)}& \le \frac{1}{(L-1)\CL^2} \left[\left\|\partial_{\theta^l} f^l(x') \Epin{\para{\partial f^l_{\theta^l}(x)}^{\intercal} \left(\W_L^{l+1}(x)\right)^{\intercal} \delta_t^L(x)}\right\|\right]\\
    &\le\frac{1}{(L-1)\CL^2}  \mathbb{E}_{x \sim p^{in}}\bigg[\sum_{i=0}^{l-1} \left\|( \para{\sigma(f^i(x'))}^{\intercal}\sigma(f^i(x))+1)\W_l^{i+2}(x')(\W_L^{i+2}(x))^{\intercal}\delta_t^L(x)\right\|\bigg]\\
    &\le\frac{1}{(L-1)\CL^2} \sum_{i=0}^{l-1}\mathbb{E}_{x \sim p^{in}} \left\| \sigma(f^i(x'))^{\intercal}\|\|\sigma(f^i(x))\|\|\W_l^{i+2}(x')\|\|(\W_L^{i+2}(x))^{\intercal}\|\|\delta_t^L(x)\right\|\\
    &\quad +\frac{1}{(L-1)\CL^2}  \sum_{i=0}^{l-1}\Epin{\left\|\W_l^{i+2}(x')\|\|(\W_L^{i+2}(x))^{\intercal}\|\|\delta_t^L(x)\right\|}\\
    &\le\frac{N^{1/2}}{(L-1)\CL^2} \sum_{i=0}^{l-1} \EP{ \sigma(f^i(x)}\ab{ \sigma(f^i(x')}\ab{\W_l^{i+2}(x')}\EP{(\W_L^{i+2}(x))^{\intercal}} \EP{\delta_t^L(x)}\\
    &\quad+\frac{1}{(L-1)\CL^2}  \sum_{i=0}^{l-1}\ab{\W_l^{i+2}(x')}\EP{(\W_L^{i+2})^{\intercal}} \EP{\delta_t^L(x)}.
\end{align*}
\end{proof}

%%%%%%%%%%%%%%%%%%%%%%%%%%%%%%%%%%%%%%%%%%%%%%%%%%%%%%%%%%%%%%%%%%%%
%%% Lemma for proposition 1
%%%%%%%%%%%%%%%%%%%%%%%%%%%%%%%%%%%%%%%%%%%%%%%%%%%%%%%%%%%%%%%%%%%%
\section{Proof of Lemma~\ref{lem:pin-bound} for Proposition~\ref{prp:pin-norm}}\label{app:prp1}

\begingroup
\def\thetheorem{\ref{lem:pin-bound}}
\begin{lemma} 
For $t \geq 0$, if $f^l_{\theta^{(l)}(t)}(x)$ is element-wise nonzero for $1 \le l \le L-1$ and $x\in\{x_1,\dots, x_N\}$, then
\begin{align*}
    \partial_t \Phi_{L,j}(t) &\le \frac{2jN^{3/2}(L-1)}{\CL}\Phi_{L,j-1}(t)\Phi_{L,8}(t)\Psi_{L, 2}(t)\Psi_{L,8}(t)\EP{\dLt}\\
    \partial_t \Psi_{L,2}(t)& \le \frac{2N(L-1)}{\CL}\Phi_{L,4}(t) \para{\Psi_{L,2}(t)}^{2}(t)\Psi_{L,8}(t)\EP{\dLt}\\
    \partial_t \Psi_{L,8}(t) & \le \frac{8N(L-1)}{\CL}\Phi_{L,4}(t) \Psi_{L,2}(t)\left(\Psi_{L,8}(t)\right)^{2}\EP{\dLt},
\end{align*}
for $1\le j\leq 8$. 
\end{lemma}
\addtocounter{theorem}{-1}
\endgroup

\begin{proof}[Proof of Lemma~\ref{lem:pin-bound}]
% Prob 1 part 1
Consider the derivative of $\Psi_{L, 8}$.
In Section~\ref{app:not crossing}, we show that the intermediate pre-activation values never reach zero while training,
and therefore we can apply Lemma~\ref{lem:bound derivative of pin norm}.
\begin{align*}
    &\partial_t \Psi_{L, 8}(t)  \\
    & \le \frac{8N}{(L-1)^2\CL^2}  \sum_{l =2 }^{L} \sum_{l=i}^{L} \EP{\W^{l+1}_{L}(t)}^2\EP{\W^i_{l-1}(t)}\left(\EP{\W_L^i(0)}+\EP{\W_L^i(t)-\W_L^i(0)}\right)^7 \\&\quad \times \EP{\fultm(x)}\EP{\dLt}.
\end{align*}
First, the following rough bound holds.
\begin{align*}
    & \sum_{l =2 }^{L} \sum_{l=i}^{L} \EP{\W^{l+1}_{L}(t)}^2\EP{\W^i_{l-1}(t)}\left(\EP{\W_L^i(0)}+\EP{\W_L^i(t)-\W_L^i(0)}\right)^7 \EP{\fultm(x)}\\
    & \le \Bigg( \sum_{i =2 }^{L} \sum_{l=i}^{L} \EP{\W^i_{l-1}(t)} \left(\EP{\fulzm}+\EP{\fultm-\fulzm}\right)\\
    &\quad \times \left(\EP{\W_L^{l+1}(0)}+\EP{\W_L^{l+1}(t)-\W_L^{l+1}(0)}\right)^2\Bigg)\sum_{i= 2}^L \left(\EP{\W_L^i(0)}+\EP{\W_L^i(t)-\W_L^i(0)}\right)^8\\
    &\le (L-1)\Psi_{L, 8}(L-1)^2\CL\para{\Psi_{L, 2}}^{(1/2)}\para{\Phi_{L, 4}}^{(1/4)}\para{\Psi_{L, 8}}^{(1/4)},
\end{align*}
where the last inequality is from 
\begin{align*}
    & \para{\sum_{i =2 }^{L} \sum_{l=i}^{L} \EP{\W^i_{l-1}}\para{\EP{\fulzm}+\EP{\fultm-\fulzm}} \left(\EP{\W_L^{l+1}(0)}+\EP{\W_L^{l+1}(t)-\W_L^{l+1}(0)}\right)^2}^2\\
    &\le \left(\sum_{i =2 }^{L} \sum_{l=i}^{L}\frac{L-1}{l-1} \EP{\W^i_{l}}^2\right) \\
    &\quad\times \left((L-1)\sum^L_{l=2} \left(\EP{\fulzm}+\EP{\fultm-\fulzm}\right)^2 \left(\EP{\W_L^{l+1}(0)}+\EP{\W_L^{l+1}(t)-\W_L^{l+1}(0)}\right)^4\right)\\
    &\le (L-1)^2 \Psi_{L, 2}(L-1)\\
    & \quad \times \sqrt{\Bigg(\sum^{L-1}_{l=1} \left(\EP{\fulz}+\EP{\fult-\fulz}\right)^4\Bigg)\Bigg(\sum^L_{l=2} \left(\EP{\W_L^l(0)}+\EP{\W_L^l(t)-\W_L^l(0)}\right)^8\Bigg)}\\
    & \le (L-1)^2 \Psi_{L, 2} (L-1) \sqrt{(L-1)^2 \CL^4\Phi_{L, 4}\Psi_{L, 8}}.
\end{align*}
Thus,
\begin{align*}
    &\partial_t \Psi_{L, 8}(t)\\
    & \le \frac{8N}{(L-1)^2\CL^2}  \sum_{l =2 }^{L} \sum_{l=i}^{L} \EP{\W^{l+1}_{L}(t)}^2\EP{\W^i_{l-1}(t)}\left(\EP{\W_L^i(0)}+\EP{\W_L^i(t)-\W_L^i(0)}\right)^7 \\&\quad \times \EP{\fultm(x)}\EP{\dLt}\\ 
    &\le \frac{8N(L-1)}{\CL} \left(\Psi_{L, 2}\right)^{1/2}\left(\Phi_{L, 4}\right)^{1/4}\left(\Psi_{L, 8}\right)^{5/4}\EP{\dLt}\\ 
    &\le \frac{8N(L-1)}{\CL} \Psi_{L, 2}\Phi_{L, 4}\left(\Psi_{L, 8}\right)^{2}\EP{\dLt}.
\end{align*}

% Prob 1 part 2
Then, let consider the derivative of $\Psi_{L, 2}$.
From Lemma~\ref{lem:bound derivative of pin norm},
\begin{align*}
    &\partial_t \Psi_{L, 2}(t)\\
    &\le \frac{2N}{(L-1)^3\CL^2} \sum^{ L}_{i=2} \sum_{ k= 2}^{i}\sum^{i}_{l=k} \frac{L-1}{i-1}\EP{\W^{l+1}_{L}}\EP{\W^{l+1}_{i}}\EP{\W^k_{l-1}}\left(\EP{\W_i^k(0)}+\EP{\W_i^k(t)-\W_i^k(0)}\right)\\
    &\quad\times\EP{f^{l-1}_{\theta^{(l-1)}(t)}}\EP{\dLt}\\
    &\le \frac{N}{(L-1)^3\CL^2}  \sum^{ L}_{i=2} \sum_{ k= 2}^{i}\sum^{i}_{l=k} \frac{L-1}{i-1}\EP{\W^{l+1}_{L}}\EP{\W^k_{l-1}}\para{\EP{\W^{l+1}_{i}}^2+\left(\EP{\W_i^k(0)}+\EP{\W_i^k(t)-\W_i^k(0)}\right)^2}\\
    &\quad\times\EP{f^{l-1}_{\theta^{(l-1)}(t)}}\EP{\dLt}\\
    &=       \frac{N}{(L-1)^3\CL^2}  \sum^{ L}_{i=2} \sum_{k=2}^{i}\sum^{i}_{l=k} \frac{L-1}{i-1}\EP{\W^{l+1}_{L}}\EP{\W^k_{l-1}}
    \EP{\W^{l+1}_{i}}^2 \EP{f^{l-1}_{\theta^{(l-1)}(t)}}\EP{\dLt}\\
    &\quad + \frac{N}{(L-1)^3\CL^2}  \sum^{ L}_{i=2} \sum_{k=2}^{i}\sum^{i}_{l=k} \frac{L-1}{i-1}\EP{\W^{l+1}_{L}} \EP{\W^k_{l-1}}
    \left(\EP{\W_i^k(0)}+\EP{\W_i^k(t)-\W_i^k(0)}\right)^2 \\& \quad \times \EP{f^{l-1}_{\theta^{(l-1)}(t)}}\EP{\dLt}.
\end{align*}
From Cauchy-Schwartz inequality,
\begin{align*}
    & \para{\sum_{k=2 }^{L} \sum_{l=k}^{L} \EP{\W^k_{l-1}}\para{\EP{\fulzm}+\EP{\fultm-\fulzm}} \left(\EP{\W_L^{l+1}(0)}+\EP{\W_L^{l+1}(t)-\W_L^{l+1}(0)}\right)}^2\\
     &\le \left(\sum_{k =2 }^{L} \sum_{l=k}^{L}\frac{L-1}{l-1} \EP{\W^k_{l}}^2\right) \\
     &\quad \times \left((L-1)\sum^L_{l=2} \left(\EP{\fulzm}+\EP{\fultm-\fulzm}\right)^2\left(\EP{\W_L^{l+1}(0)}+\EP{\W_L^{l+1}(t)-\W_L^{l+1}(0)}\right)^2\right)\\
    &\le (L-1)^2\Psi_{L, 2}(L-1)
    \\& \quad \times\sqrt{\Bigg(\sum^{L-1}_{l=1} \left(\EP{\fulz}+\EP{\fult-\fulz}\right)^4\Bigg)\Bigg(\sum^L_{l=2} \left(\EP{\W_L^l(0)}+\EP{\W_L^l(t)-\W_L^l(0)}\right)^4\Bigg)}\\
    &\le (L-1)^2\Psi_{L, 2} (L-1)
    \\& \quad \times\sqrt{\Bigg(\sum^{L-1}_{l=1} \left(\EP{\fulz}+\EP{\fult-\fulz}\right)^4\Bigg)\Bigg(\sum^L_{l=2} \left(\EP{\W_L^l(0)}+\EP{\W_L^l(t)-\W_L^l(0)}\right)^8\Bigg)}\\ 
    & \le (L-1)^2 \Psi_{L, 2} (L-1) \sqrt{(L-1)^2 \CL^4\Phi_{L, 4}\Psi_{L, 8}}\\
    & = (L-1)^4\CL^2\Psi_{L, 2}\left(\Phi_{L, 4}\right)^{(1/2)}\para{\Psi_{L, 8}}^{(1/2)}.
\end{align*}
Using the above inequality, we have the following upper bounds:
\begin{align*}
    &\sum^{ L}_{i=2} \sum_{ k= 2}^{i}\sum^{i}_{l=k} \frac{L-1}{i-1}\EP{\W^{l+1}_{L}}\EP{\W^k_{l-1}}\EP{\W^{l+1}_{i}}^2\EP{f^{l-1}_{\theta^{(l-1)}(t)}}\\
    & \le \Bigg( \sum_{k=2 }^{L} \sum_{l=k}^{L} \EP{\W^k_{l-1}}\para{\EP{\fulzm}+\EP{\fultm-\fulzm}} \\& \quad \times \left(\EP{\W_L^{l+1}(0)}+\EP{\W_L^{l+1}(t)-\W_L^{l+1}(0)}\right)\Bigg)\left(\sum_{i=2}^L \sum^i_{l=2}\frac{L-1}{i-1}\EP{\W_i^{l+1}}^2\right)\\
    &\le (L-1)^2\Psi_{L, 2}(L-1)^2\CL\para{\Psi_{L, 2}}^{(1/2)}\para{\Phi_{L, 4}}^{(1/4)}\para{\Psi_{L, 8}}^{(1/4)}\\
    & = (L-1)^4\CL \para{\Psi_{L, 2}}^{(3/2)}\para{\Phi_{L, 4}}^{(1/4)}\para{\Psi_{L, 8}}^{(1/4)}.
\end{align*}
and
\begin{align*}
    &\sum^{ L}_{i=2} \sum_{ k= 2}^{i}\sum^{i}_{l=k} \frac{L-1}{i-1}\EP{\W^{l+1}_{L}}\EP{\W^k_{l-1}}\left(\EP{\W_i^k(0)}+\EP{\W_i^k(t)-\W_i^k(0)}\right)^2\EP{f^{l-1}_{\theta^{(l-1)}(t)}}\\
    &\le \Bigg( \sum_{k=2 }^{L} \sum_{l=k}^{L} \EP{\W^k_{l-1}}\para{\EP{\fulzm}+\EP{\fultm-\fulzm}} \\
    &\quad\times \left(\EP{\W_L^{l+1}(0)}+\EP{\W_L^{l+1}(t)-\W_L^{l+1}(0)}\right)\Bigg)\left(\sum_{i=2}^L \sum^i_{k=2}\frac{L-1}{i-1}\left(\EP{\W_i^k(0)}+\EP{\W_i^k(t)-\W_i^k(0)}\right)^2\right)\\
    &\le (L-1)^2\Psi_{L, 2}(L-1)^2\CL\para{\Psi_{L, 2}}^{(1/2)}\para{\Phi_{L, 4}}^{(1/4)}\para{\Psi_{L, 8}}^{(1/4)}\\
    &= (L-1)^4\CL \para{\Psi_{L, 2}}^{(3/2)}\para{\Phi_{L, 4}}^{(1/4)}\para{\Psi_{L, 8}}^{(1/4)}.
\end{align*}

Thus, we have
\begin{align*}
    &\partial_t \Psi^{(2)}_L(t)\\
    &\le \frac{N}{(L-1)^3\CL^2}  \sum^{ L}_{i=2} \sum_{ k= 2}^{i}\sum^{i}_{l=k} \frac{L-1}{i-1}\EP{\W^{l+1}_{L}}\EP{\W^k_{l-1}}\para{\EP{\W^{l+1}_{i}}^2+\left(\EP{\W_i^k(0)}+\EP{\W_i^k(t)-\W_i^k(0)}\right)^2}\\
    &\quad \times \EP{f^{l-1}_{\theta^{(l-1)}(t)}}\EP{\dLt}\\
    & \le \frac{N2(L-1)}{\CL} \para{\Psi_{L, 2}}^{(3/2)}\para{\Phi_{L, 4}}^{(1/4)}\para{\Psi_{L, 8}}^{(1/4)}\EP{\dLt}\\
    & \le \frac{N2(L-1)}{\CL} \para{\Psi_{L, 2}}^{2}\Phi_{L, 4}\Psi_{L, 8}\EP{\dLt}.
\end{align*}

% Prop 1, part 3
Finally, let consider the derivatives of $\Phi_{L, j}$ for $1\leq j \leq 8$.
From Lemma~\ref{lem:bound derivative of pin norm},
\begin{align*}
    \partial_t \Phi_{L, j}(t) 
    &\le\frac{jN^{2}}{(L-1)^2\CL^{2+j}} \sum_{i= 1}^{L-1} \sum_{l=0}^{i-1} \left(\EP{ f^{i}_{\theta^{(i)}(0)}}+\EP{ f^{i}_{\theta^{(i)}(t)}- f^{i}_{\theta^{(i)}(0)}}\right)^{j-1}\EP{ f^{l}_{\theta^{(l)}(t)}(x)}\EP{ f^{l}_{\theta^{(l)}(t)}(x')}\\
    &\quad\quad\times \EP{\W_i^{l+2}}\EP{(\W_L^{l+2})^{\intercal}} \EP{\dLt}\\
    &\quad +\frac{jN^{2}}{(L-1)^2\CL^{2+j}} \sum_{i= 1}^{L-1} \sum_{l=0}^{i-1}\left(\EP{ f^{i}_{\theta^{(i)}(0)}}+\EP{ f^{i}_{\theta^{(i)}(t)}- f^{i}_{\theta^{(i)}(0)}}\right)^{j-1}\EP{\W_i^{l+2}}\EP{(\W_L^{l+2})^{\intercal}} \EP{\dLt}.
\end{align*}
Similar to the previous inequalities, we have
\begin{align*}
    &\bigg(\sum^{L-1}_{i=1} \sum^{i-1}_{l=0} \EP{\W_i^{l+2}} \left(\EP{\fulz}+\EP{\fult-\fulz}\right)^2 \left(\EP{\W_L^{l+2}(0)}+\EP{\W_L^{l+2}(t)-\W_L^{l+2}(0)}\right)\bigg)^2\\
    &\le\para{\sum^{L-1}_{i=1} \sum^{i-1}_{l=0}\frac{L-1}{i-1} \EP{\W_i^{l+2}}^2}\\
    &\quad \times \Bigg((L-1)\sum^{L-2}_{l=0} \left(\EP{\fulz}+\EP{\fult-\fulz}\right)^4 \left(\EP{\W_L^{l+2}(0)}+\EP{\W_L^{l+2}(t)-\W_L^{l+2}(0)}\right)^2\Bigg)\\
    &\leq (L-1)^2\Psi_{L, 2}(t)(L-1) \\& \quad \times\sqrt{\Bigg(\sum^{L-2}_{l=0} \left(\EP{\fulz}+\EP{\fult-\fulz}\right)^8\Bigg)\Bigg(\sum^L_{l=2} \left(\EP{\W_L^l(0)}+\EP{\W_L^l(t)-\W_L^l(0)}\right)^4\Bigg)}\\ 
    &\leq (L-1)^2\Psi_{L, 2}(t)(L-1)\\& \quad \times \sqrt{\Bigg(\sum^{L-2}_{l=0} \left(\EP{\fulz}+\EP{\fult-\fulz}\right)^8\Bigg)\Bigg(\sum^L_{l=2} \left(\EP{\W_L^l(0)}+\EP{\W_L^l(t)-\W_L^l(0)}\right)^8\Bigg)}\\ 
    &\leq (L-1)^2\Psi_{L, 2}(t)(L-1)\times (L-1)\CL^4\para{\Phi_{L, 8}(t)}^{(1/2)}\para{\Psi_{L, 8}(t)}^{(1/2)} \\&=(L-1)^4\CL^4\Psi_{L, 2}(t)\para{\Phi_{L, 8}(t)}^{(1/2)}\para{\Psi_{L, 8}(t)}^{(1/2)}.
\end{align*}
Thus, we have
\begin{align*}
    &\sum_{i= 1}^{L-1} \sum_{l=0}^{i-1} \left(\EP{ f^{i}_{\theta^{(i)}(0)}}+\EP{ f^{i}_{\theta^{(i)}(t)}- f^{i}_{\theta^{(i)}(0)}}\right)^{j-1}\EP{ f^{l}_{\theta^{(l)}(t)}(x)}\EP{ f^{l}_{\theta^{(l)}(t)}(x')}\EP{\W_i^{l+2}}\EP{(\W_L^{l+2})^{\intercal}}\\
    &\le \sum^{L-1}_{i=1} \sum^{i-1}_{l=0} \EP{\W_i^{l+2}} \left(\EP{\fulz}+\EP{\fult-\fulz}\right)^2 \left(\EP{\W_L^{l+2}(0)}+\EP{\W_L^{l+2}(t)-\W_L^{l+2}(0)}\right)\\
    &\quad \times \sum^{L-1}_{i=1}\left(\EP{ f^{i}_{\theta^{(i)}(0)}}+\EP{ f^{i}_{\theta^{(i)}(t)}- f^{i}_{\theta^{(i)}(0)}}\right)^{j-1}\\
    & \leq (L-1)C_L^{j-1}\Phi_{L, j - 1}(t)(L-1)^2\CL^2\para{\Psi_{L, 2}(t)}^{1/2}\para{\Phi_{L, 8}(t)}^{(1/4)}\para{\Psi_{L, 8}(t)}^{(1/4)}\\
    & = (L-1)^3C_L^{j+1}\Phi_{L, j - 1}(t)\para{\Psi_{L, 2}(t)}^{1/2}\para{\Phi_{L, 8}(t)}^{(1/4)}\para{\Psi_{L, 8}(t)}^{(1/4)}
\end{align*}
and 
\begin{align*}
    &\frac{jN^{3/2}}{(L-1)^2\CL^{2+j}} \sum_{i= 1}^{L-1} \sum_{l=0}^{i-1}\left(\EP{ f^{i}_{\theta^{(i)}(0)}}+\EP{ f^{i}_{\theta^{(i)}(t)}- f^{i}_{\theta^{(i)}(0)}}\right)^{j-1}\EP{\W_i^{l+2}}\EP{(\W_L^{l+2})^{\intercal}}\\
    &\le \frac{jN^{3/2}}{(L-1)^2\CL^{2+j}} \sum_{i= 1}^{L-1} \sum_{l=0}^{i-1} \left(\EP{ f^{i}_{\theta^{(i)}(0)}}+\EP{ f^{i}_{\theta^{(i)}(t)}- f^{i}_{\theta^{(i)}(0)}}\right)^{j-1}
    \EP{ f^{l}_{\theta^{(l)}(t)}(x)}\EP{ f^{l}_{\theta^{(l)}(t)}(x')} \\& \quad \times \EP{\W_i^{l+2}}\EP{(\W_L^{l+2})^{\intercal}} .
\end{align*}

By combining the above inequalities, we have
\begin{align*}
    \partial_t \Phi_{L, j}(t) 
    &\le\frac{jN^{3/2}}{(L-1)^2\CL^{2+j}} \sum_{i= 1}^{L-1} \sum_{l=0}^{i-1} \left(\EP{ f^{i}_{\theta^{(i)}(0)}}+\EP{ f^{i}_{\theta^{(i)}(t)}- f^{i}_{\theta^{(i)}(0)}}\right)^{j-1}\EP{ f^{l}_{\theta^{(l)}(t)}(x)}\EP{ f^{l}_{\theta^{(l)}(t)}(x')}\\
    &\quad\quad\times \EP{\W_i^{l+2}}\EP{(\W_L^{l+2})^{\intercal}} \EP{\dLt}\\
    &\quad +\frac{jN^{3/2}}{(L-1)^2\CL^{2+j}} \sum_{i= 1}^{L-1} \sum_{l=0}^{i-1}\left(\EP{ f^{i}_{\theta^{(i)}(0)}}+\EP{ f^{i}_{\theta^{(i)}(t)}- f^{i}_{\theta^{(i)}(0)}}\right)^{j-1}\EP{\W_i^{l+2}}\EP{(\W_L^{l+2})^{\intercal}} \EP{\dLt}\\
    &\le \frac{2jN^{3/2}(L-1)}{\CL}\Phi_{L, j - 1}(t)(\Psi_{L, 2}(t))^{1/2}(\Phi_{L, 8}(t))^{(1/4)}(\Psi_{L, 8}(t))^{(1/4)}\EP{\dLt}\\
    &\le \frac{2jN^{3/2}(L-1)}{\CL}\Phi_{L, j - 1}(t)\Psi_{L, 2}(t)\Phi_{L, 8}(t)\Psi_{L, 8}(t)\EP{\dLt}.
\end{align*}
\end{proof}

%%%%%%%%%%%%%%%%%%%%%%%%%%%%%%%%%%%%%%%%%%%%%%%
%%% Lemma for Proposition 2
%%%%%%%%%%%%%%%%%%%%%%%%%%%%%%%%%%%%%%%%%%%%%%%
\section{Proof of Lemma~\ref{lem:l2-bound} for Proposition~\ref{prp:l2-norm}}\label{app:prp2}

\begingroup
\def\thetheorem{\ref{lem:l2-bound}}
\begin{lemma}
For $t \geq 0$ and $x\in\fR_+^{d_{in}}$, if $f^l_{\theta^{(l)}(t)}(x)$ is element-wise nonzero for $1 \le l \le L-1$, then
\begin{align*}
    \partial_t \tilde{\Psi}_{L,2}(x,t)&\le \frac{2(L-1)}{\CL} \para{\tilde{\Psi}_{L,2}(x,t)}^{2}\Phi_{L, 4}(t)\Psi_{L,8}(t)\EP{\dLt}\\
    \partial_t \tilde{\Phi}_{L, j}(x,t) &\le \frac{2jN^{1/2}(L-1)}{\CL}\tilde{\Phi}_{L, j-1}(t)\tilde{\Phi}_{L,8}(x,t)\tilde{\Psi}_{L, 2}(x,t)\Phi_{L,8}(t)\Psi_{L,8}(t)\EP{\dLt}\\
    \partial_t \tilde{\Psi}_{L, 8}(x,t)& \le \frac{8(L-1)}{\CL}\tilde{\Phi}_{L,4}(x,t) \tilde{\Psi}_{L,2}(x,t)\para{\tilde{\Psi}_{L,8}(x,t)}^{2}\Psi_{L,8}(t)\EP{\dLt},
\end{align*}
for $1\leq j\leq 8$.
\end{lemma}
\addtocounter{theorem}{-1}
\endgroup

\begin{proof}[Proof of Lemma~\ref{lem:l2-bound}]
% Prop 2, part 1
Consider the derivative of $\tilde{\Psi}_{L, 8}$ first.
In Section~\ref{app:not crossing}, we show that the intermediate pre-activation values never reach zero while training,
and therefore we can apply Lemma~\ref{lem:bound derivative of l2 norm}.
\begin{align*}
    & \partial_t \tilde{\Psi}_{L, 8}(t)\\
    &  \le \frac{8}{(L-1)^2\CL^2}  \sum_{l =2 }^{L} \sum_{i=l}^{L} \EP{\W^{l+1}_{L}(t)}\ab{\W^{l+1}_{L}(t)}\ab{\W^i_{l-1}(t)}\left(\ab{\W_L^i(0)}+\ab{\W_L^i(t)-\W_L^i(0)}\right)^7
    \\ & \quad\times \EP{\fultm(x)}\EP{\dLt}.
\end{align*}
Then,
\begin{align*}
     &\sum_{l =2 }^{L} \sum_{i=l}^{L} \EP{\W^{l+1}_{L}(t)}\ab{\W^{l+1}_{L}(t)}\ab{\W^i_{l-1}(t)}\left(\ab{\W_L^i(0)}+\ab{\W_L^i(t)-\W_L^i(0)}\right)^7 \EP{\fultm(x)}\\
     & \le \Bigg( \sum_{i =2 }^{L} \sum_{i=l}^{L} \ab{\W^i_{l-1}} \left(\EP{\fulzm}+\EP{\fultm-\fulzm}\right) \left(\ab{\W_L^{l+1}(0)}+\ab{\W_L^{l+1}(t)-\W_L^{l+1}(0)}\right)\\
     &\quad \times\left(\EP{\W_L^{l+1}(0)}+\EP{\W_L^{l+1}(t)-\W_L^{l+1}(0)}\right)\Bigg)\Bigg(\sum_{i= 2}^L \left(\EP{\W_L^i(0)}+\EP{\W_L^i(t)-\W_L^i(0)}\right)^8\Bigg)\\
    &\le \left(\sum_{l=2 }^{L} \sum_{i=2}^{l}\frac{L-1}{l-1} \ab{\W^i_{l}}^2\right)^{1/2} \Bigg((L-1)\sum^{L}_{l=2} \left(\EP{\fulzm}+\EP{\fultm-\fulzm}\right)^2 \\
    &\quad\times \left(\ab{\W_L^{l+1}(0)}+\ab{\W_L^{l+1}(t)-\W_L^{l+1}(0)}\right)^2\left(\EP{\W_L^{l+1}(0)}+\EP{\W_L^{l+1}(t)-\W_L^{l+1}(0)}\right)^2 \Bigg)^{1/2}(L-1)\tilde{\Psi}_{L, 8}\\
    &\le \left((L-1)^2\tilde{\Psi}_{L, 2}\right)^{1/2}\sqrt{L-1}\Bigg(\sum^{L}_{l=2} \left(\EP{\fulzm}+\EP{\fultm-\fulzm}\right)^4\Bigg)^{1/4}\\
    &\quad\times \Bigg(\sum_{l=2}^{L} \left(\ab{\W_L^{l+1}(0)}+\ab{\W_L^{l+1}(t)-\W_L^{l+1}(0)}\right)^4 \left(\EP{\W_L^{l+1}(0)}+\EP{\W_L^{l+1}(t)-\W_L^{l+1}(0)}\right)^4 \Bigg)^{1/4}(L-1)\tilde{\Psi}_{L, 8}\\
    &\le \left((L-1)^2\tilde{\Psi}_{L, 2}\right)^{1/2} \sqrt{L-1} (LC_L^4)^{1/4}(\Phi_{L, 4})^{1/4}(L-1)\tilde{\Psi}_{L, 8}\\
    &\quad\times \Bigg(\sum_{l=2}^{L} \left(\ab{\W_L^{l+1}(0)}+\ab{\W_L^{l+1}(t)-\W_L^{l+1}(0)}\right)^8\Bigg)^{1/8}
    \Bigg(\sum_{l=2}^{L}\left(\EP{\W_L^{l+1}(0)}+\EP{\W_L^{l+1}(t)-\W_L^{l+1}(0)}\right)^8 \Bigg)^{1/8}\\
    &= (L-1)\tilde{\Psi}_{L, 8}(L-1)^2\CL\para{\tilde{\Psi}_{{L}, 2}}^{1/2}\para{\tilde{\Phi}_{L, 4}}^{(1/4)}\para{\Psi_{L, 8}}^{(1/8)}\para{\tilde{\Psi}_{L, 8}}^{(1/8)}
\end{align*}

Thus, we have
\begin{align*}
    &\partial_t \tilde{\Psi}_{L, 8}(t)
    \\ & \le \frac{8}{(L-1)^2\CL^2}  \sum_{l =2 }^{L} \sum_{i=l}^{L} \EP{\W^{l+1}_{L}(t)}\ab{\W^{l+1}_{L}(t)}\ab{\W^i_{l-1}(t)}\left(\ab{\W_L^i(0)}+\ab{\W_L^i(t)-\W_L^i(0)}\right)^7 \\ & \quad\times  \EP{\fultm(x)}\EP{\dLt}\\ 
    &   \le \frac{8(L-1)}{\CL} \para{\tilde{\Psi}_{{L}, 2}}^{1/2}\para{\tilde{\Phi}_{L, 4}}^{1/4}\para{\Psi_{L, 8}}^{1/8}\para{\tilde{\Psi}_{L, 8}}^{9/8}\EP{\dLt}\\ 
    & \le \frac{8(L-1)}{\CL} \tilde{\Psi}_{{L}, 2}\tilde{\Phi}_{L, 4}\Psi_{L, 8}\para{\tilde{\Psi}_{L, 8}}^{2}\EP{\dLt}.
\end{align*}

% Prop 2, part 2
Now, let consider the derivative of $\tilde{\Psi}_{L, 2}$.
From Lemma~\ref{lem:bound derivative of l2 norm},
\begin{align*}
    & \partial_t \tilde{\Psi}_{L, 2}(t)
    \\& \le \frac{2}{(L-1)^3\CL^2} \sum^{ L}_{i=2} \sum_{ k= 2}^{i}\sum^{i}_{l=k} \frac{L-1}{i-1}\EP{\W^{l+1}_{L}}\ab{\W^{l+1}_{i}}\ab{\W^k_{l-1}}\left(\ab{\W_i^k(0)}+\ab{\W_i^k(t)-\W_i^k(0)}\right) \\ & \quad\times \EP{f^{l-1}_{\theta^{(l-1)}(t)}}\EP{\dLt}\\
    & \le \frac{1}{(L-1)^3\CL^2}  \sum^{ L}_{i=2} \sum_{ k= 2}^{i}\sum^{i}_{l=k} \frac{L-1}{i-1}\EP{\W^{l+1}_{L}}\ab{\W^k_{l-1}}\para{\ab{\W^{l+1}_{i}}^2+\left(\ab{\W_i^k(0)}+\ab{\W_i^k(t)-\W_i^k(0)}\right)^2}  \\ & \quad\times \EP{f^{l-1}_{\theta^{(l-1)}(t)}}\EP{\dLt}\\
    & =\frac{1}{(L-1)^3\CL^2}  \sum^{ L}_{i=2} \sum_{ k= 2}^{i}\sum^{i}_{l=k} \frac{L-1}{i-1}\EP{\W^{l+1}_{L}}\ab{\W^k_{l-1}}\ab{\W^{l+1}_{i}}^2\EP{f^{l-1}_{\theta^{(l-1)}(t)}}\EP{\dLt}\\
    &\quad +\frac{1}{(L-1)^3\CL^2}  \sum^{ L}_{i=2} \sum_{ k= 2}^{i}\sum^{i}_{l=k} \frac{L-1}{i-1}\EP{\W^{l+1}_{L}}\ab{\W^k_{l-1}}\left(\ab{\W_i^k(0)}+\ab{\W_i^k(t)-\W_i^k(0)}\right)^2\EP{f^{l-1}_{\theta^{(l-1)}(t)}}\EP{\dLt}.
\end{align*}

By Cauchy-Schwartz inequality,
% \begin{align*}
%     L*(L-1)\CL^4\Phi^{(4)}_L(t)\Psi^{(8)}_L(t) \ge \Bigg(\sum^{L}_{l=2} \left(\EP{f^{l-1}_{\theta^{(l-1)}(0)}}+\EP{f^{l-1}_{\theta^{(l-1)}(0)}-f^{l-1}_{\theta^{(l-1)}(0)}}\right)^2\left(\EP{\W_L^{l+1}(0)}+\EP{\W_L^{l+1}(t)-\W_L^{l+1}(0)}\right)^2\Bigg)^2,
% \end{align*} 
% and
\begin{align*}
    &\para{\sum_{k=2 }^{L} \sum_{l=k}^{L} \ab{\W^k_{l-1}}\para{\EP{\fulzm}+\EP{\fultm-\fulzm}} \left(\EP{\W_L^{l+1}(0)}+\EP{\W_L^{l+1}(t)-\W_L^{l+1}(0)}\right)}^2\\
    &\le\left(\sum_{k =2 }^{L} \sum_{l=k}^{L}\frac{L-1}{l-1} \ab{\W^k_{l}}^2\right) \\
    &\quad\times \left((L-1)\sum^L_{l=2} \left(\EP{\fulzm}+\EP{\fultm-\fulzm}\right)^2\left(\EP{\W_L^{l+1}(0)}+\EP{\W_L^{l+1}(t)-\W_L^{l+1}(0)}\right)^2\right)\\
    &\le (L-1)^2 \tilde{\Psi}_{L, 2}(L-1) \sqrt{(L-1) \CL^4\Phi_{L, 4}L\Psi_{L, 8}}\\
    & \le (L-1)^4\CL^2\tilde{\Psi}_{L, 2} \left(\Phi_{L, 4}\right)^{(1/2)}\para{\Psi_{L, 8}}^{(1/2)}.
\end{align*}

Then, we have
\begin{align*}
    &\sum^{ L}_{i=2} \sum_{ k= 2}^{i}\sum^{i}_{l=k} \frac{L-1}{i-1}\EP{\W^{l+1}_{L}}\ab{\W^k_{l-1}}\ab{\W^{l+1}_{i}}^2\EP{f^{l-1}_{\theta^{(l-1)}(t)}}\\
    &\le \Bigg( \sum_{k=2 }^{L} \sum_{l=k}^{L} \ab{\W^k_{l-1}}\para{\EP{\fulzm}+\EP{\fultm-\fulzm}} \left(\EP{\W_L^{l+1}(0)}+\EP{\W_L^{l+1}(t)-\W_L^{l+1}(0)}\right)\Bigg)\\
    &\quad\times \left(\sum_{i=2}^L \sum^i_{l=2}\frac{L-1}{i-1}\ab{\W_i^{l+1}}^2\right)\\
    &\le (L-1)^2\tilde{\Psi}_{L, 2}(L-1)^2\CL\para{\tilde{\Psi}_{L, 2}}^{(1/2)}\para{\Phi_{L, 2}}^{(1/4)}\para{\Psi_{L, 8}}^{(1/4)}\\
    &=(L-1)^4\CL \para{\tilde{\Psi}_{L, 2}}^{(3/2)}\para{\Phi_{L, 4}}^{(1/2)}\para{\Psi_{L, 8}}^{(1/4)}
\end{align*}
and
\begin{align*}
    &\sum^{ L}_{i=2} \sum_{ k= 2}^{i}\sum^{i}_{l=k} \frac{L-1}{i-1}\EP{\W^{l+1}_{L}}\ab{\W^k_{l-1}}\left(\ab{\W_i^k(0)}+\ab{\W_i^k(t)-\W_i^k(0)}\right)^2\EP{f^{l-1}_{\theta^{(l-1)}(t)}}\\
    &\le\Bigg( \sum_{k=2 }^{L} \sum_{l=k}^{L} \ab{\W^k_{l-1}}\para{\EP{\fulzm}+\EP{\fultm-\fulzm}} \left(\EP{\W_L^{l+1}(0)}+\EP{\W_L^{l+1}(t)-\W_L^{l+1}(0)}\right)\Bigg)\\
    &\quad \times\left(\sum_{i=2}^L \sum^i_{k=2}\frac{L-1}{i-1}\left(\ab{\W_i^k(0)}+\ab{\W_i^k(t)-\W_i^k(0)}\right)^2\right)\\
    &\le (L-1)^2\tilde{\Psi}_{L, 2}(L-1)^2\CL\para{\tilde{\Psi}_{L, 2}}^{(1/2)}\para{\Phi_{L, 4}}^{(1/4)}\para{\Psi_{L, 8}}^{(1/4)}\\
    &=(L-1)^4\CL \para{\tilde{\Psi}_{L, 2}}^{(3/2)}\para{\Phi_{L, 4}}^{(1/4)}\para{\Psi_{L, 8}}^{(1/4)}.
\end{align*}

Thus, by combining the above two inequalities,
\begin{align*}
    & \partial_t \tilde{\Psi}_{L, 2}(t)
    \\ &\le \frac{1}{(L-1)^3\CL^2}  \sum^{ L}_{i=2} \sum_{ k= 2}^{i}\sum^{i}_{l=k} \frac{L-1}{i-1}\EP{\W^{l+1}_{L}}\ab{\W^k_{l-1}}\para{\ab{\W^{l+1}_{i}}^2+\left(\ab{\W_i^k(0)}+\ab{\W_i^k(t)-\W_i^k(0)}\right)^2} \\ & \quad\times \EP{f^{l-1}_{\theta^{(l-1)}(t)}}\EP{\dLt}\\
    & \le \frac{2(L-1)}{\CL} \para{\tilde{\Psi}_{L, 2}}^{3/2}\para{\Phi_{L, 4}}^{1/4}\para{\Psi_{L, 8}}^{1/4}\EP{\dLt}\\
    &  \le \frac{2(L-1)}{\CL} \para{\tilde{\Psi}_{L, 2}}^{2}\Phi_{L, 4}\Psi_{L, 8}\EP{\dLt}
\end{align*}

% Prop 2, part 3
Finally, let consider the derivatives of $\tilde{\Phi}_{L, j}$ for $1\leq j\leq 8$.
From Lemma~\ref{lem:bound derivative of l2 norm},
\begin{align*}
    &\partial_t \tilde{\Phi}_{L, j}(t) 
    \\ &  \le\frac{jN^{1/2}}{(L-1)^2\CL^{2+j}} \sum_{i= 1}^{L-1} \sum_{l=0}^{i-1} \left(\ab{ f^{i}_{\theta^{(i)}(0)}}+\ab{ f^{i}_{\theta^{(i)}(t)}- f^{i}_{\theta^{(i)}(0)}}\right)^{j-1}\EP{ f^{l}_{\theta^{(l)}(t)}(x)}\ab{ f^{l}_{\theta^{(l)}(t)}(x')} \\ & \quad\times \ab{\W_i^{l+2}}\EP{(\W_L^{l+2})^{\intercal}} \EP{\dLt}\\&\quad +\frac{j}{(L-1)^2\CL^{2+j}} \sum_{i= 1}^{L-1} \sum_{l=0}^{i-1}\left(\ab{ f^{i}_{\theta^{(i)}(0)}}+\ab{ f^{i}_{\theta^{(i)}(t)}- f^{i}_{\theta^{(i)}(0)}}\right)^{j-1} \ab{\W_i^{l+2}}\EP{(\W_L^{l+2})^{\intercal}} \EP{\dLt}\\
    & \le \frac{2jN^{1/2}}{(L-1)^2\CL^{2+j}} \sum_{i= 1}^{L-1} \sum_{l=0}^{i-1} \left(\ab{ f^{i}_{\theta^{(i)}(0)}}+\ab{ f^{i}_{\theta^{(i)}(t)}- f^{i}_{\theta^{(i)}(0)}}\right)^{j-1}\EP{ f^{l}_{\theta^{(l)}(t)}(x)}\ab{ f^{l}_{\theta^{(l)}(t)}(x')} \\ & \quad\times \ab{\W_i^{l+2}}\EP{(\W_L^{l+2})^{\intercal}} \EP{\dLt}
\end{align*}
since $\ab{f^l_{\theta^{(l)}(t)}(x)} \geq 1$.

By Cauchy-Schwartz inequality,
\begin{align*}
    &\quad \Bigg(\sum^{L-2}_{l=0} \left(\EP{\fulz}+\EP{\fult-\fulz}\right)^2 \left(\ab{\fulz}+\ab{\fult-\fulz}\right)^2 \\& \quad  \times \left(\EP{\W_L^{l+2}(0)}+\EP{\W_L^{l+2}(t)-\W_L^{l+2}(0)}\right)^2\Bigg)^{2}\\
    &\quad \le \Bigg(\sum^{L-2}_{l=0} \left(\EP{\fulz}+\EP{\fult-\fulz}\right)^4 \left(\ab{\fulz}+\ab{\fult-\fulz}\right)^4\Bigg)\\
    % &\quad\times \Bigg(\sum^L_{l=2} \left(\EP{\W_L^l(0)}+\EP{\W_L^l(t)-\W_L^l(0)}\right)^4\Bigg)\\
    &\quad \le\Bigg(\sum^{L-2}_{l=0} \left(\EP{\fulz}+\EP{\fult-\fulz}\right)^4 \left(\ab{\fulz}+\ab{\fult-\fulz}\right)^4\Bigg)\\
    &\quad\times \Bigg(\sum^L_{l=2} \left(\EP{\W_L^l(0)}+\EP{\W_L^l(t)-\W_L^l(0)}\right)^4\Bigg)\\
    &\quad \le\Bigg(\sum^{L-2}_{l=0} \left(\EP{\fulz}+\EP{\fult-\fulz}\right)^8\Bigg)^{1/2} \Bigg(\left(\ab{\fulz}+\ab{\fult-\fulz}\right)^8\Bigg)^{1/2}\\
    &\quad\times \Bigg(\sum^L_{l=2} \left(\EP{\W_L^l(0)}+\EP{\W_L^l(t)-\W_L^l(0)}\right)^8\Bigg)\\
    &\quad \le  (L-1)\Psi_{L, 8} (L-1)\CL^8\para{\Phi_{L, 8}}^{1/2}\para{\tilde{\Phi}_{L, 8}}^{1/2}.
\end{align*} 

Then, we have
\begingroup
\allowdisplaybreaks
\begin{align*}
    &\sum_{i= 1}^{L-1} \sum_{l=0}^{i-1} \left(\ab{ f^{i}_{\theta^{(i)}(0)}}+\ab{ f^{i}_{\theta^{(i)}(t)}- f^{i}_{\theta^{(i)}(0)}}\right)^{j-1}\EP{ f^{l}_{\theta^{(l)}(t)}(x)}\ab{ f^{l}_{\theta^{(l)}(t)}(x')}\ab{\W_i^{l+2}}\EP{(\W_L^{l+2})^{\intercal}}\\
    &\le \para{\sum_{i= 1}^{L-1} \left(\ab{ f^{i}_{\theta^{(i)}(0)}}+\ab{ f^{i}_{\theta^{(i)}(t)}- f^{i}_{\theta^{(i)}(0)}}\right)^{j-1}}\Bigg(\sum^{L-1}_{i=1} \sum^{i-1}_{l=0}\frac{L-1}{i-1} \ab{\W_i^{l+2}} \left(\EP{\fulz}+\EP{\fult-\fulz}\right)\\
    &\quad \times \left(\ab{\fulz}+\ab{\fult-\fulz}\right) \left(\EP{\W_L^{l+2}(0)}+\EP{\W_L^{l+2}(t)-\W_L^{l+2}(0)}\right)\Bigg)\\
    &\le (L-1)\CL^{j-1}\tilde{\Phi}_{L, j-1} \Bigg(\sum^{L-1}_{i=1} \sum^{i-1}_{l=0}\frac{L-1}{i-1} \ab{\W_i^{l+2}} \left(\EP{\fulz}+\EP{\fult-\fulz}\right)\\
    &\quad \times \left(\ab{\fulz}+\ab{\fult-\fulz}\right) \left(\EP{\W_L^{l+2}(0)}+\EP{\W_L^{l+2}(t)-\W_L^{l+2}(0)}\right)\Bigg)\\
    &\le (L-1)\CL^{j-1}\tilde{\Phi}_{L, j-1} \para{\sum^{L-1}_{i=1} \sum^{i-1}_{l=0}\frac{L-1}{i-1} \ab{\W_i^{l+2}}^2}^{1/2} \Bigg((L-1)\sum^{L-2}_{l=0} \left(\EP{\fulz}+\EP{\fult-\fulz}\right)^2 \\&\quad\left(\ab{\fulz}+\ab{\fult-\fulz}\right)^2\times  \left(\EP{\W_L^{l+2}(0)}+\EP{\W_L^{l+2}(t)-\W_L^{l+2}(0)}\right)^2\Bigg)^{1/2}\\
    &\le (L-1)\CL^{j-1} \tilde{\Phi}_{L, j-1} \left((L-1)^2\tilde{\Psi}_{L, 2}\right)^{1/2}
    \left((L-1)^2\CL^4\para{\Phi_{L, 8}}^{1/4}\para{\Psi_{L, 8}}^{1/2}\para{\tilde{\Phi}_{L, 8}}^{1/4}\right)^{1/2}\\
    &=(L-1)^3\CL^{j+1}\tilde{\Phi}_{{L}, j - 1}\para{\tilde{\Psi}_{L, 2}}^{1/2}\para{\Phi_{L, 8}}^{1/8}\para{\Psi_{L, 8}}^{1/4}\para{\tilde{\Phi}_{L, 8}}^{1/8} 
\end{align*}
\endgroup

Thus, by combining the above inqualities,
\begin{align*}
    &\partial_t \tilde{\Phi}_{L, j}(t) 
    \\& \le2 \frac{jN^{1/2}}{(L-1)^2\CL^{2+j}} \sum_{i= 1}^{L-1} \sum_{l=0}^{i-1} \left(\ab{ f^{i}_{\theta^{(i)}(0)}}+\ab{ f^{i}_{\theta^{(i)}(t)}- f^{i}_{\theta^{(i)}(0)}}\right)^{j-1}\EP{ f^{l}_{\theta^{(l)}(t)}(x)}\ab{ f^{l}_{\theta^{(l)}(t)}(x')} \\ & \quad\times \ab{\W_i^{l+2}}\EP{(\W_L^{l+2})^{\intercal}} \EP{\dLt}\\
    & \le \frac{2jN^{1/2}(L-1)}{\CL}\tilde{\Phi}_{{L},  j - 1}\para{\tilde{\Psi}_{L, 2}}^{1/2}\para{\Phi_{L, 8}}^{1/8}\para{\Psi_{L, 8}}^{1/4}\para{\tilde{\Phi}_{L, 8}}^{1/8}\EP{\dLt}\\
    & \le \frac{2jN^{1/2}(L-1)}{\CL}\tilde{\Phi}_{{L}, j - 1}\tilde{\Psi}_{L, 2}\Phi_{L, 8}\Psi_{L, 8}\tilde{\Phi}_{L, 8}\EP{\dLt}.
\end{align*}
\end{proof}

%%%%%%%%%%%%%%%%%%%%%%%%%%%%%%%%%%%%%%%%%%%%%%%%%%%%%
%%% Lemma for Proposition 3
%%%%%%%%%%%%%%%%%%%%%%%%%%%%%%%%%%%%%%%%%%%%%%%%%%%%%
\section{Proof of Lemma~\ref{lem:layer-bound} for Proposition~\ref{prp:layerwise}}\label{app:prp3}
\begingroup
\def\thetheorem{\ref{lem:layer-bound}}
\begin{lemma}
For $t \geq 0$ and $x\in\fR_+^{d_{in}}$, if $f^l_{\theta^{(l)}(t)}(x)$ is element-wise nonzero for $1 \le l \le L-1$, then
\begin{align*}
    \partial_t\tilde{\Psi}^{k}_L(x,t) &\le \frac{L-1}{\CL} \Psi_{L,8}(t)\tilde{\Phi}_{L,4}(x,t)\tilde{\Psi}_{L,2}(x,t)\tilde{\Psi}_{L,8}(x,t)\EP{\dLt}\\
    \partial_t\tilde{\Phi}^{l}_L(x,t) &\le \frac{2N^{1/2}(L-1)}{\CL}\Phi_{L,8}(t)\Psi_{L,8}(t)\tilde{\Phi}_{L,8}(x,t)\tilde{\Psi}_{L, 2}(x,t)\EP{\dLt},
\end{align*}
for $1 \le l \le L$ and $2 \le k \le L$.
\end{lemma}
\addtocounter{theorem}{-1}
\endgroup
\begin{proof}[Proof of Lemma~\ref{lem:layer-bound}]
Consider the derivative of $\tilde{\Psi}_L^k$ first.
In the proof of Lemma~\ref{lem:l2-bound}, we showed
\begin{align*}
    &\Bigg(\sum_{i =2 }^{L} \sum_{l=i}^{L} \ab{\W^i_{l-1}} \left(\EP{\fulzm}+\EP{\fultm-\fulzm}\right) \left(\ab{\W_L^{l+1}(0)}+\ab{\W_L^{l+1}(t)-\W_L^{l+1}(0)}\right)\\
    &\quad \times\left(\EP{\W_L^{l+1}(0)}+\EP{\W_L^{l+1}(t)-\W_L^{l+1}(0)}\right)\Bigg)^2\\
    &\le (L-1)^2 \tilde{\Psi}_{L, 2} (L-1)\times (L-1)\CL^2\para{\Phi_{L, 4}}^{1/2}\para{\Psi_{L, 8}(t)}^{1/4}\para{\tilde{\Psi}_{L, 8}(t)}^{1/4}.
\end{align*}
Thus, from Lemma~\ref{lem:bound derivative of l2 norm},
\begin{align*}
    \partial_t \tilde{\Psi}^{k}_L(t)
    & \le \frac{1}{(L-1)\CL^2}  \sum_{l=k}^{L} \EP{\W^{l+1}_{L}(t)}\ab{\W^{l+1}_{L}(t)}\ab{\W^k_{l-1}(t)} \EP{\fultm(x)}\EP{\dLt}\\ 
    &\le \frac{L-1}{\CL} \para{\tilde{\Psi}_{{L}, 2}}^{1/2}\para{\tilde{\Phi}_{L, 4}}^{(1/4)}\para{\Psi_{L, 8}}^{(1/8)}\para{\tilde{\Psi}_{L, 8}}^{(1/8)}\EP{\dLt}\\ 
    &\le \frac{L-1}{\CL} \tilde{\Psi}_{L, 2}\tilde{\Phi}_{L, 4}\Psi_{L, 8}\tilde{\Psi}_{L, 8}\EP{\dLt}.
\end{align*}

On the other hand, let consider the derivative of $\tilde{\Phi}^l_L(t)$.
We also showed the following inequality from the proof of Lemma~\ref{lem:l2-bound}.
\begin{align*}
    &\sum^{L-1}_{i=1} \sum^{i-1}_{l=0}\frac{L-1}{i-1} \ab{\W_i^{l+2}} \left(\EP{\fulz}+\EP{\fult-\fulz}\right)\left(\ab{\fulz}+\ab{\fult-\fulz}\right)\\
    &\quad\times \left(\EP{\W_L^{l+2}(0)}+\EP{\W_L^{l+2}(t)-\W_L^{l+2}(0)}\right)\\
    &\le (L-1)^2\CL^2\para{\tilde{\Psi}_{L, 2}}^{1/2}\para{\Phi_{L, 8}}^{1/8}\para{\Psi_{L, 8}}^{1/4}\para{\tilde{\Phi}_{L, 8}}^{1/8}.
\end{align*}
Thus, from Lemma~\ref{lem:bound derivative of l2 norm},
\begin{align*}
    \partial_t \tilde{\Phi}^{l}_L(t) &\le\frac{N^{1/2}}{(L-1)\CL^{3}} \sum_{i=0}^{l-1} \EP{ f^{i}_{\theta^{(i)}(t)}(x)}\ab{ f^{i}_{\theta^{(i)}(t)}(x')}\ab{\W_l^{i+2}}\EP{(\W_L^{i+2})^{\intercal}} \EP{\dLt}\\
    &\quad +\frac{N^{1/2}}{(L-1)\CL^{3}} \sum_{i=0}^{l-1}\ab{\W_l^{i+2}}\EP{(\W_L^{i+2})^{\intercal}} \EP{\dLt}\\
    &\le \frac{2N^{1/2}(L-1)}{\CL}\para{\tilde{\Psi}_{L, 2}}^{1/2}\para{\Phi_{L, 8}}^{1/8}\para{\Psi_{L, 8}}^{1/4}\para{\tilde{\Phi}_{L, 8}}^{1/8}\EP{\dLt}\\
    &\le \frac{2N^{1/2}(L-1)}{\CL}\tilde{\Psi}_{L, 2}\Phi_{L, 8}\Psi_{L, 8}\tilde{\Phi}_{L, 8}\EP{\dLt}.
\end{align*}
\end{proof}

%%%%%%%%%%%%%%%%%%%%%%%%%%%%%%%%%%%%%%%%%%%%%%%%%%%%%%%%%%%%%%%%%%%%%%%%%%%%%%%%%%%%%%%%%%
%%% CNTK
%%%%%%%%%%%%%%%%%%%%%%%%%%%%%%%%%%%%%%%%%%%%%%%%%%%%%%%%%%%%%%%%%%%%%%%%%%%%%%%%%%%%%%%%%%
\newpage
\section{Proof of Theorems of NTK for CNN}
% Define operator $\phi_{ij}(x) = [x]_{i-1:i+1, j-1:j+ 1}$ and $x_{\psi_{tu}} = [x]_{t-1:d+t-1, u-1:d+u-1}$ where $\psi_{tu}$ is set of coordinate and $x_{p, q}=0, \phi_{p,q}(y)=0$ if at least one of $p$ or $q$ is less than 1 or greater than $d$.
\subsection{Preliminaries}\label{pre::index}
We define the Jacobian of tensors through appropriate vectorization:
if $u \in \fR^{P \times Q \times R}$ and $w \in \fR^{P \times Q \times R \times S}$, then 
\[
\partial_{w} u = \partial_{\ve{w}}\ve{u},
\]
where $\ve{\cdot}$ that maps a tensor to a vector element-wise.
In the following, we use this vectorization and the Jacobian defined through vectorization implicitly.
% We will not always clarify this vectorization in calculation or proof but Jacobian should follow dimension of $\partial_{\ve{w}}\ve{u}$. 
% For instance, if we write $\partial_u w = v$ where $u \in R$, $w \in \fR^{P \times Q \times R \times S}$ and $v \in \fR^{P \times Q \times R}$, previous formula should be read as $\partial_{\ve{u}}\ve{w} = \ve{v}$.

To index 3D and 4D tensors, we use notation inspired by the slice notation of PyTorch.
However, our notation differs in that indices starts from $1$ and our slice includes the last element (i.e., $w_{p,q,r_1:r_2,s_1:s_2 }$ contains $w_{p, q, r_2, s_2}$).
For a 4D tensor $w \in \fR^{P \times Q \times R \times S}$, 
we define a slice $w_{p,q,r_1:r_2,s_1:s_2 } \in \fR^{\para{r_2-r_1+1} \times \para{s_2-s_1+1}}$,
where $\para{w_{p,q,r_1:r_2,s_1:s_2}}_{i,j}=w_{p,q,r_1-1+i,s_2-1+j}$.
Also define $w_{p, :, :, :} = w_{p, 1:Q, 1:R, 1:S}$.
Slicing for 3D tensors is defined analogously.
For a function $f$, where the output is a 3D tensor (i.e., $f(x) \in \fR^{P \times Q \times R}$ for an input $x$), 
we define a slice of function $f$ similarly, i.e., 
$f_{p,q_1:q_2,r_1:r_2} (x) = \para{f(x)}_{p, q_1:q_2, r_1:r_2}\in \fR^{(q_2-q_1+1) \times (r_2-r_1+1)}$ and $f_{p, :, :}(x) = f_{p,1:Q,1:R} (x)$.
For the sake of notational simplicity, we often omit commas (i.e., $w_{p,q,r,s} = w_{pqrs}$),
and colons (i.e., $f_{p,:,:}(x) = f_p(x)$). 

% and 2) our index starts from 1.
% If the slice contains all indices, we omit start and stop indices, i.e., $w_{p,q,:,:} = w_{p,q,1:R,1:S}$.
%
% Note that our definition of indexing and slicing differs from PyTorch convention in minor details:

% , \quad \partial_{w^l_{c'ctu}} f^l=\mathrm{vec}\para{\sigma(f^{l}(x))_{\psi_{tu}} \odot M_r} $\odot$ is element wise matrix product, and $\para{M_c}_{s,:,:}=\mathbf{1}_{\{c=s\}}$.
% Thus we have 
% \[sum_{c',c,t,u} \partial_{w^{l+1}_{c',c,t,u}} f^{l+1}_{\theta^{(l+1)}(t)}(x) \para{\partial_{w^{l+1}_{c',c,t,u}}f^{l+1}_{\theta^{(l+1)}(t)}(x')}^{\intercal}\]

% and this implies 
% \begin{align*}
% \sum_i\partial_{b_i^{l+1}} f^{l+1}(x) \para{\partial_{b_i^{l+1}}f^{l+1}(x')}^{\intercal}= \begin{bmatrix} \mathbbm{1}_{d^2 \times d^2} & \textbf{0} &\textbf{0}\\ \textbf{0} & \mathbbm{1}_{d^2 \times d^2} & \textbf{0} \\ \textbf{0} & \textbf{0} & \mathbbm{1}_{d^2 \times d^2}  \end{bmatrix}   
% \end{align*}

\subsection{Proof of Theorem~\ref{thm:4}}
For simplicity, we use $f^l\equiv f^l_{\theta^{(l)}}$ if it is clear from the context. 
\begin{proof}[Proof of Theorem~\ref{thm:4}]
For notational simpicity, we let
\[\W^k_l(x,t) = \partial_{f^{l-1}}f^l(x)\cdots \partial_{f^{k-1}}f^k(x). \]
for $2 \le k, l \le L$, where $\W^k_l(x,t)=1$ if $k>l$. Similar to MLP, we have a recursive relation for kernel 
\begin{align*}
    \tilde{\Theta}^L_t(x, x')=&\frac{1}{L\CL^2}\sum^{L-1}_{l=0} \Bigg[\W_L^{l+2}(x,t) \Bigg(\sum_{c',c,s,u} \partial_{w^{l+1}_{c'csu}} f^{l+1}_{\theta^{(l+1)}(t)}(x) \para{\partial_{w^{l+1}_{c'csu}}f^{l+1}_{\theta^{(l+1)}(t)}(x')}^{\intercal}
    \\& +\sum_i\partial_{b_i^{l+1}} f^{l+1}_{\theta^{(l+1)}(t)}(x) \para{\partial_{b_i^{l+1}}f^{l+1}_{\theta^{(l+1)}(t)}(x')}^{\intercal}\Bigg)\W_L^{l+2}(x',t)^{\intercal}.\Bigg]
\end{align*}

By calculation, the following holds: 
\begin{align*}
    \partial_{w^l_{c'csu}} f^l_{r, i,j} &= \partial_{w^l_{c'csu}} \sum_{s=1}^{n_{l-1}}\para{ w^l _{r,s}*\sigma(f^{l-1}_s)+\mathbbm{1}_{h \times h}b^l_r}_{i,j}\\
    &=\partial_{w^l_{c',c,s,u}} \sum_{s=1}^{n_{l-1}} \sum_{q=-1}^1 \sum_{p=-1}^1 w_{c',s,p+2, q+2}\sigma(f^{l-1}_{s,p+i, q+j}) \mathbf{1}_{\{r=c'\}}\\
    &=\partial_{w^l_{c',c,s,u}}\sum_{q=-1}^1 \sum_{p=-1}^1 w_{c',c,p+2, q+2}\sigma(f^{l-1}_{c,p+i, q+j})\mathbf{1}_{\{r=c'\}}\\
    &=\sigma(f^{l-1}_{c,s+i-2, u+j-2})\mathbf{1}_{\{r=c'\}}.
\end{align*}
for $1 \le l \le L$, where $f_{pqr}=0$ if the index is out of bounds.
This implies 
\[\partial_{w^l_{c'csu}} f^l_{r}= \sigma(f^{l-1}_{c}(x))_{\psi_{su}}\mathbf{1}_{\{r=c'\}}, \quad \partial_{w^l_{c'csu}} f^l= P_{c'}\para{\sigma(f_c^{l-1}(x))_{\psi_{su}}},\]
where $P_{c'}(\cdot): \fR^{d \times d} \rightarrow \fR^{n_l \times d \times d}$ is operation such that $\para{P_{c'}(f^{l-1}_{c})}_q=f^{l-1}_{c}(x)\mathbf{1}_{\{c'=q\}}$. (As mentioned in~\ref{pre::index} Preliminaries, we denote $P_{c'}(M) = \ve{P_{c'}(M)}$.)
Also
\begin{align*}
    \partial_{b^l_k} f^l_{r, i,j} &= \partial_{b^l_k} \sum_{s=1}^{n_{l-1}}\para{ w^l _{r,s}*\sigma(f^{l-1}_s)+\mathbbm{1}_{d \times d}b^l_r}_{i,j}\\
    &= \mathbf{1}_{\{r=k\}}\\
\end{align*}
for $1 \le l \le L$. Then if we let  
\begin{align*}
B^l=\sum_i\partial_{b_i^{l+1}} f^{l+1}(x) \para{\partial_{b_i^{l+1}}f^{l+1}(x')}^{\intercal}= \begin{bmatrix} \mathbbm{1}_{d^2 \times d^2} & \textbf{0} &\textbf{0}\\ \textbf{0} & \mathbbm{1}_{d^2 \times d^2} & \textbf{0} \\ \textbf{0} & \textbf{0} & \mathbbm{1}_{d^2 \times d^2}  \end{bmatrix}, \quad B^{L-1}= \mathbbm{1}_{d^2 \times d^2}  
\end{align*}
for $0 \le l \le L-2$, we have following lemma.
\begin{lemma}\label{lemm:recursive}
For any $x, x'$, and $t\ge 0$,
\begin{align*}
    \tilde{\Theta}^L_t(x, x')
    &=\frac{1}{L\CL^2}\sum^{L-1}_{l=0} \sum_{c'csu}\W_L^{l+2}(x,t)  P_{c'}\para{\sigma\para{\para{f_{\theta(t)}^{l-1}(x)}_c}_{\psi_{su}}} \para{\frac{1}{\CL}P_{c'}\para{\sigma\para{\para{f_{\theta(t)}^{l-1}(x')}_c}_{\psi_{su}}}}^{\intercal}\W_L^{l+2}(x',t)^{\intercal}\\&+\frac{1}{L\CL^2}\sum^{L-1}_{l=0} \W_L^{l+2}(x,t)B^l\W_L^{l+2}(x',t)^{\intercal},
\end{align*}
where we define $\sigma(f^{0}(x))=x$ for convenience.
\end{lemma}
% \begin{proof}[Proof of Lemma~\ref{lemm:recursive}]
% By definition, 
% \begin{align*}
%     &\tilde{\Theta}^L_t(x, x')\\
%     &=\frac{1}{L\CL^2}\sum^{L-1}_{l=0} \left[\W_L^{l+2}(x,t) \left(\sum_{c'cpq} \partial_{w^{l+1}_{c'cpq}} f^{l+1}_{\theta^{(l+1)}(t)}(x) \para{\partial_{w^{l+1}_{c'cpq}}f^{l+1}_{\theta^{(l+1)}(t)}(x')}^{\intercal}
%     + \sum_i\partial_{b_i^{l+1}} f^{l+1}_{\theta^{(l+1)}(t)}(x) \para{\partial_{b_i^{l+1}}f^{l+1}_{\theta^{(l+1)}(t)}(x')}^{\intercal}\right)\W_L^{l+2}(x',t)^{\intercal}\right]\\
%     &=\frac{1}{L\CL^2}\sum^{L-1}_{l=0} \sum_{c'cpq}\W_L^{l+2}(x,t)  P_{c'c}\para{\sigma(f^{l}(x))_{\psi_{tu}}} \para{P_{c'c}\para{\sigma(f^{l1}(x'))_{\psi_{tu}}}}^{\intercal}\W_L^{l+2}(x',t)^{\intercal}+\frac{1}{L\CL^2}\sum^{L-1}_{l=0} \W_L^{l+2}(x,t)B^l\W_L^{l+2}(x',t)^{\intercal}
%     \end{align*}
% \end{proof}
% where $\tilde{\Theta}^L_t in \mathbb{R}^{d^2 \times d^2}$.
Also our initialization ($t=0$) implies
\begin{align}
f^1_{1} &=C_L x, \quad f^1_{2}=w^1_{2,1}*x+\mathbbm{1}_{d \times d}v^1 , \quad f^1_{3}=\CL\mathbbm{1}_{d \times d} \nonumber\\
f^l_{1} &=C_L x, \quad f^l_{2}=w^l_{2,1}*C_L x+\mathbbm{1}_{d \times d}v^l, \quad  f^l_{3}=\CL \mathbbm{1}_{d \times d} \label{eq:initialization f}\\
f^L_{1} &= 0_{d \times d} \nonumber
\end{align}
for $2 \le l \le L - 1$. Note that $\mathbbm{1}_{d\times d}\in \fR^{d\times d}$ is a matrix whose entries are all 1.

Since, for $1 \le l \le L$,
\begin{align*}
    \partial_{f^{l-1}_{c,i,j}}f^l_{c',s,u}&=\partial_{f^{l-1}_{c,i,j}}\sum_{c=1}^{n_{l-1}} \sum_{q=-1}^1 \sum_{p=-1}^1 w^l_{c',c,p+2, q+2}\sigma(f^{l-1}_{c, p+s, q+u}(x))+b^l_{c'}\\
    &=w^l_{c',c,i-s+2, j-u+2}\dot{\sigma}(f^{l-1}_{c, i, j})\mathbf{1}_{\{s-1\le i\le s+1\}}
\mathbf{1}_{\{u-1\le j\le u+1\}}
\end{align*}
is nonzero for $(c',c)=(1,1), (2,1), (3,3)$ at initialization, we have
\[\partial_{f^{L-1}_{\theta^{L-1}(0)}}f^L_{\theta^{L}(0)}(x) = \begin{bmatrix} 0_{d^2 \times d^2} \quad 0_{d^2 \times d^2} \quad I_{d^2}  \end{bmatrix}, \quad \partial_{f^{l-1}_{\theta^{l-1}(0)}}f^l_{\theta^l(0)}(x)= \begin{bmatrix} I_{d^2} & 0_{d^2 \times d^2} &0_{d^2 \times d^2}\\ D^l & 0_{d^2 \times d^2} & 0_{d^2 \times d^2} \\ 0_{d^2 \times d^2} & 0_{d^2 \times d^2} & I_{d^2}  \end{bmatrix}\]
for $2 \le l \le L - 1 $ and some $D^l \in \fR^{d^2 \times d^2}$. 
% which satisfy 
% $D^l_{ijtu} = w^l_{2,1,i-t+2, j-u+2}\dot{\sigma}(f^{l-1}_{c, i, j}(x'))\mathbf{1}_{\{t-1\le i\le t+1\}}
% \mathbf{1}_{\{u-1\le j\le u+1\}}$. 
This implies 
\begin{align}
\W^k_l(0,x)=\partial_{f^{l-1}_{\theta^{l-1}(0)}}f^l_{\theta^l(0)}(x) \label{eq:initial w}    
\end{align}
for $ 2 \le l \le L$.
Thus, the scaled NTK is given by
\begin{align*}
    \tilde{\Theta}^L_0=\frac{1}{L\CL^2}\sum_{l=0}^{L-1}\para{\mathbbm{1}_{d^2 \times d^2}+\sum_{c,s,u}\sigma(f^{l}_c(x))_{\psi_{su}}\sigma(f^{l}_c(x'))_{\psi_{su}}^{\intercal}},
\end{align*}
where $\psi_{su}$ is set of coordinate which satisfy $x_{\psi_{su}} = [x]_{s-1:d+s-2, u-1:d+u-2}$ and $x_{pq}=0$ if the index is out of bounds and we define $\sigma(f^{0}(x))=x$ for convenience. Since we apply global averaging, we have to cosider $S\para{\tilde{\Theta}^L}$. 
\begin{align*}
    S\para{\tilde{\Theta}^L}=\frac{1}{L\CL^2}\sum_{l=1}^{L-1}&\Bigg(d^2+\CL^2\sum_{s,u}S\para{(\mathbbm{1}_{d \times d})_{\psi_{su}}}^2 +\CL^2 \sum_{s,u} S\para{x_{\psi_{su}}}S\para{x'_{\psi_{su}}} \\&+\sum_{t,u} S\para{\sigma \para{w^l_{2,1}*\CL x+v^l\mathbbm{1}_{d \times d}}_{\psi_{su}}}S\para{\sigma \para{w^l_{2,1}*\CL x'+v^l\mathbbm{1}_{d \times d}}_{\psi_{su}}} \Bigg)\\&
    +\frac{1}{L\CL^2}\para{d^2+ \sum_{s,u} S\para{x_{\psi_{su}}}S\para{x'_{\psi_{su}}}}
\end{align*}
Let $g^l(x)= \left\langle  w^l_{2,1}, x\right\rangle+v^l/\CL$, where $x \in \fR^{3 \times 3}$. Then $\{g^l(x)\}_{l=0}^{L-1}$ are i.i.d.\ Gaussian process,
where $g^l_{i,j}  \sim \mathcal{GP} ( 0 ,\rho^2 \left\langle  x, x'\right\rangle+\beta^2)$.
% This is because
% \[\mathbb{E}_{w^l_{i,j},v^l}[g_{i,j}]=0, \quad \mathbb{E}_{w_{i,j}}[g^j(x)g^j(x')]= \rho^2 \left\langle  x, x'\right\rangle+\beta^2\]
% This is because
% \[\mathbb{E}_{u^j,v^j}[g^j]=0\]
% and
% \begin{align*}
% \mathbb{E}_{u^j,v^j}[g^j(x)g^j(x')]&=\mathbb{E}_{u^j,v^j}\left[\left((u^{j})^{\intercal} x+v^{j}/\CL\right) \left((u^{j})^{\intercal} x'+v^{j}/\CL \right)\right]\\
% &=\mathbb{E}_{u^j}\left[\left((u^j)^{\intercal} x\right)\left((u^j)^{\intercal} x' \right) \right]+\beta^2\\
% &= \frac{\rho^2}{d} x^{\intercal}x'+\beta^2.
% \end{align*}
% Define operator $s(X) = \sum_{i,j}X_{i,j}$ where $X \in \fR^{p \times q}$. 
Thus, as $L \rightarrow \infty$, by the law of large number, we have 
 \begin{align*}
    S\para{\tilde{\Theta}^L(x,x')} \rightarrow
    % &  \frac{1}{d^2}\sum_{t,u}\Bigg(\left\langle\mathbbm{1}_{d \times d}, (\mathbbm{1}_{d \times d})_{\psi_{tu}}\right\rangle^2+ \left\langle\mathbbm{1}_{d \times d}, x_{\psi_{tu}}\right\rangle\left\langle\mathbbm{1}_{d \times d}, x'_{\psi_{tu}}\right\rangle\\&+ \sum_{i,j \in \psi_{tu}}\sum_{ i',j'\in \psi_{tu}}\mathbb{E}_{ g \sim \mathcal{GP} ( 0 ,\rho^2 \left\langle\phi_{i,j}(x), \phi_{i',j'}(x')\right\rangle +\beta^2)}[\sigma(g(x))\sigma(g(x'))]\Bigg)\\
    %  & =
\frac{1}{d^2}\sum_{s=1}^3\sum_{u=1}^3\bigg(P_d(s,u)+d^2S(x_{\psi_{su}})S(x'_{\psi_{su}})+\sum_{ij\in \psi_{su}}\sum_{i'j'\in \psi_{su}}\mathbb{E}[\sigma(g(\phi_{ij}(x))\sigma(g(\phi_{i'j'}(x'))]\bigg),
\end{align*}
$g \sim \mathcal{GP} ( 0 ,\rho^2 \left\langle x, x'\right\rangle +\beta^2))$,  $\phi_{mn}(x)=0$ if the index is out of bounds and
\begin{align*}
p_d(s,u)=\left\{
\begin{array}{cl}
&\hspace{-0.15in}d^4, \,\,\, (s,u)=(2,2),\\
&\hspace{-0.15in}d^2(d-1)^2, \,\,\, |s-u|=1,\\
&\hspace{-0.15in}(d-1)^4, \,\,\, \text{else}.
\end{array}\right.
\end{align*}
This concludes the proof.

% Note that 
% Thus scaled NTK is given by 
% \begin{align*}
%      \tilde{\Theta}^{L}_0(x,x')=\frac{1}{(L-1)\CL^2}\para{x^{\intercal} x'+1+\sum_{j=0}^{L-1} (\CL^2x^{\intercal} x'+\sigma(\CL (u^{j})^{\intercal} x+v^{j})\sigma(\CL (u^{j})^{\intercal} x'+v^{j})+B^2+1)}I_{d_{\mathrm{out}}}.
% \end{align*}
% Because of $\Y^k_L$, we just consider part of matrix where $r=c'=3$. Then
% \begin{align*}
% &\para{\sum_{c,t,u}\psi_{tu}(\sigma(y^{l-1}_{c}(x)))\para{\psi_{tu}(\sigma(y^{l-1}_{c}(x)))}^{\intercal}}_{i,i}= 0 \, \\
%  &\text{or} \, \sum_{t,u}\para{\sigma\para{\sum_{q=-1}^1 \sum_{p=-1}^1 u^{l-1}_{p+2, q+2} \CL x_{p+i, q+i}+v^l}}\para{\sigma\para{\sum_{q=-1}^1 \sum_{p=-1}^1 u^{l-1}_{p+2, q+2} \CL x_{p+i, q+i}+v^l}}^{\intercal}
% \end{align*}
% and
\end{proof}

\subsection{Proof of Theorem~\ref{thm:5}}

Similar to MLP, we define Lyapunov functions to track intermediate weights and pre-activation values of the network. Again, for the sake of simplicity, we set the scaling factor of gradient flow by $\frac{1}{(L-1)\CL^2}$ throughout the proof.
Also, without loss of generality, we assume the norms of vectorized data points $\ab{\ve{x_i}}$ are bounded by 1 for all $1\leq i\leq N$,
and we further assume that all other vectorized inputs $x\notin\{x_1, \dots, x_N\}$ also have bounded norm, i.e., $\ab{\ve{x}}\leq 1$.
% Again, \jm{without loss of generality, we assume $\EP{x}$, $\ab{x} \le 1$. Also for convenience of calculation, we use $L-1$ instead $L$ in gradient flow. }

For $ 1 \le j \le 8$ and  $\CL>0$ satisfying $L^2/\CL\rightarrow 0$,
\begin{align*}
    \Phi_{L, j}(t) & = \frac{1}{(L-1) \CL^j}\sum^{L-1}_{l=1} \left(\EP{\fulz(x)}+\EP{\fult(x)-\fulz(x)}\right)^j\\
    \Psi_{L, 2}(t)& = \frac{1}{(L-1)^2}\sum^{ L}_{l=2} \frac{L-1}{l-1} \sum_{ i=2}^{l} \left(\EP{\W_l^i(x, 0)}+\EP{\W_l^i(x, t)-\W_l^i(x, 0)}\right)^2\\
    \Psi_{L,8}(t)& = \frac{1}{(L-1)} \sum_{l= 2}^L \left(\EP{\W_L^l(x, 0)}+\EP{\W_L^l(x, t)-\W_L^l(x, 0)}\right)^8,
\end{align*}
where $\Phi_{L, 0} =1$ and we define
\begin{align*}
    &\left(\EP{\fulz(x)}+\EP{\fult(x)-\fulz(x)}\right)^j \\ &=\left(\EP{\ve{\fulz(x)}}+\EP{\ve{\fult(x)}-\ve{\fulz(x)}}\right)^j.
\end{align*}
for all $j$ and $l$ in  $\Phi_{L,j}(t)$.
At initialization ($t=0$), we have
\begin{align*}
    \Phi_{L, j}(0) & = \frac{1}{(L-1) \CL^j}\sum^{L-1}_{l=1} \left(\EP{f^l(x)}\right)^j\\
    \Psi_{L, 2}(0)& = \frac{1}{(L-1)^2}\sum^{ L}_{l=2} \frac{L-1}{l-1} \sum_{ i=2}^{l} \left(\EP{\W_l^i(x, 0)}\right)^2\\
    \Psi_{L,8}(0)& = \frac{1}{L-1} \sum_{l= 2}^L \left(\EP{\W_L^l(x, 0)}\right)^8.
\end{align*}
From \eqref{eq:initial w} and \eqref{eq:initialization f} and imply that
$\Phi_{L,j}(0)$ and $\Psi_{L,2}(0)$ converges (in probability) to constant values $\Phi_{\infty, j}(0)$ and $\Psi_{\infty, 2}(0)$ 
by the law of large number, respectively.
Furthermore, $\Psi_{L, 8}(0)$ also converges (in probability) to constant value $\Psi_{\infty, 8}(0)$ since $\W_L^l(x, 0) = \W_L^L(x, 0)$. 
Now, we are going to prove following proposition which implies Lyapunov function remains constant during training.

% prp 1
\begin{proposition} \label{prop:pin-norm}
For $1 \le j \le 8$ and any $T>0$,
\[ \Phi_{L,j}(t)\convinp \Phi_{\infty, j}(0), \quad \Psi_{L,2}(t) \convinp\Psi_{\infty, 2}(0), \quad \Psi_{L, 8}(t) \convinp\Psi_{\infty, 8}(0)\]
uniformly for $t \in [0,T]$ as $ L \rightarrow \infty$.
\end{proposition}

The following lemma is a key step to prove the proposition which allows us to apply Gr\"onwall's Lemma.
\begin{lemma}\label{lemm:pin-bound}
For $t \geq 0$, if $f^l_{\theta^{(l)}(t)}(x)$ is element-wise nonzero for $1 \le l \le L-1$ and $x\in\{x_1,\dots, x_N\}$, then
\begin{align*}
    \partial_t \Phi_{L,j}(t) &\le \frac{162\max_l\{ n_l^4\}d^4jN^{3/2}(L-1)}{\CL}\Phi_{L,j-1}(t)\Phi_{L,8}(t)\Psi_{L, 2}(t)\Psi_{L,8}(t)\EP{\dLt}\\
    \partial_t \Psi_{L,2}(t)& \le \frac{18d^4N(L-1)}{\CL}\Phi_{L,4}(t) \para{\Psi_{L,2}(t)}^{2}(t)\Psi_{L,8}(t)\EP{\dLt}\\
    \partial_t \Psi_{L,8}(t) & \le \frac{72Nd^4(L-1)}{\CL}\Phi_{L,4}(t) \Psi_{L,2}(t)\left(\Psi_{L,8}(t)\right)^{2}\EP{\dLt}
\end{align*}
for $1 \le j \le 8$.
\end{lemma}
The proof of Lemma~\ref{lemm:pin-bound} is given in Section~\ref{app:prp4}.

\begin{proof}[Proof of Proposition~\ref{prop:pin-norm}]
% ~\ref{prp:pin-norm}]
Like proof of Theorem~\ref{thm:2}, we need to show that $\fult(x)$ is element-wise nonzero for large enough $L$ for all $t\in[0, T]$, $1 \le l \le L-1$, and $x\in\{x_1,\dots, x_N\}$. This element-wise nonzero assumption also holds if $L$ is large enough, and the proof is given in Section~\ref{appe:not crossing}. From Lemma~\ref{app:prp4}, we get 
\begin{align*}
    \partial_t \left(\sum_{j=1}^8\Phi_{L,j}(t)+ \Psi_{L,2}(t)+  \Psi_{L,8}(t)\right) 
    &\le \frac{1296\max_l\{ n_l^4\}d^4N^{3/2}(L-1)}{\CL} \left(\sum_{j=1}^8\Phi_{L,j}(t)+ \Psi_{L,2}(t)+ \Psi_{L,8}(t)\right)^4\EP{\dLt},
\end{align*}
which implies 
\begin{align*}
     \Gamma_L(t) &\le \Gamma_L(0)+ \frac{1296\max_l\{ n_l^4\}d^4jN^{3/2}(L-1)}{\CL}\int_0^{t} \Gamma_L(s)^4 \EP{\delta_s^L}\,ds,
\end{align*}
where $\Gamma_L(t)= \sum_{j=1}^8\Phi_{L,j}(t) + \Psi_{L,2}(t)+ \Psi_{L,8}(t)$.
By Gr\"onwall's lemma with $z(t)=\Gamma_L(t), \, a(t)=\Gamma_L(0), \, b(t)=\frac{1296\max_l\{ n_l^4\}d^4N^{3/2}(L-1)}{\CL}, \, k(s)=\EP{\delta^L_s} $,
\[\Gamma_L(t) \le \Gamma_L(0) \left\{ 1- \frac{3888\max_l\{ n_l^4\}d^4N^{3/2}(L-1)}{\CL}\Gamma_L(0)^{3} \int_0^{t} \EP{\delta^L_s}\,ds  \right\}^{-\frac{1}{3}} \]
for $0 \le t\leq \beta_L$.
Recall that $\Gamma_L(0)$ converge to constant and we assumed stochastically
boundness of $\int_0^{T} \EP{\delta_s^L}\,ds$.
Thus, if $L^2/\CL\rightarrow 0$, then $\frac{3888\max_l\{ n_l^4\}d^4N^{3/2}(L-1)}{\CL}\Gamma_L(0)^{3} \int_0^{t} \EP{\delta^L_s}\, ds$ converge to $0$ in probability,
which implies $\Gamma_\infty(t) \le \Gamma_\infty(0)$.
On the other hand, $\Gamma_L(t) \ge \Gamma_L(0)$ by construction and  $\beta_L\rightarrow \infty$ as $L \rightarrow \infty$.
Thus, $\Gamma_L(t)$ converges to $\Gamma_\infty(0)$ uniformly in probability for $t\in[0, T]$.
This concludes the proof.
\end{proof}

Next, we consider similar Lyapunov functions but under $\ell_2$ norm.
This is because scaled NTK is not restricted to the dataset.
For $1 \le j \le 8$,
\begin{align*}
    \tilde{\Phi}_{L, j}(x, t) & = \frac{1}{(L-1) \CL^j}\sum^{L-1}_{l=1} \left(\ab{\fulz(x)}+\ab{\fult(x)\fulz(x)}\right)^j\\
    \tilde{\Psi}_{L, 2}(x,t)& = \frac{1}{(L-1)^2}\sum^{ L}_{l=2} \frac{L-1}{l-1} \sum_{ i=2}^{l} \left(\ab{\W_l^i(x, 0)}+\ab{\W_l^i(x, t)-\W_l^i(x, 0)}\right)^2\\
    \tilde{\Psi}_{L,8}(x, t)& = \frac{1}{L-1} \sum_{l= 2}^L \left(\ab{\W_L^l(x,0)}+\ab{\W_L^l(x,t)-\W_L^l(x,0)}\right)^8,
\end{align*}
where $\tilde{\Phi}_{L, 0} =1$ and we define  
\[\left(\ab{\fulz(x)}+\ab{\fult(x)-\fulz(x)}\right)^j = \left(\ab{\ve{\fulz(x)}}+\ab{\ve{\fult(x)}-\ve{\fulz(x)}}\right)^j.\]
for all $j$ and $l$ in $\tilde{\Phi}_{L,j}(t)$.

Note that Lyapunov functions under $\ell_2$-norm are functions of $(x, t)$.
Similar to Lyapunov functions under $p^{in}$-norm, 
at initialization ($t=0$), we have $\tilde{\Phi}_{L,j}(x, 0)\convinp \tilde{\Phi}_{\infty, j}(x, 0), \, \tilde{\Psi}_{L,2}(x, 0) \convinp \tilde{\Psi}_{\infty, 2}(x, 0), \, \tilde{\Psi}_{L, 8}(x, 0) \convinp \tilde{\Psi}_{\infty, 8}(x, 0)$, where $\tilde{\Phi}_{L, \infty}(x, 0), \tilde{\Psi}_{\infty, 2}(x, 0), \tilde{\Psi}_{\infty, 8}(x, 0)$ are constant by the law of large number.
The similar proposition holds, which implies the intermediate weights and pre-activation values are invariant during training in $\ell_2$-norm sense.
% prp 2
\begin{proposition}\label{prop:l2-norm}
% \label{prp:l2-norm}
For $1 \le j \le 8$ and any $T>0$,
\[ \tilde{\Phi}_{L,j}(x, t) \convinp \tilde{\Phi}_{\infty, j}(x, 0), \quad \tilde{\Psi}_{L,2}(x, t) \convinp\tilde{\Psi}_{\infty, 2}(x, 0), \quad \tilde{\Psi}_{L, 8}(x, t) \convinp \tilde{\Psi}_{\infty, 8}(x, 0)\]
uniformly  for $t \in [0,T]$ as $ L \rightarrow \infty$.
\end{proposition}
The following lemma, which corresponds to Lemma~\ref{lemm:pin-bound}
, allows us to apply Gr\"onwall's lemma.
\begin{lemma}\label{lemm:l2-bound}
For $t \geq 0$ and $1 \le j \le 8$, and $x\in\fR_+^{d \times d}$, if $f^l_{\theta^{(l)}(t)}(x)$ is element-wise nonzero for all $1 \le l \le L-1$, then
\begin{align*}
    \partial_t \tilde{\Psi}_{L,2}(x,t)&\le \frac{18d^4(L-1)}{\CL} \para{\tilde{\Psi}_{L,2}(x,t)}^{2}\Phi_{L, 4}(t)\Psi_{L,8}(t)\EP{\dLt}\\
    \partial_t \tilde{\Phi}_{L, j}(x,t) &\le \frac{j162d^4\max_l\{ n_l^4\}N^{1/2}(L-1)}{\CL}\tilde{\Phi}_{L, j-1}(t)\tilde{\Phi}_{L,8}(x,t)\tilde{\Psi}_{L, 2}(x,t)\Phi_{L,8}(t)\Psi_{L,8}(t)\EP{\dLt}\\
    \partial_t \tilde{\Psi}_{L, 8}(x,t)& \le \frac{72d ^4(L-1)}{\CL}\tilde{\Phi}_{L,4}(x,t) \tilde{\Psi}_{L,2}(x,t)\para{\tilde{\Psi}_{L,8}(x,t)}^{2}\Psi_{L,8}(t)\EP{\dLt}.
\end{align*}
\end{lemma}
The proof of Lemma~\ref{lemm:l2-bound} 
is given in Section~\ref{app:prp5}.

\begin{proof}[Proof of Proposition~\ref{prop:l2-norm}]
As we discussed in the proof of Proposition~\ref{prop:pin-norm}, $\fult(x)$ is element-wise nonzero for large enough $L$ for all $t\in[0, T]$, $1 \le l \le L-1$, and $x\in\fR_+^{d \times d}$. From Lemma~\ref{lemm:l2-bound},
\begin{align*}
\tilde{\Psi}_{L,2}(x, t) \le \tilde{\Psi}_{L,2}(x, 0) +\frac{18d^4(L-1)}{\CL}\int^{t}_0{\para{\tilde{\Psi}_{L,2}(x, s)}^{2}\Phi_{L, 4}(s)\Psi_{L,8}(s)\EP{\delta^L_s}\,ds.}
\end{align*}
Then using Gr\"onwall's lemma where $z(t)=\tilde{\Psi}_{L,2}(x, t), \, a(t)=\tilde{\Psi}_{L,2}(x, 0), \, b(t)=\frac{18d^4(L-1)}{\CL}, \, k(s)=\Phi_{L,4}(s)\Psi_{L,8}(s)\EP{\delta^L_s} $, we get
\[\tilde{\Psi}_{L, 2}(x, t) \le \tilde{\Psi}_{L, 2}(x, 0) \left\{ 1-\frac{18d^4(L-1)}{\CL}\tilde{\Psi}_{L, 2}(x, 0) \int^t_0 \Phi_{L, 4}(s)\Psi_{L, 4}(s)\EP{\delta^L_s}\,ds  \right\}^{-1} \]
for all $0 \le t \le \beta_L$.
Similar to the proof of Proposition ~\ref{prop:pin-norm}
, if $L^2/\CL\rightarrow 0$, then
we get $\tilde{\Psi}_{L,2}(t) \convinp \tilde{\Psi}_{\infty,2}(0)$ uniformly for $t\in[0, T]$.

On the other hand, Lemma~\ref{lemm:l2-bound} 
also implies
\begin{align*}
&\sum^8_{j=1}\tilde{\Phi}_{L,j}(x, t) \\&\quad \le \sum^8_{j=1}\tilde{\Phi}_{L,j}(x, 0) +\int^{t}_0{ \frac{1296d^4\max_l\{ n_l^4\}N^{1/2}(L-1)}{\CL} \para{ \sum^8_{j=1}\tilde{\Phi}_{L, j-1}(x, s)}\tilde{\Phi}_{L,8}(x, s)\tilde{\Psi}_{L, 2}(x, s)\Phi_{L,8}(s)\Psi_{L,8}(s)\EP{\delta^L_s}\,ds}\\
&\quad \le \sum^8_{j=1}\tilde{\Phi}_{L,j}(x, 0) +\int^{t}_0{ \frac{1296d^42\max_l\{ n_l^4\}N^{1/2}(L-1)}{\CL} \para{ \sum^8_{j=1}\tilde{\Phi}_{L, j}(x, s)}^2\tilde{\Psi}_{L, 2}(x,s)\Phi_{L,8}(s)\Psi_{L,8}(s)\EP{\delta^L_s}\,ds}
\\
& \tilde{\Psi}_{L,8}(x,t) \le \tilde{\Psi}_{L,8}(x, 0) +\int^{t}_0{ \frac{72d^4(L-1)}{\CL}\para{\tilde{\Psi}_{L,8}(x, s)}^{2}\tilde{\Phi}_{L,4}(x, s) \tilde{\Psi}_{L,2}(x, s)\Psi_{L,8}(s)\EP{\delta^L_s}\,ds}.
\end{align*}
We can similarly apply Gr\"onwall's lemma to show that $\sum_{j=1}^8 \tilde{\Phi}_{L, j}(x, t)$ and $\tilde{\Psi}_{L, 8}(x, t)$
converge to constant values.
\end{proof}

Now, we consider our last and core Lyapunov function which contains intermediate weight and pre-activation values from a single layer.
Define
\begin{align*}
    \tilde{\Psi}_L^{k}(x,t)& = \ab{\W_L^k(x,0)}+\ab{\W_L^k(x,t)-\W_L^k(x,0)}\\
    \tilde{\Phi}_L^{l}(x,t) & = \frac{1}{ \CL}\para{\ab{\fulz(x)}+\ab{\fult(x)-\fulz(x)}}
\end{align*}
for $1 \le l \le L$ and $2 \le k \le L$ and we define 
\[\ab{\fulz(x)}+\ab{\fult(x)-\fulz(x)} =\ab{\ve{\fulz(x)}}+\EP{\ve{\fult(x)}-\ve{\fulz(x)}}.\]
for all $l$ in $\tilde{\Phi}^l_{L}(x,t)$.

At initialization, $\tilde{\Psi}^k_L(x, 0)$ and $\tilde{\Phi}_L^l(x, 0)$ are stochastically bounded.
Then the following proposition implies the invariance of Lyapunov functions during training.
% prp 3
\begin{proposition}\label{prop:layerwise}
For $1 \le l \le L$, $2 \le k \le L$ and any $T>0$,
\[\tilde{\Psi}^{k}_L(x,t) \convinp \tilde{\Psi}^{k}_{\infty}(x,0) \quad \tilde{\Phi}^{l}_{L}(x,t) \rightarrow \tilde{\Phi}^{l}_{\infty}(x,0).\]
uniformly for $t\in[0, T]$ as $L \rightarrow \infty$.
\end{proposition}

% This implies individual intermediate weights and values are effectively unchanged at each layer. 
% Recall that this layer-wise invariance is straightforward in an infinite width network with finite depth.
% However, in our setting, an infinitely deep network has composition of infinitely many weights,
% which requires much careful analysis including the following lemma as well as previous propositions.
\begin{lemma}\label{lemm:layer-bound}
For $t \geq 0$ and $x\in\fR_+^{d \times d}$, if $f^l_{\theta^{(l)}(t)}(x)$ is element-wise nonzero for all $1 \le l \le L-1$, then
\begin{align*}
    \partial_t\tilde{\Psi}^{k}_L(x,t) &\le\frac{9d^4(L-1)}{\CL} \Psi_{L,8}(t)\tilde{\Phi}_{L,4}(x,t)\tilde{\Psi}_{L,2}(x,t)\tilde{\Psi}_{L,8}(x,t)\EP{\dLt}\\
    \partial_t\tilde{\Phi}^{l}_L(x,t) &\le \frac{81d^4\max_l\{n_{l}^4\}N^{1/2}(L-1)}{\CL}\Phi_{L,8}(t)\Psi_{L,8}(t)\tilde{\Phi}_{L,8}(x,t)\tilde{\Psi}_{L, 2}(x,t)\EP{\dLt}.
\end{align*}
for $1 \le l \le L$ and $2 \le k \le L$.
\end{lemma}
The proof of Lemma~\ref{lemm:layer-bound}
is given in Section~\ref{app:prop6}.
\begin{proof}[Proof of Proposition~\ref{prop:layerwise}]
 Again, we can assume that $\fult(x)$ is element-wise nonzero for large enough $L$ for all $t\in[0, T]$, $1 \le l \le L-1$, and $x\in\fR_+^{d \times d}$. Lemma~\ref{lemm:layer-bound} 
implies
\begin{align*}
\tilde{\Psi}^{k}_L(x, t) &\le \tilde{\Psi}^{k}_L(x, 0) +\int^{t}_0{ \frac{9d^4(L-1)}{\CL} \Psi_{L,8}(t)\tilde{\Phi}_{L,4}(x,t)\tilde{\Psi}_{L,2}(x,t)\tilde{\Psi}_{L,8}(x,t)\EP{\dLt}\,ds}\\
\tilde{\Phi}^{l}_L(x, t)&\le \tilde{\Phi}^{l}_L(x, 0)+ \int_0^{t} \frac{81d^4\max_l\{n_{l}^4\}N^{1/2}(L-1)}{\CL}\Phi_{L,8}(t)\Psi_{L,8}(t)\tilde{\Phi}_{L,8}(x,t)\tilde{\Psi}_{L, 2}(x,t)\EP{\dLt}\,ds.
\end{align*}

Unike previous proofs, we do not need Gr\"onwall's Lemma.
From Proposition~\ref{prop:pin-norm} 
and Proposition~\ref{prop:l2-norm}
, all terms in RHS 
(such as $\Psi_{L, 8}(t), \tilde{\Phi}_{L, 4}(t), \tilde{\Psi}_{L, 2}(t)$, etc.) converge to constants in probability.
Since $L^2/\CL\rightarrow 0$, integral terms converge to zero as $L \rightarrow \infty$. 
On the other hand,
it is clear that $\tilde{\Psi}_L^k(x, t) \ge \tilde{\Psi}_L^k(x, 0)$ and $\tilde{\Phi}_L^l(x, t) \ge \tilde{\Phi}_L^l(x, 0)$ by construction. 
Thus, $\tilde{\Psi}_L^k(x, t)$ and $\tilde{\Phi}_L^l(x, t)$ converges to $\tilde{\Psi}_L^k(x, 0)$ and $\tilde{\Phi}_L^l(x, 0)$ in probability, respectively.
Since the integral terms are independent from the choice of $k$ and $l$, we get uniform convergence.

\end{proof}

Proposition~\ref{prop:layerwise} implies the following desired result 
which implies that the variation of intermediate pre-activation values and weights must be negligible.

\begin{corollary}\label{corr:finite}
For any $x$, as $L\rightarrow \infty$ , we have
\begin{align*}
    \sup_{1\leq l\leq L, t\in[0, T]} \frac{1}{ \CL}\ab{\ve{\fult(x)}-\ve{\fulz(x)}} \convinp& 0\\
      \sup_{2\leq k\leq L, t\in[0, T]} \ab{\W_L^{k}(x,t)-\W_L^{k}(x,0)} \convinp & 0.
\end{align*}
\end{corollary}
With this corollary, we are now ready to prove our main theorem, the invariance of scaled NTK.

\begingroup
\def\thetheorem{\ref{thm:5}}
\begin{theorem}[Invariance of scaled NTK] 
Let $T>0$.
Suppose $\int_0^T\EP{\delta\vert_{f_\theta^L}}\,dt$ is stochastically bounded as $L\rightarrow\infty$.
Then, for any $x, x'\in\fR^{d \times d}$, 
\[S\para{\tilde{\Theta}^L_t(x,x')}
\convinp
\tilde{\Theta}^{\infty}(x,x')
\]
uniformly for $t\in [0,T]$ as $L \rightarrow \infty$.
\end{theorem}
\addtocounter{theorem}{-1}
\endgroup

\begin{proof}
% [Proof of Theorem~\ref{thm:2}]
By definition,
\begin{align*}
    \tilde{\Theta}^L_t(x, x')
    &=\frac{1}{L\CL^2}\sum^{L-1}_{l=0} \sum_{c'csu}\W_L^{l+2}(x,t)  P_{c'}\para{\sigma\para{\para{f_{\theta(t)}^{l-1}(x)}_c}_{\psi_{su}}} \para{\frac{1}{\CL}P_{c'}\para{\sigma\para{\para{f_{\theta(t)}^{l-1}(x')}_c}_{\psi_{su}}}}^{\intercal}\W_L^{l+2}(x',t)^{\intercal}\\&+\frac{1}{L\CL^2}\sum^{L-1}_{l=0} \W_L^{l+2}(x,t)B^l\W_L^{l+2}(x',t)^{\intercal}
\end{align*}
Informally, Proposition
% ~\ref{prp:layerwise} 
implies that $f^l_{\theta^{(l)}(t)}(x)$ and $\W^{l+2}_L(x, t)$ are effectively invariant,
and therefore the kernel $\tilde{\Theta}_t^L(x, x')$ is invariant.
An additional effort is required to handle the summation of $L$ terms since $L$ also increases.
More formal proof is given in the following.

Since ReLU is 1-Lipshitz, Corollary~\ref{corr:finite} 
implies
\begin{align*}
    \sup_{1\leq l\leq L, 1\leq c\leq n_l, t\in[0, T]}
    &\Bigg\|\frac{1}{\CL}P_{c'}\para{\sigma\para{\para{f_{\theta(t)}^{l-1}(x)}_c}_{\psi_{su}}} \para{\frac{1}{\CL}P_{c'}\para{\sigma\para{\para{f_{\theta(t)}^{l-1}(x')}_c}_{\psi_{su}}}}^{\intercal}-\\&\frac{1}{\CL}P_{c'}\para{\sigma\para{\para{f_{\theta(0)}^{l-1}(x)}_c}_{\psi_{su}}} \para{\frac{1}{\CL}P_{c'}\para{\sigma\para{\para{f_{\theta(0)}^{l-1}(x')}_c}_{\psi_{su}}}}^{\intercal}\Bigg\| \convinp 0
    % \sup_{2\leq k\leq L, t\in[0, T]}
    % \ab{\W_L^k(t,x)\W_L^k(t,x')-\W_L^k(0,x)\W_L^k(0,x')} \rightarrow & 0
\end{align*}
as $L\rightarrow \infty$ for all $u$,$s$,$x$ and $x'$.
On the other hand, the norm of the difference between kernels is bounded by
\begin{align*}
    &\ab{\tilde{\Theta}^L_t(x,x')- \tilde{\Theta}^L_0(x,x')}\\ 
    & \le \frac{1}{L}\sum_{l=0}^L \sum_{c'csu}\Bigg\|\W_L^{l+2}(x,t)\left(\frac{1}{\CL}P_{c'}\para{\sigma\para{\para{f_{\theta(t)}^{l-1}(x)}_c}_{\psi_{su}}} \para{\frac{1}{\CL}P_{c'}\para{\sigma\para{\para{f_{\theta(t)}^{l-1}(x')}_c}_{\psi_{su}}}}^{\intercal}\right)\W_L^{l+2}(x',t)^{\intercal}\\
    &\quad\quad-\W_L^{l+2}(x,0)\left( \frac{1}{\CL}P_{c'}\para{\sigma\para{\para{f_{\theta(0)}^{l-1}(x)}_c}_{\psi_{su}}} \para{\frac{1}{\CL}P_{c'}\para{\sigma\para{\para{f_{\theta(0)}^{l-1}(x')}_c}_{\psi_{su}}}}^{\intercal}\right)\W_L^{l+2}(x',0)^{\intercal}\Bigg\|\\
    &\quad+\frac{1}{L\CL^2}\sum_{l=0}^{L-1}\ab{ \left[\W_L^{l+2}(x,t)B^l\W_L^{l+2}(x',t)^{\intercal}\right]-\left[\W_L^{l+2}(x,0)B^l\W_L^{l+2}(x',0)^{\intercal}\right]}\\
    & \le 27\max_{l,c}\Bigg\|\W_L^{l+2}(x,t)\left(\frac{1}{\CL}P_{c'}\para{\sigma\para{\para{f_{\theta(t)}^{l-1}(x)}_c}_{\psi_{su}}} \para{\frac{1}{\CL}P_{c'}\para{\sigma\para{\para{f_{\theta(t)}^{l-1}(x')}_c}_{\psi_{su}}}}^{\intercal}\right)\W_L^{l+2}(x',t)^{\intercal}\\
    &\quad\quad-\W_L^{l+2}(x,0)\left( \frac{1}{\CL}P_{c'}\para{\sigma\para{\para{f_{\theta(0)}^{l-1}(x)}_c}_{\psi_{su}}} \para{\frac{1}{\CL}P_{c'}\para{\sigma\para{\para{f_{\theta(0)}^{l-1}(x')}_c}_{\psi_{su}}}}^{\intercal}\right)\W_L^{l+2}(x',0)^{\intercal}\Bigg\|\\
    &\quad+\frac{1}{\CL^2}\max_l\ab{ \left[\W_L^{l+2}(x,t)B^l\W_L^{l+2}(x',t)^{\intercal}\right]-\left[\W_L^{l+2}(x,0)B^l\W_L^{l+2}(x',0)^{\intercal}\right]}.
\end{align*}
Thus, as $L\rightarrow \infty$, we have
\begin{align*}
   \sup_{t\in[0, T]} \ab{\tilde{\Theta}^L_t(x,x')- \tilde{\Theta}^L_0(x,x')} \convinp 0.
\end{align*}
Finally, Theorem~\ref{thm:5} implies $\tilde{\Theta}^L_0$ converges to $\tilde{\Theta}^{\infty}_0$ in probability,
and therefore $\tilde{\Theta}_t^L$ converges to $\tilde{\Theta}^{\infty}_0$ in probability. 
This also implies 
\begin{align*}
   \sup_{t\in[0, T]} \ab{S\para{\tilde{\Theta}^L_t(x,x')}- \para{\tilde{\Theta}^L_0(x,x')}} \convinp 0.
\end{align*}
\end{proof}
%  We briefly note that zero crossing can be proved using the similar argument in Section~\ref{app:not crossing}.
\subsection{Proof of Theorem~\ref{thm:6}}

 When the loss function is quadratic loss, stochasically bounded assumption of Theorem~\ref{thm:5} also can be proved by Lemma~\ref{lemm:stochastically bound}.

\begingroup
\def\thetheorem{\ref{thm:6}}
\begin{theorem}[Equivalence between deep CNN and kernel regression]
Let $\loss(f) = \frac{1}{2}\EP{f-f^\star}^2$. Let $\tilde{\Theta}^{\infty}$ be positive definite.
If $f_t$ follows the kernel gradient flow with respect to $\tilde{\Theta}^{\infty}$, then for any $x \in \fR^{d\times d}_+$,
% \vspace{-0.02in}
\[
\lim_{t \rightarrow \infty}f_t (x) = f_{\mathrm{ntk}}(x).
\]
\end{theorem}
\addtocounter{theorem}{-1}
\endgroup
\begin{proof}[Proof of Theorem ~\ref{thm:6}]
This can be proved using the exact same argument in the proof of Theorem~\ref{thm:3}.
The only difference is the definition of $\tilde{\Theta}^{\infty}$, which does not affect the proof.
\end{proof}

\section{Zero-crossing does not happen during training CNN}\label{appe:not crossing}
Similar to Section~\ref{app:not crossing}, 
% While proving key statements in the proof of Theorem~\ref{thm:5} (Proposition~\ref{prop:pin-norm}, Proposition~\ref{prop:l2-norm}, and Proposition~\ref{prop:layerwise}),
% we assumed that all elements of $\fult(x)$ are nonzero for $x\in\{x_1, \dots, x_N\}$, $1 \le l \le L-1$, and any input $x\in\fR_+^{d_{in}}$ while training.%  
we now show that all entries of $\{\fult(x_1), \dots \fult(x_N), \fult(x'), \fult(x'')\}$ are nonzero with high probability for any $x', x''\in \fR_+^{d\times d}$ and $T>0$.
For the sake of notational simplicity, we set $x_{N+1}=x'$ and $x_{N+2} = x''$ in the following lemma.

\begin{lemma}\label{lemm:not crossing} 
Suppose $\int_0^T\EP{\delta\vert_{f_\theta^L}}\,dt$ is stochastically bounded as $L\rightarrow \infty$.
For any $T>0$, let 
\begin{align*}
    T_L = \bigwedge^{N+2}_{i=1} \inf_t \left\{\para{\fult(x_i)}_{rpq}=0 \mbox{ for some $1 \le l \le L-1$, $r$, $p$, and $q$}\right\} \bigwedge T,
\end{align*}
where $x_{N+1}, x_{N+2}\in\fR_+^{d\times d}$ are arbitrary inputs.
Then, for any  $\epsilon>0$, there exists $L_{\max}$ such that $\Pr{T_L= T} \ge 1-\epsilon$ for all $L > L_{\max}$.
\end{lemma}
\begin{proof}[Proof of Lemma~\ref{lem:not crossing}]
For $\epsilon'>0$, let $M = 1+\epsilon'$.
Since $\int_0^T\EP{\delta\vert_{f_\theta^L}}\,dt$ is stochastically bounded as $L\rightarrow \infty$,
there exists $K_0>0$ and large enough $L_0>0$ such that
\begin{align*}
\Pr{\int_0^T\EP{\delta\vert_{f_\theta^L}}\,dt > K_0} < \frac{\epsilon}{3}
\end{align*}
for all $L\ge L_0$.
We define a complement of such event by
\begin{align*}
    E_0 =  \left\{\int_0^T\EP{\delta\vert_{f_\theta^L}}\,dt \leq K_0\right\},
\end{align*}
where $\Pr{E_0} \geq 1-\frac{\epsilon}{3}$.

On the other hand, recall that 
$\tilde{\Psi}_{\infty, 2}(x_i, 0), \tilde{\Psi}_{\infty, 8}(x, 0), \tilde{\Phi}_{\infty, j}(x_i, 0), 
\Gamma_{\infty}(0)$ are all (non-random) constants for all $1\leq j\leq 8$ and $1\leq i\leq N+2$.
Since $L^2/\CL\rightarrow 0$ as $L\rightarrow \infty$, we have
\begin{align*}
 \frac{3888\max_l\{ n_l^4\}d^4N^{3/2}M^3(L-1)}{\CL}\Gamma_{\infty}(0)^{3}K_0\rightarrow& 0, \\
\frac{18d^4M^5(L-1)}{\CL}\tilde{\Psi}_{\infty, 2}(x_i, 0) \Gamma^2_{\infty}(0)K_0 \rightarrow& 0, \\
\frac{M^71296d^42\max_l\{ n_l^4\}N^{1/2}(L-1)}{\CL}\para{ \sum^8_{j=1}\tilde{\Phi}_{\infty, j}(x_i, 0)}\tilde{\Psi}_{\infty, 2}(0)\Gamma^2_{\infty}(0)K_0 \rightarrow& 0,\\
\frac{M^772d^4(L-1)}{\CL}\tilde{\Psi}_{\infty,8}(x_i, 0)\para{\sum^8_{j=1}\tilde{\Phi}_{\infty, j}(x_i, 0)} \tilde{\Psi}_{\infty,2}(x_i, 0)\Gamma_{\infty}(0)K_0\rightarrow& 0,
\end{align*}
as $L\rightarrow \infty$.
Thus, there exists $L_1$ such that 
\begin{align*}
\frac{3888\max_l\{ n_l^4\}d^4N^{3/2}M^3(L-1)}{\CL}\Gamma_{\infty}(0)^{3}K_0 <& 1-M^{-3}, \\
\frac{18d^4M^5(L-1)}{\CL}\tilde{\Psi}_{\infty, 2}(x_i, 0) \Gamma^2_{\infty}(0)K_0 <& 1-M^{-1}, \\
\frac{M^7 1296d^42\max_l\{ n_l^4\}N^{1/2}(L-1)}{\CL} \para{ \sum^8_{j=1}\tilde{\Phi}_{\infty, j}(x_i, 0)}\tilde{\Psi}_{\infty, 2}(0)\Gamma^2_{\infty}(0)K_0 <& 1-M^{-1},\\
\frac{M^772d^4(L-1)}{\CL}\tilde{\Psi}_{\infty,8}(x_i, 0)\para{\sum^8_{j=1}\tilde{\Phi}_{\infty, j}(x_i, 0)} \tilde{\Psi}_{\infty,2}(x_i, 0)\Gamma_{\infty}(0)K_0 <& 1-M^{-1}, 
\end{align*}
for all $L\ge L_1$.

Now, we define following events which indicate Lyapunov functions converge to constant values by the law of large numbers.
\begin{align*}
    E^{(L)}_{1} =& \left\{\Gamma_{L}(0) \leq M\Gamma_{\infty}(0)\right\} \\
    E^{(L)}_{2} =& \bigcup^{N+2}_{i=1}  \left\{\tilde{\Psi}_{L,8}(x_i,0)\leq M\tilde{\Psi}_{\infty,8}(x_i,0)\right\}\\
    E^{(L)}_{3} =& \bigcup^{N+2}_{i=1}  \left\{\tilde{\Psi}_{L,2}(x_i, 0) \leq M\tilde{\Psi}_{\infty,2}(x_i,0)\right\}\\
    E^{(L)}_{4} =& \bigcup^{N+2}_{i=1}\left\{\sum^8_{j=1}\tilde{\Phi}_{L,j}(x_i, 0) \leq M\sum^8_{j=1}\tilde{\Phi}_{\infty,j}(x_i, 0)\right\}.
\end{align*} 
By the law of large numbers, all events $E^{(L)}_{1}, E^{(L)}_{2}, E^{(L)}_{3}, E^{(L)}_{4}$ have probabilities converge to 1 as $L\rightarrow\infty$.
Thus, there exists $L_2$ such that 
\begin{align*}
    \Pr{E^{(L)}_{1}\cap E^{(L)}_{2}\cap E^{(L)}_{3}\cap E^{(L)}_{4}} \ge 1-\frac{\epsilon}{3}
\end{align*}
for all $L\ge L_2$.

Finally, we are interested in the event
\begin{align*}
E_5 = \bigcup^{N+2}_{i=1} \left\{\left|\para{f_{\theta(0)}^l(x_i)}_{rpq}\right| \geq \frac{L}{\CL}(K_1+1)\mbox{ for all $1 \le l \le L-1$,   $r$, $p$, and $q$}\right\},
\end{align*} 
where we specify $K_1>0$ later.
Consider a random element of intermediate pre-activation values
\begin{align*}
g^1_{pq}(x) =& \left(f^1_{\theta(0)}(x)\right)_{2pq} = \frac{1}{\CL}\left \langle  w^1_{2,1}, \phi_{pq}(x) \right \rangle +\frac{1}{\CL}v^1\\
g^l_{pq}(x) =& \left(\fulz(x)\right)_{2pq} = \left \langle  w^l_{2,1}, \phi_{pq}(x) \right \rangle+\frac{1}{\CL}v^l \mbox{ for $2 \le l \le L-1$}.
\end{align*}
Then, $g^1_{pq} \sim \mathcal{N}(0, \sigma_{1pq}^2)$ and $g^l_{pq} \sim \mathcal{N}(0, \sigma_{gpq}^2)$ for $2 \le l \le L-1$, $p$, and $q$ 
are independent Gaussian random variables with finite variances.
Since
\begin{align*}
    \Pr{|g^l_{pq}| < \frac{L}{\CL}(K+1)\mbox{ for some $1 \le l \le L-1$}}
    &\leq 2d^2(K+1)\frac{L(L-2)}{\CL}\frac{1}{\sqrt{2\pi\sigma_g^2}}+2d^2(K+1)\frac{L}{\CL}\frac{1}{\sqrt{2\pi\sigma_{1}^2}}
\end{align*}
by union bound, the probability converges to 0 as $L\rightarrow \infty$.
Thus, there exists $L_3$ such that $\Pr{E_5}\geq 1-\frac{\epsilon}{3}$ holds for all $L\ge L_3$.

Then, we would like to show the following claim.
% claim
\begin{claim}\label{claim2}
If $L\geq L_1$, and all events $E_0, E^{(L)}_1, E^{(L)}_2, E^{(L)}_3, E^{(L)}_4, E_5$ are given,
then all elements of intermediate pre-activation values never cross zeros while training for $t\in[0, T]$.
\end{claim}
If the claim holds, then for $L\geq L_{\max}=\max\{L_0, L_1, L_2, L_3\}$,
\begin{align*}
    \Pr{E_0\cap E^{(L)}_1\cap E^{(L)}_2\cap E^{(L)}_3\cap E^{(L)}_4\cap E_5}\geq 1-\epsilon,
\end{align*}
which concludes the proof of Lemma~\ref{lem:not crossing}.
In the rest of the proof, we will prove the above claim.
Suppose $L\geq L_1$ and all events $E_0, E^{(L)}_1, E^{(L)}_2, E^{(L)}_3, E^{(L)}_4, E_5$ are given.
Since all intermediate pre-activation values at initialization 
\begin{align*}
f^1_{1} &=C_L x, \quad f^1_{2}=w^1_{2,1}*x+\mathbbm{1}_{d \times d}v^1 , \quad f^1_{3}=\CL\mathbbm{1}_{d \times d} \nonumber\\
f^l_{1} &=C_L x, \quad f^l_{2}=w^l_{2,1}*C_L x+\mathbbm{1}_{d \times d}v^l, \quad  f^l_{3}=\CL \mathbbm{1}_{d \times d} 
\end{align*}
are nonzero for $2 \le l \le L-1$ and $1 \le i \le N+2$ with probability 1. 
Then, for all $t\in[0,T_L]$, Lemma~\ref{lem:pin-bound} holds which is essential in the proof of Proposition~\ref{prp:pin-norm}.
In the proof, using Gr\"onwall's lemma, we showed  
\[
\Gamma_L(t) \le \Gamma_L(0) \left\{ 1- \frac{3888\max_l\{ n_l^4\}d^4N^{3/2}(L-1)}{\CL}\Gamma_L(0)^{3}\Gamma_L(0)^{3} \int_0^{\beta_L} \EP{\delta^L_s}\,ds  \right\}^{-\frac{1}{3}}
\]
for $t \in \left[0, \beta^{(1)}_L \right]$, where 
\[
\beta^{(1)}_L = \mathrm{sup}\left\{t: \frac{3888\max_l\{ n_l^4\}d^4N^{3/2}(L-1)}{\CL}\Gamma_L(0)^{3} \int_0^{t} \EP{\delta^L_s}\, ds <1\right\}.
\]
Since $E^{(L)}_1$ is given, $ \Gamma_L(0) \le M \Gamma_{\infty}(0)$.
Then, since $E_0$ is given,
\begin{align*}
\frac{3888\max_l\{ n_l^4\}d^4N^{3/2}(L-1)}{\CL}\Gamma_{L}(0)^{3} \int_0^{t} \EP{\delta^L_s}\, ds  
\leq &\frac{3888M^3\max_l\{ n_l^4\}d^4N^{3/2}(L-1)}{\CL}\Gamma_{\infty}(0)^{3} K_0 \\
<& 1-M^{-3}
\end{align*} 
by definition of $L_1$.
This implies that $\beta_L^{(1)}=T_L$ and $\Gamma_L(t) \le M^2 \Gamma_{\infty}(0)$ if $E_0$ and $E^{(L)}_1$ are given and $L\geq L_1$.
Recall that $\Gamma_L(t)= \sum_{j=1}^8\Phi_{L,j}(t) + \Psi_{L,2}(t)+ \Psi_{L,8}(t)$,
and therefore $\Gamma_L(t)\leq M^2\Gamma_{\infty}(0)$ also implies 
$\Psi_{L,8}(t) \le M^2 \Gamma_{\infty}(0)$, $\Psi_{L,2}(t) \le M^2 \Gamma_{\infty}(0)$,
and $\Phi_{L,j}(t) \le M^2 \Gamma_{\infty}(0)$ for all $1 \le j \le 8$.

Similarly, in the proof of Proposition~\ref{prp:l2-norm}, we proved 
\[
\tilde{\Psi}_{L, 2}(x_i, t) \le \tilde{\Psi}_{L, 2}(x_i, 0) 
\left\{ 1-\frac{18d^4(L-1)}{\CL}\tilde{\Psi}_{L, 2}(x_i, 0) \int^t_0 \Phi_{L, 4}(s)\Psi_{L, 4}(s)\EP{\delta^L_s}\,ds  \right\}^{-1} 
\]
for all $ t \in \left[0, \beta^{(2)}_L \right]$, where
\[
\beta^{(2)}_L 
= \mathrm{sup}\left\{t: \frac{18d^4(L-1)}{\CL}\tilde{\Psi}_{L, 2}(x_i, 0) \int^t_0 \Phi_{L, 4}(s)\Psi_{L, 8}(s)\EP{\delta^L_s}\,ds <1\right\}.
\]
We have $\tilde{\Psi}_{L,8}(x_i,0) < M\tilde{\Psi}_{\infty,2}(x_i,0)$ if $E^{(L)}_3$ is given.
In this case, with a definition of $L_1$,
 \begin{align*}
     \frac{18d^4(L-1)}{\CL}\tilde{\Psi}_{L, 2}(x_i, 0) \int^t_0 \Phi_{L, 4}(s)\Psi_{L, 8}(s)\EP{\delta^L_s}\,ds 
     & \le  \frac{18M^5d^4(L-1)}{\CL}\tilde{\Psi}_{\infty, 2}(x_i, 0) \Gamma^2_{\infty}(0)K_0\\ 
     & <1-M^{-1}
 \end{align*}
 if $E_0$ is also given.
This implies that $\beta^{(2)}_L=T_L$ and  $\tilde{\Psi}_{L,2}(x_i, t) \le M^2 \tilde{\Psi}_{\infty,2}(x_i, 0)$,
if $E^{(L)}_3$ and $E_0$ are given and $L\geq L_1$.

In similar way,  we obtain $\sum^8_{j=1}\tilde{\Phi}_{\infty, j}(x_i, t) \le M^2 \sum^8_{j=1}\tilde{\Phi}_{\infty, j}(x_i, 0)$ for $t \in [0, T_L]$ 
if $E^{(L)}_4$ and $E_0$ are given and $L\geq L_1$.
Also, we have $\tilde{\Psi}_{L,8}(x_i, t) \le M^2 \tilde{\Psi}_{\infty,8}(x_i, 0)$ for $t \in [0, T_L]$ 
if $E^{(L)}_2$ and $E_0$ are given and $L\geq L_1$.

Finally, in the proof of Proposition~\ref{prp:layerwise}, we proved 
\begin{align*}
\tilde{\Phi}^{l}_L(x, t)&\le \tilde{\Phi}^{l}_L(x, 0)+ \int_0^{t} \frac{81d^4\max_l\{n_{l}^4\}N^{1/2}(L-1)}{\CL}\Phi_{L,8}(t)\Psi_{L,8}(t)\tilde{\Phi}_{L,8}(x,t)\tilde{\Psi}_{L, 2}(x,t)\EP{\dLt}\,ds.
\end{align*}
Then, if $E_0, E^{(L)}_1, E^{(L)}_2, E^{(L)}_3, E^{(L)}_4$ are given and $L\geq L_1$,
\begin{align*}
    &\int^{T_L}_0{ \frac{81d^4\max_l\{n_{l}^4\}N^{1/2}(L-1)}{\CL}\Phi_{L,8}(t)\Psi_{L,8}(t)\tilde{\Phi}_{L,8}(x,t)\tilde{\Psi}_{L, 2}(x,t)\EP{\dLt}\,ds.} \\
    & \quad \le \frac{81M^8d^4\max_l\{n_{l}^4\}N^{1/2}(L-1)}{\CL} \Gamma^2_{\infty}(0)
 \left(\sum_{j=1}^8\tilde{\Phi}_{\infty, j}(x_i, 0)\right) \tilde{\Psi}_{\infty, 2}(x_i, 0)K_0. 
\end{align*} 
If we let $K_1=\sup_{1\le i\le N+2} 81M^8d^4\max_l\{n_{l}^4\}N^{1/2}\Gamma^2_{\infty}(0)
 \left(\sum_{j=1}^8\tilde{\Phi}_{\infty, j}(x_i, 0)\right) \tilde{\Psi}_{\infty, 2}(x_i, 0)K_0$,
then this inequality implies
\[
\sup_{1\leq l\leq {L-1}, t\in[0, T_L]} \frac{1}{ \CL}\ab{\ve{\fult(x_i)}-\ve{\fulz(x_i)}} < \frac{L}{ \CL}K_1
\]
for all $i$.
On the other hand, if $E_5$ is given,
\[
\frac{1}{\CL}\left|\para{f_{\theta(0)}^l(x_i)}_{rpq}\right| > \frac{L}{\CL}(K_1+1)
\]
for all $i, l, r, p$, and $q$, and therefore
\[
\inf_{1\leq l\leq L-1, t\in[0, T_L]} \left|\para{\fult(x_i)}_{rpq}\right| > L
\]
for all $i,r,p,$ and $q$.
This implies $T_L=T$, which concludes the proof of Claim~\ref{claim2}.
\end{proof}

% prove theorem 2 again
Equipped with Lemma~\ref{lemm:not crossing}, we provide a brief outline of the rigorous proof of Theorem~\ref{thm:5}.
For any $T>0$, let 
\begin{align*}
    T_L = \bigwedge^{N+2}_{i=1} \inf_t \left\{\para{\fult(x_i)}_r=0 \mbox{ for some $1 \le l \le L-1$ and $r$}\right\} \bigwedge T,
\end{align*}
where $x_{N+1}, x_{N+2}\in\fR^{d_{in}}_+$ are arbitrary inputs.
Also, for $\epsilon, \epsilon'>0$, where $M=1+\epsilon'$, we can define events $E_0, E^{(L)}_1, E^{(L)}_2, E^{(L)}_3, E^{(L)}_4, E_5$
and constants $K_0, L_0, L_1, L_2, L_3$ as defined in the proof of Lemma~\ref{lemm:not crossing}.
We further let 
\begin{align*}
    K_2 =  \max\left\{K_1, \sup_{1\leq i\leq N+2}M^89d^4\Gamma_{\infty}(0)\para{\sum^8_{j=1}\tilde{\Phi}_{\infty, j}(x_i, 0)}\tilde{\Psi}_{\infty,2}(x_i,0)\tilde{\Psi}_{\infty,8}(x_i,0) K_0\right\}.
\end{align*}
and
\begin{align*}
    E_6 = 
    & \Bigg\{\frac{1}{L}\sum_l\Bigg[\left( 1 + \frac{L}{C_L}K_2+\Bigg\| \frac{1}{\CL}P_{c'}\para{\sigma\para{\para{f_{\theta(0)}^{l-1}(x)}_c}_{\psi_{su}}}\Bigg\|\right)
        \left( 1 + \frac{L}{C_L}K_2+\Bigg\| \frac{1}{\CL}P_{c'}\para{\sigma\para{\para{f_{\theta(0)}^{l-1}(x')}_c}_{\psi_{su}}}\Bigg\|\right)\\
        &\quad\quad\quad\times  \left( 1 + \frac{L}{C_L}K_2+\Bigg\| \W_L^{l+2}(x, 0)\Bigg\|\right) \left( 1 + \frac{L}{C_L}K_2+\Bigg\| \W_L^{l+2}(x', 0)\Bigg\|\right)\Bigg] \times \frac{L}{C_L}K_2\\
        &\quad+\frac{1}{L\CL^2}\sum_l\left(1 + \frac{L}{\CL}K_2+ \ab{\W_L^{l+2}(x, 0)}\right) 
            \left(1 + \frac{L}{\CL}K_2+ \ab{\W_L^{l+2}(x', 0)}\right)\times\frac{L}{C_L}K_2  < \epsilon''\Bigg\}
\end{align*}
for $\epsilon''>0$.
Since $L^2/\CL\rightarrow 0$ as $L\rightarrow\infty$, the probability of $E^{(L)}_6$ converges to  $1$ by the law of large numbers,
and there exists $L_4$ such that
\begin{align*}
    \Pr{E^{(L)}_6} > 1-\epsilon
\end{align*}
for all $L\ge L_4$.
Suppose all events $E_0, E^{(L)}_1, E^{(L)}_2, E^{(L)}_3, E^{(L)}_4, E_5, E^{(L)}_6$ are given, and $L\geq L_{\max}=\max\{L_0, L_1, L_2, L_3, L_4\}$.
Then, instead of Proposition~\ref{prop:pin-norm}, the following inequalities hold without having $L\rightarrow\infty$:
\begin{align*}
    \Phi_{L, j}(t) & \leq M^2 \Gamma_{\infty}(0)\\
    \Psi_{L, 2}(t) & \leq M^2 \Gamma_{\infty}(0)\\
    \Psi_{L, 8}(t) & \leq M^2 \Gamma_{\infty}(0)
\end{align*}
for all $1\leq j\leq 8$ and $t\in[0, T_L]$ since $\beta_L^{(1)}=T_L$.
Similarly, we can rewrite Proposition~\ref{prop:l2-norm} without having $L\rightarrow\infty$:
\begin{align*}
    \sum_{j=1}^8\tilde{\Phi}_{\infty, j}(x_i, 0)\leq& \sum_{j=1}^8\tilde{\Phi}_{L, j}(x_i, t) \leq M^2 \sum_{j=1}^8\tilde{\Phi}_{\infty, j}(x_i, 0)\\
    \tilde{\Psi}_{\infty, 2}(x_i, 0)\leq& \tilde{\Psi}_{L, 2}(x_i, t) \leq M^2 \tilde{\Psi}_{\infty, 2}(x_i, 0)\\
    \tilde{\Psi}_{\infty, 8}(x_i, 0)\leq& \tilde{\Psi}_{L, 8}(x_i, t) \leq M^2 \tilde{\Psi}_{\infty, 8}(x_i, 0)
\end{align*}
for all $1\leq i\leq N+2$ and $t\in[0, T_L]$.
Finally, Corollary~\ref{corr:finite} can be replaced by
\begin{align*}
    \sup_{1\leq l\leq L-1, t\in[0, T_L]} \ab{P_{c'}\para{\sigma\para{\para{f_{\theta(t)}^{l-1}(x)}_c}_{\psi_{su}}}-P_{c'}\para{\sigma\para{\para{f_{\theta(0)}^{l-1}(x)}_c}_{\psi_{su}}} }< \frac{L}{\CL}K_1
\end{align*}
for all $i, c, c', s, u$, where $K_1$ is as defined in the proof of Lemma~\ref{lemm:not crossing}.
This already implies $T_L=T$.

In addition, by Lemma~\ref{lemm:layer-bound},
\begin{align*}
    \sup_{1\leq l\leq L-1, t\in[0, T]} \ab{\W^k_L(x_i, t)-\W^k_L(x_i, 0)} < \frac{L}{C_L}K_2.
\end{align*}
Thus, for $x, x'\in\{x_1, \dots, x_{N+2}\}$, we can bound the deviation of $\tilde{\Theta}^L_t(x, x')$ by
\begin{align*}
    &\ab{\tilde{\Theta}^L_t(x,x')- \tilde{\Theta}^L_0(x,x')}\\ 
    & \le \frac{1}{L}\sum_{l=0}^L \sum_{c'csu}\Bigg\|\W_L^{l+2}(x,t)\left(\frac{1}{\CL}P_{c'}\para{\sigma\para{\para{f_{\theta(t)}^{l-1}(x)}_c}_{\psi_{su}}} \para{\frac{1}{\CL}P_{c'}\para{\sigma\para{\para{f_{\theta(t)}^{l-1}(x')}_c}_{\psi_{su}}}}^{\intercal}\right)\W_L^{l+2}(x',t)^{\intercal}\\
    &\quad\quad-\W_L^{l+2}(x,0)\left( \frac{1}{\CL}P_{c'}\para{\sigma\para{\para{f_{\theta(0)}^{l-1}(x)}_c}_{\psi_{su}}} \para{\frac{1}{\CL}P_{c'}\para{\sigma\para{\para{f_{\theta(0)}^{l-1}(x')}_c}_{\psi_{su}}}}^{\intercal}\right)\W_L^{l+2}(x',0)^{\intercal}\Bigg\|\\
    &\quad+\frac{1}{L\CL^2}\sum_{l=0}^{L-1}\ab{ \left[\W_L^{l+2}(x,t)B^l\W_L^{l+2}(x',t)^{\intercal}\right]-\left[\W_L^{l+2}(x,0)B^l\W_L^{l+2}(x',0)^{\intercal}\right]}\\
    & \le \frac{1}{L}\sum_l\Bigg[\left( \frac{L}{C_L}K_2+\Bigg\| \frac{1}{\CL}P_{c'}\para{\sigma\para{\para{f_{\theta(t)}^{l-1}(x)}_c}_{\psi_{su}}}\Bigg\|\right)
        \left(  \frac{L}{C_L}K_2+\Bigg\| \frac{1}{\CL}P_{c'}\para{\sigma\para{\para{f_{\theta(t)}^{l-1}(x')}_c}_{\psi_{su}}}\Bigg\|\right)\\
        &\quad\quad\quad\times  \left(\frac{L}{C_L}K_2+\Bigg\| \W_L^{l+2}(x, 0)\Bigg\|\right) \left(\frac{L}{C_L}K_2+\Bigg\| \W_L^{l+2}(x', 0)\Bigg\|\right)\\
        &\quad\quad\quad-\ab{\frac{1}{\CL}P_{c'}\para{\sigma\para{\para{f_{\theta(0)}^{l-1}(x)}_c}_{\psi_{su}}}} \ab{\frac{1}{\CL}P_{c'}\para{\sigma\para{\para{f_{\theta(0)}^{l-1}(x)}_c}_{\psi_{su}}}} 
            \ab{\W_L^{l+2}(x, 0)} \ab{\W_L^{l+2}(x', 0)} \Bigg]\\
        &\quad+\frac{1}{L\CL^2}\sum_l\Bigg[\left(\frac{L}{\CL}K_2+ \ab{\W_L^{l+2}(x, 0)}\right) \left(\frac{L}{\CL}K_2+ \ab{\W_L^{l+2}(x', 0)}\right) \\
        &\quad\quad\quad\quad -  \ab{\W_L^{l+2}(x, 0)} \ab{\W_L^{l+2}(x', 0)} \Bigg]\\
    & \le \frac{1}{L}\sum_l\Bigg[\left( 1 + \frac{L}{C_L}K_2+\Bigg\| \frac{1}{\CL}P_{c'}\para{\sigma\para{\para{f_{\theta(0)}^{l-1}(x)}_c}_{\psi_{su}}}\Bigg\|\right)
        \left( 1 + \frac{L}{C_L}K_2+\Bigg\| \frac{1}{\CL}P_{c'}\para{\sigma\para{\para{f_{\theta(0)}^{l-1}(x')}_c}_{\psi_{su}}}\Bigg\|\right)\\
        &\quad\quad\quad\times  \left( 1 + \frac{L}{C_L}K_2+\Bigg\| \W_L^{l+2}(x, 0)\Bigg\|\right) \left( 1 + \frac{L}{C_L}K_2+\Bigg\| \W_L^{l+2}(x', 0)\Bigg\|\right)\Bigg] \times \frac{L}{C_L}K_2\\
        &\quad+\frac{1}{L\CL^2}\sum_l\left(1 + \frac{L}{\CL}K_2+ \ab{\W_L^{l+2}(x, 0)}\right) 
            \left(1 + \frac{L}{\CL}K_2+ \ab{\W_L^{l+2}(x', 0)}\right)\times\frac{L}{C_L}K_2\\
    &\leq \epsilon''
\end{align*}
Finally,
\begin{align*}
    \Pr{ \ab{\tilde{\Theta}^L_t(x,x')- \tilde{\Theta}^L_0(x,x')} < \epsilon''} \ge 1-2\epsilon
\end{align*}
for all $L\ge L_{\max}'$.
Since $\epsilon, \epsilon'$, and $\epsilon''$ are arbitrary,
we can conclude $S\para{\tilde{\Theta}_t^L(x, x')}\convinp S\para{\tilde{\Theta}^{\infty}(x, x')}$.

\section{Equalities and inequalities of sub layers for CNN}\label{sec:ineq of CNN}
In this section, we introduce some useful equalities and inequalities in our problem setting. Like Section~\ref{sec:ineq of MLP}, we use $f^l\equiv f^l_{\theta^{(l)}}$ if it is clear from the context.

First, following lemma provides bound of Jacobian. 

\begin{lemma}\label{ineq:bound jacobian of f}
For $1\leq l\leq L$,
\[\max \left\{\EP{\partial_{w^l} f^l}, \ab{\partial_{w^l} f^l} \right\} \le 9n_ln_{l-1}\EP{\para{\sigma(f^{l-1}(x))})}\]
\end{lemma}
\begin{proof}[Proof of Lemma~\ref{ineq:bound jacobian of f}]
\begin{align*}
    \EP{\partial_{w^l} f^l}& \le 
    \sum_{c'csu}\EP{\partial_{w_{c'csu}} f^l}\\& \le \sum_{c'csu} \EP{P_{c'}\para{\sigma(f_c^{l-1}(x))_{\psi_{su}}}}\\
    & \le 9n_ln_{l-1}\EP{{\sigma(f^{l-1}(x)))}}
\end{align*}
We can bound $\ab{\partial_{w^l} f^l}$ similarly.
\end{proof}
\begin{lemma}\label{ineq:bound jacobian of fb}
For $1\leq l\leq L$,
\begin{align*}
    \max \left \{\EP{\partial_{b^l} f^l}, \ab{\partial_{b^l} f^l} \right \}\le d^2.
\end{align*}
\end{lemma}

As previously mentioned, we set the scaling factor of gradient flow by $\frac{1}{(L-1)\CL^2}$ throughout the proof. The gradient flow is given by
\begin{align*}
    \partial_t \theta^l &=  -\frac{1}{(L-1)\CL^2}\mathbb{E}_{x \sim p^{in}}[(\partial_{\theta^l} f^l(x))^{\intercal} \delta^l_t]\\
    \partial_t f^l(x') &=-\frac{1}{(L-1)\CL^2}\mathbb{E}_{x \sim p^{ in}}[ \partial_{\theta^l}f^l(x') (\partial_{\theta^l} f^l(x))^{\intercal} \delta^l_t],
\end{align*}
where $\delta^j_t = \para{\W^{j+1}_L(x,t)}^{\intercal}\delta^L_t $ since
\begin{align*}
\partial_{\theta^l}f^{L}= \W^{l+1}_L(x,t)\partial_{\theta^l} f^{l} \quad \text{for} \, L \ge l.
\end{align*}
Then, by the chain rule, we obtain the following lemma.
\begin{lemma}\label{lemm:bound weight derivatives}
For $t \le 0$ and $1\leq l\leq L$,
\[\EP{\partial_t w^l} \le \frac{9n_ln_{l-1}}{(L-1)\CL^2}\EP{ \delta^l_t}\EP{\sigma(f^{l-1})}, \quad \EP{\partial_t b^l} \le \frac{d^2}{(L-1)\CL^2}\EP{ \delta^l_t}\]
\end{lemma}
\begin{proof}[Proof of Lemma~\ref{lemm:bound weight derivatives}]
\begin{align*}
\EP{\partial_t w^l}&=  \ab{\frac{1}{(L-1)\CL^2}\mathbb{E}_{x \sim p^{in}}[(\partial_{w^l} f^l(x))^{\intercal} \delta^l_t]}\\
& \le \frac{1}{(L-1)\CL^2}\Epin{ \|(\partial_{w^l} f^l(x))^{\intercal} \delta^l_t\|}\\
& \le \frac{1}{(L-1)\CL^2}\Epin{\|\partial_{w^l} f^l(x)\| \|\delta^l_t\| }\\
&\le \frac{1}{(L-1)\CL^2}\sqrt{\Epin{\|\partial_{w^l} f^l(x)\|^2}}\sqrt{\Epin{\|\delta^l_t\| ^2}}\\
&= \frac{1}{(L-1)\CL^2}\EP{\partial_{w^l} f^l(x)}\EP{\delta^l_t}\\
&\le \frac{9n_ln_{l-1}}{(L-1)\CL^2}\EP{\sigma(f^{l-1})}\EP{\delta^l_t},
\end{align*}
where the first inequality is from Jensen's inequality, third inequality comes from Cauchy's inequality and last inequality follows from Lemma~\ref{ineq:bound jacobian of f}. Finally, we can bound $\EP{\partial_t b^l}$ similarly.
\end{proof}

Also we can induce following lemma. 
\begin{lemma}\label{ineq:bound of derivative of w}
Suppose all elements of $f^i_{\theta^{(i)}(t)}(x')$ are nonzero for all $1 \le l \le L-1$ and $x'$ in dataset. Then, for $2 \le l \le L$, 
\[ \max \left \{ \EP{\partial_t\W^l_l}, \ab{\partial_t\W^l_l}\right\}\le \frac{9d^4}{(L-1)\CL^2}\EP{\sigma(f^{l-1}(x))}\EP{\delta^l_t}.\]
\end{lemma}
\begin{proof}[Proof of Lemma~\ref{ineq:bound of derivative of w}]
By the chain rule, we have
\begin{align*}
    &\EP{\partial_t\para{\partial_{f^{l-1}_{c,i,j}}f^l_{c',s,u}(x)}}\\ &=\EP{\partial_t \para{w^l_{c',c,i-s+2, j-u+2}\dot{\sigma}(f^{l-1}_{c, i, j})\mathbf{1}_{\{s-1\le i\le s+1\}}
\mathbf{1}_{\{u-1\le j\le u+1\}}}}\\
    &=\frac{1}{(L-1)\CL^2}\ab{\Epin{ (\partial_{w^l_{c',c,i-s+2, j-u+2}} f^l(x))^{\intercal} \delta^l_t}}\EP{\dot{\sigma}(f^{l-1}_{c, i, j}(x))}\mathbf{1}_{\{s-1\le i\le s+1\}}
\mathbf{1}_{\{u-1\le j\le u+1\}}\\
    &\le \frac{1}{(L-1)\CL^2}\EP{\sigma(f^{l-1}_{c}(x))_{\psi_{i-s+2, j-u+2}}}\EP{\delta^l_t}\mathbf{1}_{\{s-1\le i\le s+1\}}
\mathbf{1}_{\{u-1\le j\le u+1\}}.
\end{align*}
Hence 
\begin{align*}
    \EP{\partial_t\W^l_l} \le \frac{9d^4}{(L-1)\CL^2}\EP{\sigma(f^{l-1}(x))}\EP{\delta^l_t} 
\end{align*}
and we can bound $\EP{\partial_t\W^l_l}$ similarly.
\end{proof}

From gradient flow, we get following lemma provides a bound of the norm.
\begin{lemma}\label{lemm:bound derivative of intermediate layer}
For $1\leq l\leq L$ and $t \le 0$,
\[\EP{\partial_t f^l(x')} \le \frac{N}{(L-1)\CL^2}\EE{\left\|(\partial_{\theta^l} f^l(x')) (\partial f^l_{\theta^l}(x))^{\intercal} \dLt(x)\right\|}\]
\end{lemma}
Following lemma is the key lemma to prove invariances of Lyapunov functions and it's only factor different compared to MLP setting.
\begin{lemma}\label{lemm:bound derivative of pin norm}
Suppose all elements of $f^i_{\theta^{(i)}(t)}(x')$ are nonzero for all $1 \le i \le L-1$ and $x'$ in dataset. 
Then, for $t \in [0,T]$, $1\leq k \le j\le L$, and $1\leq i \leq L$, 
\begin{align*}
    \partial_t \EP{\W^k_j(x,t)} 
    &\le \frac{9d^4N}{(L-1)\CL^2}\sum^j_{i = k}\EP{\W^{i+1}_{j}(x,t)}\EP{\W^k_{i-1}(x,t)}\EP{\dit} \EP{\sigma\para{f^{i-1}_{\theta^{(i-1)}(t)}(x)}}\\
    \partial_t \EP{f^i_{\theta^{(i)}(t)}(x')}
    &\le \frac{81n_i^2n_{i-1}^2N^{3/2}}{(L-1)\CL^{2}} \sum_{l=0}^{i-1} \EP{ f^{l}_{\theta^{(l)}(t)}(x)}\EP{ f^{l}_{\theta^{(l)}(t)}(x')}\EP{\W_i^{l+2}}\EP{(\W_L^{l+2})^{\intercal}} \EP{\dLt}\\
    &\quad +\frac{d^4N}{(L-1)\CL^{2}}  \sum_{l=0}^{i-1}\EP{\W_i^{l+2}}\EP{(\W_L^{l+2})^{\intercal}} \EP{\dLt}.
  \end{align*}
\end{lemma}
\begin{proof}[Proof of Lemma~\ref{lem:bound derivative of pin norm} ]
By the chain rule,
\begin{align*}
    \partial_t \EP{\W^k_j(x,t)} &\le  \sum^j_{i = k} \EP{\W^{i + 1}_j \partial_t \W^{i}_i \W_{i - 1}^k }\\
    &\le  N\sum^j_{i = k} \EP{\W^{i + 1}_j} \EP{ \partial_t \W^{i}_i }\EP{\W_{i - 1}^k },
 %     +\sum^j_{i = k} \EP{\W^{i+1}_j W^i(t)\textcolor{red}{\partial_t\diag(\dot{\sigma}(f^{i}_{\theta^{(i)}(t)}(x))}\W_{i-1}^k}
\end{align*}
where $\W^{i+1}_j$ denotes $\W^{i+1}_j(x, t)$ if it is clear from the context.
Then, by previous Lemma~\ref{ineq:bound of derivative of w},
\begin{align*}
    \partial_t \EP{\W^k_j(x,t)}
    &\le  \frac{9d^4N}{(L-1)\CL^2}\sum^j_{i = k}\EP{\W^{i+1}_{j}(x,t)}\EP{\W^k_{i-1}(x,t)}\EP{\dit} \EP{\sigma(f^{i-1}_{\theta^{(i-1)}(t)}(x)}.
  \end{align*}

On the other hand, from Lemma~\ref{lemm:bound derivative of intermediate layer} and \ref{lemm:recursive},
\begin{align*}
      \EP{\partial_t  \fult(x)}    & \le \frac{N}{(L-1)\CL^2}  \mathbb{E}_{x, x' \sim p^{in}}\left[\left\|\partial_{\theta^l} \fult(x') \para{\partial_{\theta^l} \fult(x)}^{\intercal} \left(\W_L^{l+1}(x)\right)^{\intercal} \dLt(x)\right\|\right]\\
    &\le\frac{N}{(L-1)\CL^2}  \mathbb{E}_{x, x' \sim p^{in}}\bigg[\sum_{i=0}^{l-1} \bigg\| \sum_{c'csu}\W_L^{l+2}(x,t)  P_{c'}\para{\sigma(f_c^{l}(x))_{\psi_{su}}} \para{P_{c'}\para{\sigma(f_c^{l}(x'))_{\psi_{su}}}}^{\intercal}\W_L^{l+2}(x',t)^{\intercal}+\\& \W_L^{l+2}(x,t)B^l\W_L^{l+2}(x',t)^{\intercal} \dLt(x)\bigg\|\bigg]\\
    &\le\frac{81\max_l\{ n_l^4\}N}{(L-1)\CL^2}  \sum_{i=0}^{l-1}\mathbb{E}_{x, x' \sim p^{in}} \big[\|( \sigma(f^i(x))\|\|\sigma(f^i(x'))\|\|\W_l^{i+2}(x')\|\|(\W_L^{i+2}(x))^{\intercal}\| \|\dLt(x)\|\big]\\
    &\quad+\frac{d^4N}{(L-1)\CL^2}  \sum_{i=0}^{l-1}\EE{\left\|\W_l^{i+2}(x')\|\|(\W_L^{i+2}(x))^{\intercal}\| \|\dLt(x)\right\|}\\
    &\le\frac{81\max_l\{ n_l^4\}N^{3/2}}{(L-1)\CL^2} \sum_{i=0}^{l-1} \EP{ \sigma(f^i(x))}\EP{ \sigma(f^i(x'))}\EP{\W_l^{i+2}}\EP{(\W_L^{i+2})^{\intercal}} \EP{\dLt(x)}\\
    &\quad+\frac{d^4N}{(L-1)\CL^2}  \sum_{i=0}^{l-1}\EP{\W_l^{i+2}}\EP{(\W_L^{i+2})^{\intercal}} \EP{\dLt(x)},
\end{align*}
where third inequality comes from Lemma~\ref{ineq:bound jacobian of f} 
and ~\ref{ineq:bound jacobian of fb} and the last inequality holds due to the property of $p^{in}$-norm
\begin{align*}
\Epin{\|( \sigma(f^i(x))\| \|(\W_L^{i+2})^{\intercal}\| \|\dLt(x)\|}
    \le \sqrt{N}\EP{( \sigma(f^i(x))}\EP{(\W_L^{i+2})^{\intercal}} \EP{\dLt(x)}.
\end{align*} 
% \begin{align*}
% \ab{\sum_{c',c,t,u} \partial_{w^{l}_{c',c,t,u}} f^{l}(x) \para{\partial_{w^{l}_{c',c,t,u}}f^{l}(x')}^{\intercal}} \le 81n_l^2 \ab {\sigma(f^{l-1}(x)))}\ab{\para{(\sigma(f^{l-1}(x')))}^{\intercal}}
% \end{align*}
% \begin{align*}
%     \ab{\sum_i\partial_{b_i^{l+1}} y^{l+1}(x) \para{\partial_{b_i^{l+1}}y^{l+1}(x')}^{\intercal}} \le d^2
% \end{align*}
\end{proof}

We have a similar bounds for derivatives of $\ell_2$-norms.
\begin{lemma}\label{lemm:bound derivative of l2 norm}
For $x'\in\fR_+^{d \times d}$, if all elements of $f^i_{\theta^{(i)}(t)}(x')$ are nonzero for all $1 \le i \le L-1$, 
then, for $t \in [0,T]$, $1\leq l \le j\le L$, and $1\leq i \leq L$
\begin{align*}
    \partial_t \ab{\W^k_j(x',t)} 
    & \le \frac{9d^4}{(L-1)\CL^2}\sum_{l \ge i \ge k} \ab{\W^{i+1}_{l}(x')}\ab{\W^k_{i-1}(x')}\EP{\delta^i_{t}} \EP{\sigma(f^{i-1})^{\intercal}}\\
    \partial_t \ab{f^i_{\theta^{(i)}(t)}(x')} 
    &\le\frac{81n_i^2n_{i-1}^2N^{1/2}}{(L-1)\CL^2} \sum_{i=0}^{l-1} \EP{ \sigma(f^i(x)}\ab{ \sigma(f^i(x')}\ab{\W_l^{i+2}(x')}\EP{(\W_L^{i+2})^{\intercal}} \EP{\delta_t^L(x)}\\
    &\quad +\frac{d^4}{(L-1)\CL^2}  \sum_{i=0}^{l-1}\ab{\W_l^{i+2}(x')}\EP{(\W_L^{i+2})^{\intercal}} \EP{\delta_t^L(x)}.
  \end{align*}
\end{lemma}

\begin{proof}[Proof of Lemma~\ref{lemm:bound derivative of l2 norm}]
The proof is similar to that of Lemma~\ref{lemm:bound derivative of pin norm}.
\begin{align*}
    \partial_t \ab{\W^k_l(x)} &\le \ab{\partial_t \W_l^k(x)}\\
    &\le  \sum^j_{i = k} \ab{\W^{i + 1}_j(x)} \EP{ \partial_t \W^{i}_i(x) }\ab{\W_{i - 1}^k (x)}\\
    &\le  \frac{9d^4}{(L-1)\CL^2}\sum^j_{i = k}\EP{\W^{i+1}_{j}(x,t)}\EP{\W^k_{i-1}(x,t)}\EP{\dit} \EP{\sigma(f^{i-1}_{\theta^{(i-1)}(t)}(x)}
\end{align*}

On the other hand,
\begin{align*}
 \ab{\partial_t  f^l(t)}& \le \frac{1}{(L-1)\CL^2} \left[\left\|\partial_{\theta^l} f^l(x') \Epin{\para{\partial f^l_{\theta^l}(x)}^{\intercal} \left(\W_L^{l+1}(x)\right)^{\intercal} \delta_t^L(x)}\right\|\right]\\
    &\le\frac{1}{(L-1)\CL^2}  \mathbb{E}_{x \sim p^{in}}\bigg[\sum_{i=0}^{l-1} \bigg\| \sum_{c'csu}\W_L^{l+2}(x,t)  P_{c'}\para{\sigma(f_c^{l}(x))_{\psi_{su}}} \para{P_{c'}\para{\sigma(f_c^{l}(x'))_{\psi_{su}}}}^{\intercal}\W_L^{l+2}(x',t)^{\intercal}+\\& \W_L^{l+2}(x,t)B^l\W_L^{l+2}(x',t)^{\intercal} \dLt(x)\bigg\|\bigg]\\
   &\le\frac{81\max_l\{ n_l^4\}}{(L-1)\CL^2}  \sum_{i=0}^{l-1}\mathbb{E}_{x \sim p^{in}} \big[\|( \sigma(f^i(x))\|\|\sigma(f^i(x')\|\|\W_l^{i+2}(x')\|\|(\W_L^{i+2}(x))^{\intercal}\| \|\dLt(x)\|\big]\\
    &\quad+\frac{d^4 }{(L-1)\CL^2}  \sum_{i=0}^{l-1}\Epin{\left\|\W_l^{i+2}(x')\|\|(\W_L^{i+2}(x))^{\intercal}\| \|\dLt(x)\right\|}\\
    &\le\frac{81\max_l\{ n_l^4\}N^{1/2}}{(L-1)\CL^2} \sum_{i=0}^{l-1} \EP{ \sigma(f^i(x)}\EP{ \sigma(f^i(x')}\EP{\W_l^{i+2}}\EP{(\W_L^{i+2})^{\intercal}} \EP{\dLt(x)}\\
    &\quad+\frac{d^4}{(L-1)\CL^2}  \sum_{i=0}^{l-1}\EP{\W_l^{i+2}}\EP{(\W_L^{i+2})^{\intercal}} \EP{\dLt(x)}.
\end{align*}
\end{proof}

\section{Proof of Lemma~\ref{lemm:pin-bound} for Proposition~\ref{prop:pin-norm}}\label{app:prp4}

\begingroup
\def\thetheorem{\ref{lemm:pin-bound}}
\begin{lemma}
For $t \geq 0$, if $f^l_{\theta^{(l)}(t)}(x)$ is element-wise nonzero for all $1 \le l \le L-1$ and $x\in\{x_1,\dots, x_N\}$, then
\begin{align*}
    \partial_t \Phi_{L,j}(t) &\le \frac{162\max_l\{ n_l^4\}d^4jN^{3/2}(L-1)}{\CL}\Phi_{L,j-1}(t)\Phi_{L,8}(t)\Psi_{L, 2}(t)\Psi_{L,8}(t)\EP{\dLt}\\
    \partial_t \Psi_{L,2}(t)& \le \frac{18d^4N(L-1)}{\CL}\Phi_{L,4}(t) \para{\Psi_{L,2}(t)}^{2}(t)\Psi_{L,8}(t)\EP{\dLt}\\
    \partial_t \Psi_{L,8}(t) & \le \frac{72d^4N(L-1)}{\CL}\Phi_{L,4}(t) \Psi_{L,2}(t)\left(\Psi_{L,8}(t)\right)^{2}\EP{\dLt}
\end{align*}
$1 \le j \le 8$.
\end{lemma}
\addtocounter{theorem}{-1}
\endgroup

\begin{proof}[Proof of Lemma~\ref{lemm:pin-bound}]
% ~\ref{lem:pin-bound}
% Prob 1 part 1
Consider the derivative of $\Psi_{L, 8}$.
Like Section~\ref{app:not crossing}, we show that the intermediate pre-activation values never reach zero while training,
and therefore we can apply Lemma~\ref{lemm:bound derivative of pin norm}.
% ~\ref{lem:bound derivative of pin norm}.
\begin{align*}
    &\partial_t \Psi_{L, 8}(t)\\
    & \le \frac{72Nd^4}{(L-1)^2\CL^2}  \sum_{l =2 }^{L} \sum_{l=i}^{L} \EP{\W^{l+1}_{L}(t)}^2\EP{\W^i_{l-1}(t)}\left(\EP{\W_L^i(0)}+\EP{\W_L^i(t)-\W_L^i(0)}\right)^7\\&\quad \times\EP{\fultm(x)}\EP{\dLt}.
\end{align*}
First, the following rough bound holds.
\begin{align*}
    & \sum_{l =2 }^{L} \sum_{l=i}^{L} \EP{\W^{l+1}_{L}(t)}^2\EP{\W^i_{l-1}(t)}\left(\EP{\W_L^i(0)}+\EP{\W_L^i(t)-\W_L^i(0)}\right)^7 \EP{\fultm(x)}\\
    & \le \Bigg( \sum_{i =2 }^{L} \sum_{l=i}^{L} \EP{\W^i_{l-1}(t)} \left(\EP{\fulzm}+\EP{\fultm-\fulzm}\right)\\
    &\quad \times \left(\EP{\W_L^{l+1}(0)}+\EP{\W_L^{l+1}(t)-\W_L^{l+1}(0)}\right)^2\Bigg) \sum_{i= 2}^L \left(\EP{\W_L^i(0)}+\EP{\W_L^i(t)-\W_L^i(0)}\right)^8\\
    &\le (L-1)\Psi_{L, 8}(L-1)^2\CL\para{\Psi_{L, 2}}^{(1/2)}\para{\Phi_{L, 4}}^{(1/4)}\para{\Psi_{L, 8}}^{(1/4)},
\end{align*}
where the last inequality is from 
\begin{align*}
    & \para{\sum_{i =2 }^{L} \sum_{l=i}^{L} \EP{\W^i_{l-1}}\para{\EP{\fulzm}+\EP{\fultm-\fulzm}} \left(\EP{\W_L^{l+1}(0)}+\EP{\W_L^{l+1}(t)-\W_L^{l+1}(0)}\right)^2}^2\\
    &\le \left(\sum_{i =2 }^{L} \sum_{l=i}^{L}\frac{L-1}{l-1} \EP{\W^i_{l}}^2\right) \\
    &\quad\times \left((L-1)\sum^L_{l=2} \left(\EP{\fulzm}+\EP{\fultm-\fulzm}\right)^2 \left(\EP{\W_L^{l+1}(0)}+\EP{\W_L^{l+1}(t)-\W_L^{l+1}(0)}\right)^4\right)\\
    &\le (L-1)^2 \Psi_{L, 2}(L-1)
    \\
    &\quad \times\sqrt{\Bigg(\sum^{L-1}_{l=1} \left(\EP{\fulz}+\EP{\fult-\fulz}\right)^4\Bigg)\Bigg(\sum^L_{l=2} \left(\EP{\W_L^l(0)}+\EP{\W_L^l(t)-\W_L^l(0)}\right)^8\Bigg)}\\
    & \le (L-1)^2 \Psi_{L, 2} (L-1) \sqrt{(L-1)^2 \CL^4\Phi_{L, 4}\Psi_{L, 8}}.
\end{align*}
Thus,
\begin{align*}
    &\partial_t \Psi_{L, 8}(t)\\
    & \le \frac{72Nd^4}{(L-1)^2\CL^2}  \sum_{l =2 }^{L} \sum_{l=i}^{L} \EP{\W^{l+1}_{L}(t)}^2\EP{\W^i_{l-1}(t)}\left(\EP{\W_L^i(0)}+\EP{\W_L^i(t)-\W_L^i(0)}\right)^7 \\
    &\quad \times\EP{\fultm(x)}\EP{\dLt}\\ 
    &\le \frac{72Nd^4(L-1)}{\CL} \left(\Psi_{L, 2}\right)^{1/2}\left(\Phi_{L, 4}\right)^{1/4}\left(\Psi_{L, 8}\right)^{5/4}\EP{\dLt}\\ 
    &\le \frac{72Nd^4(L-1)}{\CL} \Psi_{L, 2}\Phi_{L, 4}\left(\Psi_{L, 8}\right)^{2}\EP{\dLt}.
\end{align*}

% Prob 1 part 2
Then, let consider the derivative of $\Psi_{L, 2}$.
From Lemma~\ref{lemm:bound derivative of pin norm},
\begin{align*}
    &\partial_t \Psi_{L, 2}(t)\\
    &\le \frac{18d^4N}{(L-1)^3\CL^2} \sum^{ L}_{i=2} \sum_{ k= 2}^{i}\sum^{i}_{l=k} \frac{L-1}{i-1}\EP{\W^{l+1}_{L}}\EP{\W^{l+1}_{i}}\EP{\W^k_{l-1}}\left(\EP{\W_i^k(0)}+\EP{\W_i^k(t)-\W_i^k(0)}\right)\\
    &\quad\times\EP{f^{l-1}_{\theta^{(l-1)}(t)}}\EP{\dLt}\\
    &\le \frac{9d^4N}{(L-1)^3\CL^2}  \sum^{ L}_{i=2} \sum_{ k= 2}^{i}\sum^{i}_{l=k} \frac{L-1}{i-1}\EP{\W^{l+1}_{L}}\EP{\W^k_{l-1}}\para{\EP{\W^{l+1}_{i}}^2+\left(\EP{\W_i^k(0)}+\EP{\W_i^k(t)-\W_i^k(0)}\right)^2}\\
    &\quad\times\EP{f^{l-1}_{\theta^{(l-1)}(t)}}\EP{\dLt}\\
    &=       \frac{9d^4N}{(L-1)^3\CL^2}  \sum^{ L}_{i=2} \sum_{k=2}^{i}\sum^{i}_{l=k} \frac{L-1}{i-1}\EP{\W^{l+1}_{L}}\EP{\W^k_{l-1}}
    \EP{\W^{l+1}_{i}}^2 \EP{f^{l-1}_{\theta^{(l-1)}(t)}}\EP{\dLt}\\
    &\quad + \frac{9d^4N}{(L-1)^3\CL^2}  \sum^{ L}_{i=2} \sum_{k=2}^{i}\sum^{i}_{l=k} \frac{L-1}{i-1}\EP{\W^{l+1}_{L}} \EP{\W^k_{l-1}}
    \left(\EP{\W_i^k(0)}+\EP{\W_i^k(t)-\W_i^k(0)}\right)^2\\
    &\quad \times \EP{f^{l-1}_{\theta^{(l-1)}(t)}}\EP{\dLt}.
\end{align*}
From Cauchy-Schwartz inequality,
\begin{align*}
    & \para{\sum_{k=2 }^{L} \sum_{l=k}^{L} \EP{\W^k_{l-1}}\para{\EP{\fulzm}+\EP{\fultm-\fulzm}} \left(\EP{\W_L^{l+1}(0)}+\EP{\W_L^{l+1}(t)-\W_L^{l+1}(0)}\right)}^2\\
     &\le \left(\sum_{k =2 }^{L} \sum_{l=k}^{L}\frac{L-1}{l-1} \EP{\W^k_{l}}^2\right) \\
     &\quad \times \left((L-1)\sum^L_{l=2} \left(\EP{\fulzm}+\EP{\fultm-\fulzm}\right)^2\left(\EP{\W_L^{l+1}(0)}+\EP{\W_L^{l+1}(t)-\W_L^{l+1}(0)}\right)^2\right)\\
    &\le (L-1)^2\Psi_{L, 2}(L-1) \\
    &\quad \times\sqrt{\Bigg(\sum^{L-1}_{l=1} \left(\EP{\fulz}+\EP{\fult-\fulz}\right)^4\Bigg)\Bigg(\sum^L_{l=2} \left(\EP{\W_L^l(0)}+\EP{\W_L^l(t)-\W_L^l(0)}\right)^4\Bigg)}\\
    &\le (L-1)^2\Psi_{L, 2} (L-1)\\
    &\quad \times\sqrt{\Bigg(\sum^{L-1}_{l=1} \left(\EP{\fulz}+\EP{\fult-\fulz}\right)^4\Bigg)\Bigg(\sum^L_{l=2} \left(\EP{\W_L^l(0)}+\EP{\W_L^l(t)-\W_L^l(0)}\right)^8\Bigg)}\\ 
    & \le (L-1)^2 \Psi_{L, 2} (L-1) \sqrt{(L-1)^2 \CL^4\Phi_{L, 4}\Psi_{L, 8}}\\
    & = (L-1)^4\CL^2\Psi_{L, 2}\left(\Phi_{L, 4}\right)^{(1/2)}\para{\Psi_{L, 8}}^{(1/2)}.
\end{align*}
Using the above inequality, we have the following upper bounds:
\begin{align*}
    &\sum^{ L}_{i=2} \sum_{ k= 2}^{i}\sum^{i}_{l=k} \frac{L-1}{i-1}\EP{\W^{l+1}_{L}}\EP{\W^k_{l-1}}\EP{\W^{l+1}_{i}}^2\EP{f^{l-1}_{\theta^{(l-1)}(t)}}\\
    & \le \Bigg( \sum_{k=2 }^{L} \sum_{l=k}^{L} \EP{\W^k_{l-1}}\para{\EP{\fulzm}+\EP{\fultm-\fulzm}}\\
    &\quad \times  \left(\EP{\W_L^{l+1}(0)}+\EP{\W_L^{l+1}(t)-\W_L^{l+1}(0)}\right)\Bigg) \left(\sum_{i=2}^L \sum^i_{l=2}\frac{L-1}{i-1}\EP{\W_i^{l+1}}^2\right)\\
    &\le (L-1)^2\Psi_{L, 2}(L-1)^2\CL\para{\Psi_{L, 2}}^{(1/2)}\para{\Phi_{L, 4}}^{(1/4)}\para{\Psi_{L, 8}}^{(1/4)}\\
    & = (L-1)^4\CL \para{\Psi_{L, 2}}^{(3/2)}\para{\Phi_{L, 4}}^{(1/4)}\para{\Psi_{L, 8}}^{(1/4)}.
\end{align*}
and
\begin{align*}
    &\sum^{ L}_{i=2} \sum_{ k= 2}^{i}\sum^{i}_{l=k} \frac{L-1}{i-1}\EP{\W^{l+1}_{L}}\EP{\W^k_{l-1}}\left(\EP{\W_i^k(0)}+\EP{\W_i^k(t)-\W_i^k(0)}\right)^2\EP{f^{l-1}_{\theta^{(l-1)}(t)}}\\
    &\le \Bigg( \sum_{k=2 }^{L} \sum_{l=k}^{L} \EP{\W^k_{l-1}}\para{\EP{\fulzm}+\EP{\fultm-\fulzm}}\\
    &\quad\times  \left(\EP{\W_L^{l+1}(0)}+\EP{\W_L^{l+1}(t)-\W_L^{l+1}(0)}\right)\Bigg)\left(\sum_{i=2}^L \sum^i_{k=2}\frac{L-1}{i-1}\left(\EP{\W_i^k(0)}+\EP{\W_i^k(t)-\W_i^k(0)}\right)^2\right)\\
    &\le (L-1)^2\Psi_{L, 2}(L-1)^2\CL\para{\Psi_{L, 2}}^{(1/2)}\para{\Phi_{L, 4}}^{(1/4)}\para{\Psi_{L, 8}}^{(1/4)}\\
    &= (L-1)^4\CL \para{\Psi_{L, 2}}^{(3/2)}\para{\Phi_{L, 4}}^{(1/4)}\para{\Psi_{L, 8}}^{(1/4)}.
\end{align*}

Thus, we have
\begin{align*}
    \partial_t \Psi^{(2)}_L(t)
    &\le \frac{9d^4N}{(L-1)^3\CL^2}  \sum^{ L}_{i=2} \sum_{ k= 2}^{i}\sum^{i}_{l=k} \frac{L-1}{i-1}\EP{\W^{l+1}_{L}}\EP{\W^k_{l-1}}\\
    &\quad \times \para{\EP{\W^{l+1}_{i}}^2+\left(\EP{\W_i^k(0)}+\EP{\W_i^k(t)-\W_i^k(0)}\right)^2}\EP{f^{l-1}_{\theta^{(l-1)}(t)}}\EP{\dLt}\\
    & \le \frac{18d^4N(L-1)}{\CL} \para{\Psi_{L, 2}}^{(3/2)}\para{\Phi_{L, 4}}^{(1/4)}\para{\Psi_{L, 8}}^{(1/4)}\EP{\dLt}\\
    & \le \frac{18d^4N(L-1)}{\CL} \para{\Psi_{L, 2}}^{2}\Phi_{L, 4}\Psi_{L, 8}\EP{\dLt}.
\end{align*}

% Prop 1, part 3
Finally, let consider the derivatives of $\Phi_{L, j}$ for $1\leq j \leq 8$.
From Lemma~\ref{lemm:bound derivative of pin norm},
\begin{align*}
    \partial_t \Phi_{L, j}(t) 
    &\le\frac{81\max_l\{ n_l^4\}jN^{3/2}}{(L-1)^2\CL^{2+j}} \sum_{i= 1}^{L-1} \sum_{l=0}^{i-1} \left(\EP{ f^{i}_{\theta^{(i)}(0)}}+\EP{ f^{i}_{\theta^{(i)}(t)}- f^{i}_{\theta^{(i)}(0)}}\right)^{j-1}\EP{ f^{l}_{\theta^{(l)}(t)}(x)}\EP{ f^{l}_{\theta^{(l)}(t)}(x')}\\
    &\quad\quad\times \EP{\W_i^{l+2}}\EP{(\W_L^{l+2})^{\intercal}} \EP{\dLt}\\
    &\quad +\frac{d^4jN}{(L-1)^2\CL^{2+j}} \sum_{i= 1}^{L-1} \sum_{l=0}^{i-1}\left(\EP{ f^{i}_{\theta^{(i)}(0)}}+\EP{ f^{i}_{\theta^{(i)}(t)}- f^{i}_{\theta^{(i)}(0)}}\right)^{j-1}\EP{\W_i^{l+2}}\EP{(\W_L^{l+2})^{\intercal}} \EP{\dLt}.
\end{align*}
Similar to the previous inequalities, we have
\begin{align*}
    &\bigg(\sum^{L-1}_{i=1} \sum^{i-1}_{l=0} \EP{\W_i^{l+2}} \left(\EP{\fulz}+\EP{\fult-\fulz}\right)^2 \left(\EP{\W_L^{l+2}(0)}+\EP{\W_L^{l+2}(t)-\W_L^{l+2}(0)}\right)\bigg)^2\\
    &\le\para{\sum^{L-1}_{i=1} \sum^{i-1}_{l=0}\frac{L-1}{i-1} \EP{\W_i^{l+2}}^2}\\
    &\quad \times \Bigg((L-1)\sum^{L-2}_{l=0} \left(\EP{\fulz}+\EP{\fult-\fulz}\right)^4 \left(\EP{\W_L^{l+2}(0)}+\EP{\W_L^{l+2}(t)-\W_L^{l+2}(0)}\right)^2\Bigg)\\
    &\leq (L-1)^2\Psi_{L, 2}(t)(L-1) \\ &\quad \times\sqrt{\Bigg(\sum^{L-2}_{l=0} \left(\EP{\fulz}+\EP{\fult-\fulz}\right)^8\Bigg)\Bigg(\sum^L_{l=2} \left(\EP{\W_L^l(0)}+\EP{\W_L^l(t)-\W_L^l(0)}\right)^4\Bigg)}\\ 
    &\leq (L-1)^2\Psi_{L, 2}(t)(L-1) \\ &\quad \times \sqrt{\Bigg(\sum^{L-2}_{l=0} \left(\EP{\fulz}+\EP{\fult-\fulz}\right)^8\Bigg)\Bigg(\sum^L_{l=2} \left(\EP{\W_L^l(0)}+\EP{\W_L^l(t)-\W_L^l(0)}\right)^8\Bigg)}\\ 
    &\leq (L-1)^2\Psi_{L, 2}(t)(L-1)\times (L-1)\CL^4\para{\Phi_{L, 8}(t)}^{(1/2)}\para{\Psi_{L, 8}(t)}^{(1/2)} \\ &=(L-1)^4\CL^4\Psi_{L, 2}(t)\para{\Phi_{L, 8}(t)}^{(1/2)}\para{\Psi_{L, 8}(t)}^{(1/2)}.
\end{align*}
Thus, we have
\begin{align*}
    &\sum_{i= 1}^{L-1} \sum_{l=0}^{i-1} \left(\EP{ f^{i}_{\theta^{(i)}(0)}}+\EP{ f^{i}_{\theta^{(i)}(t)}- f^{i}_{\theta^{(i)}(0)}}\right)^{j-1}\EP{ f^{l}_{\theta^{(l)}(t)}(x)}\EP{ f^{l}_{\theta^{(l)}(t)}(x')}\EP{\W_i^{l+2}}\EP{(\W_L^{l+2})^{\intercal}}\\
    &\le \sum^{L-1}_{i=1} \sum^{i-1}_{l=0} \EP{\W_i^{l+2}} \left(\EP{\fulz}+\EP{\fult-\fulz}\right)^2 \left(\EP{\W_L^{l+2}(0)}+\EP{\W_L^{l+2}(t)-\W_L^{l+2}(0)}\right)\\
    &\quad \times \sum^{L-1}_{i=1}\left(\EP{ f^{i}_{\theta^{(i)}(0)}}+\EP{ f^{i}_{\theta^{(i)}(t)}- f^{i}_{\theta^{(i)}(0)}}\right)^{j-1}\\
    & \leq (L-1)C_L^{j-1}\Phi_{L, j - 1}(t)(L-1)^2\CL^2\para{\Psi_{L, 2}(t)}^{1/2}\para{\Phi_{L, 8}(t)}^{(1/4)}\para{\Psi_{L, 8}(t)}^{(1/4)}\\
    & = (L-1)^3C_L^{j+1}\Phi_{L, j - 1}(t)\para{\Psi_{L, 2}(t)}^{1/2}\para{\Phi_{L, 8}(t)}^{(1/4)}\para{\Psi_{L, 8}(t)}^{(1/4)}
\end{align*}
and 
\begin{align*}
    &\frac{d^4jN}{(L-1)^2\CL^{2+j}} \sum_{i= 1}^{L-1} \sum_{l=0}^{i-1}\left(\EP{ f^{i}_{\theta^{(i)}(0)}}+\EP{ f^{i}_{\theta^{(i)}(t)}- f^{i}_{\theta^{(i)}(0)}}\right)^{j-1}\EP{\W_i^{l+2}}\EP{(\W_L^{l+2})^{\intercal}}\\
    &\le \frac{81\max_l\{ n_l^4\}d^4jN^{3/2}}{(L-1)^2\CL^{2+j}} \sum_{i= 1}^{L-1} \\& \quad \times \sum_{l=0}^{i-1} \left(\EP{ f^{i}_{\theta^{(i)}(0)}}+\EP{ f^{i}_{\theta^{(i)}(t)}- f^{i}_{\theta^{(i)}(0)}}\right)^{j-1}
    \EP{ f^{l}_{\theta^{(l)}(t)}(x)}\EP{ f^{l}_{\theta^{(l)}(t)}(x')} \EP{\W_i^{l+2}}\EP{(\W_L^{l+2})^{\intercal}} .
\end{align*}

By combining the above inequalities, we have
\begin{align*}
    \partial_t \Phi_{L, j}(t) 
    &\le\frac{81\max_l\{ n_l^4\}jN^{3/2}}{(L-1)^2\CL^{2+j}} \sum_{i= 1}^{L-1} \sum_{l=0}^{i-1} \left(\EP{ f^{i}_{\theta^{(i)}(0)}}+\EP{ f^{i}_{\theta^{(i)}(t)}- f^{i}_{\theta^{(i)}(0)}}\right)^{j-1}\EP{ f^{l}_{\theta^{(l)}(t)}(x)}\EP{ f^{l}_{\theta^{(l)}(t)}(x')}\\
    &\quad\quad\times \EP{\W_i^{l+2}}\EP{(\W_L^{l+2})^{\intercal}} \EP{\dLt}\\
    &\quad +\frac{d^4jN}{(L-1)^2\CL^{2+j}} \sum_{i= 1}^{L-1} \sum_{l=0}^{i-1}\left(\EP{ f^{i}_{\theta^{(i)}(0)}}+\EP{ f^{i}_{\theta^{(i)}(t)}- f^{i}_{\theta^{(i)}(0)}}\right)^{j-1}\EP{\W_i^{l+2}}\EP{(\W_L^{l+2})^{\intercal}} \EP{\dLt}\\
    &\le \frac{162\max_l\{ n_l^4\}d^4jN^{3/2}(L-1)}{\CL}\Phi_{L, j - 1}(t)(\Psi_{L, 2}(t))^{1/2}(\Phi_{L, 8}(t))^{(1/4)}(\Psi_{L, 8}(t))^{(1/4)}\EP{\dLt}\\
    &\le \frac{162\max_l\{ n_l^4\}d^4jN^{3/2}(L-1)}{\CL}\Phi_{L, j - 1}(t)\Psi_{L, 2}(t)\Phi_{L, 8}(t)\Psi_{L, 8}(t)\EP{\dLt}.
\end{align*}
\end{proof}

\section{Proof of Lemma~\ref{lemm:l2-bound} for Proposition~\ref{prop:l2-norm}}\label{app:prp5}

\begingroup
\def\thetheorem{\ref{lemm:l2-bound}}
\begin{lemma}
For $t \geq 0$ and $x\in\fR_+^{d \times d}$, if $f^l_{\theta^{(l)}(t)}(x)$ is element-wise nonzero for all $1 \le l \le L-1$, then
\begin{align*}
    \partial_t \tilde{\Psi}_{L,2}(x,t)&\le \frac{j162d^4\max_l\{ n_l^4\}N^{1/2}(L-1)}{\CL} \para{\tilde{\Psi}_{L,2}(x,t)}^{2}\Phi_{L, 4}(t)\Psi_{L,8}(t)\EP{\dLt}\\
    \partial_t \tilde{\Phi}_{L, j}(x,t) &\le \frac{18d^4(L-1)}{\CL}\tilde{\Phi}_{L, j-1}(t)\tilde{\Phi}_{L,8}(x,t)\tilde{\Psi}_{L, 2}(x,t)\Phi_{L,8}(t)\Psi_{L,8}(t)\EP{\dLt}\\
    \partial_t \tilde{\Psi}_{L, 8}(x,t)& \le \frac{72d ^4(L-1)}{\CL}\tilde{\Phi}_{L,4}(x,t) \tilde{\Psi}_{L,2}(x,t)\para{\tilde{\Psi}_{L,8}(x,t)}^{2}\Psi_{L,8}(t)\EP{\dLt}.
\end{align*}
for $1\leq j\leq 8$.
\end{lemma}
\addtocounter{theorem}{-1}
\endgroup
\begin{proof}[Proof of Lemma~\ref{lemm:l2-bound}]
% Prop 2, part 1
Consider the derivative of $\tilde{\Psi}_{L, 8}$ first.
Like Section~\ref{app:not crossing}, we show that the intermediate pre-activation values never reach zero while training,
and therefore we can apply Lemma~\ref{lemm:bound derivative of l2 norm}.
\begin{align*}
    & \partial_t \tilde{\Psi}_{L, 8}(t)\\
    & \le \frac{72d^4}{(L-1)^2\CL^2}  \sum_{l =2 }^{L} \sum_{i=l}^{L} \EP{\W^{l+1}_{L}(t)}\ab{\W^{l+1}_{L}(t)}\ab{\W^i_{l-1}(t)}\left(\ab{\W_L^i(0)}+\ab{\W_L^i(t)-\W_L^i(0)}\right)^7 \\ &\quad \times\EP{\fultm(x)}\EP{\dLt}.
\end{align*}
Then,
\begin{align*}
     &\sum_{l =2 }^{L} \sum_{i=l}^{L} \EP{\W^{l+1}_{L}(t)}\ab{\W^{l+1}_{L}(t)}\ab{\W^i_{l-1}(t)}\left(\ab{\W_L^i(0)}+\ab{\W_L^i(t)-\W_L^i(0)}\right)^7 \EP{\fultm(x)}\\
     & \le \Bigg( \sum_{i =2 }^{L} \sum_{i=l}^{L} \ab{\W^i_{l-1}} \left(\EP{\fulzm}+\EP{\fultm-\fulzm}\right) \left(\ab{\W_L^{l+1}(0)}+\ab{\W_L^{l+1}(t)-\W_L^{l+1}(0)}\right)\\
     &\quad \times\left(\EP{\W_L^{l+1}(0)}+\EP{\W_L^{l+1}(t)-\W_L^{l+1}(0)}\right)\Bigg)\Bigg(\sum_{i= 2}^L \left(\EP{\W_L^i(0)}+\EP{\W_L^i(t)-\W_L^i(0)}\right)^8\Bigg)\\
    &\le \left(\sum_{l=2 }^{L} \sum_{i=2}^{l}\frac{L-1}{l-1} \ab{\W^i_{l}}^2\right)^{1/2} \Bigg((L-1)\sum^{L}_{l=2} \left(\EP{\fulzm}+\EP{\fultm-\fulzm}\right)^2 \\
    &\quad\times\left(\ab{\W_L^{l+1}(0)}+\ab{\W_L^{l+1}(t)-\W_L^{l+1}(0)}\right)^2\left(\EP{\W_L^{l+1}(0)}+\EP{\W_L^{l+1}(t)-\W_L^{l+1}(0)}\right)^2 \Bigg)^{1/2}(L-1)\tilde{\Psi}_{L, 8}\\
    &\le \left((L-1)^2\tilde{\Psi}_{L, 2}\right)^{1/2}\sqrt{L-1}\Bigg(\sum^{L}_{l=2} \left(\EP{\fulzm}+\EP{\fultm-\fulzm}\right)^4\Bigg)^{1/4}\\
    &\quad\times \Bigg(\sum_{l=2}^{L} \left(\ab{\W_L^{l+1}(0)}+\ab{\W_L^{l+1}(t)-\W_L^{l+1}(0)}\right)^4 \left(\EP{\W_L^{l+1}(0)}+\EP{\W_L^{l+1}(t)-\W_L^{l+1}(0)}\right)^4 \Bigg)^{1/4}(L-1)\tilde{\Psi}_{L, 8}\\
    &\le \left((L-1)^2\tilde{\Psi}_{L, 2}\right)^{1/2} \sqrt{L-1} (LC_L^4)^{1/4}(\Phi_{L, 4})^{1/4}(L-1)\tilde{\Psi}_{L, 8}\\
    &\quad\times \Bigg(\sum_{l=2}^{L} \left(\ab{\W_L^{l+1}(0)}+\ab{\W_L^{l+1}(t)-\W_L^{l+1}(0)}\right)^8\Bigg)^{1/8}
    \Bigg(\sum_{l=2}^{L}\left(\EP{\W_L^{l+1}(0)}+\EP{\W_L^{l+1}(t)-\W_L^{l+1}(0)}\right)^8 \Bigg)^{1/8}\\
    &= (L-1)\tilde{\Psi}_{L, 8}(L-1)^2\CL\para{\tilde{\Psi}_{{L}, 2}}^{1/2}\para{\tilde{\Phi}_{L, 4}}^{(1/4)}\para{\Psi_{L, 8}}^{(1/8)}\para{\tilde{\Psi}_{L, 8}}^{(1/8)}
\end{align*}

Thus, we have
\begin{align*}
    &\partial_t \tilde{\Psi}_{L, 8}(t)\\
    & \le \frac{72d^4}{(L-1)^2\CL^2}  \sum_{l =2 }^{L} \sum_{i=l}^{L} \EP{\W^{l+1}_{L}(t)}\ab{\W^{l+1}_{L}(t)}\ab{\W^i_{l-1}(t)}\left(\ab{\W_L^i(0)}+\ab{\W_L^i(t)-\W_L^i(0)}\right)^7 \EP{\fultm(x)}\EP{\dLt}\\ 
    &\le \frac{72d^4(L-1)}{\CL} \para{\tilde{\Psi}_{{L}, 2}}^{1/2}\para{\tilde{\Phi}_{L, 4}}^{1/4}\para{\Psi_{L, 8}}^{1/8}\para{\tilde{\Psi}_{L, 8}}^{9/8}\EP{\dLt}\\ 
    &\le \frac{72d^4 L}{\CL} \tilde{\Psi}_{{L}, 2}\tilde{\Phi}_{L, 4}\Psi_{L, 8}\para{\tilde{\Psi}_{L, 8}}^{2}\EP{\dLt}.
\end{align*}

Now, let consider the derivative of $\tilde{\Psi}_{L, 2}$.
From Lemma~\ref{lemm:bound derivative of l2 norm},
\begin{align*}
    &\partial_t \tilde{\Psi}_{L, 2}(t)\\ & \le \frac{18d^4}{(L-1)^3\CL^2} \sum^{ L}_{i=2} \sum_{ k= 2}^{i}\sum^{i}_{l=k} \frac{L-1}{i-1}\EP{\W^{l+1}_{L}}\ab{\W^{l+1}_{i}}\ab{\W^k_{l-1}}\left(\ab{\W_i^k(0)}+\ab{\W_i^k(t)-\W_i^k(0)}\right)\\ &\quad \times\EP{f^{l-1}_{\theta^{(l-1)}(t)}}\EP{\dLt}\\
    &\le \frac{9d^4}{(L-1)^3\CL^2}  \sum^{ L}_{i=2} \sum_{ k= 2}^{i}\sum^{i}_{l=k} \frac{L-1}{i-1}\EP{\W^{l+1}_{L}}\ab{\W^k_{l-1}}\para{\ab{\W^{l+1}_{i}}^2+\left(\ab{\W_i^k(0)}+\ab{\W_i^k(t)-\W_i^k(0)}\right)^2}\\ &\quad \times \EP{f^{l-1}_{\theta^{(l-1)}(t)}}\EP{\dLt}\\
    & =\frac{9d^4}{(L-1)^3\CL^2}  \sum^{ L}_{i=2} \sum_{ k= 2}^{i}\sum^{i}_{l=k} \frac{L-1}{i-1}\EP{\W^{l+1}_{L}}\ab{\W^k_{l-1}}\ab{\W^{l+1}_{i}}^2\EP{f^{l-1}_{\theta^{(l-1)}(t)}}\EP{\dLt}\\
    &\quad +\frac{9d^4}{(L-1)^3\CL^2}  \sum^{ L}_{i=2} \sum_{ k= 2}^{i}\sum^{i}_{l=k} \frac{L-1}{i-1}\EP{\W^{l+1}_{L}}\ab{\W^k_{l-1}}\left(\ab{\W_i^k(0)}+\ab{\W_i^k(t)-\W_i^k(0)}\right)^2\EP{f^{l-1}_{\theta^{(l-1)}(t)}}\EP{\dLt}.
\end{align*}

By Cauchy-Schwartz inequality,
% \begin{align*}
%     L*(L-1)\CL^4\Phi^{(4)}_L(t)\Psi^{(8)}_L(t) \ge \Bigg(\sum^{L}_{l=2} \left(\EP{f^{l-1}_{\theta^{(l-1)}(0)}}+\EP{f^{l-1}_{\theta^{(l-1)}(0)}-f^{l-1}_{\theta^{(l-1)}(0)}}\right)^2\left(\EP{\W_L^{l+1}(0)}+\EP{\W_L^{l+1}(t)-\W_L^{l+1}(0)}\right)^2\Bigg)^2,
% \end{align*} 
% and
\begin{align*}
    &\para{\sum_{k=2 }^{L} \sum_{l=k}^{L} \ab{\W^k_{l-1}}\para{\EP{\fulzm}+\EP{\fultm-\fulzm}} \left(\EP{\W_L^{l+1}(0)}+\EP{\W_L^{l+1}(t)-\W_L^{l+1}(0)}\right)}^2\\
    &\le\left(\sum_{k =2 }^{L} \sum_{l=k}^{L}\frac{L-1}{l-1} \ab{\W^k_{l}}^2\right) \\
    &\quad\times \left((L-1)\sum^L_{l=2} \left(\EP{\fulzm}+\EP{\fultm-\fulzm}\right)^2\left(\EP{\W_L^{l+1}(0)}+\EP{\W_L^{l+1}(t)-\W_L^{l+1}(0)}\right)^2\right)\\
    &\le (L-1)^2 \tilde{\Psi}_{L, 2}(L-1) \sqrt{(L-1) \CL^4\Phi_{L, 4}L\Psi_{L, 8}}\\
    & \le (L-1)^4\CL^2\tilde{\Psi}_{L, 2} \left(\Phi_{L, 4}\right)^{(1/2)}\para{\Psi_{L, 8}}^{(1/2)}.
\end{align*}

Then, we have
\begin{align*}
    &\sum^{ L}_{i=2} \sum_{ k= 2}^{i}\sum^{i}_{l=k} \frac{L-1}{i-1}\EP{\W^{l+1}_{L}}\ab{\W^k_{l-1}}\ab{\W^{l+1}_{i}}^2\EP{f^{l-1}_{\theta^{(l-1)}(t)}}\\
    &\le \Bigg( \sum_{k=2 }^{L} \sum_{l=k}^{L} \ab{\W^k_{l-1}}\para{\EP{\fulzm}+\EP{\fultm-\fulzm}} \left(\EP{\W_L^{l+1}(0)}+\EP{\W_L^{l+1}(t)-\W_L^{l+1}(0)}\right)\Bigg)\\
    &\quad\times \left(\sum_{i=2}^L \sum^i_{l=2}\frac{L-1}{i-1}\ab{\W_i^{l+1}}^2\right)\\
    &\le (L-1)^2\tilde{\Psi}_{L, 2}(L-1)^2\CL\para{\tilde{\Psi}_{L, 2}}^{(1/2)}\para{\Phi_{L, 2}}^{(1/4)}\para{\Psi_{L, 8}}^{(1/4)}\\
    &=(L-1)^4\CL \para{\tilde{\Psi}_{L, 2}}^{(3/2)}\para{\Phi_{L, 4}}^{(1/2)}\para{\Psi_{L, 8}}^{(1/4)}
\end{align*}
and
\begin{align*}
    &\sum^{ L}_{i=2} \sum_{ k= 2}^{i}\sum^{i}_{l=k} \frac{L-1}{i-1}\EP{\W^{l+1}_{L}}\ab{\W^k_{l-1}}\left(\ab{\W_i^k(0)}+\ab{\W_i^k(t)-\W_i^k(0)}\right)^2\EP{f^{l-1}_{\theta^{(l-1)}(t)}}\\
    &\le\Bigg( \sum_{k=2 }^{L} \sum_{l=k}^{L} \ab{\W^k_{l-1}}\para{\EP{\fulzm}+\EP{\fultm-\fulzm}} \left(\EP{\W_L^{l+1}(0)}+\EP{\W_L^{l+1}(t)-\W_L^{l+1}(0)}\right)\Bigg)\\
    &\quad \times\left(\sum_{i=2}^L \sum^i_{k=2}\frac{L-1}{i-1}\left(\ab{\W_i^k(0)}+\ab{\W_i^k(t)-\W_i^k(0)}\right)^2\right)\\
    &\le (L-1)^2\tilde{\Psi}_{L, 2}(L-1)^2\CL\para{\tilde{\Psi}_{L, 2}}^{(1/2)}\para{\Phi_{L, 4}}^{(1/4)}\para{\Psi_{L, 8}}^{(1/4)}\\
    &=(L-1)^4\CL \para{\tilde{\Psi}_{L, 2}}^{(3/2)}\para{\Phi_{L, 4}}^{(1/4)}\para{\Psi_{L, 8}}^{(1/4)}.
\end{align*}

Thus, by combining the above two inequalities,
\begin{align*}
    &\partial_t \tilde{\Psi}_{L, 2}(t)\\
    & \le \frac{9d^4}{(L-1)^3\CL^2}  \sum^{ L}_{i=2} \sum_{ k= 2}^{i}\sum^{i}_{l=k} \frac{L-1}{i-1}\EP{\W^{l+1}_{L}}\ab{\W^k_{l-1}}\para{\ab{\W^{l+1}_{i}}^2+\left(\ab{\W_i^k(0)}+\ab{\W_i^k(t)-\W_i^k(0)}\right)^2}\\ &\quad \times\EP{f^{l-1}_{\theta^{(l-1)}(t)}}\EP{\dLt}\\
    & \le \frac{18d^4(L-1)}{\CL} \para{\tilde{\Psi}_{L, 2}}^{3/2}\para{\Phi_{L, 4}}^{1/4}\para{\Psi_{L, 8}}^{1/4}\EP{\dLt}\\
    & \le \frac{18d^4(L-1)}{\CL} \para{\tilde{\Psi}_{L, 2}}^{2}\Phi_{L, 4}\Psi_{L, 8}\EP{\dLt}
\end{align*}

Finally, let consider the derivatives of $\tilde{\Phi}_{L, j}$ for $1\leq j\leq 8$.
From Lemma~\ref{lemm:bound derivative of l2 norm},
\begin{align*}
    &\partial_t \tilde{\Phi}_{L, j}(t) 
    \\& \le\frac{j81\max_l\{ n_l^4\}N^{1/2}}{(L-1)^2\CL^{2+j}} \sum_{i= 1}^{L-1} \sum_{l=0}^{i-1} \left(\ab{ f^{i}_{\theta^{(i)}(0)}}+\ab{ f^{i}_{\theta^{(i)}(t)}- f^{i}_{\theta^{(i)}(0)}}\right)^{j-1}\EP{ f^{l}_{\theta^{(l)}(t)}(x)}\ab{ f^{l}_{\theta^{(l)}(t)}(x')}\\ &\quad \times \ab{\W_i^{l+2}}\EP{(\W_L^{l+2})^{\intercal}} \EP{\dLt}\\
    &\quad +\frac{jd^4}{(L-1)^2\CL^{2+j}} \sum_{i= 1}^{L-1} \sum_{l=0}^{i-1}\left(\ab{ f^{i}_{\theta^{(i)}(0)}}+\ab{ f^{i}_{\theta^{(i)}(t)}- f^{i}_{\theta^{(i)}(0)}}\right)^{j-1}\ab{\W_i^{l+2}}\EP{(\W_L^{l+2})^{\intercal}} \EP{\dLt}\\
    &\le \frac{j162d^4\max_l\{ n_l^4\}N^{1/2}}{(L-1)^2\CL^{2+j}} \sum_{i= 1}^{L-1} \sum_{l=0}^{i-1} \left(\ab{ f^{i}_{\theta^{(i)}(0)}}+\ab{ f^{i}_{\theta^{(i)}(t)}- f^{i}_{\theta^{(i)}(0)}}\right)^{j-1}\EP{ f^{l}_{\theta^{(l)}(t)}(x)}\ab{ f^{l}_{\theta^{(l)}(t)}(x')}\\ &\quad \times\ab{\W_i^{l+2}}\EP{(\W_L^{l+2})^{\intercal}} \EP{\dLt}
\end{align*}
since $\ab{f^l_{\theta^{(l)}(t)}(x)} \geq 1$.

By Cauchy-Schwartz inequality,
\begin{align*}
    &\Bigg(\sum^{L-2}_{l=0} \left(\EP{\fulz}+\EP{\fult-\fulz}\right)^2 \left(\ab{\fulz}+\ab{\fult-\fulz}\right)^2\\& \quad \times \left(\EP{\W_L^{l+2}(0)}+\EP{\W_L^{l+2}(t)-\W_L^{l+2}(0)}\right)^2\Bigg)^{2}
    \\&\le \Bigg(\sum^{L-2}_{l=0} \left(\EP{\fulz}+\EP{\fult-\fulz}\right)^4 \left(\ab{\fulz}+\ab{\fult-\fulz}\right)^4\Bigg)\\
    % &\quad\times \Bigg(\sum^L_{l=2} \left(\EP{\W_L^l(0)}+\EP{\W_L^l(t)-\W_L^l(0)}\right)^4\Bigg)\\
    &\le\Bigg(\sum^{L-2}_{l=0} \left(\EP{\fulz}+\EP{\fult-\fulz}\right)^4 \left(\ab{\fulz}+\ab{\fult-\fulz}\right)^4\Bigg)\\
    &\quad\times \Bigg(\sum^L_{l=2} \left(\EP{\W_L^l(0)}+\EP{\W_L^l(t)-\W_L^l(0)}\right)^4\Bigg)\\
    &\le\Bigg(\sum^{L-2}_{l=0} \left(\EP{\fulz}+\EP{\fult-\fulz}\right)^8\Bigg)^{1/2} \Bigg(\left(\ab{\fulz}+\ab{\fult-\fulz}\right)^8\Bigg)^{1/2}\\
    &\quad\times \Bigg(\sum^L_{l=2} \left(\EP{\W_L^l(0)}+\EP{\W_L^l(t)-\W_L^l(0)}\right)^8\Bigg)\\
    &\le  (L-1)\Psi_{L, 8} (L-1)\CL^8\para{\Phi_{L, 8}}^{1/2}\para{\tilde{\Phi}_{L, 8}}^{1/2}.
\end{align*} 

Then, we have
\begingroup
\allowdisplaybreaks
\begin{align*}
    &\sum_{i= 1}^{L-1} \sum_{l=0}^{i-1} \left(\ab{ f^{i}_{\theta^{(i)}(0)}}+\ab{ f^{i}_{\theta^{(i)}(t)}- f^{i}_{\theta^{(i)}(0)}}\right)^{j-1}\EP{ f^{l}_{\theta^{(l)}(t)}(x)}\ab{ f^{l}_{\theta^{(l)}(t)}(x')}\ab{\W_i^{l+2}}\EP{(\W_L^{l+2})^{\intercal}}\\
    &\le \para{\sum_{i= 1}^{L-1} \left(\ab{ f^{i}_{\theta^{(i)}(0)}}+\ab{ f^{i}_{\theta^{(i)}(t)}- f^{i}_{\theta^{(i)}(0)}}\right)^{j-1}}\Bigg(\sum^{L-1}_{i=1} \sum^{i-1}_{l=0}\frac{L-1}{i-1} \ab{\W_i^{l+2}} \left(\EP{\fulz}+\EP{\fult-\fulz}\right)\\
    &\quad \times \left(\ab{\fulz}+\ab{\fult-\fulz}\right) \left(\EP{\W_L^{l+2}(0)}+\EP{\W_L^{l+2}(t)-\W_L^{l+2}(0)}\right)\Bigg)\\
    &\le (L-1)\CL^{j-1}\tilde{\Phi}_{L, j-1} \Bigg(\sum^{L-1}_{i=1} \sum^{i-1}_{l=0}\frac{L-1}{i-1} \ab{\W_i^{l+2}} \left(\EP{\fulz}+\EP{\fult-\fulz}\right)\\
    &\quad \times \left(\ab{\fulz}+\ab{\fult-\fulz}\right) \left(\EP{\W_L^{l+2}(0)}+\EP{\W_L^{l+2}(t)-\W_L^{l+2}(0)}\right)\Bigg)\\
    &\le (L-1)\CL^{j-1}\tilde{\Phi}_{L, j-1} \para{\sum^{L-1}_{i=1} \sum^{i-1}_{l=0}\frac{L-1}{i-1} \ab{\W_i^{l+2}}^2}^{1/2} \Bigg((L-1)\sum^{L-2}_{l=0} \left(\EP{\fulz}+\EP{\fult-\fulz}\right)^2 \\
    &\quad \left(\ab{\fulz}+\ab{\fult-\fulz}\right)^2\times  \left(\EP{\W_L^{l+2}(0)}+\EP{\W_L^{l+2}(t)-\W_L^{l+2}(0)}\right)^2\Bigg)^{1/2}\\
    &\le (L-1)\CL^{j-1} \tilde{\Phi}_{L, j-1} \left((L-1)^2\tilde{\Psi}_{L, 2}\right)^{1/2}
    \left((L-1)^2\CL^4\para{\Phi_{L, 8}}^{1/4}\para{\Psi_{L, 8}}^{1/2}\para{\tilde{\Phi}_{L, 8}}^{1/4}\right)^{1/2}\\
    &=(L-1)^3\CL^{j+1}\tilde{\Phi}_{{L}, j - 1}\para{\tilde{\Psi}_{L, 2}}^{1/2}\para{\Phi_{L, 8}}^{1/8}\para{\Psi_{L, 8}}^{1/4}\para{\tilde{\Phi}_{L, 8}}^{1/8} 
\end{align*}
\endgroup

Thus, by combining the above inqualities,
\begin{align*}
    &\partial_t \tilde{\Phi}_{L, j}(t) \\
    &\le \frac{j162d^2\max_l\{ n_l^4\}N^{1/2}}{(L-1)^2\CL^{2+j}} \sum_{i= 1}^{L-1} \sum_{l=0}^{i-1} \left(\ab{ f^{i}_{\theta^{(i)}(0)}}+\ab{ f^{i}_{\theta^{(i)}(t)}- f^{i}_{\theta^{(i)}(0)}}\right)^{j-1}\EP{ f^{l}_{\theta^{(l)}(t)}(x)}\ab{ f^{l}_{\theta^{(l)}(t)}(x')}\\ &\quad \times\ab{\W_i^{l+2}}\EP{(\W_L^{l+2})^{\intercal}} \EP{\dLt}\\
    &\le \frac{j162d^2\max_l\{ n_l^4\}N^{1/2}(L-1)}{\CL}\tilde{\Phi}_{{L},  j - 1}\para{\tilde{\Psi}_{L, 2}}^{1/2}\para{\Phi_{L, 8}}^{1/8}\para{\Psi_{L, 8}}^{1/4}\para{\tilde{\Phi}_{L, 8}}^{1/8}\EP{\dLt}\\
    &\le \frac{j162d^2\max_l\{ n_l^4\}N^{1/2}(L-1)}{\CL}\tilde{\Phi}_{{L}, j - 1}\tilde{\Psi}_{L, 2}\Phi_{L, 8}\Psi_{L, 8}\tilde{\Phi}_{L, 8}\EP{\dLt}.
\end{align*}
\end{proof}

\section{Proof of Lemma~\ref{lemm:layer-bound} for Proposition~\ref{prop:layerwise}}\label{app:prop6}
\begingroup
\def\thetheorem{\ref{lemm:layer-bound}}
\begin{lemma}
For $t \geq 0$ and $x\in\fR_+^{d \times d}$, if $f^l_{\theta^{(l)}(t)}(x)$ is element-wise nonzero for all $1 \le l \le L-1$, then
\begin{align*}
    \partial_t\tilde{\Psi}^{k}_L(x,t) &\le\frac{9d^4(L-1)}{\CL} \Psi_{L,8}(t)\tilde{\Phi}_{L,4}(x,t)\tilde{\Psi}_{L,2}(x,t)\tilde{\Psi}_{L,8}(x,t)\EP{\dLt}\\
    \partial_t\tilde{\Phi}^{l}_L(x,t) &\le \frac{81d^4\max_l\{n_{l}^4\}N^{1/2}(L-1)}{\CL}\Phi_{L,8}(t)\Psi_{L,8}(t)\tilde{\Phi}_{L,8}(x,t)\tilde{\Psi}_{L, 2}(x,t)\EP{\dLt}.
\end{align*}
for $1 \le l \le L$ and $2 \le k \le L$.
\end{lemma}
\addtocounter{theorem}{-1}
\endgroup

\begin{proof}[Proof of Lemma~\ref{lemm:layer-bound}]
Consider the derivative of $\tilde{\Psi}_L^k$ first.
In the proof of Lemma~\ref{lemm:l2-bound}, 
we showed
\begin{align*}
    &\Bigg(\sum_{i =2 }^{L} \sum_{l=i}^{L} \ab{\W^i_{l-1}} \left(\EP{\fulzm}+\EP{\fultm-\fulzm}\right) \left(\ab{\W_L^{l+1}(0)}+\ab{\W_L^{l+1}(t)-\W_L^{l+1}(0)}\right)\\
    &\quad \times\left(\EP{\W_L^{l+1}(0)}+\EP{\W_L^{l+1}(t)-\W_L^{l+1}(0)}\right)\Bigg)^2\\
    &\le (L-1)^2 \tilde{\Psi}_{L, 2} (L-1)\times (L-1)\CL^2\para{\Phi_{L, 4}}^{1/2}\para{\Psi_{L, 8}(t)}^{1/4}\para{\tilde{\Psi}_{L, 8}(t)}^{1/4}.
\end{align*}
Thus, from Lemma~\ref{lemm:bound derivative of l2 norm},
\begin{align*}
    \partial_t \tilde{\Psi}^{k}_L(t)
    & \le  \frac{9d^4}{(L-1)\CL^2}  \sum_{l=k}^{L} \EP{\W^{l+1}_{L}(t)}\ab{\W^{l+1}_{L}(t)}\ab{\W^k_{l-1}(t)} \EP{\fultm(x)}\EP{\dLt}\\ 
    &\le  \frac{9d^4(L-1)}{\CL} \para{\tilde{\Psi}_{{L}, 2}}^{1/2}\para{\tilde{\Phi}_{L, 4}}^{(1/4)}\para{\Psi_{L, 8}}^{(1/8)}\para{\tilde{\Psi}_{L, 8}}^{(1/8)}\EP{\dLt}\\ 
    &\le \frac{9d^4(L-1)}{\CL} \tilde{\Psi}_{L, 2}\tilde{\Phi}_{L, 4}\Psi_{L, 8}\tilde{\Psi}_{L, 8}\EP{\dLt}.
\end{align*}

On the other hand, let consider the derivative of $\tilde{\Phi}^l_L(t)$.
We also showed the following inequality from the proof of Lemma~\ref{lemm:l2-bound}.
\begin{align*}
    &\sum^{L-1}_{i=1} \sum^{i-1}_{l=0}\frac{L-1}{i-1} \ab{\W_i^{l+2}} \left(\EP{\fulz}+\EP{\fult-\fulz}\right)\left(\ab{\fulz}+\ab{\fult-\fulz}\right)\\
    &\quad\times \left(\EP{\W_L^{l+2}(0)}+\EP{\W_L^{l+2}(t)-\W_L^{l+2}(0)}\right)\\
    &\le (L-1)^2\CL^2\para{\tilde{\Psi}_{L, 2}}^{1/2}\para{\Phi_{L, 8}}^{1/8}\para{\Psi_{L, 8}}^{1/4}\para{\tilde{\Phi}_{L, 8}}^{1/8}.
\end{align*}
Thus, from Lemma~\ref{lemm:bound derivative of l2 norm},
\begin{align*}
    \partial_t \tilde{\Phi}^{l}_L(t) &\le\frac{81n_i^2n_{i-1}^2N^{1/2}}{(L-1)\CL^3} \sum_{i=0}^{l-1} \EP{ f^{i}_{\theta^{(i)}(t)}(x)}\ab{ f^{i}_{\theta^{(i)}(t)}(x')}\ab{\W_l^{i+2}}\EP{(\W_L^{i+2})^{\intercal}} \EP{\dLt}\\
    &\quad +\frac{d^4}{(L-1)\CL^3} \sum_{i=0}^{l-1}\ab{\W_l^{i+2}}\EP{(\W_L^{i+2})^{\intercal}} \EP{\dLt}\\
    &\le \frac{81d^4\max_l\{n_{l}^4\}N^{1/2}(L-1)}{\CL}\para{\tilde{\Psi}_{L, 2}}^{1/2}\para{\Phi_{L, 8}}^{1/8}\para{\Psi_{L, 8}}^{1/4}\para{\tilde{\Phi}_{L, 8}}^{1/8}\EP{\dLt}\\
    &\le \frac{81d^4\max_l\{n_{l}^4\}N^{1/2}(L-1)}{\CL}\tilde{\Psi}_{L, 2}\Phi_{L, 8}\Psi_{L, 8}\tilde{\Phi}_{L, 8}\EP{\dLt}.
\end{align*}
\end{proof}
\end{document}